\def\fileversion{v2.25a} \def\filedate{2009/10/10}

\documentclass[edeposit,fullpage]{uiucthesis2009}

\usepackage[pagebackref=true,breaklinks=true,letterpaper=true,colorlinks,linkcolor=blue,linktoc=page,citecolor=blue,bookmarks=false]{hyperref}
\usepackage{rotating}
\usepackage[abs]{overpic}
\usepackage{graphicx}
\usepackage{times}
\usepackage{amsmath}
\usepackage{amssymb}
\usepackage{subfigure}
\usepackage{colortbl}
\usepackage{xcolor}
\usepackage{color}
\usepackage{multirow}
\usepackage{relsize}
\usepackage{epsfig}
\usepackage{float}
\usepackage[]{algorithm2e}

\definecolor{darkyellow}{rgb}{1,.9,0}
\definecolor{dblue}{RGB}{67,88,178}
\definecolor{dyellow}{RGB}{178,146,50}

\newcommand{\urlwofont}[1] {\urlstyle{same}\url{#1} }
\newcommand{\mb}[1]{\mathbf{#1}}
\newcommand{\tb}[1]{\textbf{#1}}
\newcommand{\boldhead}[1]{\vspace{0.05in}\noindent\textbf{#1.}}
\newcommand{\todo}[1]{{ \bf TODO:} {\it #1}}
\newcommand{\ea}[0]{et al.}
\newcommand{\ve}[1]{{\bf #1}}
\newcommand{\comment}[1]{}
\newcommand{\hili}[1]{{\bf #1}}

\newcommand{\supi}[1]{{#1}^{ {\left( i \right)} }}
\newcommand{\round}[1]{ \operatorname{round}\left( #1 \right) }
\providecommand{\norm}[1]{\left\lVert#1\right\rVert}

\def\argmax{\mathop{\rm argmax}}
\def\argmin{\mathop{\rm argmin}}
\def\argmaxu{\argmax\limits}
\def\argminu{\argmin\limits}
\def\maxu{\max\limits}
\def\sumu{\displaystyle\sum}

\begin{document}

\title{Inverse Rendering Techniques for Physically Grounded Image Editing}
\author{Kevin Karsch}
\department{Computer Science}
\schools{B.S., University of Missouri, 2009}
\phdthesis
\advisor{David Forsyth}
\degreeyear{2015}
\committee{Professor David Forsyth, Chair\\Assistant Professor Derek Hoiem\\Professor John Hart\\Principal Researcher Sing Bing Kang, Microsoft Research}
\maketitle

\frontmatter

\begin{abstract}
From a single picture of a scene, people can typically grasp the spatial layout immediately and even make good guesses at materials properties and where light is coming from to illuminate the scene. For example, we can reliably tell which objects occlude others, what an object is made of and its rough shape, regions that are illuminated or in shadow, and so on. It is interesting how little is known about our ability to make these determinations; as such, we are still not able to robustly "teach" computers to make the same high-level observations as people.

This document presents algorithms for understanding intrinsic scene properties from single images. The goal of these {\it inverse rendering} techniques is to estimate the configurations of scene elements (geometry, materials, luminaires, camera parameters, etc) using only information visible in an image. Such algorithms have applications in robotics and computer graphics. One such application is in {\it physically grounded image editing}: photo editing made easier by leveraging knowledge of the physical space. These applications allow sophisticated editing operations to be performed in a matter of seconds, enabling seamless addition, removal, or relocation of objects in images.
\end{abstract}


\chapter*{Acknowledgments}
I would like to thank all of the people and agencies who have contributed to this thesis.

First, I thank my thesis advisor David Forsyth for his guidance and supervision throughout my doctoral studies. David's seemingly endless supply of ideas insured that we were never stuck on a problem for long, and his fervor for research made our work extremely enjoyable. Derek Hoiem also provided a great deal of support and guidance that complemented David's many virtues. Together, their intellectual and emotional support could not have been more ideal for my studies. I am grateful for the opportunity I had to work with both David and Derek and proud of our accomplishments thus far.

I also thank John Hart and Sing Bing Kang, members of my thesis committee, who have  provided invaluable advice in my studies and led me to the completion of several of my most valued projects.

Through various internships and external collaborations I have had many indispensable mentors. These outside projects have broadened my experiences and helped define my past and future research objectives, and I thank Ce Liu, Mani Golparvar-Fard, Kalyan Sunkavalli, Sunil Hadap, Nathan Carr, and Hailin Jin.

I am grateful for the opportunities I have had to work with many great peer collaborators, including Zicheng Liao, Varsha Hedau, Jason Rock, Jon Barron, Rafael Fonte, Michael Sittig, and Keunhong Park, Brett Jones, Rajinder Sodhi, as well as the innumerable friends and students at the University of Illinois, in particular the members of the computer vision and graphics groups and at the University of Missouri.

I also thank Ye Duan, my undergraduate mentor, for allowing me to explore various research areas early in my career and also teaching me many concepts that transferred directly to my graduate studies, as well as Don Sievert for his general guidance and overall wisdom.

Finally, I thank my parents, Linda and Robert, for their many years of love and support. I owe my deepest gratitude to my fianc\'{e}e, Lindsey Lanfersieck, whose encouragement, patience, and love has continually motivated me throughout my career and enabled me to achieve a doctoral degree.

\tableofcontents

\comment{
\chapter{List of Abbreviations}
\begin{symbollist*}
\item[CA] Caffeine Addict.
\item[CD] Coffee Drinker.
\end{symbollist*}
}

\comment{
\chapter{List of Symbols}
\begin{symbollist}[0.7in]
\item[$\tau$] Time taken to drink one cup of coffee.
\item[$\mu$g] Micrograms (of caffeine, generally).
\end{symbollist}
}


\mainmatter

\chapter{Introduction}

From a single picture of a scene, people can typically grasp an image's spatial layout immediately and even make educated guesses at material properties and where light is coming from to illuminate the scene. For example, we can reliably tell which objects occlude others, the rough shape of objects, what things are made of, what regions are illuminated or in shadow, and so on. It is interesting how little is known about our ability to make these determinations; as such, we are still not able to robustly ``teach'' computers to make the same high-level observations as people.

In this proposal, we explore algorithms for understanding intrinsic scene properties from single images. We pose these questions from a computer graphics standpoint: given a digital photograph, what are the most likely configurations of scene elements (geometry, materials, light sources, camera parameters, etc)? The process of estimating these properties from images has come to be known as {\it inverse rendering}. Clearly an underconstrained problem, many questions must be answered to form reasonable solutions: How do we represent the space of a scene? What cues allow us to estimate these properties? Can we exploit relational knowledge for improved inference?

A better understanding of this process is critical to advancing computer vision and machine understanding. Reliably estimating these properties can allow us to answer questions such as: Where can I walk? What can I grab? What can I move? Where are the light sources? We show that this inferred scene knowledge is also extremely useful for image editing and visualization and focus our attention on such applications in this proposal. We use the term {\it physically grounded image editing} to imply the editing of images imbued with knowledge of the physical space, and we show that sophisticated editing operations (object insertion, removal, relighting, etc) can be performed in a matter of seconds. For example, we demonstrate image editors that can perform scene completion, segmentation, texture synthesis, relighting, and object scaling that take into account the room space, lighting, and contact surfaces. These tools allow seamless addition, removal, or relocation of objects in images of indoor scenes.

\section{Digital image formation}
Digital images are created through the physical interaction of many factors. Light traverses a scene, reflecting off and passing through surfaces, including the camera lens(es), until the light arrives at the at the digital sensor where it is recorded and accumulated. As light travels around a scene, many different interactions take place. Photons, the smallest entities of light, are absorbed, reflected, refracted, diffracted (bent around surfaces), as well as even more complex interactions like fluorescence and phosphorescence. These interactions are dictated by the surfaces in the scene -- primarily the geometry and the materials they are composed of.

From a computational perspective, these interactions must all be quantized and simulated. Furthermore, in order to simplify these processes, many of the more complicated interactions are ignored (e.g. diffraction). For example, in photorealistic computer graphics rendering, light is typically quantized as individual rays that are ``traced'' around the scene. These rays are reflected and refracted according other quantized scene elements: surface normals\footnote{Surface normals (the direction perpendicular to the surface at a given point) are used in simulating the flow of light around a scene. They are computed from synthetic surfaces that are typically represented as either implicit functions or explicit polygons.} and bidirectional scattering distribution functions\footnote{BSDFs are a common way to encode materials: given an incoming light direction and a surface normal, as well as the outgoing direction of light, a BSDF returns the light reflected along the outgoing direction} (BSDFs) of the synthetic surfaces in the scene.

\section{Inverse rendering}
Our goal is to reverse the image formation process. In other words, we attempt to explain the 3D properties and interactions of a scene that is depicted in a single picture. This process has been called {\it inverse rendering}, or the process of recovering intrinsic properties of a 3D scene through photometric cues. For example, given an image, we attempt to find the most likely configuration of the 3D scene (geometry, materials, light sources, camera parameters, etc).

In general, human observers, if shown a picture, are typically capable of determining the shape of objects, distances and heights, material properties, dominant light sources, and so on. This is believed to be due to the fact that humans are processing scenes with many tasks in mind, such as detection (which pixels belong to which object), recognition (what object does a group of pixels represent), reconstruction (what shape is depicted by a group of pixels), and many others. This joint processing, coupled with a person's years of experience in observing 3D environments, allows for a robust determination of these intrinsic scene properties.

\begin{figure*}
\includegraphics[width=\linewidth]{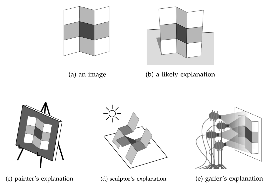}
\caption{The workshop metaphor from Adelson and Pentland~\cite{Adelson:1996} illustrating a few fundamental ambiguities that complicates inverse rendering. The image (a) is probably depicting a folded, multi-colored sheet (b), but it could simply be a painting (c), a complex sculpture (d), or a lighting setup (e), among other arrangements.}
\label{fig:workshop}
\end{figure*}

However, teaching a computer to make these determinations has proven difficult and is one of the fundamental problems that computer vision strives to solve. There are many reasons for this. For one, there are many pieces of a scene (the shape of all objects, their distance from the camera, their materials, where the lights are, camera settings, and many more) that must be estimated from just a single picture, which, when input to a computer, is nothing more than a grid of RGB pixel values. It is also difficult to tractably represent all of these components computationally, and even more difficult to estimate them. Underlying all of these issues are the ambiguities associated with image formation and perspective projection: for nearly all scenes, an infinite number of configurations of geometry, material, and light, can produce {\it exactly the same image}. Figure~\ref{fig:workshop} demonstrates a few common examples of these ambiguities, but there are many others (consider a flat mirror reflecting part of a scene, or that curved 3D lines/shapes can appear flat in perspective photos, and so on).

We believe that the keys to solving such issues lie in the joint estimation of scene properties, as well as providing significant evidence to a computer of other scenes in a wide variety of configuration (``training examples''). The human vision and understanding pipeline also provides many clues for achieving these goals. We survey existing scene representations and estimation techniques in this thesis, and in each chapter, propose several new ideas or improvements to existing methods for inverse rendering.

\section{Thesis overview and contributions}
Each chapter of this thesis examines one or more of the problems of inverse rendering and demonstrates the use of these recovered properties for various computer graphics and computer vision applications. Chapter~\ref{chap:rsolp} motivates the goal of physically grounded image editing and discusses a semiautomatic framework for inverse rendering that allows for quick and easy 3D object insertion~\cite{Karsch:SA2011}. In addition to this system, we present a semiautomatic algorithm for estimating a physical lighting model from a single image. Our method can generate a full lighting model that is demonstrated to be physically meaningful through a ground truth evaluation. We also introduce a novel image decomposition algorithm that uses geometry to improve lightness estimates, and we show in another evaluation to be state-of-the-art for single image reflectance estimation. A user study shows that the results of our method are confusable with real scenes, even for people who believe they are good at telling the difference. Our study also shows that our method is competitive with other insertion methods while requiring less scene information.

Representations of shape and various geometry estimation procedures are explored in Chapter~\ref{chap:shape}. We show that our shape estimation technique surpasses existing methods~\cite{Karsch:CVPR13} and is especially useful for image fragment relighting~\cite{Liao:CVPR15}, and that human observers prefer our shape estimates over other existing methods for the use case of physically grounded image editing. Another contribution in this chapter is in evaluating the importance of various boundary and shading cues for shape reconstruction and shape-based recognition. We extend Barron and Malik's shape from shading and silhouette method~\cite{Barron2012B} to include interior occlusions with figure/ground labels, folds, and sharp/soft boundary labels.  We also introduce perceptual and recognition-based measures of reconstruction quality. Our findings suggest a 3D representation that incorporates interior occlusions and folds might benefit such existing systems.

Estimating materials and reflectance properties of objects from a single view is discussed in Chapter~\ref{chap:mats}. Our technique recovers spatially varying materials, and we demonstrate applications of re-rendering (material transfer) and material classification~\cite{Karsch:SA_ISU:14}. Our primary contribution in this chapter is a technique for extracting spatially varying material reflectance (beyond diffuse parameters) directly from an object's appearance in a single photograph {\it without requiring any knowledge of the object's shape or scene illumination}. We use a low-order parameterization of material, and develop a new model of illumination that can be described also with only a few parameters, allowing for efficient rendering. Because our model has few parameters, we tend to get low variance and thus robustness in our material estimates (e.g. bias-variance tradeoff). By design, our material model is the same that is used throughout the 3D art/design community, and describes a large family of real-world materials. We show how to efficiently estimate materials from plausible initializations of lighting and shape, and propose novel priors that are crucial in estimating material robustly. Our material estimates perform favorably to baseline methods and measure well with ground truth, and we demonstrate results for both synthetic and real images. Such material estimates have applications in both the domains of vision and graphics.

In Chapter~\ref{chap:illum}, we demonstrate a fully automatic system for inverse rendering of a single image. Our primary contribution is a completely automatic algorithm for estimating a full 3D scene model from a single LDR photograph, which can be used to quickly and intuitively perform physically grounded image edits~\cite{Karsch:TOG:2014}. We have developed a novel, data-driven illumination estimation procedure that automatically estimates a physical lighting model for the entire scene (including out-of-view light sources). This estimation is aided by our single-image light classifier to detect emitting pixels, which we believe is the first of its kind. We also demonstrate state-of-the-art depth estimates by combining data-driven depth inference with geometric reasoning. We have created an intuitive interface for inserting 3D models seamlessly into photographs, using our scene approximation method to relight the object and facilitate drag-and-drop insertion. Our interface also supports other physically grounded image editing operations, such as post-process depth-of-field and lighting changes. In a user study, we show that our system is capable of making photorealistic edits: in side-by-side comparisons of ground truth photos with photos edited by our software, subjects had a difficult time choosing the ground truth.

Finally, we discuss advanced image editing and visualization applications of inverse rendering in Chapter~\ref{chap:applications}, including techniques for visualizing and monitoring construction sites automatically~\cite{Karsch:SA:14}. We also demonstrate a new, user-assisted Structure-from-Motion method, which leverages 2D-3D point correspondences between a mesh model and one image in the collection. We propose new objective functions for the classical point-$n$-perspective and bundle adjustment problems, and demonstrate that our SfM method outperforms existing approaches.

A discussion concludes the thesis in Chapter~\ref{chap:conclusion}.

\section{Other work not in thesis}
Throughout the course of my doctoral studies, I have also explored areas outside of single image inference in a variety of fields. Under the supervision of Sing Bing Kang and Ce Liu at Microsoft Research, I developed a technique for estimating dense depth from images and videos and converting such media automatically into stereoscopic 3D~\cite{Karsch:TPAMI:14,Karsch:ECCV12}. With John Hart, I implemented a system to extract vector art from meshed objects~\cite{karsch:npar11}. At the Naval Research Laboratory, I constructed and tested an augmented reality system to improve the well-being of infantry with Mark Livingston and Zhuming Ai~\cite{livingston:ar11}. I have also studied computer vision techniques for medical image segmentation and analysis with Ye Duan~\cite{Karsch:BB:09,Karsch:EMBS:08,he:ijdmb11,he:ijcbb:09,He:SPIE:10}.

\chapter{Physically grounded image editing: Rendering synthetic objects into legacy photographs}
\label{chap:rsolp}
\comment{
\begin{abstract}
We propose a method to realistically insert synthetic objects into existing photographs without requiring access to the scene or any additional scene measurements.  With a single image and a small amount of annotation, our method creates a physical model of the scene that is suitable for realistically rendering synthetic objects with diffuse, specular, and even glowing materials while accounting for lighting interactions between the objects and the scene. We demonstrate in a user study that synthetic images produced by our method are confusable with real scenes, even for people who believe they are good at telling the difference. Further, our study shows that our method is competitive with other insertion methods while requiring less scene information. We also collected new illumination and reflectance datasets; renderings produced by our system compare well to ground truth. Our system has applications in the movie and gaming industry, as well as home decorating and user content creation, among others.
\end{abstract}
}

\begin{figure}[t]
\begin{center}
\includegraphics[width=0.49\linewidth]{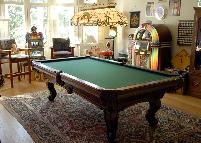}
\includegraphics[width=0.49\linewidth]{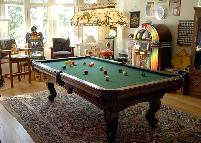}
\\ \vspace{1mm}
\includegraphics[width=0.49\linewidth]{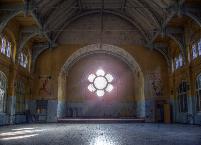}
\includegraphics[width=0.49\linewidth]{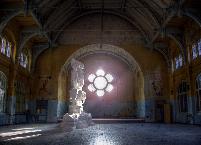}
\end{center}
\caption{With only a small amount of user interaction, our system allows objects to be inserted into legacy images so that perspective, occlusion, and lighting of inserted objects adhere to the physical properties of the scene. Our method works with only a single LDR photograph, and no access to the scene is required.
}
\label{fig:rsolp:teaser}
\end{figure}

\section{Introduction}
Many applications require a user to insert 3D meshed characters, props, or other synthetic objects into images and videos. Currently, to insert objects into the scene, some scene geometry must be manually created, and lighting models may be produced by photographing mirrored light probes placed in the scene, taking multiple photographs of the scene, or even modeling the sources manually. Either way, the process is painstaking and requires expertise.

\begin{figure*}[htpb!]
\centerline{
\includegraphics[width=167mm]{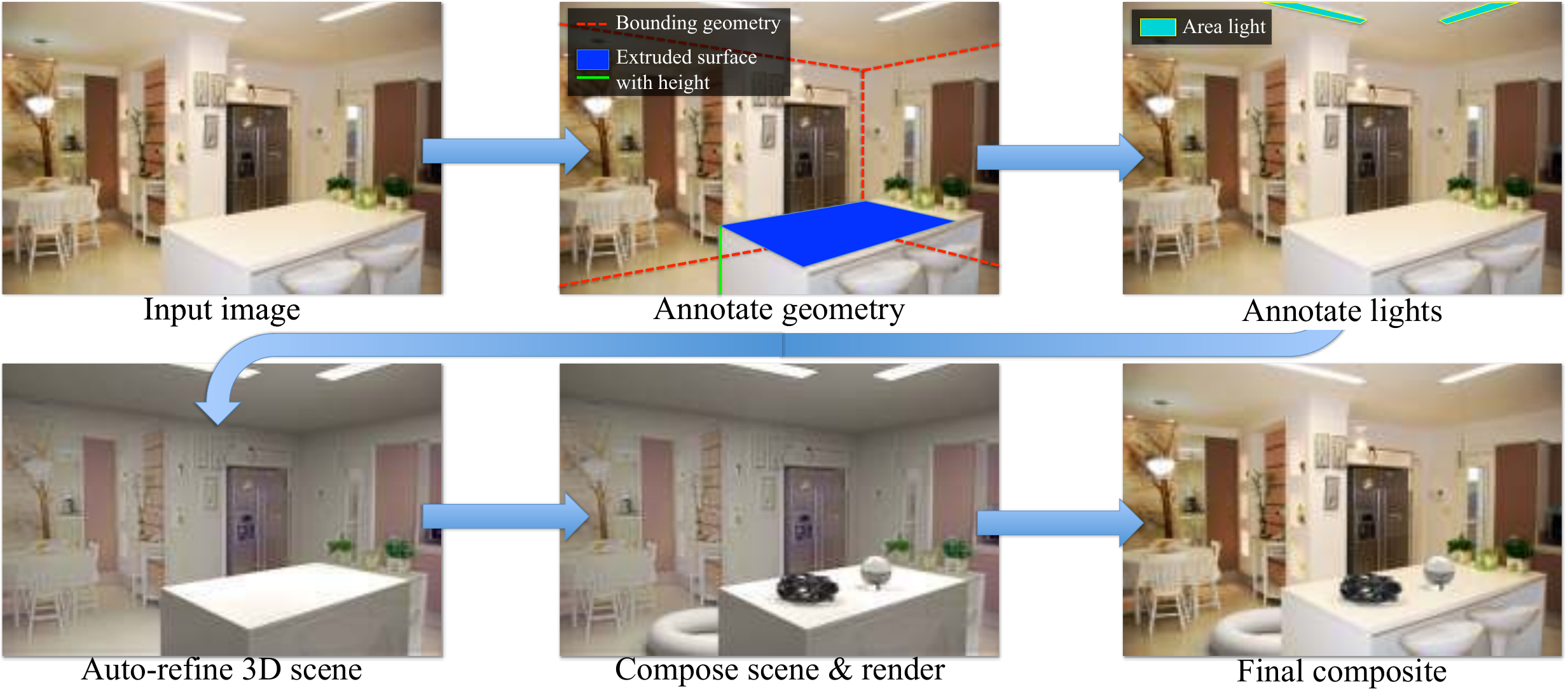}
}
\caption{ Our method for inserting synthetic objects into legacy photographs. From an input image \emph{(top left)}, initial geometry is estimated and a user annotates other necessary geometry \emph{(top middle)} as well as light positions \emph{(top right)}. From this input, our system automatically computes a 3D scene, including a physical light model, surface materials, and camera parameters \emph{(bottom left}). After a user places synthetic objects in the scene \emph{(bottom middle)}, objects are rendered and composited into the original image \emph{(bottom right)}.  Objects appear naturally lit and adhere to the perspective and geometry of the physical scene. From our experience, the markup procedure takes only a minute or two, and the user can begin inserting objects and authoring scenes in a matter of minutes.
}
\label{fig:systemdemo}
\end{figure*}

We propose a method to realistically insert synthetic objects into existing photographs without requiring access to the scene, special equipment, multiple photographs, time lapses, or any other aids.  
Our approach, outlined in Figure~\ref{fig:systemdemo}, is to take advantage of small amounts of annotation to recover a simplistic model of geometry and the position, shape, and intensity of light sources.  First, we automatically estimate a rough geometric model of the scene, and ask the user to specify (through image space annotations) any additional geometry that synthetic objects should interact with. Next, the user annotates light sources and light shafts (strongly directed light) in the image. Our system automatically generates a physical model of the scene using these annotations. The models created by our method are suitable for realistically rendering synthetic objects with diffuse, specular, and even glowing materials while accounting for lighting interactions between the objects and the scene.  

In addition to our overall system, our primary technical contribution is a semiautomatic algorithm for estimating a physical lighting model from a single image. Our method can generate a full lighting model that is demonstrated to be physically meaningful through a ground truth evaluation. We also introduce a novel image decomposition algorithm that uses geometry to improve lightness estimates, and we show in another evaluation to be state-of-the-art for single image reflectance estimation. 
We demonstrate with a user study that the results of our method are confusable with real scenes, even for people who believe they are good at telling the difference. Our study also shows that our method is competitive with other insertion methods while requiring less scene information. 
This method has become possible from advances in recent literature. In the past few years, we have learned a great deal about extracting high level information from indoor scenes~\cite{hedau2009iccv,Lee:CVPR09,Lee:10}, and that detecting shadows in images is relatively straightforward~\cite{guo_cvpr11}. Grosse \ea~\cite{grosse09intrinsic} have also shown that simple lightness assumptions lead to powerful surface estimation algorithms; Retinex remains among the best methods.

\section{Related work}
Debevec's work~\cite{Debevec:98} is most closely related to ours. Debevec shows that a light probe, such as a spherical mirror, can be used to capture a physically accurate radiance map for the position where a synthetic object is to be inserted. This method requires a considerable amount of user input: HDR photographs of the probe, converting these photos into an environment map, and manual modeling of scene geometry and materials. 
More robust methods exist at the cost of more setup time (e.g. the plenopter~\cite{Mury:2009vz}). Unlike these methods and others (e.g.~\cite{fournier1992b,alnasser2006,cossairt2008,Lalonde:sa09}), we require no special equipment, measurements, or multiple photographs. Our method can be used with only a single LDR image, e.g. from Flickr, or even historical photos that cannot be recaptured.  

\boldhead{Image-based Content Creation}  
Like us, Lalonde \ea~\cite{lalonde2007} aim to allow a non-expert user populate an image with objects. Objects are segmented from a large database of images, which they automatically sort to present the user with source images that have similar lighting and geometry. Insertion is simplified by automatic blending and shadow transfer, and the object region is resized as the user moves the cursor across the ground. 
This method is only suitable if an appropriate exemplar image exists, and even in that case, the object cannot participate in the scene's illumination. Similar methods exist for translucent and refractive objects~\cite{Yeung:2011:MCT:1899404.1899406}, but in either case, inserted objects cannot reflect light onto other objects or cast caustics. Furthermore, these methods do not allow for mesh insertion, because scene illumination is not calculated. We avoid these problems by using synthetic objects (3D textured meshes, now plentiful and mostly free on sites like Google 3D Warehouse and turbosquid.com) and physical lighting models. 

\boldhead{Single-view 3D Modeling}
Several user-guided~\cite{Liebowitz99,Criminisi00:SVMetrology,Zhang01,Horry97:TiP,TIP2,Oh01:IBM,sinhaArch} or automatic~\cite{Hoiem05:Popup,Saxena:pami09} methods are able to perform 3D modeling from a single image.  These works are generally interested in constructing 3D geometric models for novel view synthesis.  Instead, we use the geometry to help infer illumination and to handle perspective and occlusion effects.  Thus, we can use simple box-like models of the scene~\cite{hedau2009iccv} with planar billboard models~\cite{TIP2} of occluding objects.  The geometry of background objects can be safely ignored.  Our ability to appropriately resize 3D objects and place them on supporting surfaces, such as table-tops, is based on the single-view metrology work of Criminisi~\cite{Criminisi00:SVMetrology}; also described by Hartley and Zisserman~\cite{hartley2004}.  We recover focal length and automatically estimate three orthogonal vanishing points, using the method from Hedau \ea~\cite{hedau2009iccv}, which is based on Rother's technique~\cite{rother2002}.

\begin{figure}
\begin{center}
\begin{minipage}{0.24\linewidth}
\includegraphics[width=\linewidth]{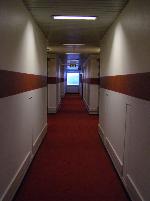}\\
\centerline{\large Input (a)}
\includegraphics[width=\linewidth]{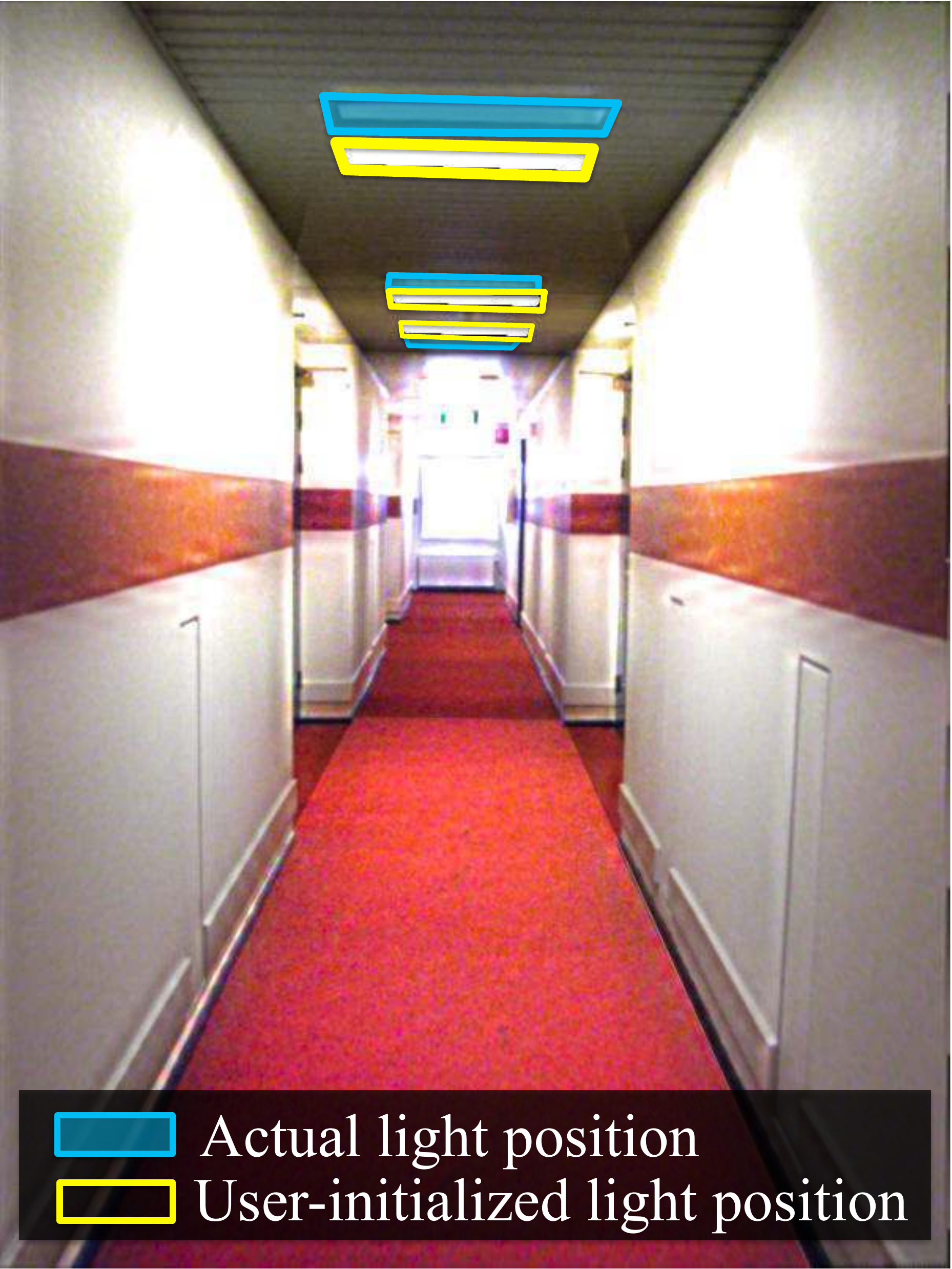}\\
\centerline{\large Initial lights (e)}
\end{minipage}
\begin{minipage}{0.24\linewidth}
\includegraphics[width=\linewidth]{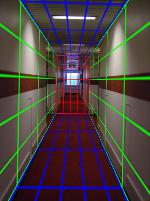}\\
\centerline{\large Geometry (b)}
\includegraphics[width=\linewidth]{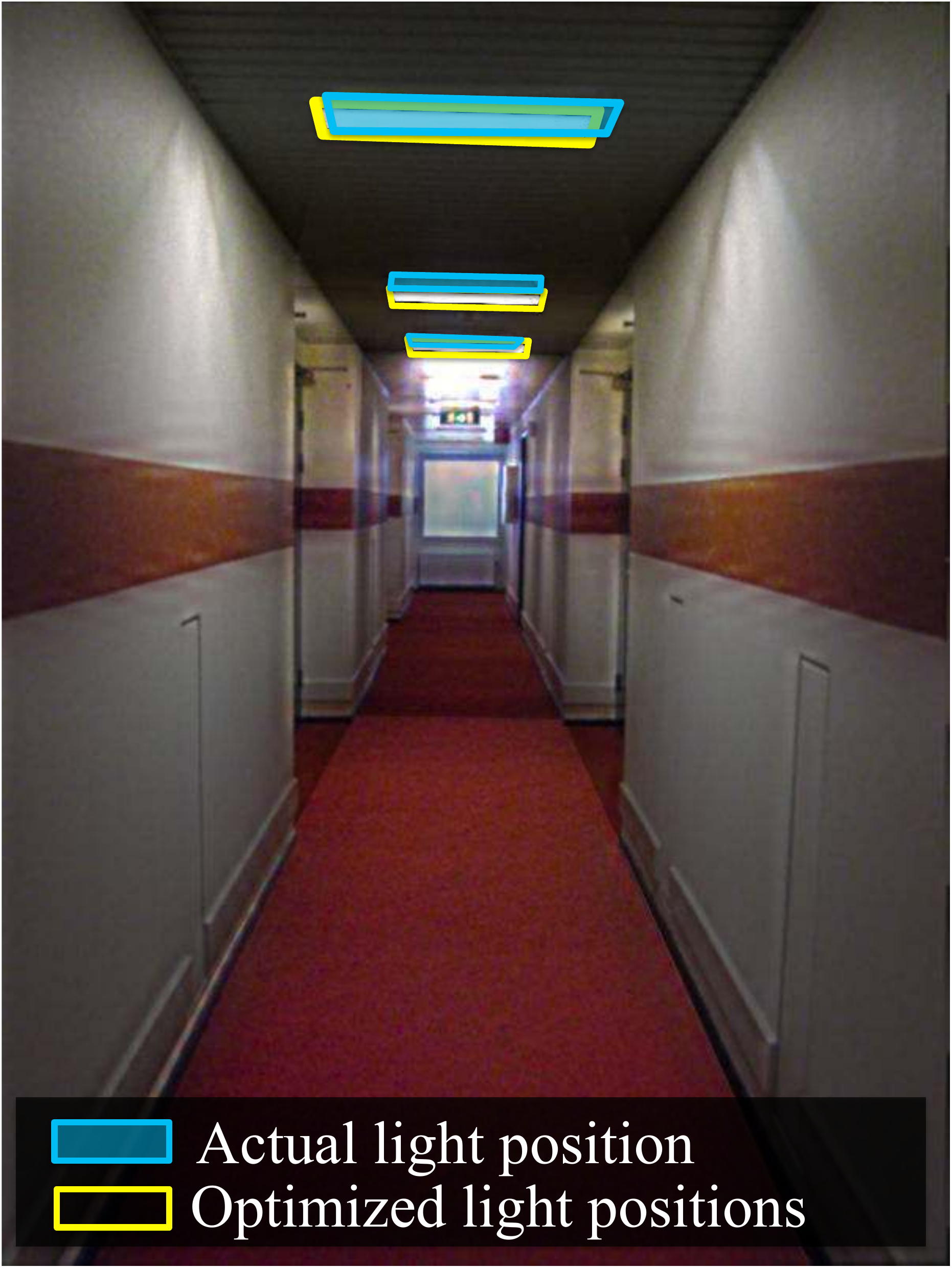}\\
\centerline{\large Refined lights (f)}
\end{minipage}
\begin{minipage}{0.24\linewidth}
\includegraphics[width=\linewidth]{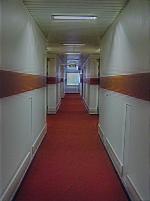}\\
\centerline{\large Albedo (c)}
\includegraphics[width=\linewidth]{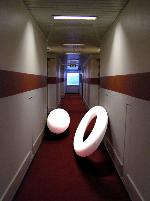}\\
\centerline{\large Initial result (g)}
\end{minipage}
\begin{minipage}{0.24\linewidth}
\includegraphics[width=\linewidth]{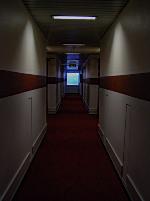}\\
\centerline{\large Direct (d)}
\includegraphics[width=\linewidth]{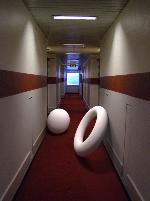}\\
\centerline{\large Refined result (h)}
\end{minipage}
\caption{Overview of our interior lighting algorithm. For an input image \emph{(a)}, we use the modeled geometry (visualization of 3D scene boundaries as colored wireframe mesh, \emph{(b)}) to decompose the image into albedo \emph{(c)} and direct reflected light \emph{(d)}. The user defines initial lighting primitives in the scene \emph{(e)}, and the light parameters are re-estimated \emph{(f)}. The effectiveness of our lighting algorithm is demonstrated by comparing a composited result  \emph{(g)} using the initial light parameters to another composited result \emph{(h)} using the optimized light parameters. Our automatic lighting refinement enhances the realism of inserted objects. Lights are initialized away from the actual sources to demonstrate the effectiveness of our refinement.
}
\label{fig:lightingdemo}
\end{center}
\end{figure}

\boldhead{Materials and Illumination}
We use an automatic decomposition of the image into albedo, direct illumination and indirect illumination terms ({\em intrinsic images}~\cite{BarTen78}). Our geometric estimates are used to improve these terms and material estimates, similar to Boivin and Gagalowicz~\cite{Boivin:2001} and Debevec~\cite{Debevec:98}, but our method improves efficiency of our illumination inference algorithm and is sufficient for realistic insertion (as demonstrated in Sections~\ref{sec:Evaluation} and~\ref{sec:rsolp:study}). 
We must work with a single legacy image, and wish to capture a physical light source estimate so that our method can be used in conjunction with any physical rendering software. Such representations as an irradiance volume do not apply~\cite{irvol}. Yu \ea  show that when a comprehensive model of geometry and luminaires is available, scenes can be relit convincingly~\cite{Yu:1999}.  We differ from them in that our estimate of geometry is coarse, and do not require multiple images.  Illumination in a room is not strongly directed, and cannot be encoded with a small set of point light sources, so the methods of Wang and Samaras~\cite{wangsamaras} and Lopez-Moreno \ea~\cite{LopezMoreno2010698} do not apply. As we show in our user study, point light models fail to achieve the realism that physical models do. We also cannot rely on having a known object present~\cite{satoikeuchi}.  

In the past, we have seen that people are unable to detect perceptual errors in lighting~\cite{LopezMoreno2010698}. Such observations allow for high level image editing using rough estimates (e.g. materials~\cite{Khan:2006:IME:1179352.1141937} and lighting~\cite{kee-farid10a}). Lalonde and Efros~\cite{Lalonde_iccv07} consider the color distribution of images to differentiate real and fake images; our user study provides human assessment on this problem as well. Other methods are focused on the computational detection of forged/synthetic images using cues obtained from inverse rendering~\cite{Kee:2013:EPM:2487228.2487236,Kee:2014:EPM}. While we don't require exact estimates, these techniques use physical properties to determine whether or not an image is forged (e.g. testing if the shading and shadow directions are consistent from the sun).


There are standard computational cues for estimating intrinsic images.  Albedo tends to display sharp, localized changes (which result in large image gradients), while shading tends to change slowly.  These rules-of-thumb inform the Retinex method~\cite{Land71} and important variants~\cite{Horn,Blake,BrelstaffBlake}. Sharp changes of shading do occur at shadow boundaries or normal discontinuities, but cues such as chromaticity~\cite{Funtlightness} or differently lit images~\cite{WeissII} can control these difficulties, as can methods that classify edges into albedo or shading~\cite{Tappenpami,Farenzena}. Tappen et al.~\cite{TappenCVPR06} assemble example patches of intrinsic image, guided by the real image, and exploiting the constraint that patches join up.  Recent work by Grosse \ea demonstrates that the color variant of Retinex is state-of-the-art for single-image decomposition methods~\cite{grosse09intrinsic}.

\section{Modeling}
\label{sec:technique}
To render synthetic objects realistically into a scene, we need estimates of geometry and lighting. At present, there are no methods for obtaining such information accurately and automatically; we incorporate user guidance to synthesize sufficient models. 

Our lighting estimation procedure is the primary technical contribution of our method. With a bit of user markup, we automatically decompose the image with a novel intrinsic image method, refine initial light sources based on this decomposition, and estimate light shafts using a shadow detection method.
Our method can be broken into three phases. The first two phases interactively create models of geometry and lighting respectively, and the final phase renders and composites the synthetic objects into the image. An overview of our method is sketched in Algorithm~\ref{alg:technique}.

\subsection{Estimating geometry and materials}
\label{sec:geometry}
To realistically insert objects into a scene, we only need enough geometry to faithfully model lighting effects. We automatically obtain a coarse geometric representation of the scene using the technique of Hedau \ea\cite{hedau2009iccv}, and estimate vanishing points to recover camera pose automatically. Our interface allows a user to correct errors in these estimates, and also create simple geometry (tables and or near-flat surfaces) through image-space annotations. If necessary, other geometry can be added manually, such as complex objects near inserted synthetic objects. However, we have found that in most cases our simple models suffice in creating realistic results; all results in this chapter require no additional complex geometry. Refer to Section~\ref{sec:details:geometry} for implementation details.


\subsection{Estimating illumination}
\label{sec:lights}
Estimating physical light sources automatically from a single image is an extremely difficult task. Instead, we describe a method to obtain a physical lighting model that, when rendered, closely resembles the original image.
We wish to reproduce two different types of lighting: {\it interior lighting}, emitters present within the scene, and {\it exterior lighting}, shafts of strongly directed light which lie outside of the immediate scene (e.g. sunlight).

\boldhead{Interior lighting}
Our geometry is generally rough and not canonical, and our lighting model should account for this; lights should be modeled such that renderings of the scene look similar to the original image. This step should be transparent to the user. We ask the user to mark intuitively where light sources should be placed, and then refine the sources so that the rendered image best matches the original image. Also, intensity estimation and color cast can be difficult to estimate, and we correct these automatically (see Fig~\ref{fig:lightingdemo}).

{\it Initializing light sources}. 
To begin, the user clicks polygons in the image corresponding to each source. These polygons are projected onto the geometry to define an area light source.  Out-of-view sources are specified with 3D modeling tools.

\floatstyle{boxed}
\newfloat{algorithm}{htp!}{loa}
\floatname{algorithm}{Algorithm}
\begin{algorithm}
\underline{\textsc{legacyInsertion}($img, \textsc{user}$)}
\begin{tabbing} 
\emph{Model}\= \emph{ geometry (Sec~\ref{sec:details:geometry}), auto-estimate materials (Sec~\ref{sec:details:materials})} \+ \\
	$geometry \gets \textsc{detectBoundaries}(img)$ \\
	$geometry \gets \textsc{user}(\text{`Correct boundaries'})$ \\
	$geometry \gets \textsc{user}(\text{`Annotate/add additional geometry'})$ \\
	$geometry_{mat} \gets \textsc{estMaterials}(img, geometry)$ \emph{[Eq~\ref{eq:decomp}]}\- \\
\emph{Refine initial lights and estimate shafts (Sec~\ref{sec:lights})} \+ \\
	$lights \gets \textsc{user}(\text{`Annotate lights/shaft bounding boxes'})$ \\
	$lights \gets \textsc{refineLights}(img, geometry)$ \emph{[Eq~\ref{eq:objfunc}]} \\
	$lights \gets \textsc{detectShafts}(img)$ \- \\
\emph{Insert objects, render and composite (Sec~\ref{sec:details:insertion})} \+ \\
	$scene \gets \textsc{createScene}(geometry, lights)$  \\
	$scene \gets \textsc{user}(\text{'Add synthetic objects'})$ \\
	return \textsc{composite}($img, \textsc{render}(scene))$  \emph{[Eq~\ref{eq:rsolp:composite}]} \-
\end{tabbing}
\caption{Our method for rendering objects into legacy images
}
\label{alg:technique}
\end{algorithm}
\floatstyle{plain}

{\it Improving light parameters}.
Our technique is to choose light parameters to minimize the squared pixel-wise differences between the rendered image (with estimated lighting and geometry) and the target image (e.g. the original image). Denoting $R({\bf L})$ as the rendered image parameterized by the current lighting parameter vector ${\bf L}$, $R^{*}$ as the target image, and ${\bf L}_0$ as the initial lighting parameters, we seek to minimize the objective
\begin{equation}
\label{eq:objfunc}
\begin{array}{c}
\displaystyle \argmin_{\bf L} \sum_{i \in \text{pixels}} \alpha_{i} (R_{i}({\bf L})-R_{i}^{*})^2 + \sum_{j \in \text{params} } w_j ( {\bf L}_j-{\bf L}_{0_j})^2 \vspace{1mm} \\
\hspace{15mm}\text{subject to: } 0 \leq {\bf L}_j \leq 1 \ \forall j
\vspace{-3.5mm}
\end{array}
\vspace{1mm}
\end{equation}

where $w$ is a weight vector that constrains lighting parameters near their initial values, and $\alpha$ is a per-pixel weighting that places less emphasis on pixels near the ground. Our geometry estimates will generally be worse near the bottom of the scene since we may not have geometry for objects near the floor.  In practice, we set $\alpha=1$ for all pixels above the spatial midpoint of the scene (height-wise), and $\alpha$ decreases quadratically from 1 to 0 at floor pixels. Also, in our implementation, ${\bf L}$ contains 6 scalars per light source: RGB intensity, and 3D position. More parameters could also be optimized. For all results, we normalize each light parameter to the range $[0,1]$, and set the corresponding values of $w$ to 10 for spatial parameters and 1 for intensity parameters. A user can also modify these weights depending on the confidence of their manual source estimates.
To render the synthetic scene and determine $R$, we must first estimate materials for all geometry in the scene.  We use our own intrinsic image decomposition algorithm to estimate surface reflectance (albedo), and the albedo is then projected onto the scene geometry as a diffuse texture map, as described in Section~\ref{sec:details:materials}.

\begin{figure}[htp]
\begin{center}
\includegraphics[width=.8\linewidth]{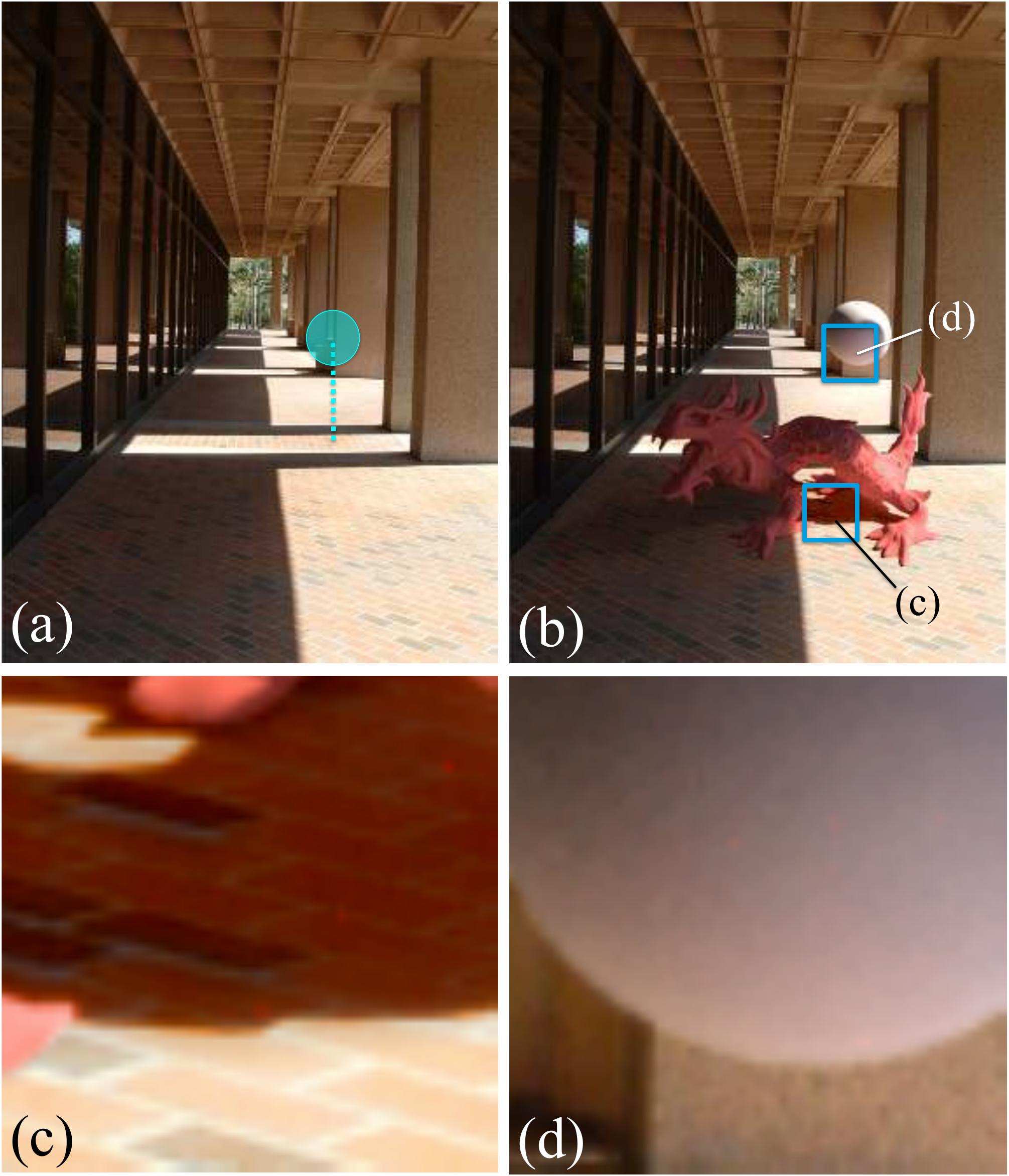}
\end{center}
\caption{Inserted objects fully participate with the scene lighting as if they were naturally a part of the image. Here, an input image \emph{(a)} is augmented with inserted objects and illuminated with a bright shaft of light \emph{(b)}. Interreflected red light from the dragon onto the brick floor is evident in \emph{(c)}, and the underside of the inserted sphere has a slight red tint from light reflecting off of the brick \emph{(d)}. A registration probe in \emph{(a)} displays the scale and location of the sphere in \emph{(b)}. Best viewed on a high resolution, high contrast display.
}
\label{fig:interreflections}
\end{figure}

{\it Intrinsic decomposition}.
Our decomposition method exploits our geometry estimates. First, indirect irradiance is computed by {\it gathering} radiance values at each 3D patch of geometry that a pixel projects onto. The gathered radiance values are obtained by sampling observed pixel values from the original image, which are projected onto geometry along the camera's viewpoint. We denote this indirect irradiance image as $\Gamma$; this term is equivalent to the integral in the radiosity equation. Given the typical Lambertian assumptions, we assume that the original image $B$ can be expressed as the product of albedo $\rho$ and shading $S$ as well as the sum of reflected direct light $D$ and reflected indirect light $I$. Furthermore, reflected gathered irradiance is equivalent to reflected indirect lighting under these assumptions. This leads to the equations
\begin{equation}
B=\rho S,\ \ \ B=D+I,\ \ \ I=\rho\Gamma,\ \ \ B = D + \rho \Gamma.
\label{eq:intr_assumptions}
\end{equation}
We use the last equation as constraints in our optimization below.

We have developed an objective function to decompose an image $B$ into albedo $\rho$ and direct light $D$ by solving
\begin{equation}
\label{eq:decomp}
\begin{split}
\argmin_{\rho, D} &\sum_{i \in \text{pixels}} |\Delta \rho|_i + \gamma_1 m_i (\nabla \rho)_i^2 + \gamma_2 (D_i-D_{0_i})^2 + \gamma_3 (\nabla D)_i^2 \\
&\text{subject to } \ \ \ B = D + \rho \Gamma, \ \ \ 0 \leq \rho \leq 1, \ \ \ 0 \leq D,
\end{split}
\vspace{-3mm}
\end{equation}
where $\gamma_1,\gamma_2,\gamma_3$ are weights, $m$ is a scalar mask taking large values where $B$ has small gradients, and small values otherwise, and $D_0$ is the initial direct lighting estimate. We define $m$ as a sigmoid applied to the gradient magnitude of $B$:  $m_i = 1-1/(1+e^{-s(||\nabla B||_i^2-c)})$, setting $s = 10.0$, $c=0.15$ in our implementation.

\begin{figure*}[htp]
\centerline{
\includegraphics[width=0.4\linewidth,clip=true,trim=0 0 0 10]{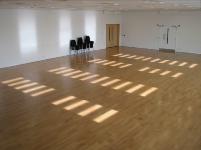}
\includegraphics[width=0.4\linewidth]{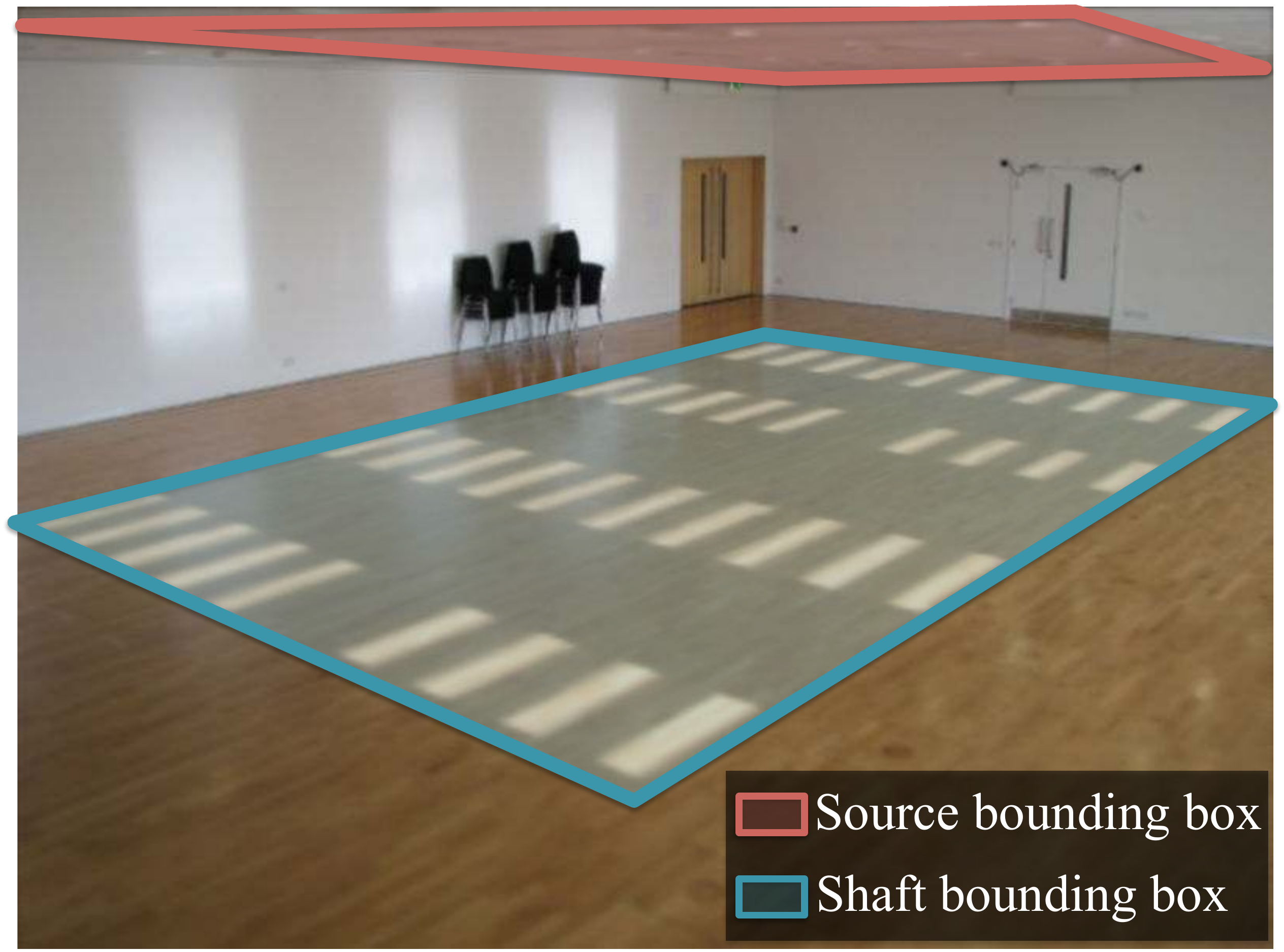}
}
\centerline{\Large (a) \hspace{60mm} (b)}
\centerline{
\includegraphics[width=0.4\linewidth]{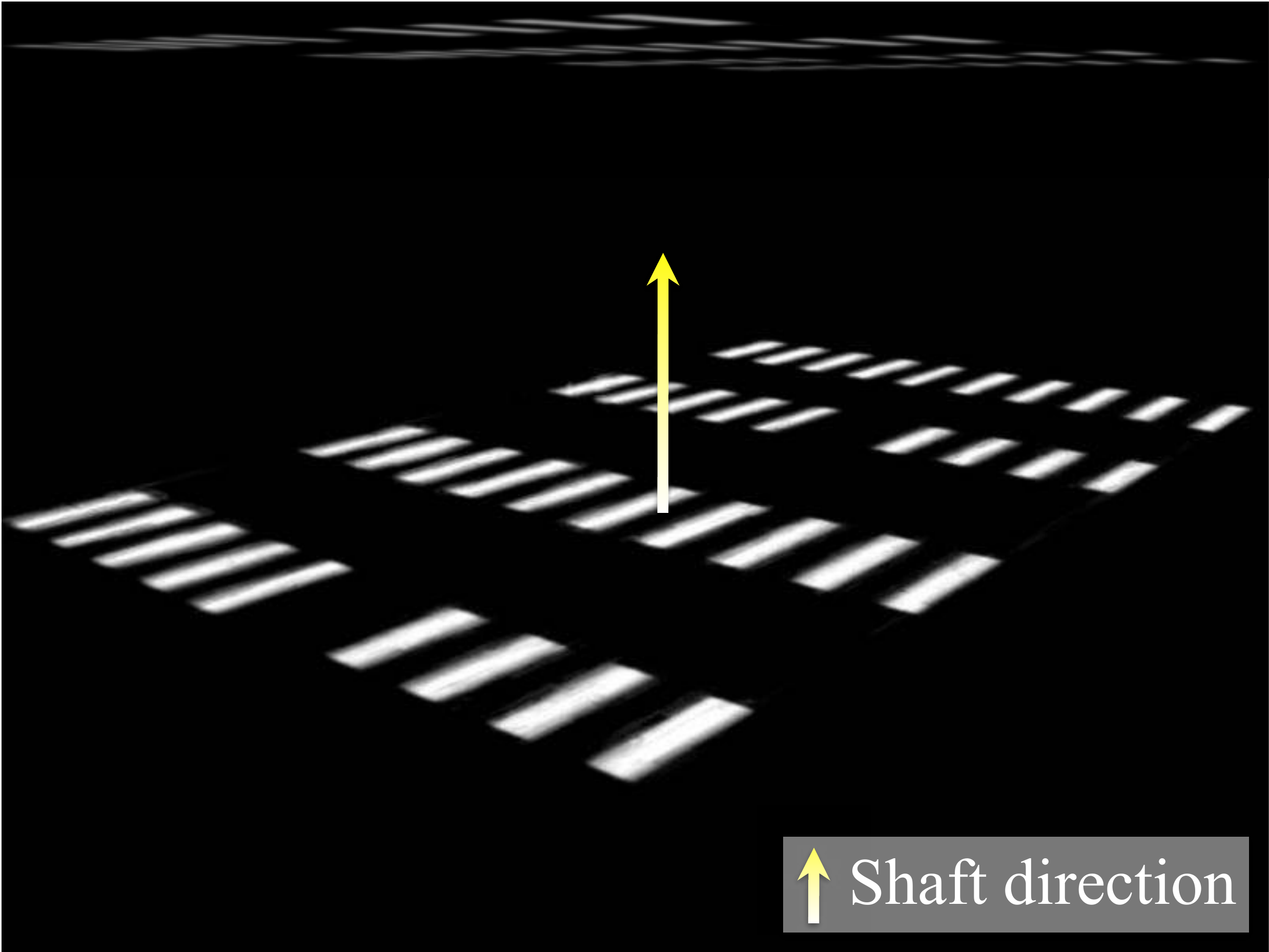}
\includegraphics[width=0.4\linewidth]{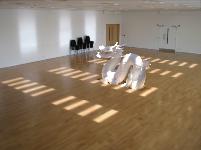}
}
\centerline{\Large (c) \hspace{60mm} (d)}
\caption{Our algorithm for estimating exterior lighting (light shafts). Given an input image \emph{(a)}, the user specifies bounding boxes around the shafts and their sources \emph{(b)}. The shafts are detected automatically, and the shaft direction is estimated using the centroid of the the bounding boxes in 3D \emph{(c)}. A physical lighting model (e.g. a masked, infinitely far spotlight) is created from this information, and objects can be rendered inserted realistically into the scene \emph{(d)}.
}
\label{fig:shafts}
\end{figure*}

Our objective function is grounded in widespread intrinsic image assumptions \cite{Land71,Blake,BrelstaffBlake}, namely that shading is spatially slow and albedo consists of piecewise constant patches with potentially sharp boundaries. The first two terms in the objective coerce $\rho$ to be piecewise constant. The first term enforces an L1 sparsity penalty on edges in $\rho$, and the second term smoothes albedo only where $B$'s gradients are small. The final two terms smooth $D$ while ensuring it stays near the initial estimate $D_0$. We set the objective weights to $\gamma_1 = 0.2$, $\gamma_2 = 0.9$, and $\gamma_3 = 0.1$. We initialize $\rho$ using the color variant of Retinex as described by Grosse \ea\cite{grosse09intrinsic}, and initialize $D$ as $D_0 = B-\rho\Gamma$ (by Eq.~\ref{eq:intr_assumptions}). This optimization problem can be solved in a variety of ways; we use an interior point method (implemented with MATLAB's optimization toolbox). 
In our implementation, to improve performance of our lighting optimization (Eq.~\ref{eq:objfunc}), we set the target image as our estimate of the direct term, and render our scene only with direct lighting (which greatly reduces the time in recalculating the rendered image). We choose our method as it utilizes the estimated scene geometry to obtain better albedo estimates, and reduces the computation cost of solving Eq.~\ref{eq:objfunc}, but any decomposition method could be used (e.g. Retinex).

\begin{figure}[tp]
\begin{center}
\includegraphics[width=0.49\linewidth]{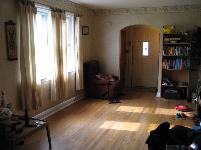}
\includegraphics[width=0.49\linewidth]{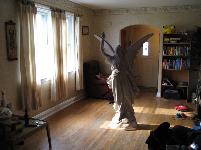}
\end{center}
\caption{A difficult image for detecting light shafts. Many pixels near the window are saturated, and some shaft patterns on the floor are occluded, as in the image on the left. However, an average of the matte produced for the floor and wall provides an acceptable estimate (used to relight the statue on the right).
}
\label{fig:shaftissues}
\end{figure}

\boldhead{Exterior lighting (light shafts)}
Light shafts are usually produced by the sun, or some other extremely far away source. Thus, the type of light we wish to model can be thought of as purely directional, and each shaft in a scene will have the same direction.

We define a light shaft with a 2D polygonal projection of the shaft  and a direction vector. In Figure~\ref{fig:shafts}, the left image shows a scene with many light shafts penetrating the ceiling and projecting onto the floor. Our idea is to detect either the source or the projections of shafts in an image and recover the shaft direction. The user first draws a bounding box encompassing shafts visible in the scene, as well as a bounding box containing shaft sources (windows, etc.).  We then use the shadow detection algorithm of Guo \ea~\cite{guo_cvpr11} to determine a scalar mask that estimates the confidence that a pixel is {\it not} illuminated by a shaft. This method models region based appearance features along with pairwise relations between regions that have similar surface material and illumination.  A graph cut inference is then performed to identify the regions that have same material and different illumination conditions, resulting in the confidence mask. The detected shadow mask is then used to recover a soft shadow matte using the spectral matting method of Levin \ea~\cite{Levin07spectralmatting}. We then use our estimate of scene geometry to recover the direction of the shafts (the direction defined by the two midpoints of the two bounding boxes). However, it may be the case that either the shaft source or the shaft projection is not visible in an image. In this case, we ask the user to provide an estimate of the direction, and automatically project the source/shaft accordingly. Figure~\ref{fig:shafts} shows an example of our shaft procedure where the direction vector is calculated automatically from the marked bounding boxes. Shafts are represented as masked spotlights for rendering.

\begin{figure}[htpb!]
\centerline{
\includegraphics[width=0.49\columnwidth]{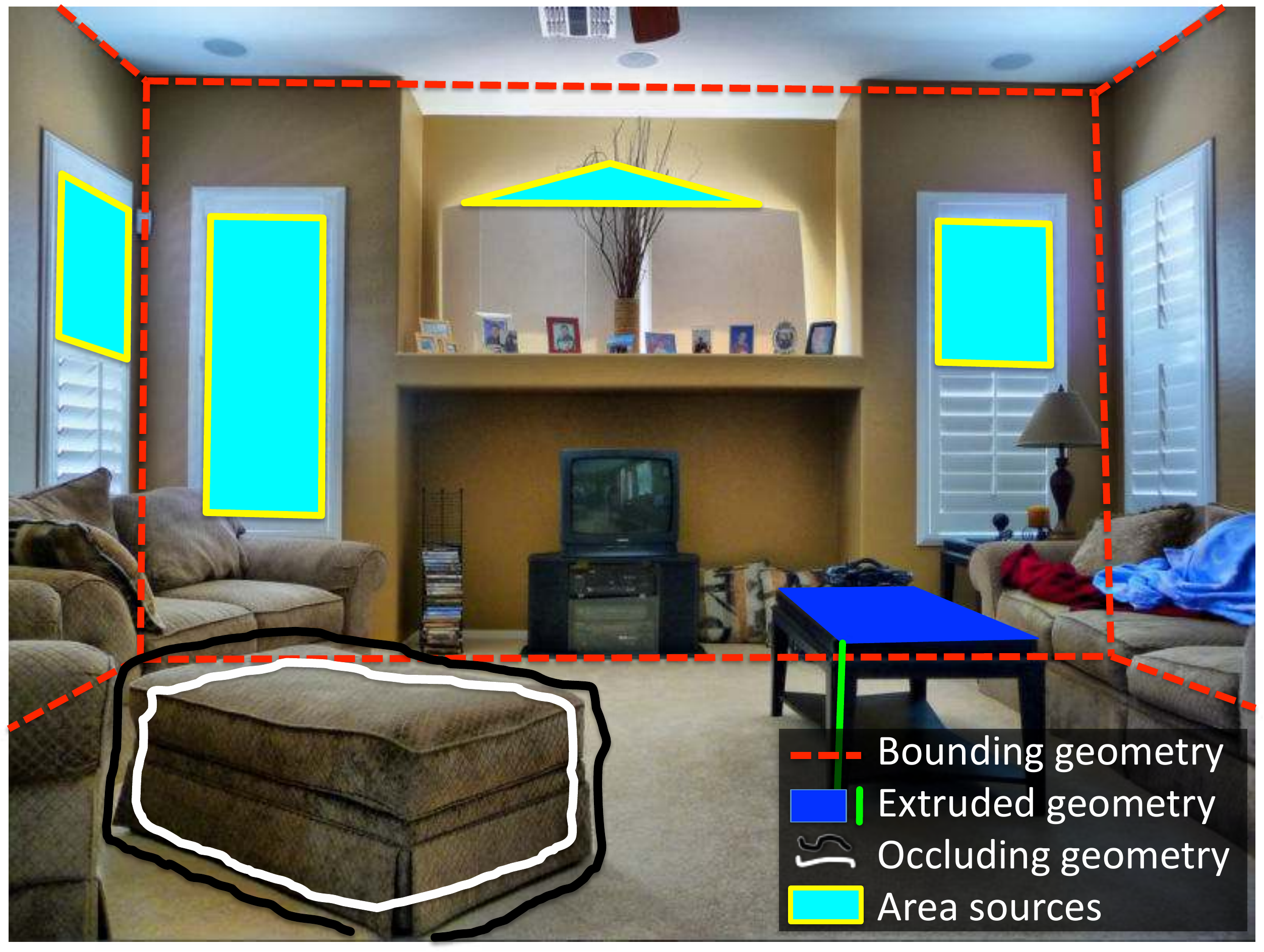}
\includegraphics[width=0.49\columnwidth]{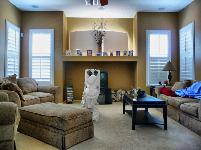}
}
\caption{Our system is intuitive and quick. This result was modeled by a user unfamiliar with our interface (after a short demonstration). From start to finish, this result was created in under 10 minutes (render time not included). User's markup shown on left.
}
\label{fig:noviceresult}
\end{figure}

In some cases, it is  difficult to recover accurate shadow mattes for a window on a wall or a shaft on the floor individually. For instance, it is difficult to detect the window in Figure~\ref{fig:shaftissues}  using only the cues from the wall.  In such cases, we project the recovered mask on the floor along the shaft direction to get the mapping on the wall and average matting results for the wall and floor to improve the results. Similarly, an accurate matte of a window can be used to improve the matte of a shaft  on the floor (as in the right image of Figure~\ref{fig:rsolp:teaser}).

\subsection{Inserting synthetic objects}
\label{sec:details:insertion}
With the lighting and geometry modeled, a user is now free to insert synthetic 3D geometry 
into the scene. Once objects have been inserted, the scene can be rendered with any suitable rendering software.\footnote{For our results, we use LuxRender (http://www.luxrender.net)}~Rendering is trivial, as all of the information required by the renderer has been estimated (lights, geometry, materials, etc).

To complete the insertion process, we composite the rendered objects back into the original photograph using the additive differential rendering method~\cite{Debevec:98}. This method renders two images: one containing synthetic objects $\mathcal{I}_{obj}$, and one without synthetic objects $\mathcal{I}_{noobj}$, as well as an object mask $M$ (scalar image that is 0 everywhere where no object is present, and $(0,1]$ otherwise). The final composite image $\mathcal{I}_{final}$ is obtained by 
\begin{equation}
\label{eq:rsolp:composite}
\mathcal{I}_{final} = M \odot \mathcal{I}_{obj} + (1-M) \odot (\mathcal{I}_b + \mathcal{I}_{obj}-\mathcal{I}_{noobj})
\end{equation}
where $\mathcal{I}_b$ is the input image, and $\odot$ is the Hadamard product.

\begin{figure}[htp]
\begin{center}
\includegraphics[width=0.49\columnwidth]{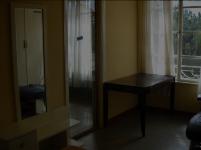}
\includegraphics[width=0.49\columnwidth]{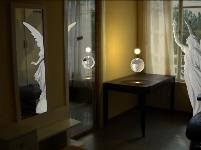}
\end{center}
\caption{Our method allows for light source insertion and easy material reassignment. Here, a glowing ball is inserted above a synthetic glass sphere, casting a caustic on the table. The mirror has been marked as reflective, allowing synthetic objects to realistically interact with the scene.
}
\label{fig:mirror}
\end{figure}

\section{Implementation details}
\label{sec:details}
\subsection{Modeling geometry}
\label{sec:details:geometry}
Rough scene boundaries ({\it bounding geometry}) are estimated first along with the camera pose, and we provide tools for correcting and supplementing these estimates. Our method also assigns materials to this geometry automatically based on our intrinsic decomposition algorithm (Sec.~\ref{sec:lights}).




\boldhead{Bounding geometry}
We model the bounding geometry as a 3D cuboid; essentially the scene is modeled as a box that circumscribes the camera so that up to five faces are visible. Using the technique of Hedau \ea~\cite{hedau2009iccv}, we automatically generate an estimate of this box layout for an input image, including camera pose. This method estimates three vanishing points for the scene (which parameterize the box's rotation), as well as a 3D translation to align the box faces with planar faces of the scene (walls, ceiling floor). However, the geometric estimate may be inaccurate, and in that case, we ask the user to manually correct the layout using a simple interface we have developed. The user drags the incorrect vertices of the box to corresponding scene corners, and manipulates vanishing points using a pair of line segments (as in the Google Sketchup\footnote{http://sketchup.google.com} interface) to fully specify the 3D box geometry.

\boldhead{Additional geometry}
We allow the user to easily model {\it extruded geometry}, i.e. geometry defined by a closed 2D curve that is extruded along some 3D vector, such as tables, stairs, and other axis-aligned surfaces. In our interface, a user sketches a 2D curve defining the surface boundary, then clicks a point in the footprint of the object which specifies the 3D height of the object~\cite{Criminisi00:SVMetrology}. Previously specified vanishing points and bounding geometry allow for these annotations to be automatically converted to a 3D model.

In our interface, users can also specify {\it occluding surfaces}, complex surfaces which will occlude inserted synthetic objects (if the inserted object is behind the occluding surface). We allow the user to create occlusion boundaries for objects using the interactive spectral matting segmentation approach~\cite{Levin07spectralmatting}. The user defines the interior and exterior of an object by scribbling, and a segmentation matte for the object is computed. These segmentations act as cardboard cutouts in the scene; if an inserted object intersects the segmentation and it is farther from the camera, then it will be occluded by the cutout.  We obtain the depth of an object by assuming the lowermost point on its boundary to be its contact point with the floor. Figures~\ref{fig:systemdemo} and~\ref{fig:noviceresult} show examples of both extruded and occluding geometry.

\begin{figure}[htp]
\begin{minipage}[b]{.65\columnwidth}
\includegraphics[width=.49\columnwidth]{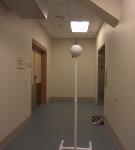}
\includegraphics[width=.49\columnwidth]{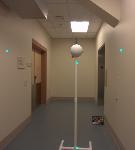}
\caption{The instrument used for collecting ground truth illumination data. The left image shows the apparatus (a white, diffuse ball resting on a plastic, height-adjustable pole). Using knowledge of the physical scene, we can align a rendered sphere over the probe for error measurements \emph{(right}). 
}
\label{fig:rsolp:rig}
\end{minipage}
\hfill
\begin{minipage}[b]{.29\columnwidth}
\includegraphics[width=\columnwidth]{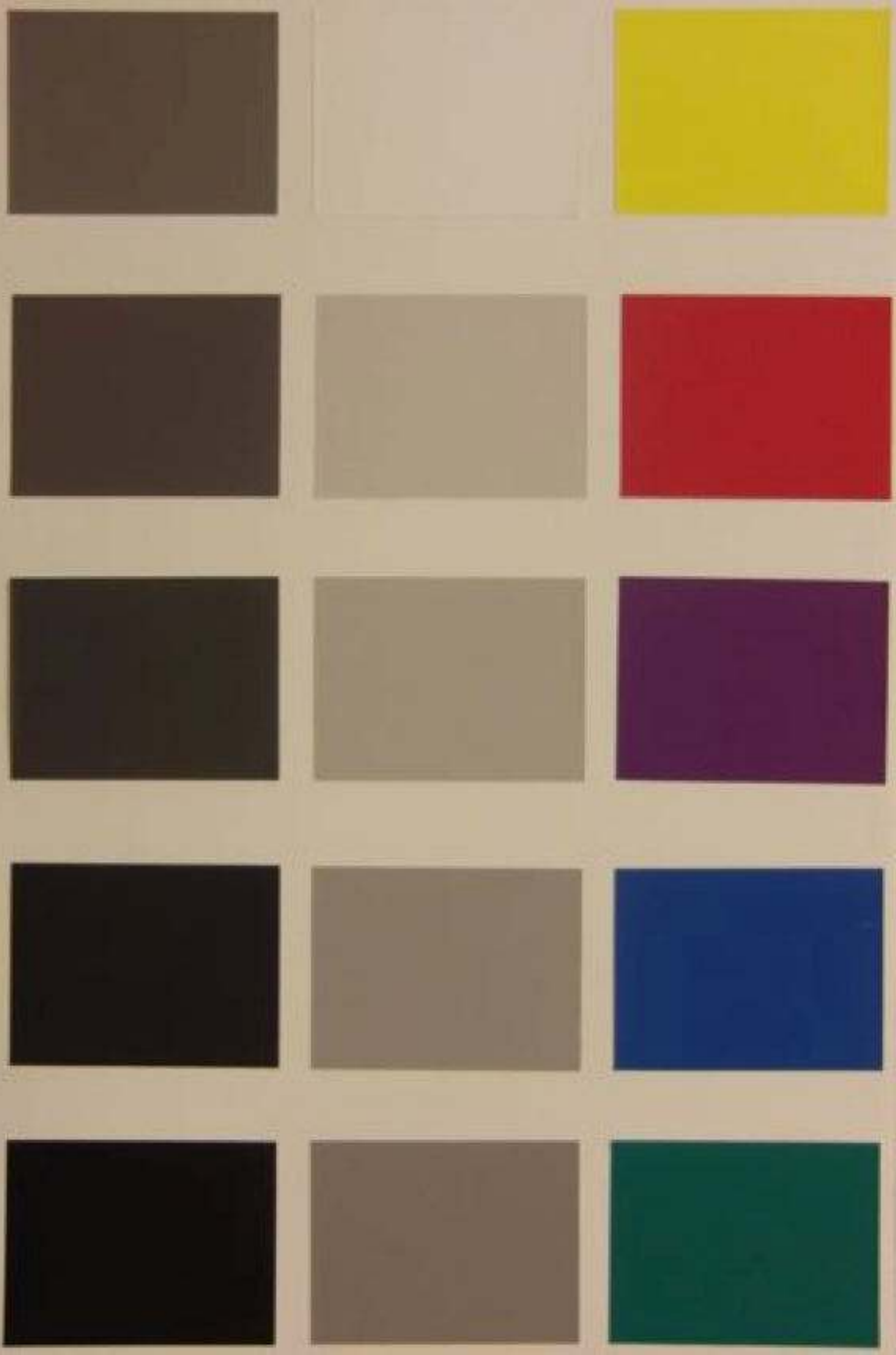}
\caption{The chart used in our ground truth reflectance experiments (Sec.~\ref{sec:eval_intrinsic}).}
\label{fig:colorchecker}
\end{minipage}
\end{figure}

\subsection{Modeling materials}
\label{sec:details:materials}
We assign a material to all estimated geometry based on the albedo estimated during intrinsic image decomposition (Sec~\ref{sec:lights}). We project the estimated albedo along the camera's view vector onto the estimated geometry, and render the objects with a diffuse texture corresponding to projected albedo. This projection applies also to out-of-view geometry (such as the wall behind the camera, or any other hidden geometry). Although unrealistic, this scheme has proven effective for rendering non-diffuse objects (it is generally difficult to tell that out-of-view materials are incorrect; see Fig~\ref{fig:results2}).


\section{Ground truth evaluations}
\label{sec:Evaluation}
Here, we evaluate the physical accuracy of lighting estimates produced by our method as well as our intrinsic decomposition algorithm. We do not strive for physical accuracy (rather, human believability), but we feel that these studies may shed light on how physical accuracy corresponds to people's perception of a real (or synthetic) image. Our studies show that our lighting models are quite accurate, but as we show later in our user study, people are not very good at detecting physical inaccuracies in lighting. Our reflectance estimates are also shown to be more accurate than the color variant of Retinex, which is currently one of the best single-image diffuse reflectance estimators.

\begin{figure}
\begin{center}
\includegraphics[width=0.7\linewidth]{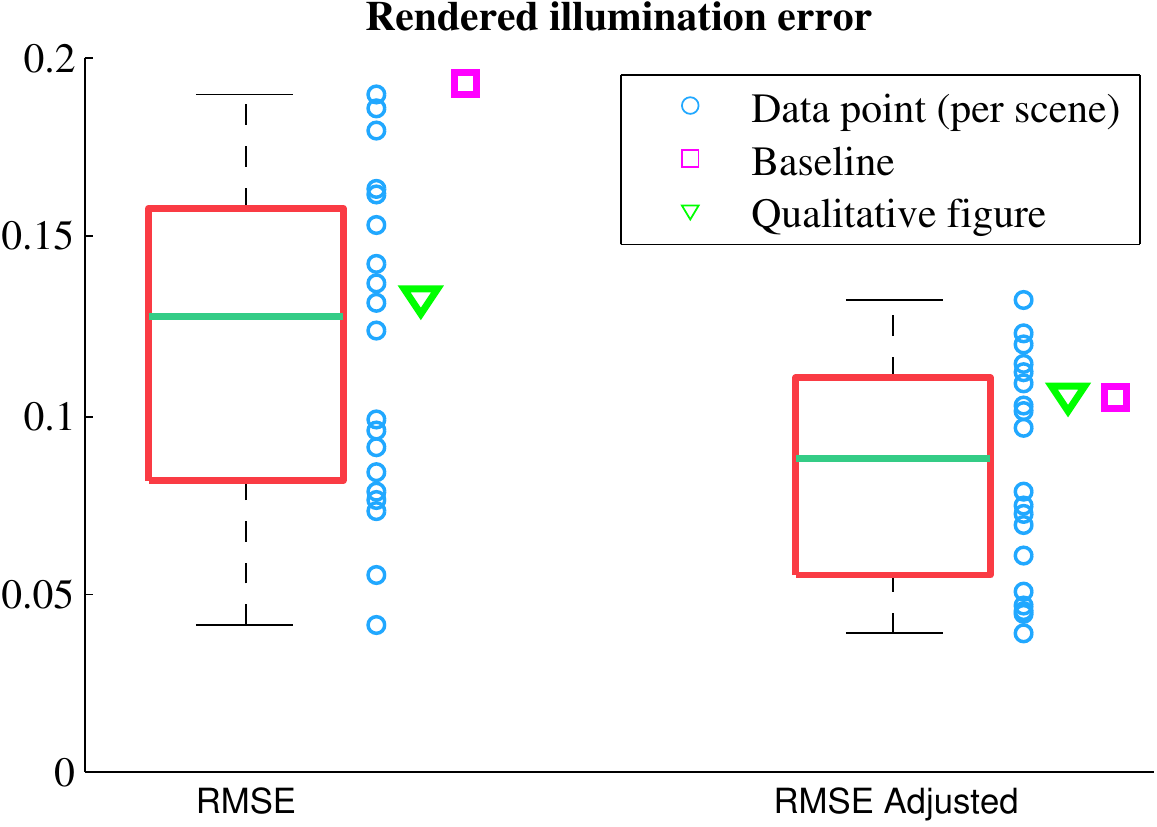}
\end{center}
\caption{We report results for both the root mean squared error (RMSE) and the RMSE after subtracting the mean intensity per sphere (RMSE adjusted). The RMSE metric illustrates how our method compares to the ground truth in an absolute metric, and the RMSE adjusted metric gives a sense of how accurate the lighting pattern is on each of the spheres (indicating whether light size/direction is correct).  For each metric, we show a box plot where the green horizontal line is the median, and the red box extends to the 25th and 75th percentiles. The averaged RMSE per scene (10 spheres are in each scene) is shown as a blue circle. A baseline (purple square) was computed by rendering all spheres with uniform intensity, and set to be the mean intensity of all images in the dataset. The green triangle indicates the error for the qualitative illustration in Fig~\ref{fig:errorvis}. No outliers exist for either metric, and image intensities range from [0,1]. 
}
\label{fig:errorgraph}
\end{figure}

\subsection{Lighting evaluation}
\label{sec:eval_lighting}
We have collected a ground truth dataset in which the surface BRDF is known for an object (a white, diffuse ball) in each image.  Using our algorithm, we estimate the lighting for each scene and insert a synthetic sphere. Because we know the rough geometry of the scene, we can place the synthetic sphere at the same spatial location as the sphere in the ground truth image. 

\boldhead{Dataset} Our dataset contains 200 images from 20 indoor scenes illuminated under varying lighting conditions. We use an inflatable ball painted with flat white paint as the object with known BRDF, which was matched and verified using a Macbeth Color Checker. The ball is suspended by a pole that protrudes from the ground and can be positioned at varying heights (see Fig~\ref{fig:rsolp:rig}).  The images were taken with a Casio EXILIM EX-FH100 using a linear camera response function ($\gamma = 1$).

\begin{figure}
\centerline{\includegraphics[width=0.5\linewidth]{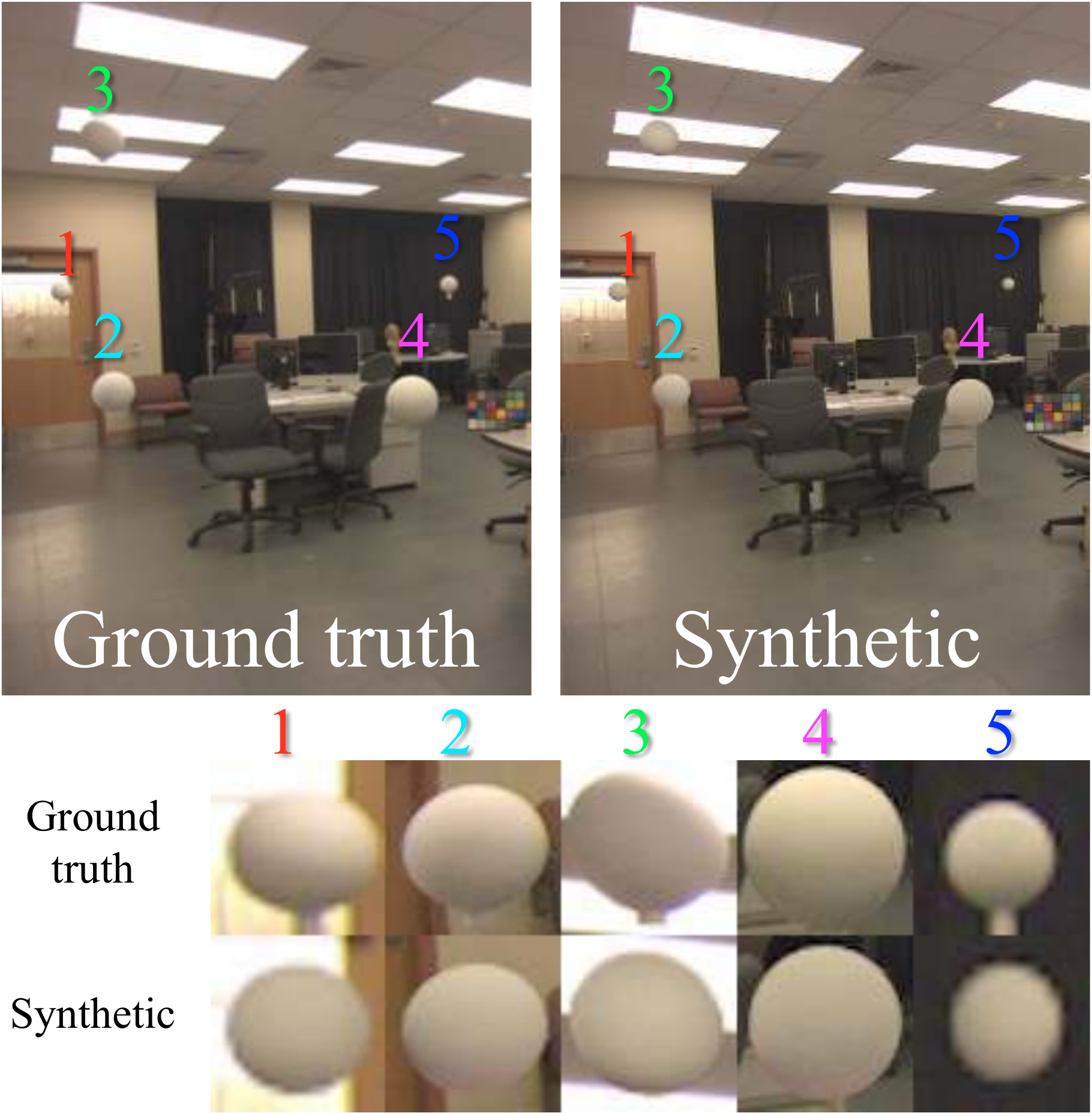} }
\caption{Qualitative comparison of our lighting algorithm to ground truth lighting. The top left image shows a scene containing five real spheres with authentic lighting (poles edited out for visual compactness). We estimate illumination using our algorithm and render the spheres into the scene at the same spatial locations \emph{(top right)}.  The bottom image matrix shows close-up views of the ground truth and rendered spheres.
See Fig~\ref{fig:errorgraph} for quantitative results. 
}
\label{fig:errorvis}
\end{figure}

\boldhead{Results} For a pair of corresponding ground truth and rendered images, we measure the error by computing the pixel-wise difference of all pixels that have known BRDF. We measure this error for each image in the dataset, and report the root mean squared error (RMSE).  Overall, we found the RMSE to be $0.12 \pm 0.049$ for images with an intensity range of $[0,1]$. For comparing lighting patterns on the spheres, we also computed the error after subtracting the mean intensity (per sphere) from each sphere. We found that this error to be $0.085\pm 0.03$. Figure~\ref{fig:errorgraph} shows the RMSE for the entire dataset, as well as the RMSE after subtracting the mean intensity (RMSE adjusted), and a baseline for each metric (comparing against a set of uniformly lit spheres with intensity set as the mean of all dataset images). Our method beats the baseline for every example in the RMSE metric, suggesting decent absolute intensity estimates, and about 70\% of our renders beat the adjusted RMSE baseline. A qualitative visualization for five spheres in one scene from the dataset is also displayed in Figure~\ref{fig:errorvis}. In general, baseline renders are not visually pleasing but still do not have tremendous error, suggesting qualitative comparisons may be more useful when evaluating lightness estimation schemes. 

\subsection{Intrinsic decomposition evaluation}
\label{sec:eval_intrinsic}
We also collected a ground truth reflectance dataset to compare to the reflectance estimates obtained from our intrinsic decomposition algorithm. We place a chart with known diffuse reflectances (ranging from dark to bright) in each scene, and measure the error in reflectance obtained by our method as well as Retinex. We show that our method achieves more accurate absolute reflectance than Retinex in nearly every scene in the dataset.

\boldhead{Dataset} Our reflectance dataset contains 80 images from different indoor scenes containing our ground truth reflectance chart (shown in Fig~\ref{fig:colorchecker}). We created the chart using 15 Color-aid papers; 10 of which are monochrome patches varying between 3\% reflectance (very dark) and 89\% reflectance (very bright). Reflectances were provided by the manufacturer. Each image in the dataset was captured by with the same camera and response as in Sec.~\ref{sec:eval_lighting}.

\begin{figure}
\begin{center}
\includegraphics[width=0.7\linewidth]{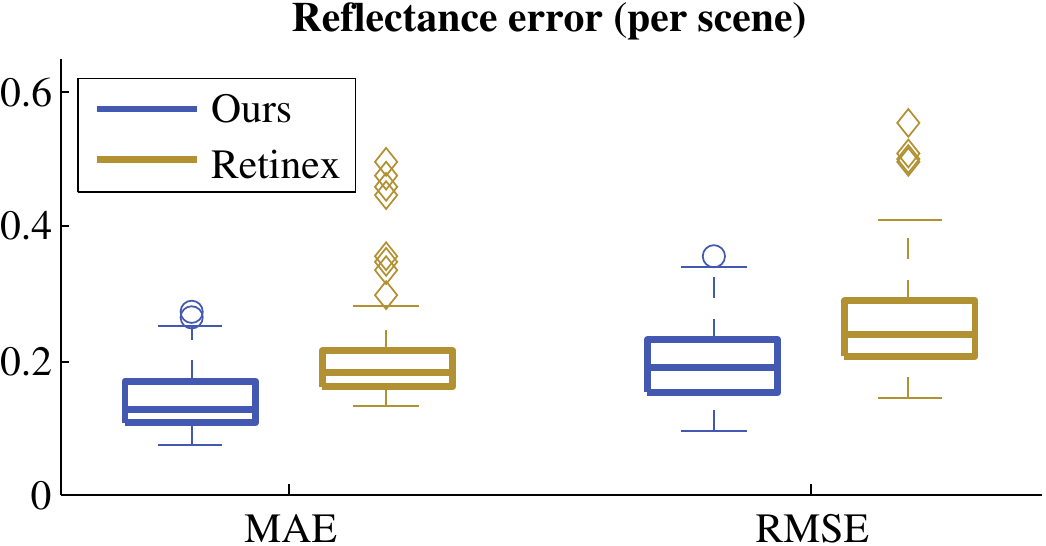} 
\end{center}
\setlength{\unitlength}{1mm}
\begin{picture}(0,0)
\color{dblue} 
\put(14.5, 35.65){\circle{2}}
\color{dyellow}
\put(13.5, 31.1){$\Diamond$}
\end{picture}
\vspace{-4mm}
\caption{Summary of the reflectance evaluation. Errors are measured per scene using a ground truth reflectance chart and reported in MAE and RMSE.  For each method and metric, a box plot is shown where the center horizontal line indicates the median, and the box extends to the 25th and 75th percentiles. Results from our decomposition method are displayed in blue (outliers as circles); Retinex results are displayed in gold (outliers as diamonds).
}
\label{fig:intrinsiccomp}
\end{figure}
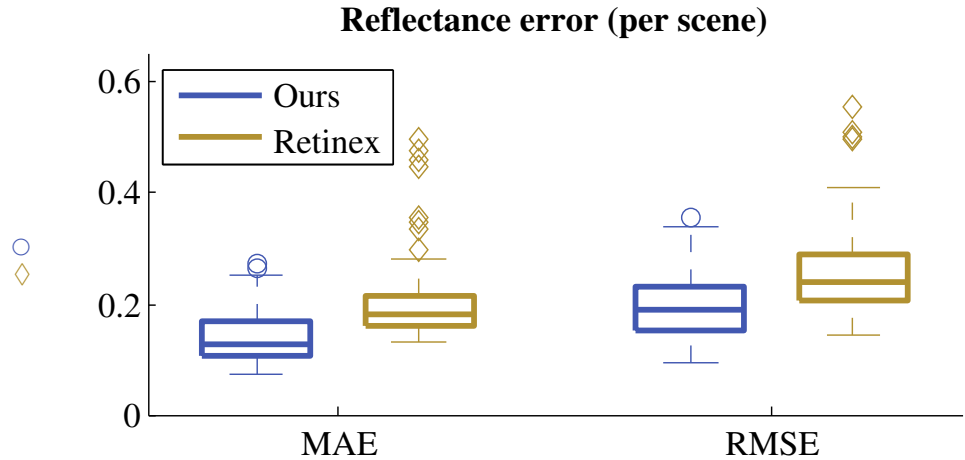

\boldhead{Results} Using our decomposition method described in Sec.~\ref{sec:lights}, we estimate the per-pixel reflectance of each scene in our dataset. We then compute the mean absolute error (MAE) and root mean squared error (RMSE) for each image over all pixels with known reflectance (i.e. only for the pixels inside monochromatic patches). For further comparison, we compute the same error measures using the color variant of Retinex (as described in Grosse \ea~\cite{grosse09intrinsic}) as another method for estimating reflectance. Figure~\ref{fig:intrinsiccomp} summarizes these results. Our decomposition method outperforms Retinex for almost a large majority of the scenes in the dataset, and when averaged over the entire dataset, our method produced an MAE and RMSE of .141 and .207 respectively, compared to Retinex's MAE of .205 and RMSE of .272. These results indicate that much improvement can be made to absolute reflectance estimates when the user supplies a small amount of rough geometry, and that our method may improve other user-aided decomposition techniques, such as the method of Carroll~\ea\cite{Carroll:sg11}.

\subsection{Physical accuracy of intermediate results}
From these studies, we conclude that our method achieves comparatively accurate illumination and reflection estimates. However, it is important to note that these estimates are heavily influenced by the rough estimates of scene geometry, and optimized to produce a perceptually plausible rendered image (with our method) rather than to achieve physical accuracy. Our method adjusts light positions so that the rendered scenes look most like the original image, and our reflectance estimates are guided by rough scene geometry. Thus, the physical accuracy of the light positions and reflectance bear little correlation on the fidelity of the final result.

\begin{figure}
\begin{center}
\includegraphics[width=0.45\linewidth]{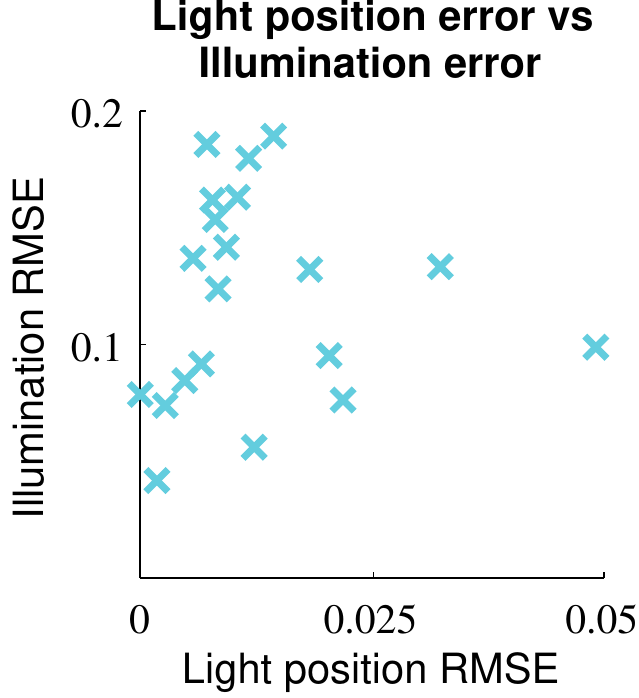} 
\hfill
\includegraphics[width=0.45\linewidth]{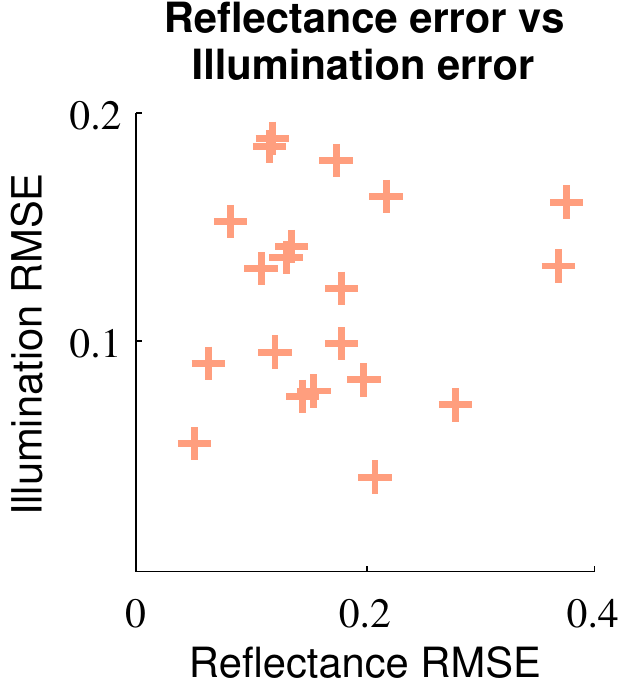} 
\end{center}
\caption{The physical accuracy of our light position estimates as well as reflectance have little influence on the accuracy of illumination. This is likely because the light positions are optimized so that the rendered scene looks most like the original image, and the reflectance estimates are biased by our rough geometry estimates.
}
\label{fig:error_correlation}
\end{figure}

To verify this point, for each of the scenes in Sec~\ref{sec:eval_lighting}, we plotted the physical accuracy of our illumination estimates versus the physical accuracy of both our light position and reflectance estimates (Fig~\ref{fig:error_correlation}).  Light positions were marked by hand and a Macbeth ColorChecker was used for ground truth reflectance. We found that the overall Pearson correlation of illumination error and lighting position error was 0.034, and the correlation between illumination error and reflectance error was 0.074. These values and plots indicate a weak relation for both comparisons. Thus, our method is particularly good at achieving the final result, but this comes at the expense of physical inaccuracies along the way.

\section{User study}
\label{sec:rsolp:study}
We also devised a user study to measure how well users can differentiate between real images and synthetic images of the same scene under varying conditions. For the study, we show each participant a sequence of images of the same background scene containing various objects. Some of these images are photographs containing no synthetic objects, and other images contain synthetic objects inserted with one of three methods: our method, a variant of Debevec's light probe method\footnote{We use Debevec's method for estimating illumination through the use of a light probe, coupled with our estimates of geometry and reflectance.}~\cite{Debevec:98}, or a baseline method (our method but with a simplified lighting model). Images are always shown in pairs and each of the paired images contain the exact same objects (although some of these objects may be synthetically inserted). The task presented to each user is a two-alternative forced choice test: for each pair of images, the user must choose (by mouse click) the image which they believe appears {\it most realistic}.

\boldhead{Methods} Our study tests three different methods for inserting synthetic objects into an image.  For the first, we use our method as described in Section~\ref{sec:technique}, which we will call {\bf ours}. We also compare to a method that uses Debevec's light probe method for estimating the illumination, combined with our coarse geometry and reflectance estimates, referred to as {\bf light probe}. To reproduce this method, we capture HDR photographs of a mirrored sphere in the scene from two angles (for removing artifacts/distortion), use these photographs to create a radiance map, model local geometry, and composite the rendered results~\cite{Debevec:98}. Much more time was spent creating scenes with the light probe method than our own. The third method, denoted as {\bf baseline}, also uses our geometry and reflectances but places a single point light source near the center of the ceiling rather than using our method for estimating light sources. {\it Note that each of these methods use identical reflectance and geometry estimates; the only change is in illumination.}

\boldhead{Variants} We also test four different variations when presenting users with the images to determine whether certain image cues are more or less helpful in completing this task. These variants are {\bf monochrome} (an image pair is converted from RGB to luminance), {\bf cropped} (shadows and regions of surface contact are cropped out of the image), {\bf clutter} (real background objects are added to the scene), and {\bf spotlight} (a strongly directed out of scene light is used rather than diffuse ceiling light). Note that the spotlight variant requires a new lighting estimate using our method, and a new radiance map to be constructed using the light probe method; also, this variant is not applicable to the baseline method. If no variant is applied, we label its variant as {\bf none}.

\begin{figure*}
\centerline{
\includegraphics[width=0.245\columnwidth]{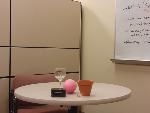}
\includegraphics[width=0.245\columnwidth]{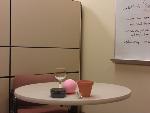}
\includegraphics[width=0.245\columnwidth]{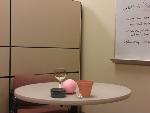}
\includegraphics[width=0.245\columnwidth]{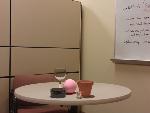} \vspace{-30mm}
}
\centerline{ \color{white} \Large \hspace{12mm} Real \hspace{30mm} Ours \hspace{24mm}  Light probe  \hspace{17mm} Baseline \hspace{10mm}}
 \vspace{23.5mm}
\centerline{
\includegraphics[width=0.245\columnwidth]{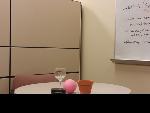}
\includegraphics[width=0.245\columnwidth]{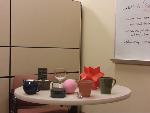}
\includegraphics[width=0.245\columnwidth]{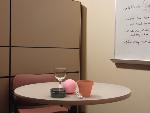}
\includegraphics[width=0.245\columnwidth]{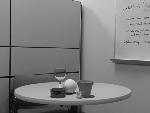} 
\vspace{-30mm}
}
\centerline{ \color{white} \Large \hspace{10mm} Cropped \hspace{24mm} Clutter \hspace{24mm}  Spotlight  \hspace{17mm} Monochrome \hspace{5mm}}
 \vspace{25.5mm}
\caption{Examples of methods and variants for Scene 10 in our user study. In the top row, from left to right, we show the real image, and synthetic images produced by our method, the light probe method, and the baseline method. In the bottom row, the four variants are shown.
}
\label{fig:study_overview}
\end{figure*}

\boldhead{Study details}  
There are 10 total scenes that are used in the study. Each scene contains the same background objects (walls, table, chairs, etc) and has the same camera pose, but the geometry within the scene changes. We use five real objects with varying geometric and material complexity (shown in Fig~\ref{fig:study_overview}), and have recreated synthetic versions of these objects with 3D modeling software. The 10 different scenes correspond to unique combinations and placements of these objects. Each method was rendered using the same software (LuxRender), and the inserted synthetic geometry/materials remained constant for each scene and method. The rendered images were tone mapped with a linear kernel, but the exposure and gamma values differed per method. Tone mapping was performed so that the set of all scenes across a particular method looked most realistic (i.e. our preparation of images was biased towards realistic appearance for a skilled viewer, rather than physical estimates).

We recruited 30 subjects for this task. All subjects had a minimal graphics background, but a majority of the participants were computer scientists and/or graduate students. Each subject sees 24 pairs of images of identical scenes. 14 of these pairs contain one real and one synthetic image. Of these 14 synthetic images, five are created using our method, five are created using the light probe method, and the remaining four are created using the baseline method. Variants are applied to these pairs of images so that each user will see exactly one combination of each method and the applicable variants.  The other 10 pairs of images shown to the subject are all synthetic; one image is created using our method, and the other using the light probe method. No variants are applied to these images.

Users are told that their times are recorded, but no time limit is enforced. We ensure that all scenes, methods, and variants are presented in a randomly permuted order, and that the image placement (left or righthand side) is randomized. In addition to the primary task of choosing the most realistic image in the image pair, users are asked to rate their ability in performing this task both before and after the study using a scale of 1 (poor) to 5 (excellent). 

\begin{figure}
\begin{center}
\begin{minipage}{0.8\linewidth}
\includegraphics[width=\linewidth]{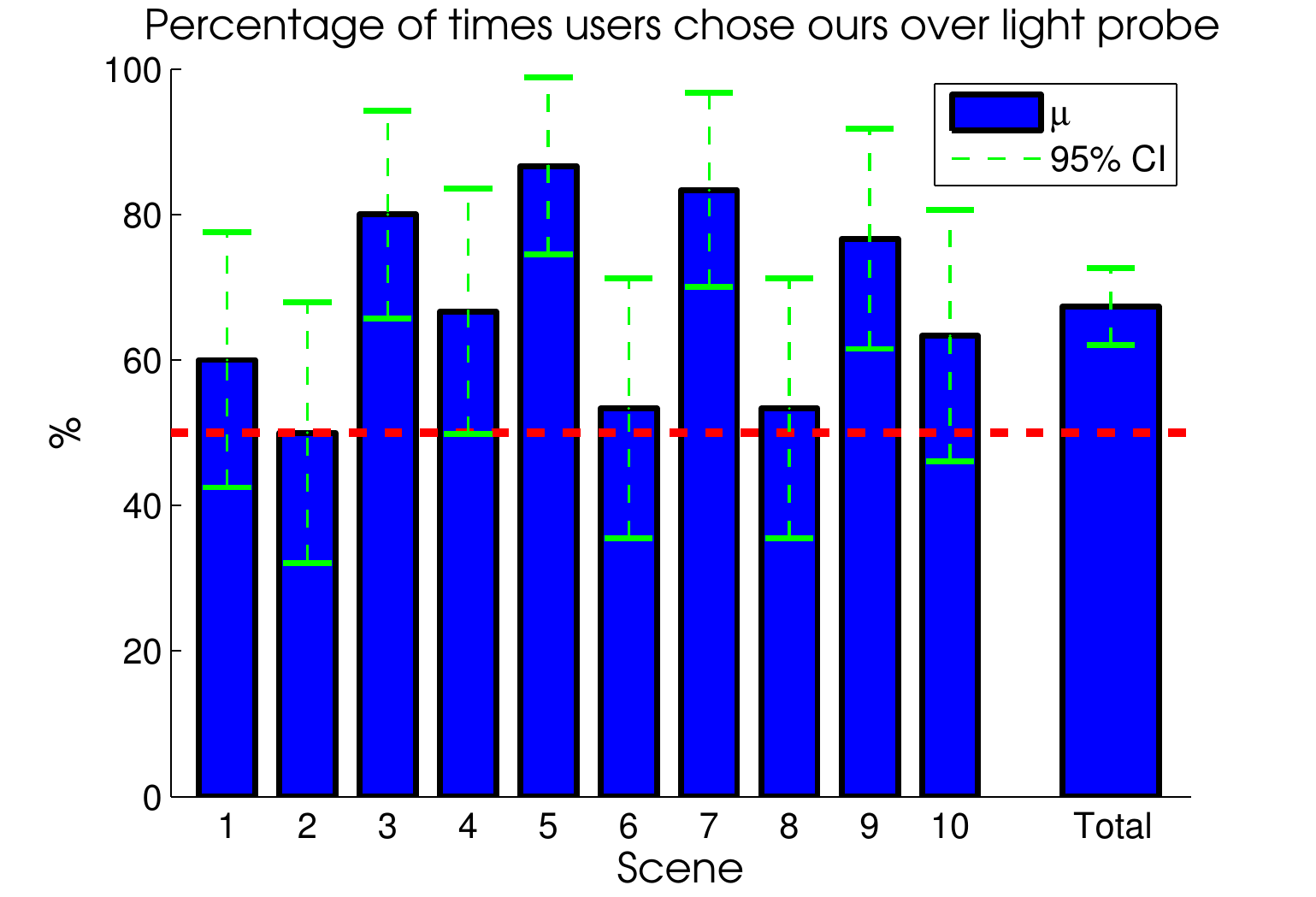} \\
\line(1,0){400}\\
\begin{minipage}[b]{0.1\linewidth}
\hfill
\begin{rotate}{90} \hspace{20mm}\Large Scene 9 \hspace{30mm} Scene 8 \end{rotate}
\end{minipage}
\begin{minipage}[b]{0.44\linewidth}
\centerline{\Large Ours} \vspace{1mm}
\includegraphics[width=\linewidth]{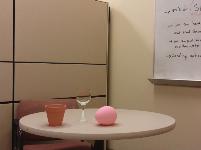}
\includegraphics[width=\linewidth]{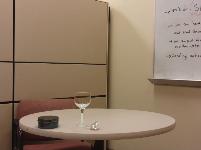}
\end{minipage}
\begin{minipage}[b]{0.44\linewidth}
\centerline{\Large Light probe} \vspace{1mm}
\includegraphics[width=\linewidth]{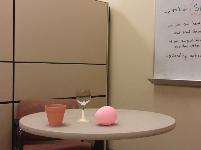}
\includegraphics[width=\linewidth]{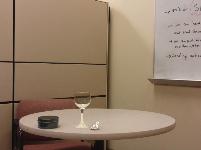}
\end{minipage}
\end{minipage}
\end{center}
\caption{When asked to pick which method appeared more realistic, subjects chose our method over the light probe method at least 50\% of the time for each scene (67\% on average), indicating a statistically significant number of users preferred our method. The blue bars represent the mean response (30 responses per bar, 300 total), and the green lines represent the 95\% confidence interval. The horizontal red line indicates the 50\% line. The images below the graph show two scenes from the study that in total contain all objects.  Scene 8 was one of the lowest scoring scenes (53\%), while scene 9 was one of the highest scoring (77\%). 
}
\label{fig:study_us_v_deb}
\end{figure}

\boldhead{Results} 
We analyze the results of the different methods versus the real pictures separately from the results of our method compared to the light probe method. When describing our results, we denote $N$ as the sample size. When asked to choose which image appeared more realistic between our method and the light probe method, participants chose our image 67\% of the time (202 of 300).  Using a one-sample, one-tailed t-test, we found that users significantly preferred our method ($p\text{-value} \ll 0.001$), and on average users preferred our method more than the light probe method for all 10 scenes (see Fig~\ref{fig:study_us_v_deb}).

\begin{figure}
\begin{center}
\includegraphics[width=.8\columnwidth]{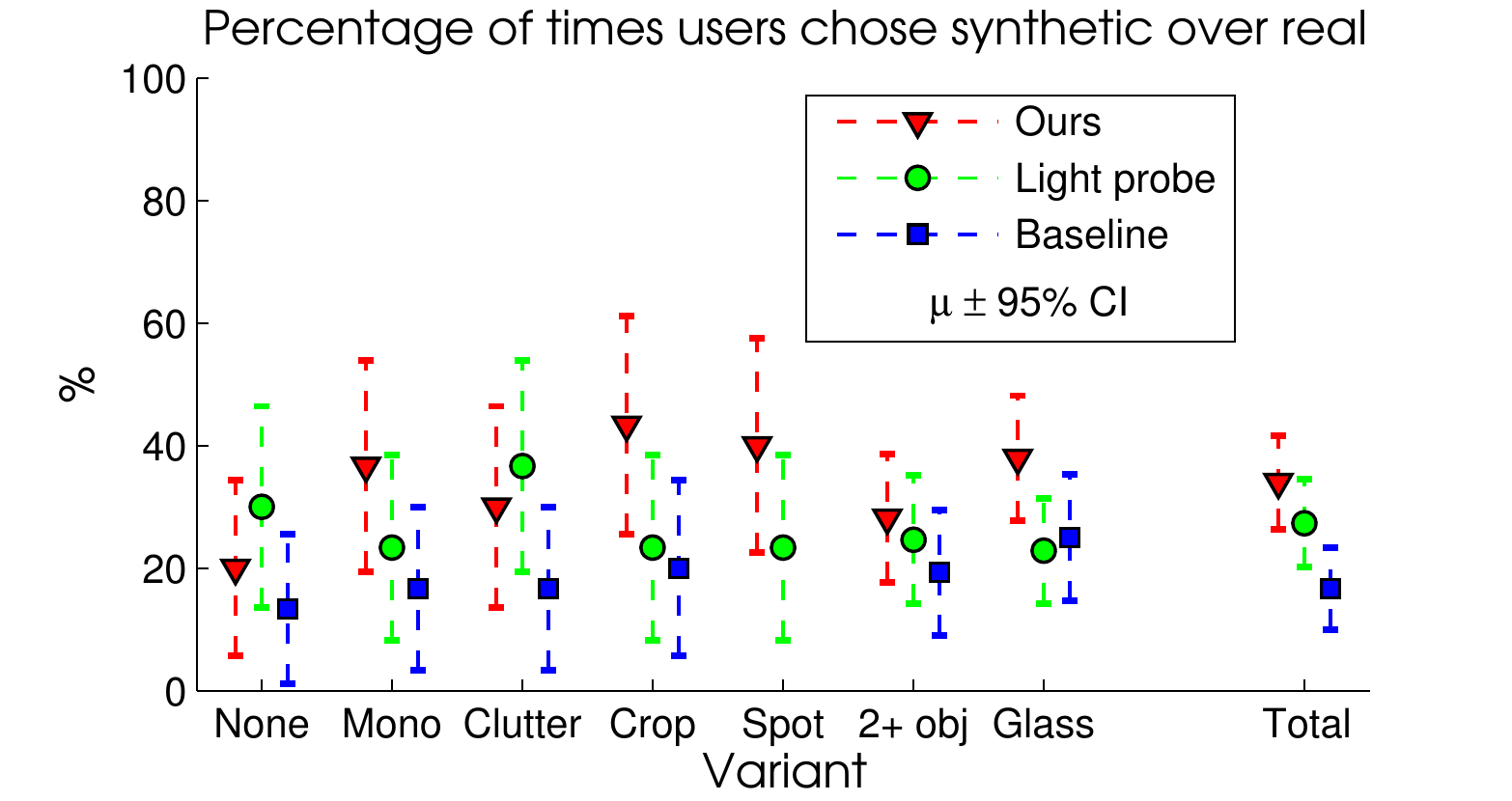} \vspace{5mm} \\
\centerline{\bf \Large Percentage of times users chose synthetic over real}
\Large
\begin{tabular}{|c||c c c| c|} \hline
  $N=30$ & ours & light probe & baseline & total \\ \hline
 none   & 20 & 30 & 13.3 & 21.1  \\
 monochrome & 36.7 & 23.3 & 16.7 & 26.6 \\
 clutter & 30 & 36.7 & 16.7 & 27.8 \\
 cropped & 43.3 & 23.3 & 20 & 28.9\\
 spotlight &  40 & 23.3 & N/A & 31.7 \\ \hline
 total & 34 & 27.3 & 16.7 & 26.7 \\ \hline
\end{tabular}
\end{center}
\vspace{1mm}
\begin{center}
\Large
\begin{tabular}{|c||c c c| c|} \hline
  & ours & light probe & baseline & total \\ \hline
 \hspace{1mm}  2+ objects \hspace{1mm}  & 28.2 & 24.6 & 19.3 & 24.4  \\
 glass & 37.9 & 22.8 & 25 & 28.7 \\ \hline
\end{tabular}
\end{center}
\caption{Results for the three methods compared to a real image. In the graph, the mean response for each method is indicated by a triangle (ours), circle (light probe), and square (baseline). The vertical bars represent the 95\% binomial confidence interval. The tables indicate the average population response for each category. We also considered the effects of inserting multiple synthetic objects and synthetic objects made of glass, and these results were consistent with other variants. Both our method and the light probe method performed similarly, indicated especially by the overlapping confidence intervals, and both methods clearly outperform the baseline. Variants do appear to have a slight affect on human perception (making it harder to differentiate real from synthetic).
}
\label{fig:study_alldata}
\end{figure}

In the synthetic versus real comparison, we found overall that people incorrectly believe the synthetic photograph produced with our method is real 34\% of the time (51 of 150), 27\% of the time with the light probe method (41 of 150), and 17\% for the baseline (20 of 120). Using a two-sample, one-tailed t-test, we found that there was not a significant difference in subjects that chose our method over the light probe method ($p = 0.106$); however, there was a significant difference in subjects choosing our method over the baseline ($p = 0.001$), and in subjects choosing the light probe method over the baseline ($p=0.012$). For real versus synthetic comparisons, we also tested the variants as described above. All variants (aside from ``none'') made subjects perform worse overall in choosing the real photo, but these changes were not statistically significant. Figure~\ref{fig:study_alldata} summarizes these results.

We also surveyed four non-na\"{\i}ve users (graphics graduate students), whose results were not included in the above comparisons. Contrary to our assumption, their results {\it were} consistent with the other 30 na\"{\i}ve subjects. These four subjects selected 2, 3, 5, and 8 synthetic photographs (out of 14 real-synthetic pairs), an average of 35\%, which is actually higher than the general population average of 27\% (averaged over all methods/variants), indicating more trouble in selecting the real photo. In the comparison of our method to the light probe method, these users chose our method 5, 7, 7, and 8 times (out of 10 pairs) for an average of 68\%, consistent with the na\"{\i}ve subject average of 67\%.
 
\boldhead{Discussion} 
From our study, we conclude that both our method and the light probe method are highly realistic, but that users can tell a real image apart from a synthetic image with probability higher than chance. However, even though users had no time restrictions, they still could not differentiate real images from both our method and the light probe method reliably. As expected, both of these synthetic methods outperform the baseline, but the baseline still did surprisingly well. Applying different variants to the pairs of images hindered subjects' ability to determine the real photograph, but this difference was not statistically significant.

When choosing between our method and the light probe method, subjects chose our method with equal or greater probability than the light probe method for each scene in the study. This trend was probably the result of our light probe method implementation, which used rough geometry and reflectance estimates produced by our algorithm, and was not performed by a visual effects or image-based lighting expert. {\it Had such an expert generated the renderings for the light probe method, the results for this method might have improved, and so led to a change in user preference for comparisons involving the light probe method. The important conclusion is that we can now achieve realistic insertions without access to the scene.} 


Surprisingly, subjects tended to do a worse job identifying the real picture as the study progressed. We think that this may have been caused by people using a particular cue to guide their selection initially, but during the study decide that this cue is unreliable or incorrect, when in fact their initial intuition was accurate. If this is the case, it further demonstrates how realistic the synthetic scenes look as well as the inability of humans to pinpoint realistic cues.

\begin{figure}[htp]
\begin{center}
\includegraphics[width=0.49\columnwidth]{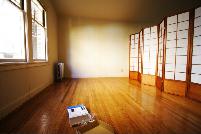}
\includegraphics[width=0.49\columnwidth]{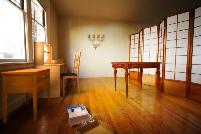}
\end{center}
\caption{Home redecorating is a natural application for our method. A user could take a picture of a room, and visualize new furniture or decorations without leaving home.
}
\label{fig:interiordecor}
\end{figure}

Many subjects commented that the task was more difficult than they thought it would be. Self assessment scores reflected these comments as self evaluations decreased for 25 of 30 subjects (i.e. a subject rated him/herself higher in the entry assessment than in the exit assessment), and in the other five subjects, the assessment remained the same. The average entry assessment was 3.9, compared to the average exit assessment of 2.8. No subject rated him/herself higher in the exit assessment than in the entry assessment.

The fact that na\"{\i}ve subjects scored comaparably to non-na\"{\i}ve subjects indicates that this test is difficult even for those familiar with computer graphics and synthetic renderings. All of these results indicate that people are not good at differentiating real from synthetic photographs, and that our method is state of the art.

\section{Results and discussion}
\begin{figure}
\begin{center}
\begin{minipage}{0.75\linewidth}
\includegraphics[width=0.49\columnwidth]{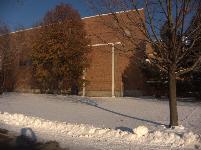}
\includegraphics[width=0.49\columnwidth]{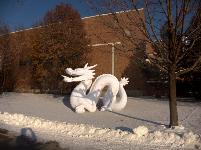}  \vspace{0.5mm} \\
\includegraphics[width=0.49\columnwidth]{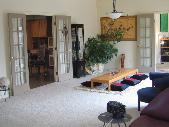}
\includegraphics[width=0.49\columnwidth]{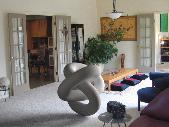}
\end{minipage}
\end{center}
\caption{Our algorithm can handle complex shadows \emph{(top)}, as well as out-of-view light sources \emph{(bottom)}.
}
\label{fig:results1}
\end{figure}

\begin{figure}
\begin{center}
\begin{minipage}{0.75\linewidth}
\includegraphics[width=0.49\columnwidth]{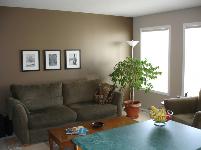}
\includegraphics[width=0.49\columnwidth]{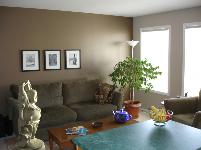} \vspace{0.5mm} \\
\includegraphics[width=0.49\columnwidth]{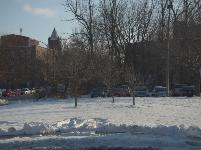}
\includegraphics[width=0.49\columnwidth]{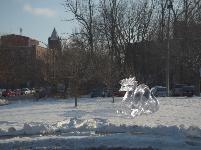}
\end{minipage}
\end{center}
\caption{Specular materials naturally reflect the scene \emph{(top)}, and translucent objects reflect the background realistically \emph{(bottom)}.
}
\label{fig:results2}
\end{figure}

\begin{figure}
\begin{center}
\includegraphics[width=0.49\columnwidth]{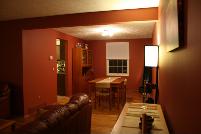}
\includegraphics[width=0.49\columnwidth]{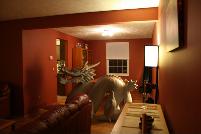}  
\end{center}
\vspace{0.5mm}
\begin{center}
\includegraphics[width=0.49\columnwidth]{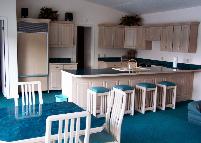}
\includegraphics[width=0.49\columnwidth]{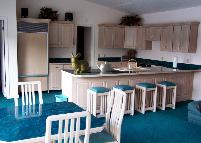}
\end{center}
\caption{Complex occluding geometry can be specified quickly via segmentation \emph{(top, couch)}, and glossy surfaces in the image reflect inserted objects \emph{(bottom, reflections under objects).}
}
\label{fig:results3}
\end{figure}

We show additional results produced with our system in Figs~\ref{fig:interiordecor}-~\ref{fig:results3}.  Lighting effects are generally compelling (even for inserted emitters, Fig~\ref{fig:mirror}), and light interplay occurs automatically (Fig~\ref{fig:interreflections}), although  result quality is dependent on inserted models/materials.  We conclude from our study that when shown to people, results produced by our method are confused with real images quite often, and compare favorably with other state-of-the-art methods.

Our interface is intuitive and easy to learn. Users unfamiliar with our system or other photo editing programs can begin inserting objects within minutes. Figure~\ref{fig:noviceresult} shows a result created by a novice user in under 10 minutes.

We have found that many scenes can be parameterized by our geometric representation. Even images without an apparent box structure (e.g. outdoor scenes) work well (see Figs~\ref{fig:results1} and~\ref{fig:results2}). 

Quantitative measures of error are reassuring; our method beats natural baselines (Fig~\ref{fig:errorvis}).  Our intrinsic decomposition method incorporates a small amount of interaction and achieves significant improvement over Retinex in a physical comparison (Fig~\ref{fig:intrinsiccomp}), and the datasets we collected (Sec~\ref{sec:Evaluation}) should aid future research in lightness and material estimation. However, it is still unclear which metrics should be used to evaluate these results, and qualitative evaluation is the most important for applications such as ours.


\subsection{Limitations and future work}
For extreme camera viewpoints (closeups, etc), our system may fail due to a lack of scene information. In these cases, luminaires may not exist in the image, and may be difficult to estimate (manually or automatically). Also, camera pose and geometry estimation might be difficult, as there may not be enough information available to determine vanishing points and scene borders.

Intrinsic image extraction may fail, either because the problem is still very difficult for diffuse scenes or because surfaces are not diffuse.  For example, specular surfaces modeled as purely diffuse may cause missed reflections. Other single material estimation schemes could be used \cite{Boivin:2001,Debevec:98}, but for specular surfaces and complex BRDFs, these methods will also likely require manual edits. It would be interesting to more accurately estimate complex surface materials automatically. Robust interactive techniques might also be a suitable alternative (i.e. \cite{Carroll:sg11}).


Insertion of synthetic objects into legacy videos is an attractive extension to our work, and could be aided, for example, by using multiple frames to automatically infer geometry~\cite{Furukawa:2010}, surface properties~\cite{Yu:1999}, or even light positions. Tone mapping rendered images can involve significant user interaction, and methods to help automate this process the would prove useful for applications such as ours.  Incorporating our technique within redecorating aids (e.g. \cite{Merrell:sg11,Yu:sg11}) could also provide a more realistic sense of interaction and visualization (as demonstrated by Fig~\ref{fig:interiordecor}).



\section{Conclusion}
We have demonstrated a system that allows a user to insert objects into legacy images. Our method only needs a few quick annotations, allowing novice users to create professional quality results, and does not require access to the scene or any other tools used previously to achieve this task. 
The results achieved by our method appear realistic, and people tend to favor our synthetic renderings over other insertion methods.

\chapter{Shape representations for realistic relighting}
\label{chap:shape}

\section{Boundary cues for 3D object shape recovery}
\subsection{Introduction}
3D object shape is a major cue to object category and function.  Early approaches to object recognition~\cite{Roberts} considered shape reconstruction as the first step. As data-driven approaches to recognition became popular, researchers began to represent shape implicitly through weighted image gradient features, rather than explicitly through reconstruction~\cite{Mundy06}.  The best current approaches recognize objects with mixtures of gradient-based templates.  Were the early researchers misguided to focus on explicit shape representation?
 
We have good reason to reconsider the importance of 3D shape.  A study by Hoiem et al.~\cite{hoiem2012eccv} provides some evidence that gradient-based features are a limiting factor in object detection performance.  Distinct architectures~\cite{felzenszwalb2010pami, vedaldi09iccv} whose main commonality is gradient-based feature representations have very similar performance characteristics.  The study also suggests that performance may be limited by heavy-tailed appearance distributions of object categories.  For example, projected dog shapes may vary due to pose, viewpoint, and high intraclass variation.  Because many examples are required to learn which boundaries are reliable (i.e., correspond to shape), dogs of unusual variety, pose, or viewpoint are poorly classified.  Representations based on 3D shape would enable more sample-efficient category learning through viewpoint robustness and a reduced need to learn stable boundaries through statistics. Beyond interest in object categorization, ability to recover 3D shape is important for inferring object pose and affordance and for manipulation tasks.
 
The importance of shape is clear, but there are many mysteries to be solved before we can recover shape.  What cues are important?  What errors in 3D shape are important?  How do we recover shape cues from an image?  How do we encode and use 3D shape for recognition?  In this chapter, we focus on improving our understanding of the importance of boundary shape cues for 3D shape reconstruction and recognition.  In particular, we consider boundaries due to object silhouette, self-occlusion (depth discontinuity) and folds (surface normal discontinuity).  We also consider 
cues for whether boundaries are soft (extrema of curved surface) or sharp.  We evaluate on a standard 3D shape dataset and a selection of PASCAL VOC object images.  On the standard dataset, reconstructions using various cues are compared via metrics of surface normal and depth accuracy.  On the VOC dataset, we evaluate reconstructions qualitatively and in terms of how well people and computers can categorize objects given the reconstructed shape.
 
\begin{figure*}
\begin{minipage}{\textwidth}
\begin{minipage}{.14\linewidth}\centerline{Input image/labels}\end{minipage}
\begin{minipage}{.139\linewidth}\centerline{silh}\end{minipage}
\begin{minipage}{.139\linewidth}\centerline{+selfocc}\end{minipage}
\begin{minipage}{.139\linewidth}\centerline{+folds}\end{minipage}
\begin{minipage}{.139\linewidth}\centerline{+occ+folds}\end{minipage}
\begin{minipage}{.135\linewidth}\centerline{+shading}\end{minipage}
\begin{minipage}{.135\linewidth}\centerline{+shading+occ+folds}\end{minipage}\\
\centerline{ \includegraphics[width=\linewidth]{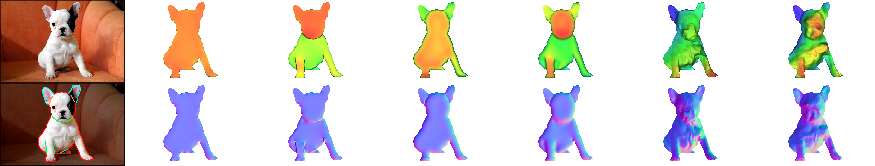}}
\caption{For a given input image, we hand-label geometric cues including: smooth silhouette contour (red), sharp silhouette contour (cyan), self occlusions (green), and folds (orange). We then use various combinations of these cues (as well as appearance-based cues) to obtain different shape reconstructions (see Sec~\ref{sec:boundarycues:eval}). We evaluate these reconstructions in a variety of tasks in order to find which set(s) of cues may be most beneficial for reconstructing shapes.}
\label{fig:sample_shapes}
\end{minipage}
\end{figure*}

\boldhead{Contributions} 
Our main contribution is to evaluate the importance of various boundary and shading cues for shape reconstruction and shape-based recognition. We extend Barron and Malik's shape from shading and silhouette method~\cite{Barron2012B} to include interior occlusions with figure/ground labels, folds, and sharp/soft boundary labels.  The standard evaluation is based on depth error, surface normal, shading, or reflectance on the MIT Intrinsic Image dataset.  We also introduce perceptual and recognition-based measures of reconstruction quality for the PASCAL VOC dataset (Fig~\ref{fig:sample_shapes} shows one example of the types of reconstructions we evaluate, and the annotation required by our algorithm). These experiments are important because they tests reconstruction of typical objects, such as cats and boats, with complex shapes and materials in natural environments, and because it can provide insight into which errors matter. Furthermore, much work has gone into shape-based representations for recognition, focusing on the cues provided by the silhouette (e.g. Ferrari et al.~\cite{ferrari:hal-00204002}).  Our findings suggest a 3D representation that incorporates interior occlusions and folds might benefit such existing systems. 
 
\boldhead{Limitations}  Our study is a good step towards understanding shape reconstruction in the context of recognition, but we must leave several aspects of this complex problem unexplored.  First, we assume boundary cues are provided.  Eventually, we will want automatic recovery of shape cues and reconstruction algorithms that handle uncertainty.  Second, cues such as ground contact points and object-level shape priors are useful but not investigated.  Third, we assume an orthographic projection which can be a poor assumption for large objects, such as busses or trains.  Finally, we recover depth maps, which provides a 2.5D reconstruction, rather than a full 3D reconstruction.


\subsection{Cues for object reconstruction}
We focus on reconstructing shape from geometric cues, revisiting early work on reconstructing shape from line drawings~\cite{MalikThesis,MalikMaydan}. Through human labeling, we collect information about an object's {\it silhouette}, {\it self-occlusions}, and {\it folds} in the surface. Since appearance can be a helpful factor in determining shape, we also investigate the benefit of shading cues using the shape-from-shading priors of Barron and Malik~\cite{Barron2012B}. Figure~\ref{fig:sample_shapes} shows reconstructions using each of these cues.

To reconstruct shapes, we extend the continuous optimization framework of Barron and Malik by building in additional constraints on the surface. Following Barron and Malik's notation, we write $Z$ for the surface (represented by a height field viewed orthographically), and $N : \mathbb{R} \rightarrow \mathbb{R}^3 $ as the function that takes a height field to surface normals (component-wise; $N = (N^x, N^y, N^z)$). We use a coordinate system such that $x$ and $y$ vary in the image plane, and negative $z$ is in the viewing direction.

Extending Barron and Malik's continuous optimization framework, we write our optimization problem as:
\begin{eqnarray}
\label{eq:objective}
\displaystyle \underset{ Z, R, L }{\operatorname{minimize}} \!\!&\!&\!\! \delta_\mathit{sfc} f_\mathit{sfc}( Z ) + \delta_\mathit{selfocc} f_\mathit{selfocc}(Z) \nonumber \\
\!\!&\!&\!\!  + \delta_\mathit{fold} f_\mathit{fold}(Z) + \delta_\mathit{reg} f_\mathit{reg}(Z) \nonumber \\
\!\!&\!&\!\!   + \delta_\mathit{sfs} (g(R)  +  h(L)) \nonumber \\
\displaystyle \operatorname{subject\;to} \!\!&\!&\!\!  c_\mathit{sfs}(Z,R,L) = 0,
\end{eqnarray}
where $f_{{}^*}$ and $c_{\mathit{sfs}}$ are sub-objective and constraint functions, $g(R)$ and $h(L)$ are priors on reflectance and illumination, and $\delta_{{}^*}$ are the weights that determine their influence. In the remainder of the section, we describe each of these functions/constraints. 

\boldhead{Silhouette} The silhouette is rich with shape information, both perceptually and geometrically~\cite{koenderink1984does}.  At the occluding contour of an object, the surface is tangent to all rays from the vantage point, unless however there is a discontinuity in surface normals across the visible and non-visible regions of the object (e.g. the edges of a cube).  We treat these two cases separately, labeling parts of the silhouette as {\it smooth} if the surface normal should lie perpendicular to both the viewing direction and image silhouette, and {\it sharp} otherwise\footnote{It is also common notation to denote smooth boundaries as ``limbs'' and sharp boundaries as ``edges'' or ``cuts''}. In the case of a smooth silhouette contour, the $z$-component of the normal is $0$, and the $x$ and $y$ components are normal to the silhouette (i.e. perpendicular to the silhouette's tangent in 2D). Denoting $(n^x,n^y)$ as normals of the silhouette contour, and $C_{\mathit{smooth}}$ as the set of pixels labelled as the smooth part of the silhouette, 
we write the silhouette constraint as:
\begin{equation}
f_{\mathit{sfc}}(Z) = \sum_{i\in C_{\mathit{smooth}}} \sqrt{ (N_i^x(Z)-n_i^x)^2 +  (N_i^y(Z)-n_i^y)^2 }.
\label{eq:sfc}
\end{equation}
This is the most typical constraint used in shape-from-contour algorithms (hence the notation $f_{\mathit{sfc}}$), and is identical to that used by Barron and Malik, with the notable exception that we only enforce the constraint when the silhouette is not sharp. If the silhouette is labelled sharp, there is no added constraint.

\boldhead{Self-occlusions} Self-occlusions can be thought of in much the same way as the silhouette. The boundary of a self-occlusion implies a {\it discontinuity in depth}, and thus the surface along the foreground boundary should be constrained to be tangent to the viewing direction. Besides knowing a self occlusion boundary, it is also mandatory to know which side of the contour is in front of the other (figure and ground labels). With this information, we impose additional surface normal constraints along self occlusion boundaries ($C_{\mathit{selfocc}}$):
\begin{equation}
f_{\mathit{selfocc}}(Z) = \hspace{-2mm} \sum_{i\in C_{\mathit{selfocc}}} \sqrt{ (N_i^x(Z)-n_i^x)^2 +  (N_i^y(Z)-n_i^y)^2 }.
\label{eq:selfocc}
\end{equation}
Notice that there is no explicit constraint to force the height of the foreground to be greater than that of the background; however, by constraining the foreground normals to be pointing outward and perpendicular to the viewing direction, the correct effect is achieved. This is due in part because we enforce integrability of the surface (since height is directly optimized).

\boldhead{Folds} A fold in the surface denotes a {\it discontinuity in surface normals} across a contour along the object, e.g. edges where faces of a cube meet. Folds can be at any angle (e.g. folds on a cube are at $90^\circ$, but this is not always the case), and can be convex (surface normals pointing away from each other) or concave (surface normals pointing towards each other). Our labels consist of fold contours and also a flag denoting whether the given fold is convex or concave. We did not annotate exact fold orientation as this task is susceptible to human error and tedious. 

We incorporate fold labels by adding another term to our objective function, developed using intuition from Malik and Maydan~\cite{MalikMaydan}. The idea is to constrain normals at pixels that lie across a fold to have convex or concave orientation (depending on the label), and to be oriented consistently in the direction of the fold. Let $\mathbf{u}=(\mathbf{u}_x,\mathbf{u}_y,0)$ be a fold's tangent vector in the image plane, and $N_i^\ell, N_i^r$ as two corresponding normals across pixel $i$ in the fold contour $C$. We write the constraint as
\begin{eqnarray}
f_{\mathit{fold}}(Z) = \sum_{i \in C} \max(0, \epsilon - (N_i^\ell \times N_i^r) \cdot \mathbf{u}),
\end{eqnarray}
and set $\epsilon = {1 \over \sqrt{2}}$ (additional details can be found in \hyperlink{appendix:foldConstraint}{Appendix A}).

\boldhead{Regularization priors} Because we only have constraints at a sparse set of points on the surface, we incorporate additional terms to guide the optimization to a plausible result. Following Barron and Malik, we impose one prior that prefers the flattest shape within the bas-relief family ($f_f$), and another that minimizes change in mean curvature ($f_k$):

\begin{align}
f_f(Z) &= -\sum_{i \in \mathit{pixels}} \log\left( N^z_{i}(Z) \right), \\
f_k(Z) &= \sum_{i \in \mathit{pixels}} \sum_{j \in \mathit{neighbors}(i)} c\left(  H(Z)_i - H(Z)_j \right),\\ 
f_{\mathit{reg}}(Z) &= \lambda_f f_f(Z) + \lambda_k f_k(Z),
\end{align}
where $c(\cdot)$ is the negative log-likelihood of a Gaussian scale mixture, $H(\cdot)$ computes mean curvature, and the neighbors are in a 5x5 window around $i$. For all of our reconstructions, we set $\lambda_f = \lambda_k = 1$ (see~\cite{Barron2012B} for implementation details).

\boldhead{Shading} We use the albedo and illumination priors of Barron and Malik to incorporate shading cues into our reconstructions. Summarizing these priors, we encourage albedo to be be piecewise smooth over space. Illumination is parameterized by second order spherical harmonics (9 coefficients per color channel), and is encouraged to match a Gaussian fit to real world spherical harmonics (regressed from an image based lighting dataset\footnote{\url{http://www.hdrlabs.com/sibl}}). For brevity, we denote priors on reflectance as $g(R)$, and priors on illumination as $h(L)$, where $R$ is log-diffuse reflectance (log-albedo) and $L$ is the 27-dimensional RGB spherical harmonic coefficient vector. We refer the reader to~\cite{Barron2012B,Barron:cvpr:12} for further details.

Jointly estimating shape along with albedo and illumination requires an additional constraint that forces a rendering of the surface to match the input image. Assuming Lambertian reflectance and disregarding occlusions, our rendering function is simply reflectance multiplied by shading (or in log space, log-reflectance plus log-shading). Denoting $I$ as the log-input image, $R$ as log-diffuse reflectance (log-albedo), and $S(Z,L)$ as the log-shaded surface $Z$ under light $L$, we write the shape-from-shading constraint as:
\begin{equation}
c_{\mathit{sfs}}(Z,R,L) = R + S(Z,L) - I.
\end{equation}
We emphasize that $R$, $I$, and $S(\cdot)$ are all in log-space, as is done in~\cite{Barron2012B} which allows us to write the rendering constraint in additive fashion.

\begin{figure*}
\begin{center}
\begin{minipage}{.98\linewidth}
\begin{minipage}{.13\linewidth}\centerline{Input +}\centerline{annotations}\end{minipage}
\hspace{5mm}
\begin{minipage}{.26\linewidth}\centerline{\bf +occ+folds} \centerline{\small View 1 \hspace{10mm} View 2} \end{minipage}
\hspace{2mm}
\begin{minipage}{.26\linewidth}\centerline{\bf +shading} \centerline{\small View 1 \hspace{10mm} View 2} \end{minipage}
\hspace{2mm}
\begin{minipage}{.26\linewidth}\centerline{\bf +shading+occ+folds} \centerline{\small View 1 \hspace{10mm} View 2} \end{minipage}
\vspace{1mm}\\
\includegraphics[width=\linewidth]{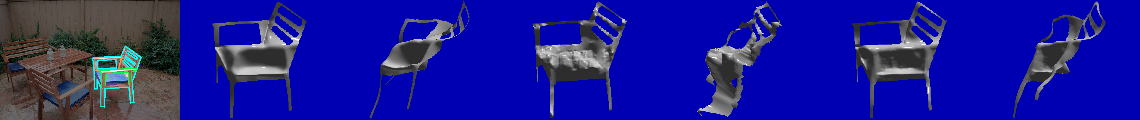}\\
\includegraphics[width=\linewidth]{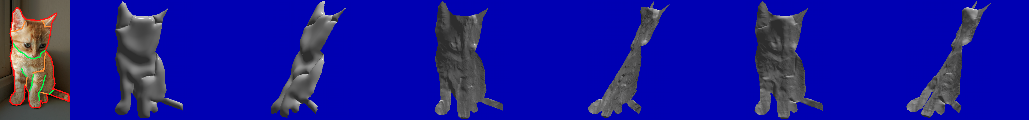}\\
\includegraphics[width=\linewidth]{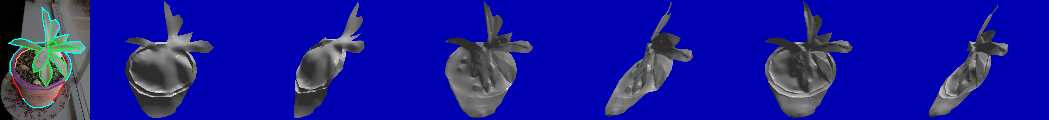}\\
\includegraphics[width=\linewidth]{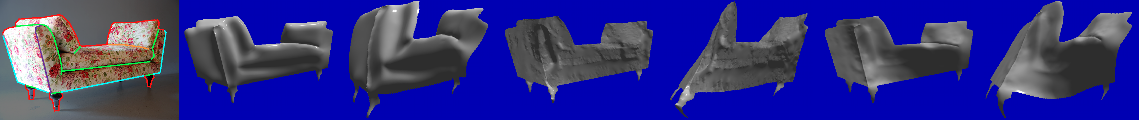}\\
\includegraphics[width=\linewidth]{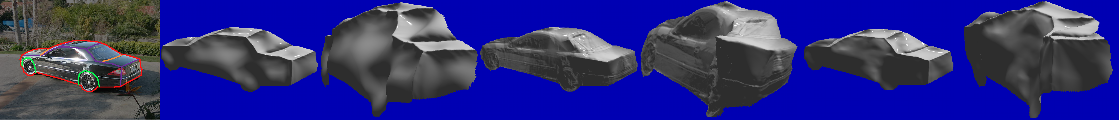}\\
\includegraphics[width=\linewidth]{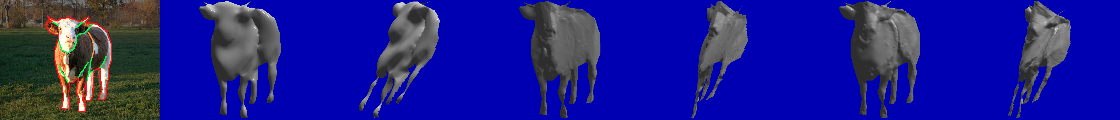}
\end{minipage}
\vspace{-0mm}
\end{center}
\caption{Several annotations and shape reconstructions used in our analyses. The annotated images (left) include: smooth silhouette contour (red), sharp silhouette contour (cyan), self occlusions (green), and folds (orange). In each row, we show the input image (with geometric labels), and the results of three reconstruction algorithms. For each algorithm, two views of the shape are shown (frontal on left, heavily rotated view on right). Notice that the reconstructed shapes look generally good frontally.  Rotated views expose that shape estimates often err towards being too flat (especially with the cow or potted plant).  This is the first publication that we know of to provide a rigorous analysis of shape reconstruction on typical objects in consumer photographs (e.g. outside of a lab setting).}
\label{fig:boundarycues:results}
\end{figure*}

\subsubsection{Optimization}
To estimate a shape given a set of labels, we solve the optimization problem in Eq.~\ref{eq:objective} using the multiscale optimization technique introduced by Barron and Malik~\cite{Barron2012B}. Notice that shading cues are only incorporated if $\delta_{\mathit{sfs}}>0$; otherwise, our reconstructions rely purely on geometric information.

\boldhead{Setting the weights ($\delta$)} Throughout our experiments, we choose each weight to be binary for two reasons. For one, each term in the objective should have an equal weighting for a fair comparison, otherwise one cue may dominate others. Second, learning these weights requires a dataset of ground truth shapes, and we have good reason to believe that weights learned from existing datasets (e.g. the MIT Intrinsic Image dataset~\cite{grosse09intrinsic}) will not generalize to shapes found in the VOC dataset (e.g. more geometric detail on VOC shapes). Furthermore, we ran the MIT-learned parameters on several of the VOC images, and noticed only slight perceptual differences in results.

\begin{figure}[t]
\centering
\includegraphics[width=0.95\linewidth]{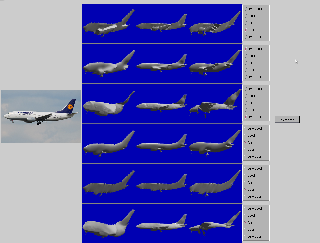}
\caption{User study interface for qualitative rating. Left: source image of the object. Middle: shape visualizations. Each row is the result of one algorithm (silh, +selfocc, etc) in random order visualized in three view angles: upper left, frontal, bottom right (from left to right). The participant rates each row as a whole.}
\label{fig:interface1}
\end{figure}

\subsection{Evaluation of shape and appearance cues}
\label{sec:boundarycues:eval}

In this section, we examine each of the cues used in our shape reconstruction method, and hope to find a cue or set of cues that lead to better shape estimates (qualitatively, and in terms of recognition ability).  Our objective function (Eq~\ref{eq:objective}) allows us to easily produce shape reconstructions for various combinations of cues by turning ``on'' and ``off'' different cues; equivalently, setting the corresponding weights to 1 (on) or 0 (off). We use six different cue combinations to see which cue or set of cues contribute most to a better reconstruction. These six combinations are:
\begin{itemize}
  \item {\bf silh}: Priors on silhouette shape and surface smoothness; i.e. shape-from-contour constraints ($\delta_{\mathit{sfc}} = 1$). 
  \item {\bf +selfocc}: Silhouette and self occlusion constraints ($\delta_{\mathit{sfc}} =  \delta_{\mathit{selfocc}} = 1$).
  \item {\bf +folds}: Silhouette and fold constraints ($\delta_{\mathit{sfc}} =  \delta_{\mathit{folds}} = 1$).
  \item {\bf +occ+folds}: Silhouette,  self occlusion and fold constraints ($\delta_{\mathit{sfc}} =  \delta_{\mathit{selfocc}} = \delta_{\mathit{folds}} = 1$).
  \item {\bf +shading}: Shape-from-shading as in~\cite{Barron2012B}; includes silh ($\delta_{\mathit{sfc}} =  \delta_{\mathit{sfs}} = 1$).
  \item {\bf +shading+occ+folds}: SFS with self occlusion and fold constraints ($\delta_{\mathit{sfc}} =  \delta_{\mathit{sfs}} = \delta_{\mathit{selfocc}} = \delta_{\mathit{folds}} = 1$).
\end{itemize}

We will refer to these as separate {\it algorithms} for the remainder of the chapter, and Fig~\ref{fig:sample_shapes} shows an example reconstruction for each of these algorithms.  Note that silh cues are present in each algorithm (hence the `+' prefix).

To find which cues are most critical for recovering shape, we evaluate each algorithm on a variety of tasks that measure {\it shape quality} and {\it shape recognition}.  We first evaluate the performance of the six algorithms on the VOC 2012 dataset. We selected 17 of the VOC categories out of the 20 (we exclude ``bicycle'', ``motorbike'' and ``person'' since we found these objects difficult to label by hand). Each class has 10 examples. Since we do not have ground truth shape for VOC objects, we conduct two user studies to evaluate qualitative performance: \emph{qualitative rating} and \emph{shape-based recognition}. Next, we evaluated the different algorithms using existing automatic recognition techniques, and compare them to the results of using RGB features (alone) and RGB+depth features. Finally, we ran a quantitative comparison of depth and surface normals using the MIT depth dataset. The remainder of this section details our results for each of these tasks, split under headings concerning shape quality and shape recognition.

\begin{figure}[t]
\begin{minipage}{0.475\linewidth}
\centering
\includegraphics[width=\linewidth]{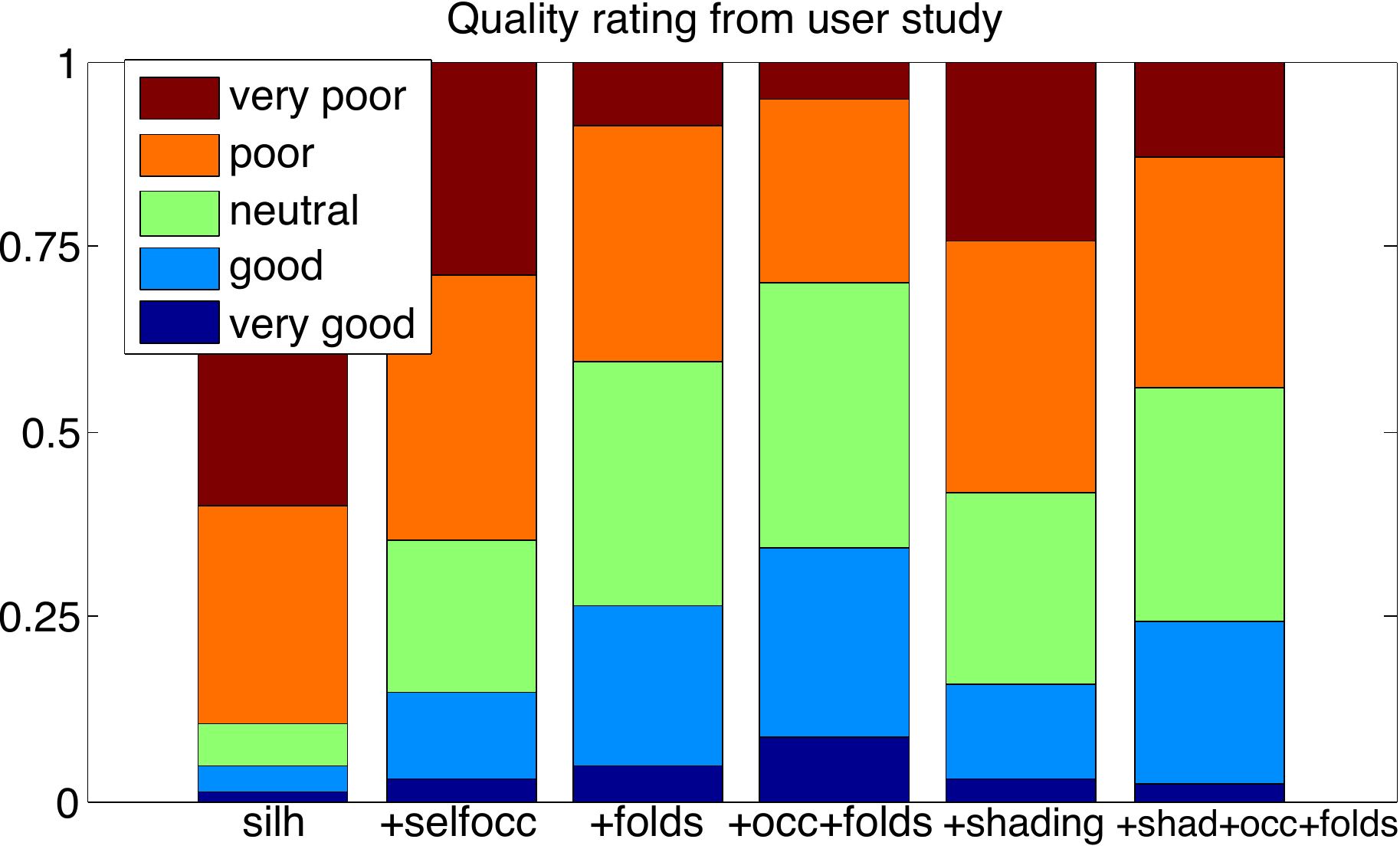}
\caption{For each algorithm, we show the percentage of times a certain rating was assigned to it during the qualitative rating user study. +occ+folds had the highest average rating, followed closely by +folds and +shading+occ+folds. Notice however that there is still much room for improvement, since the best-rated method (+occ+folds) was only chosen as ``good'' or ``very good'' less than 30\% of the time.}
\label{fig:summary}
\end{minipage}
\hfill
\begin{minipage}{0.475\linewidth}
\centering
\includegraphics[width=\linewidth]{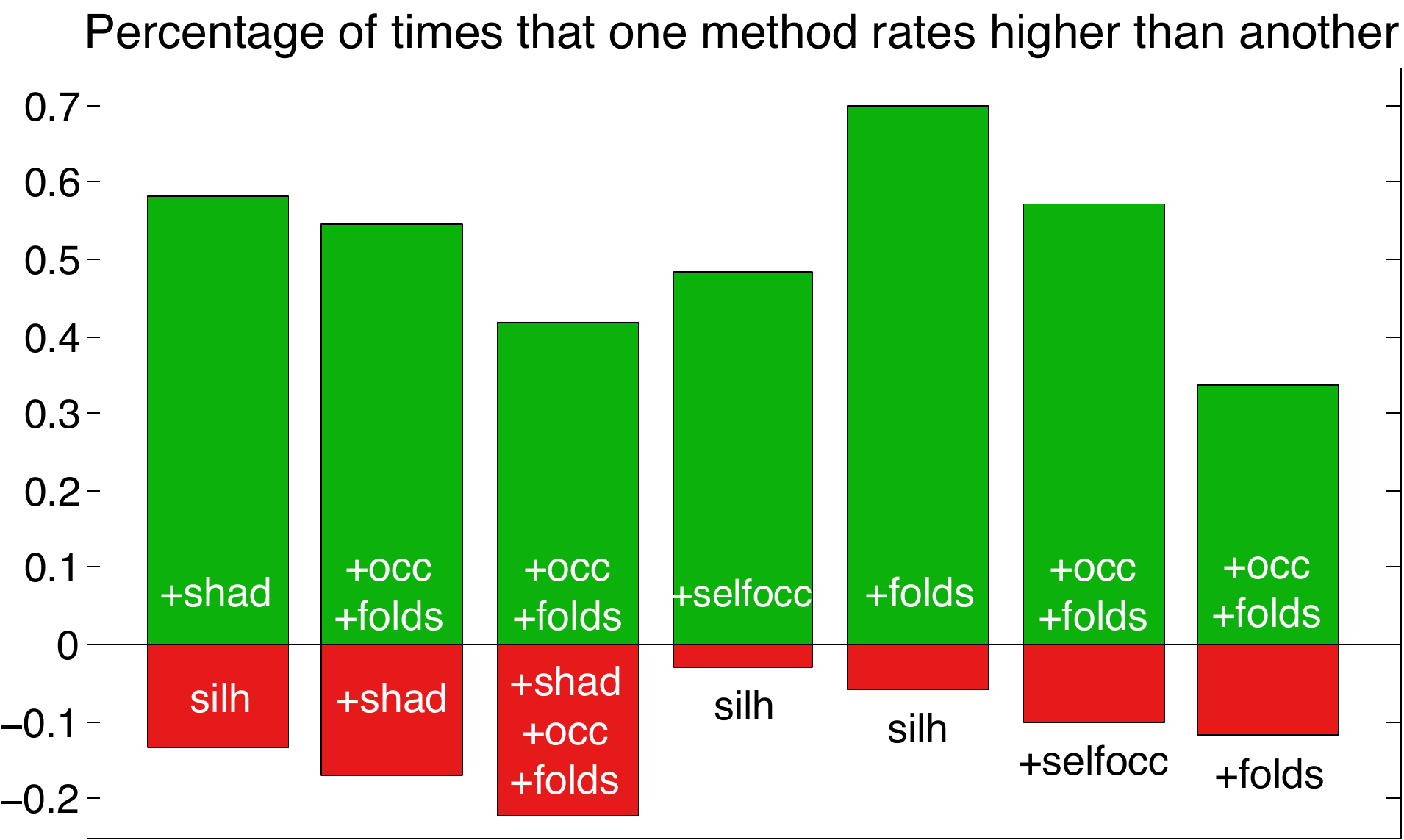}
\caption{The percentage that one algorithm rates higher (green color) or lower (red color) than another. Each column shows the result of one algorithm pair. For example, for the left most column, +shading was rated above silh approximately 60\% of the time, below silh about 10\% of the time, and rated the same otherwise. Shading seems to help when accompanied by with a silhouette cues, but when additional boundary cues are present, shading tends to produce more artifacts than improvements.  We also see a strong improvement from combining fold and occlusion contours.}
\label{fig:fraction}
\end{minipage}
\end{figure}

\subsubsection{Shape quality}
\label{sec:boundarycues:eval:quality}
Our experiments examine shape quality perceived by people (through a user study) and computers (ground truth comparison). The goal of these experiments is to find a common set of cues, or shape reconstruction algorithm(s), that consistently report the best shape. 

\boldhead{Qualitative rating on VOC}
The qualitative rating portion of the user study collected subjects' ratings for each of the six shape reconstruction algorithms. We designed an interface (Fig.~\ref{fig:interface1}) that displays the visualization of the six shape estimation results side by side on the screen and allows participants to rate the quality of each shape estimation result from scale 1 (very poor) to 5 (very good). The $17$ class $\times$ $10$ instance results are shuffled and divided into 5 groups. Each participant rated an entire group. Additional images from the study are displayed in Fig~\ref{fig:boundarycues:results}.

Our results indicate that +occ+folds is the most appealing reconstruction method to humans, followed closely by +folds and +shading+occ+folds. Figure~\ref{fig:summary} shows the averaged rating score grouped by algorithm; where a higher average rating indicates a better shape. In every case, as intuition suggests, adding more geometric cues leads to a more preferable shape. For an algorithm-by-algorithm comparison, we plot the percentage of times that one algorithm was rated higher than another (Figure~\ref{fig:fraction}). Here, we see geometric cues (other than silh) were consistently preferred over shading cues; in one example, +occ+folds was rated higher than +shading+occ+folds about 40\% of the time.





\boldhead{Ground truth comparison} 
Using ground truth shapes available from the MIT Intrinsic Image dataset~\cite{grosse09intrinsic}, we analyze our shape reconstructions using established errors metrics. We report results for both a surface normal-based error metric, $N$-MSE~\cite{Barron2012B}, as well as for a depth-based error metric, $Z$-MAE~\cite{Barron:cvpr:12}. $N$-MSE is computed as the mean squared error of the difference in normal orientation (measured in radians), and $Z$-MAE is the translation-invariant absolute error of of depth. Both metrics are averaged per-pixel, over the entire dataset of 20 objects. We also ran the same comparison, but substituted Barron and Malik's learned weights on the MIT dataset for our binary weights ($\delta_{{}^*}$); these results are in the $N$-MSE${}^\dagger$ and $Z$-MAE${}^\dagger$ columns:\\

\centerline{
\begin{tabular}{| l | c | c | c | c |} \hline
& \footnotesize $N$-MSE & \footnotesize $Z$-MAE & \footnotesize $N$-MSE${}^\dagger$ &\footnotesize $Z$-MAE${}^\dagger$ \\ \hline
{\footnotesize \bf silh} & 0.573 & 25.533 & 0.521 & 25.637 \\ \hline 
{\footnotesize \bf +selfocc} & 0.565 & 25.198 & 0.498 & 25.342 \\ \hline 
{\footnotesize \bf +folds} & 0.496 & 25.562 & 0.501 & 25.400 \\ \hline 
{\footnotesize \bf +occ+folds} & 0.487 & 25.161 & 0.482 & 24.983 \\ \hline 
{\footnotesize \bf +shading} & 0.874 & 38.968 & 0.310 & 25.793 \\ \hline 
{\footnotesize \bf +shading+occ+folds} & 0.574 & 27.379 & 0.350 & 24.492  \\ \hline
\end{tabular}
\vspace{4mm}
}

We observe that adding geometric cues generally increase quantitative performance. One notable exception is in the $N$-MSE${}^\dagger$ column, where +shading alone performs the best. This is almost certainly because all +shading reconstruction has been trained on the MIT dataset, whereas several parameters for the other algorithms have not been (e.g. $\delta_{\mathit{selfocc}}, \delta_{\mathit{fold}}$). Surprisingly, using binary weights (as in the $N$-MSE and $Z$-MAE columns) results in significantly worse +shading performance, but the geometric-based algorithms are largely unaffected. However, for non-MIT reconstructed shapes (e.g. VOC), using binary weights versus the learned weights weights gave perceptually similar results, possibly indicating that these metrics are sensitive to different criteria than human perception.

\begin{figure}[b]
\centering
\includegraphics[width=.495\linewidth]{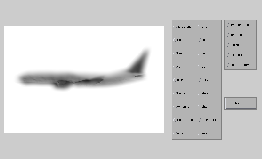}
\includegraphics[width=.495\linewidth]{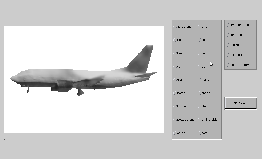}
\caption{User study interface for shape-based recognition. Participants are asked to recognize the object category using shape alone, estimated with one of the six algorithms. For each trial, a ``masked'' view (top) is displayed first to deter silhouette-based recognition, followed by the unmasked view (bottom).
\vspace{-2mm}
}
\label{fig:interface2}
\end{figure}

\subsubsection{Shape for object recognition}
\label{sec:boundarycues:eval:recog}

We are also interested in how well our shapes convey the object that has been reconstructed. Here, we describe experiments that gauge this task both through a human recognition study  and computer recognition algorithms. 

In the second task, we asked users to identify the object class based on the reconstructed shape alone. Our hypothesis is that higher class-recognition indicates better shape quality. We also consider that object silhouette could be a dominating factor for recognition; to reduce this factor, we show a  silhouette-masked view of each result first (Fig.~\ref{fig:interface2} left); and then show the result without masking (Fig.~\ref{fig:interface2} right). The task is evenly divided into 7 groups. Each group is assigned to one participant, who will go over all of the 170 objects in our test set. For each object, only one of the 7 results (generated by random permutation) is shown to one participant. The participants are also asked to rate their level of confidence on a scale from 1 (least confident) to 5 (most confident).

\begin{figure}[t]
\centering
\includegraphics[width=\linewidth]{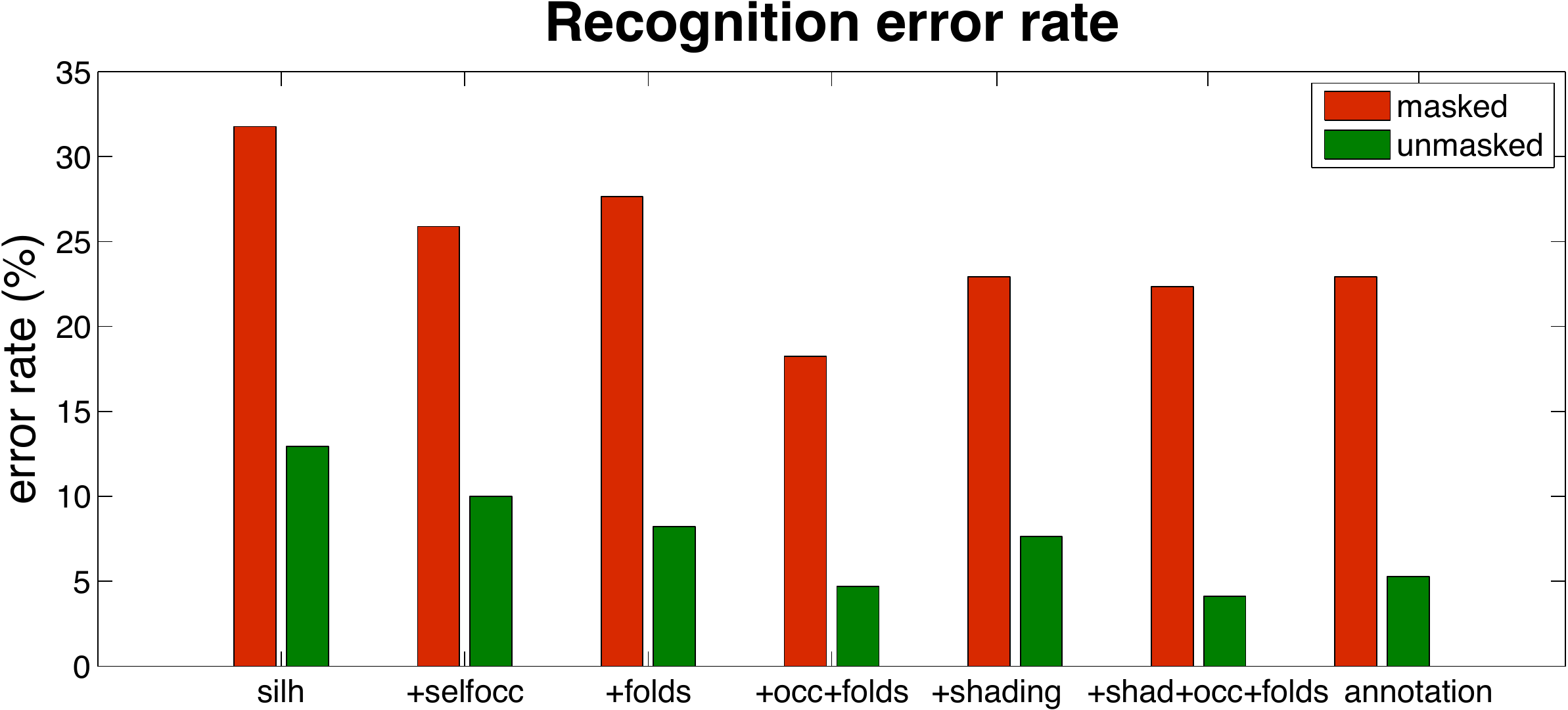}
\caption{Recognition error rate (as judged by participants in our user study) for each algorithm using the silhouette-masked and unmasked (unaltered) images.}
\label{fig:errorrate}
\end{figure}

Figure~\ref{fig:errorrate} displays the recognition error rate for each algorithm. For each algorithm, the left bar shows the result from the masked view; the right bar shows that result from the unmasked view. In the masked view, +occ+folds yields the lowest recognition error, consistent with qualitative rating portion of our user study. In the unmasked view, +shading+occ+folds performs the best, closely followed by +occ+folds.


\boldhead{Automatic recognition}
We evaluate the shapes by performing classification on the depth maps.  Outside the image, the depth is set to 0.  Since the heights inside objects are set to be fairly high, this ensures that there is a large edge at the contour.  To provide some invariance to specifics of classification methods or features, we run classification using two methods. We use a PHOW feature from~\cite{bosch2007image} and a the Pegasos SVM solver~\cite{shalev2007pegasos} with homogeneous kernel mapping~\cite{vedaldi2012efficient} as a baseline classifier (all available from VLFeat~\cite{vedaldi08vlfeat}).  It is motivated by a similar method in \cite{silberman11indoor} used to classify objects in Kinect images.  For another method, we use the RGB-D kernel match descriptors of \cite{bo2010kernel,bo2011depth} for which code is available.  Leave one out cross validation is used to determine the accuracy of classification on each reconstruction as well as {rgb}, {rgb+occ+folds}, and {rgb+shading+occ+folds} for the kernel matching method to determine if shape and shading cues add information compared to RGB alone.

Table~\ref{tab:recog} shows classifications results for each of the metrics. Our classification accuracy results are slightly different than the human ratings, although there are some similar trends.  {+occ+folds} still appears to be one of the best, though it is beaten in this case by {+selfocc}. Also expected, {+shading+occ+folds} outperforms {+shading}. The ordering of the remaining reconstructions is less consistent across the two classifiers, therefore it is difficult to draw any strong conclusions. It is interesting that {+folds} performed poorly but was rated highly by our test subjects; this likely implies that the features used do not make use of the information available from folds.

We also show the results of an RGB classifier using \cite{bo2011depth}. While state of the art classification on VOC2012 is roughly 70\%, we see only 55\% due to the constrained dataset (few examples per class).  The shape reconstructions increase the accuracy of the result, but this could be partially due to the mask provided by the height which is not available in the RGB only method.




\subsection{Conclusion}
We demonstrate a simple and extensible technique for reconstructing shape from images, resurrecting highly informative cues from early vision work. Our method itself is an extension of Barron and Malik's~\cite{Barron2012B} reconstruction framework, and we show how additional cues can be incorporated in this framework to create improved reconstructions. 

Through our experiments, we have shown the necessity of considering cues that  go beyond typical shape-from-shading constraints. In almost every task we assessed, using more geometric cues gives better results.  For human-based tasks, shading cues seem to help when applied with to silhouette cues (+shading consistently outperforms silh), but adds little information once additional boundary cues are incorporated (+occ+folds performs similarly to +shading+occ+folds); see Figs~\ref{fig:summary} and~\ref{fig:errorrate}. Interestingly, when the boundary is not available for viewing, +occ+folds performs better than +shading+occ+folds (Fig~\ref{fig:errorrate}; masked errors), and shading cues seem to have an adverse effect on automatic recognition algorithms (Table~\ref{tab:recog}). 
As far as we know, our experiments are the first to evaluate reconstruction methods on consumer photos (e.g. PASCAL VOC).

\begin{table}
\centering
\begin{tabular}{| l | c | c |} \hline
  & \small RGB-D kernel~\cite{bo2011depth} & \small VLFeat~\cite{vedaldi08vlfeat} \\ \hline
 \small \bf rgb & 55.29 & - \\ \hline
 \small \bf +occ+folds+rgb & \bf 70.00 & - \\ \hline
 \small \bf +shading+occ+folds+rgb & 62.35 & - \\ \hline
 \small \bf +shading & 48.24 & 41.76 \\ \hline
 \small \bf +shading+occ+folds & 52.94 & 42.94 \\ \hline
 \small \bf silh & 47.06 & 45.29 \\ \hline
 \small \bf +selfocc & 65.88 & \bf 54.12 \\ \hline
 \small \bf +folds & 51.76 & 47.65 \\ \hline
 \small \bf  +occ+folds & 65.88 & 51.76 \\ \hline
\end{tabular}
\vspace{0mm}
\caption{Average recognition accuracy for different sets of features using existing, automatic recognition methods. rgb implies that image appearance was used as a feature (row 1), and compared against rgb+depth (rows 2 and 3), as well as using depth alone (remaining rows). As one might expect, adding geometric features to the existing rgb information improves recognition accuracy, and shape tends to be more revealing than appearance alone. VLFeat offers only depth classification, hence the missing entries.
\vspace{-2mm}
}
\label{tab:recog}
\end{table}

One interesting observation from our experiments is that our shading cues tend to confound boundary cues; e.g. +occ+folds outperforms +shading+occ+folds in each task except (unmasked) human recognition (Sec ~\ref{sec:boundarycues:eval:recog}). It seems counterintuitive that incorporating shading information would degrade reconstructions, and we offer several possible causes. Foremost is the fact that we weight all terms equally, whereas learning these weights from ground truth will lead to better shading reconstructions (evidenced especially by our quantitative results on the MIT Intrinsic dataset in Sec~\ref{sec:boundarycues:eval:quality}). Second, this observation may be in part due to the inherent assumptions of existing shape-from-shading algorithms, including our own (e.g. 2.5D shape, orthographic camera, Lambertian reflectance, and smooth and infinitely distant illumination). Our tests use real  images from PASCAL, and some contain significant perspective, as well as complex reflectance and illumination. Relaxing these assumptions, as well as developing and enforcing stronger shape priors, are difficult but interesting problems for future research.

Our evaluations show that self occlusion and fold cues are undoubtedly helpful, and most importantly, point in many directions for improving existing shape reconstruction algorithms. Extracting boundary cues, such as folds and self occlusions, automatically from photographs is a logical next step. It is also evident that shape-from-shading algorithms can be improved by incorporating additional geometric cues, and additional research should go into extending shape-from-shading to real world (rather than lab) images. In terms of reconstructing shapes, considering perspective projections (rather than orthographic) may help, as well as extending surface representations beyond 2.5D and into 3D. By exploring these directions, we believe significant steps can be taken in the longstanding vision goal of reconstructing shape in the wild.

\section{Shading field decomposition for object relighting}

\subsection{Introduction}
An important task of image composition is to take an existing \emph{image fragment} and insert it into another scene. This task is appealing because 3D models are difficult to build, and image fragments carry real texture and material effects that achieve realism in a data-driven manner.

Relighting is generally necessary in the process. To relight the object, we need to know its shape and material properties. Image-based composition methods~\cite{BurtAdelson:1983,Perez:2003:poisson,Agarwala:2004:montage}, on the other hand, avoid the relighting process, totally relying on the artist's discretion for determining shading-compatible image fragments and limiting the range of data that can be used for a particular scene. Despite such limitations, image-based methods have been largely preferred to relighting-based methods, because shape estimation (for the latter) remains an extremely challenging task. State-of-the-art algorithms, such as the SIRFS method of Barron et al.~\cite{Barron2012B} still produce weak shapes and do not work well on complex materials. Is there a compromise between these two spaces that allows for improved image editing?

We propose such an approach by exploring an \emph{approximate shading model}. The model circumvents the formidable 3D reconstruction problem, yet is reshadable and allows a much wider range of objects to be inserted into a target scene.

There are good reasons to consider approximate shading models in image relighting. Evidence shows that human visual system (HVS) can tolerate certain degrees of shading inaccuracy~\cite{Ostrovsky:2005,Sinha:2006:face,Conway:2007:PSA}. Psychologists describe this phenomenon with the term \emph{alternative physics}~\cite{Cavanagh:2005}, explaining that the brain employs a set of rules that are not strict physics when interpreting a scene from retina image. When these rules are violated, a perception alarm is fired, or recognition is negatively effected~\cite{Tarr:1999}. Otherwise, visual plausibility is maintained without having to adhere strictly to physical correctness. Our model tries to exploit the inherent ambiguity of HVS based on this line of reasoning. Methods bearing the same spirit have been found in material editing~\cite{Khan:2006:IME:1179352.1141937} and illumination estimation~\cite{Karsch:SA2011}. Second, the illumination cone theory~\cite{Belhumeur:1998:illcone} states that shading of a Lambertian surface lies close to a very low dimensional space. A 9-Spherical Harmonics (SH) illumination can account for up to 98\% of shading variation~\cite{Basri:2003}. This indicates that shading can be expressed as linear combination of a very few number of components.

\begin{figure}[t]
  \centering
  \includegraphics[width=0.9\linewidth]{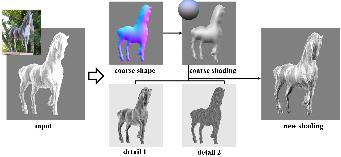}\\
  \caption{
  Given a shading image estimated from a single image of an object (left), our approximate shading model (middle) can reshade the object under new illumination and produce a new shading on the right. Shape/detail images are rescaled for visualization.
  }\label{fig:shadingdecomp:illustration}
\end{figure}

Based on these assumptions, we propose an approximate shading model as follows: shading is decomposed into 3 components (basis): a smooth component captured by a coarse shape $h$, and two shading detail layers: parametric residual $S_p$ and geometric detail $S_g$. A new shading is expressed as the coarse shading plus a weighted combination of the two detail layers:
\begin{equation}\label{eq:model}
  S(h, S_p, S_g) = shade(h,L), + w_pS_p + w_gS_g
\end{equation}
where $L$ is (new) illumination, $w_p$ and $w_g$ are scalar weights. See illustration in Fig.~\ref{fig:shadingdecomp:illustration}.

The coarse shape produces a smoothing shading that captures directional and coarse-scale illumination effects that are critical for perceptual consistency. The shape is purely based on contour constraint (e.g. surface normals at the silhouette are perpendicular to the viewing direction), easy to construct and robust to moderate perturbation of view direction. The two ``detail'' layers ($S_p$ and $S_g$) account for visual complexity of the object. They encode the middle and high frequencies of the shading signal left out by the smooth shading component. While image-based composition of the detail layers (equation~\ref{eq:model}) is not physically-based, in practice it yields surprisingly good results for a variety of object and material types. With this model, we implement an image relighting system that supports object insertion with little user input. Figure~\ref{fig:shadingdecomp:illustration} shows our model of pipeline.

To evaluate the model, we compare our relighting results with a state-of-the-art shape reconstruction model~\cite{Barron2012B} in two tasks: (a) re-rendering MSE on the MIT intrinsic image dataset, and (b) user study in which we ask subjects on Amazon Mechanical Turk to rate the relative realism of results by both methods as well as against real scenes. In the first task, our method yields slightly lower MSE on the MIT Lab illumination dataset. The user study is more compelling as qualitative realism is our primary focus in image composition. The study showed that subjects preferred our insertions over that of Barron and Malik by a margin of 20\%, and over that of Karsch et al.~\cite{Karsch:SA2011} by a margin of 14\%.


\subsection{Related work}\label{sec:relatedwork}
{\bf Object insertion}
takes an object from one source and sticks it into a target scene. Pure image-based methods, like pyramid blending~\cite{BurtAdelson:1983}, Poisson editing~\cite{Perez:2003:poisson} and graph cut-based method~\cite{Agarwala:2004:montage}, totally rely on artist's discretion for shading consistency. \cite{lalonde2007,Chen:2009:sketch2photo} take a data driven approach, searching in a large database for compatible source. A relighting procedure would significantly expand the range of images to composite with. Khan et al.~\cite{Khan:2006:IME:1179352.1141937} show a straightforward method that simulate changes in the apparent material of objects, given an approximate normal field and environment map. Recently, Karsch et al.~\cite{Karsch:SA2011} introduce a technique that reconstructs a 3D scene from a single indoor image and allows synthetic objects to be inserted in. Our object insertion system utilizes this technique for scene modeling. The major advance is that we insert objects from images instead of existing 3D models.

{\bf Shape estimation:}
Current methods are still unable to recover accurate shape from a single image, even inferring satisfactory approximate shape is difficult. Methods that recover \emph{shape from shading}(SfS) are unstable as well as inaccurate, particularly in the absence of reliable albedo and illumination~\cite{Zhang:1999:survey,Durou:2008:survey,Forsyth:2011:VSSA}. A more sophisticated approach is to jointly estimate shape, illumination and albedo from a single image~\cite{Barron2012B}. An alternative is to recover \emph{shape from contour} cues alone, for example, by building a surface that is smooth and compelled to have the normal constraints along the contour~\cite{Johnston:2002:lumo,Prasad:2006:SVR}. Yet another alternative is to assume that lighter pixels are nearer and darker pixels are further away~\cite{Khan:2006:IME:1179352.1141937}.

For re-rendering and other predictive purposes, an alternative to a shape estimate would be {\bf illumination cone} (a representation of all images of an object in a fixed configuration, as lighting changes). This cone is known to be convex~\cite{Belhumeur:1998:illcone} and to lie close to a low dimensional space~\cite{Basri:2003}, suggesting that quite low-dimensional image based reshading methods are available. Standard methods to variation in appearance caused by lighting estimate a low dimensional representation of this cone (e.g. for face recognition~\cite{Georg:2001}).  Our representation could be seen as a hybrid of a shape and an illumination cone representation.

{\bf Material and Illumination}
Our method needs to decompose an input image into albedo and shading ({\em intrinsic images}~\cite{BarTen78}) as a first step.
The standard methods for estimating intrinsic images assume sharp changes come from albedo, and slow changes come from shading.  The Retinex method~\cite{Land71} is derived from this assumption. Recent work by Grosse {\em et al.}~\cite{grosse09intrinsic} demonstrates that the color variant of Retinex is state-of-the-art for single-image decomposition methods. 
We use a variant of the color Retinex algorithm for our albedo and shading estimation. Liao et al.~\cite{Liao:cvpr13a} recently propose a method that decomposes an image into albedo, smooth shading, and shading detail caused by high frequency surface geometry. We use this method to derive our geometric detail layer.


\subsection{Creating the model}\label{sec:decomposition}
Our object model has four components.  We compute a coarse 3D shape estimate, then compute three maps: the albedo, a parametric shading residual, and a geometric detail layer. We refer to the ``coarse shading'' by the shape, the ``parametric shading residual'' and the ``geometric detail'' as the three shading components.

\begin{figure}
\centering
\subfigure[normal]{\includegraphics[height=1in]{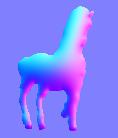}}\hspace{.5mm}
\subfigure[3D shape]{\includegraphics[height=1in]{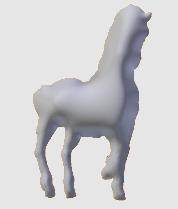}}\hspace{.5mm}
\subfigure[view2]{\includegraphics[height=1in]{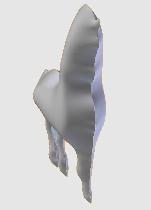}}\hspace{.5mm}
\subfigure[view3]{\includegraphics[height=1in]{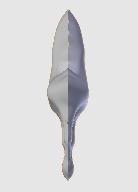}}
\caption{Normal field and shape reconstruction. Our reconstructions are simple but typically robust to large errors that may manifest in state-of-the-art SfS algorithms (see Fig.~\ref{fig:shadingdecomp:cube}). This benefit is key to our goal of image fragment insertion.}\label{fig:shadingdecomp:shape}
\vspace{-3mm}
\end{figure}

\subsubsection{Coarse shape}\label{subsec:coarseshading}
We assume the object to be inserted is an image fragment, and wish to estimate what its appearance is under new illumination. Exact shape is ideal but unavailable. We need a representation capable of handling extreme shading effects.  For example, a vertical cylinder with light from the left will be light on left, dark on right.  Moving the light to the right will cause it to become dark on left, light on right. We also want our reconstruction to be consistent with a generic view assumption. This implies that (a) the outline should not shift too much if the view shifts, and (b) there should not be large bumps in the shape that are concealed by the view direction (Fig.~\ref{fig:shadingdecomp:cube} demonstrates these kinds of mistakes typically generated by more complicated SfS methods). To support these, we use a simple shape from contour method with stable outline and smooth surface (Fig.~\ref{fig:shadingdecomp:shape}).

First, we create a normal field by constraining normals on the object boundary to be perpendicular to the view direction, and interpolating them from the boundary to the interior region, similar to Johnston's Lumo technique~\cite{Johnston:2002:lumo}. Let $N$ be the normal field, $\Omega$ and $\partial\Omega$ be the set of pixels in the object and on boundary, respectively, and $N_{\bot}^{i}$ be the tangent of mask boundary at pixel $i$. We compute $N$ by the following optimization:
\begin{equation}\label{eq:normal}
\begin{aligned}
& \underset{N}{\text{min}} & & \sum_{\Omega} {||\nabla {\bf N}||^2 + (||{\bf N}||-1)^2} \\
& \text{subject to} & & N_{z}^{i} = 0 
\hspace{3mm} \forall i \in \partial\Omega
\end{aligned}
\end{equation}

We then reconstruct a height field from the normal. Reconstructing an exact shape with vertical boundary is tricky; Wu et al.~\cite{Wu:2008:SFS} describes a method for it. Instead, we reconstruct an approximate height field $h$ by minimizing:
\begin{equation}
\sum_{\Omega}{ ||(\frac{\partial h}{\partial x} - \frac{N_{x}}{max(\epsilon, N_z)})||^2 +
||(\frac{\partial h}{\partial y} - \frac{N_{y}}{max(\epsilon, N_z)})||^2}
\end{equation}
subject to $h_{i}=0$ for boundary pixels (stable outline). The reconstructed height field is flipped to make a symmetric full 3D shape (Fig.~\ref{fig:shadingdecomp:shape}). The threshold $\epsilon = 0.1$ avoids numerical issues near the boundary and forces the reconstructed object to have a crease along its boundary. This crease is very useful for the support of generic view direction, as it allows slight change of view direction without exposing the back of the object and causing self-occlusion.

\begin{figure}
\centering
\includegraphics[width=0.6\textwidth]{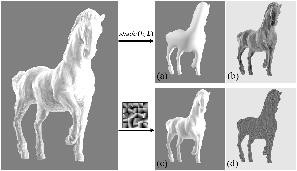}
\caption{From a shading image (left), the upper right row shows the parametric fitting procedure to compute the best fit shading (a) from the shape and the parametric shading residual (b); the bottom right row shows the non-parametric patch-based filtering procedure to compute the filtered shading image (c) and the residual known as geometric detail (d).}
\label{fig:shadingdecomp:detailmaps}
\vspace{-3mm}
\end{figure}

\vspace{-3mm}
\subsubsection{Albedo and Parametric shading residual}\label{subsec:parresidual}
The coarse shape can recover gross changes in shading caused by lighting. However, it cannot represent finer detail. We use shading detail maps to represent this detail. We define the shading detail maps as a representation of the residual incurred by shading the coarse shape with some model. We use two shading details in our model: \emph{parametric shading residual} ($S_p$ in equation~\ref{eq:model}) that encodes object level features (silhouettes, crease and folds, etc.), and \emph{geometric detail} ($S_g$ in equation~\ref{eq:model}) that encodes short scale effects. We refer to the former as \emph{detail 1} and the latter as \emph{detail 2} throughout this chapter.

First, we use a standard color Retinex algorithm~\cite{grosse09intrinsic} to get an initial albedo $\rho$ and shading $S$ estimates from the input image: $I = \rho \cdot S$. We then use a parametric illumination model $L(\theta)$ to shade the estimated shape model and compute the shading residual by solving:
\begin{equation}\label{eq:argminL}
\hat{\theta}: \underset{\theta}{\operatorname{argmin}} \sum{||S - \mbox{\emph{shade}}(h, L(\theta))||^2}
\end{equation}
The optimized illumination $\hat{\theta}$ is substituted to obtain the parametric shading residual:
\begin{equation}\label{eq:computePS}
S_p = S- \mbox{Shade}(h, L(\hat{\theta})).
\end{equation}
Many parametric illuminations are possible (i.e., spherical harmonics). We used a mixture of 5 point sources, the parameters being the position and intensity of each source, forming a 20-dimensional representation.

Figure~\ref{fig:shadingdecomp:detailmaps} top shows an example of the best fit coarse shading and the resultant parametric shading detail. Note that the directional shading is effectively removed, leaving shading cues of object level features.

\subsubsection{Geometric detail}\label{subsec:geometricdetail}
The parametric shading residual is computed by a \emph{global} shape and illumination parameterization, and contains all the shading details missed by the shape. Now we wish a compute another layer that contains only fine-scale details. We borrow a technique from Liao et al.~\cite{Liao:cvpr13a}, in which they extract very fine-scale geometric details with a \emph{local} patch-based non-parametric filter. The resultant geometric detail represents high frequency shading signal caused by local surface geometry like bumps and grooves and is insensitive to gross shading and higher-level object features such as silhouettes (see the difference to the parametric shading residual in Fig.~\ref{fig:shadingdecomp:detailmaps}).

The filtering procedure uses a set of shading patches learned from smooth shading images to reconstruct an input shading image. Because geometric detail signals are poorly encoded by the smooth shading dictionary, they are effectively left out. See Liao et al.~\cite{Liao:cvpr13a} for more details. In the experiment we use dictionary size of 500 with patch size $12 \times 12$.

\vspace{5mm}
\begin{figure}[h]
\centering
\includegraphics[width=\textwidth]{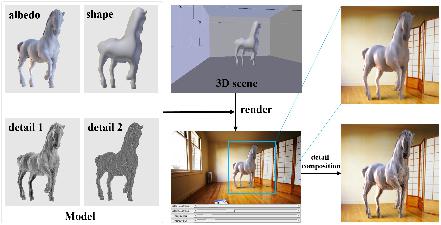}
\caption{Given the object model, an artist places the object into a 3D scene, render it with a physically-based renderer, and then composite it with the detail layers to generate the final result. Notice the difference on the horse before and after the detail composition.}
\label{fig:shadingdecomp:rendercomposite}
\end{figure}
\subsection{A relighting system}\label{sec:relighting}
With the object model, we develop a system that relights an object from image into a new scene. The system combines interactive scene modeling, physically-based rendering and image-based detail composition (Fig.~\ref{fig:shadingdecomp:rendercomposite}).

\begin{figure}[h]
\begin{minipage}{0.32\linewidth}
\centering{\bf Inset} \\
\includegraphics[width=\linewidth]{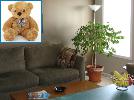}\\\vspace{0.5mm}
\includegraphics[width=\linewidth]{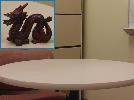}
\end{minipage}
\begin{minipage}{0.32\linewidth}
\centering{\bf Barron and Malik} \\
\includegraphics[width=\linewidth]{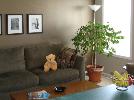}\\\vspace{0.5mm}
\includegraphics[width=\linewidth,clip,trim=100 0 168 201]{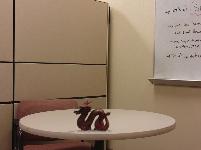}
\end{minipage}
\begin{minipage}{0.32\linewidth}
\centering{\bf Ours} \\
\includegraphics[width=\linewidth]{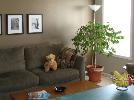}\\\vspace{0.5mm}
\includegraphics[width=\linewidth,clip,trim=100 0 168 201]{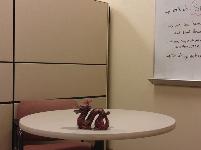}
\end{minipage}
\vspace{0mm}
\caption{
Our relighting system adjusts the shading on the object for a variety of scenes with different illumination conditions. Detail composition simulates complex surface geometry and materials properties that is difficult to achieve by physically-based modeling. Best viewed in color at high-resolution.
}
\label{fig:shadingdecomp:result1}
\end{figure}

\vspace{-2mm}
\subsubsection{Modeling and Rendering}
We use the technique from Xia et al.~\cite{Xia:2009} to build a sparse mesh object with proper boundary conditions from the height field. The target scene can be existing 3D scenes, or built from an image (Karsch et al.~\cite{Karsch:SA2011}, Hedau et al.~\cite{hedau2009iccv}, etc.). The artist then selects an object and places it into the scene, adjusting its scale and orientation, and making sure the view is roughly the same as that of the object in the original image. The model is then rendered with the estimated albedo. For all the results in this chapter, we use Blender (http://blender.org) for modeling and LuxRender (http://luxrender.net) for rendering.

Our shape model assumes an orthographic camera. However, most rendering systems use a perspective camera. This will cause texture distortion. We use a simple ``easing'' method to avoid it. Besides, the flipped shape model is thin along the base and can cause light leaks and/or skinny lateral shadows. We created a simple user-controllable extrusion procedure to handle such cases; refer to supplemental material for details.

\begin{figure}[!th]
\begin{minipage}{0.32\linewidth}
\centering{\bf No detail} \\
\includegraphics[width=\linewidth]{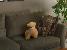}\\\vspace{0.5mm}
\includegraphics[width=\linewidth]{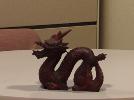}
\end{minipage}
\begin{minipage}{0.32\linewidth}
\centering{\bf Detail layer 1} \\
\includegraphics[width=\linewidth]{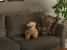}\\\vspace{0.5mm}
\includegraphics[width=\linewidth]{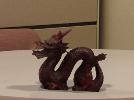}
\end{minipage}
\begin{minipage}{0.32\linewidth}
\centering{\bf Detail layer 1 + 2} \\
\includegraphics[width=\linewidth]{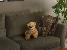}\\\vspace{0.5mm}
\includegraphics[width=\linewidth]{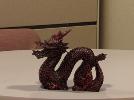}
\end{minipage}
\vspace{-2mm}
\caption{
Relighting and detail composition. The left column displays relighting results with our coarse shape model and estimated albedo. The middle column displays results compositing with only the parametric shading residual. Notice how this component adds object level shading cues and improves realism of perception. The right column are results compositing with both detail layers. Fine-scale surface detail is further enhanced (see the dragon). Best viewed in color at high-resolution.
}
\label{fig:shadingdecomp:detailedit}
\end{figure}

\vspace{-2mm}
\subsubsection{Detail composition}\label{subsec:compositing}
We then composite the rendered scene with the two detail maps and original scene to produce final result. See Fig.~\ref{fig:shadingdecomp:result1} for examples (more in supplemental material).

First, we composite the two shading detail images with the shading field of the rendered image (similar to the material editing technique by Liao et al.~\cite{Liao:cvpr13a})
\begin{equation}
\label{eq:detail_composite}
C = \rho(S + w_pS_p + w_gS_g)
\end{equation}
where $S=R/\rho$ is the shading field, $R$ is the rendered image. The weights $w_p$ and $w_g$ can be automatically determined by regression (section~\ref{subsec:compare1}) or manually adjusted by artist with a slider control. Compositing the two details significantly improves the visual realism of object and adds flexibility to the rendered image (Fig.~\ref{fig:shadingdecomp:detailedit}).

Second, we use standard techniques (e.g.~\cite{Debevec:98,Karsch:SA2011}) to composite $C$ with the original image of the target scene. This produces the final result. Write $I$ for the target image, $E$ for the empty rendered scene {\it without} the inserted object, and $M$ for the object matte (0 where no object is present, and $(0,1]$ otherwise). The final composite image $C$ is obtained by:
\begin{equation}
\label{eq:shadingdecomp:composite}
C_{\text{final}} = M \odot C + (1-M) \odot (I + R - E).
\end{equation}

\subsection{Evaluation}
Our assumption is that the shading decomposition model can capture major effects of illumination change of an object. To evaluate this, we compare our representation with state-of-the-art shape reconstructions by Barron and Malik~\cite{Barron2012B} on a re-rendering metric (Sec.~\ref{subsec:compare1}). We also conducted a user study to evaluate the realism of our relighting results (Sec.~\ref{subsec:userstudy}).

\subsubsection{Re-rendering Error}\label{subsec:compare1}
The re-rendering metric measures the error of relighting an estimated shape. On a canonical shape representation (a depth field), the metric is defined as
\begin{equation}\label{MSEmetric2}
\mbox{IMSE}_{rerender} = \frac{1}{n}||I - k\hat{\rho} \mbox{ReShade}(\hat{h}, L)||^2
\end{equation}
where $\hat{\rho}$ and $\hat{h}$ are estimated albedo and shape, $I$ is the re-rendering with the ground truth shape $h^*$ and albedo $\rho^*$: $I = \rho^*\mbox{ReShade}(h^*,L)$, $n$ is the number of pixels, $k$ is a scaling factor.

With our model, 
write $S_c = shade(h,L), S_p, S_g$ for the coarse shading, parametric shading detail and the geometric detail, respectively, and rewrite Equation~\ref{eq:model} as $\mbox{ReShade}(S(L), w) = S_c + w_pS_p + w_gS_g$ for some choice of weight vector $w=(1,w_p,w_g)$. The re-rendering metric is:
\begin{equation}\label{eq:MSEmetric}
\mbox{IMSE}^{'}_{rerender} = \frac{1}{n}||I - k\hat{\rho} \mbox{ReShade}(S(L),w)||^2
\end{equation}

\begin{table*}[h]
\begin{center}
\begin{tabular}{|l|cc|cc|}
\hline
Method &
\multicolumn{2}{|c|}{\begin{tabular}[x]{@{}c@{}}``Natural'' Illumination\\ \scriptsize{(No strong shadows)}\end{tabular}} &
\multicolumn{2}{|c|}{\begin{tabular}[x]{@{}c@{}}Lab Illumination\\ \scriptsize{(With strong shadows)}\end{tabular}}\\
\hline\hline
{\em Barron 2012}                     & \multicolumn{2}{|c|}{0.0172}      & \multicolumn{2}{|c|}{0.0372} \\
\hline
\hline
 {\em Ours} & {\em LSQ}  & {\em Regression}  & {\em LSQ}  & {\em Regression} \\

{(a) default}                & 0.0329    & 0.0358 & 0.0586   & 0.0641 \\
{(b) Barron \& Malik $S$}        & 0.0274    & 0.0320 & 0.0341   & 0.0360 \\
{(c) GT $S$}                   & 0.0206    & 0.0243 & 0.0228   & 0.0240 \\
{(d) GT $S \& L$}              & 0.0149    & 0.0219 &  NA  & NA \\
\hline
\end{tabular}
\end{center}
\caption{
Re-rendering error of our method compared to Barron \& Malik~\protect{\cite{Barron2012B}}. Our automatic weights (regression) can generate slightly lower MSE on the real Lab illumination dataset.}
\vspace{-5mm}
\label{tab:mse}
\end{table*}

\begin{figure}
\centering{
\begin{overpic}[width=.3\linewidth,clip,trim=36 50 100 37]{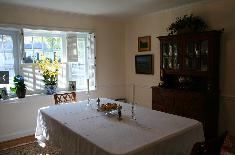}\put(88,2){\colorbox{white}{\small A1}}
\end{overpic}
\begin{overpic}[width=.3\linewidth]{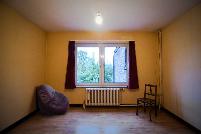}\put(88,2){\colorbox{white}{\small B1}}
\end{overpic}
\begin{overpic}[width=.3\linewidth,clip,trim=0 0 0 33]{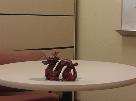}\put(88,2){\colorbox{white}{\small C1}}
\end{overpic}
}

\centering{
\begin{overpic}[width=.3\linewidth,clip,trim=0 0 0 0]{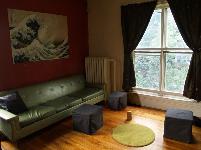}\put(88,2){\colorbox{white}{\small A2}}
\end{overpic}
\begin{overpic}[width=.3\linewidth]{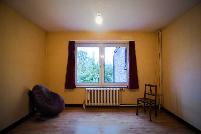}\put(88,2){\colorbox{white}{\small B2}}
\end{overpic}
\begin{overpic}[width=.3\linewidth,clip,trim=0 0 0 0]{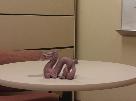}\put(88,2){\colorbox{white}{\small C1}}
\end{overpic}
}
\caption{Example trial pairs from our user study. The left column shows an insertion result by our method and a real image (task 1). The middle column shows that from our method and the method of Barron and Malik (task 3). And the right column shows that from out method and the method of Karsch et al.~\protect{\cite{Karsch:SA2011}}. Users were instructed to choose the picture from the pair that looked the most realistic. For each row, which image would you choose? Best viewed in color at high-resolution.
\protect\begin{turn}{180} \centerline{\small \it A2,B1,C1: our results (ottoman inserted near window in A2); A1: real image; B2: Barron and Malik; C3: Karsch et al.}\protect\end{turn}}
\label{fig:shadingdecomp:studypics}
\end{figure}

We offer three methods to select $w$. An {\em oracle} could determine the values by least square fitting that leads to best MSE. {\em Regression} could offer a value based on past experience. We learn a simple linear regression model to predict the weights from illumination. Lastly, an artist could {\em manually} choose the weights, as demonstrated in our relighting system (Sec.~\ref{subsec:compositing}).


\vspace{3mm}
\noindent{\bf Experiment}
We run the evaluation on the augmented MIT Intrinsic image dataset~\cite{Barron2012B}.
To generate the target images, we re-render each of the 20 object by 20 randomized monochrome ($9\times1)$ SH illuminations, forming a $20\times20$ image set.
We then measure the re-rendering error using our representation and the shape estimation by Barron and Malik. For our method, we compare models built from the Natural Illumination dataset and Lab Illumination dataset separately. The models are built (a) in the default setting, (b) using Barron and Malik's shading and albedo estimation, (c) using the ground truth shading, and (d) using both ground truth shading and illumination. See Table~\ref{tab:mse} for the results. To learn the regression model, for each object we draw 100 nearest neighbors (in terms of Illumination) from the other 380 data points (leave-1-out scheme), and fit a linear model to their LSQ coefficients.

The result shows that when the shape estimation is accurate (on the ``Natural'' Illumination dataset, a synthetic dataset by the same shading model used in their optimization), our approximate shading performs less as well. This is reasonable, because a perfect shape is supposed to produce zero error in the re-rendering metric. However, when the shape estimation is inaccurate (on the ``Lab'' Illumination dataset, real images with strong shadows taken in lab environment), our approximate shading model can produce lower error with both regressed weights and oracle's setting. With better detail layers (when ground truth shading is used to derive them), our model achieves significantly lower errors, indicating space of improvement with a better intrinsic image decomposition algorithm or alternative detail layer definitions.

\begin{figure}[h]
\includegraphics[width=\linewidth]{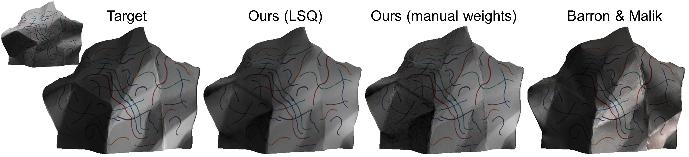}
\caption{
Left: target shading 
 (original image in upper left); Middle left: our reshading by LSQ fitting. Middle right: our reshading by user adjusted weight (for the geometric detail); Right: reshading by shape estimation from~\protect{\cite{Barron2012B}}. Notice the user adjusted weight makes a more realistic result, though not in the measure of least MSE.
}\label{fig:shadingdecomp:MSEillustration}
\end{figure}

It is worth noting that MSE is not geared toward visual realism (image features takes little weight; non-linearity of visual perception on light intensity, etc.). As a result, the shading images fit by LSQ or regression do not always emphasize the shading details as much as we expect (Fig.~\ref{fig:shadingdecomp:MSEillustration}). 
To demonstrate the real potential of our model in an interactive object insertion setting, 
we employed a user study to evaluate the ``realism'' achieved by our insertion technique.

\vspace{3mm}
\subsubsection{User study}\label{subsec:userstudy}
In the study, each subject is shown series of two-alternative forced choice tests, where the subject chooses between a pair of images which he/she feels the most realistic. We tested four different tasks: (1) our method against real images, (2) the method of Barron and Malik against real images, (3) our method against Barron and Malik, and (4) our method against Karsch et al.~\cite{Karsch:SA2011}. Figure~\ref{fig:shadingdecomp:studypics} shows example trials from the first and third tasks.

\vspace{2mm}
\noindent{\bf Experiment setup} For each task, we created 10 different insertion results using a particular method (ours, Barron and Malik, or Karsch et al.). For the results of Barron and Malik, we ensuring the same object was inserted at roughly the same location as our results. This is not the case for the results of Karsch et al., as synthetic models are not all available for the objects we modeled from image for the other two methods.
We also collected 10 real scenes (similar to the ones with insertion) for the tasks involving real images. Each subject viewed all 10 pairs of images for one but only one of the three tasks. 
For the 10 results by our method, the detail layer weights were manually selected (it is hard to apply the regression model as in Section~\ref{subsec:compare1} to the real scene illuminations) while the other two methods do not have such options. 

\begin{table}[!t]
\centerline{\bf Fraction of times subjects chose an insertion result over a real image in the study}
\vspace{1mm}
\centerline{
\begin{tabular}{| l c || c | c |} \hline
Subpopulation & \# of trials & \hspace{10mm} ours \hspace{10mm} & \hspace{5mm} Barron and Malik~\cite{Barron2012B} \hspace{5mm} \\ \hline
all & 1040 & \bf 0.440$\pm$0.015 &  0.418$\pm$0.016\\ \hline
expert & 200 & \bf 0.435$\pm$0.034 &  0.362$\pm$0.037\\
non-expert & 840 & \bf 0.442$\pm$0.017 &  0.429$\pm$0.018\\ \hline
male & 680 & \bf 0.447$\pm$0.019 &  0.426$\pm$0.018\\
female & 360 & \bf 0.428$\pm$0.024 &  0.390$\pm$0.035\\ \hline
age ($\leq$ 25) & 380 & \bf 0.432$\pm$0.025 &  0.417$\pm$0.025\\
age ($>$25) & 660 & \bf 0.445$\pm$0.019 &  0.419$\pm$0.020\\ \hline
passed p-s tests & 740 & \bf 0.442$\pm$0.018 &  0.405$\pm$0.021\\
failed p-s tests & 300 &  0.437$\pm$0.027 & \bf 0.439$\pm$0.025\\ \hline
first half & 520 & \bf 0.456$\pm$0.022 &  0.430$\pm$0.023\\
second half & 520 & \bf 0.425$\pm$0.020 &  0.418$\pm$0.021\\ \hline
\end{tabular}}
\vspace{2mm}
\label{tab:study_vs_real}
\caption{Overall, users confused our insertion results with real pictures 44\% of the time, while confusing the results of Barron and Malik with real images 42\% of the time. Interestingly, for the subpopulation of ``expert'' subjects, this difference became more pronounced (44\% vs 36\%). Each cell shows the mean standard deviation.
}
\vspace{-5mm}
\end{table}

We polled 300 subjects using Mechanical Turk. In an attempt to avoid inattentive subjects, each task also included four ``qualification'' image pairs (a cartoon picture next to a real image). Subjects who incorrectly chose any of the four cartoon picture as realistic were removed from our findings (6 in total, leaving 294 studies with usable data). At the end of the study, we showed subjects two additional image pairs: a pair containing rendered spheres (one a physically plausible, the other not), and a pair containing line drawings of a scene (one with proper vanishing point perspective, the other not). For each pair, subjects chose the image they felt looked most realistic. Then, each subject completed a brief questionnaire, listing demographics, expertise, and voluntary comments.

These answers allowed us to separate subjects into subpopulations: {\bf male/female}, {\bf age $<25$ / $\geq 25$}, whether or not the subject correctly identified both the physically accurate sphere {\it and} the proper-perspective line drawing at the end of the study ({\bf passed/failed perspective-shading (p-s) tests}), and also {\bf expert/non-expert} (subjects were classified as experts only if they passed the perspective-shading tests {\it and} indicated that they had expertise in art/graphics). We also attempted to quantify any learning effects by grouping responses into the {\bf first half} (first five images shown to a subject) and the {\bf second half} (last five images shown). We would like to make the userstudy image set and the collected data publicly available.

\begin{figure}[t]
\centering{
\includegraphics[width=0.75\linewidth]{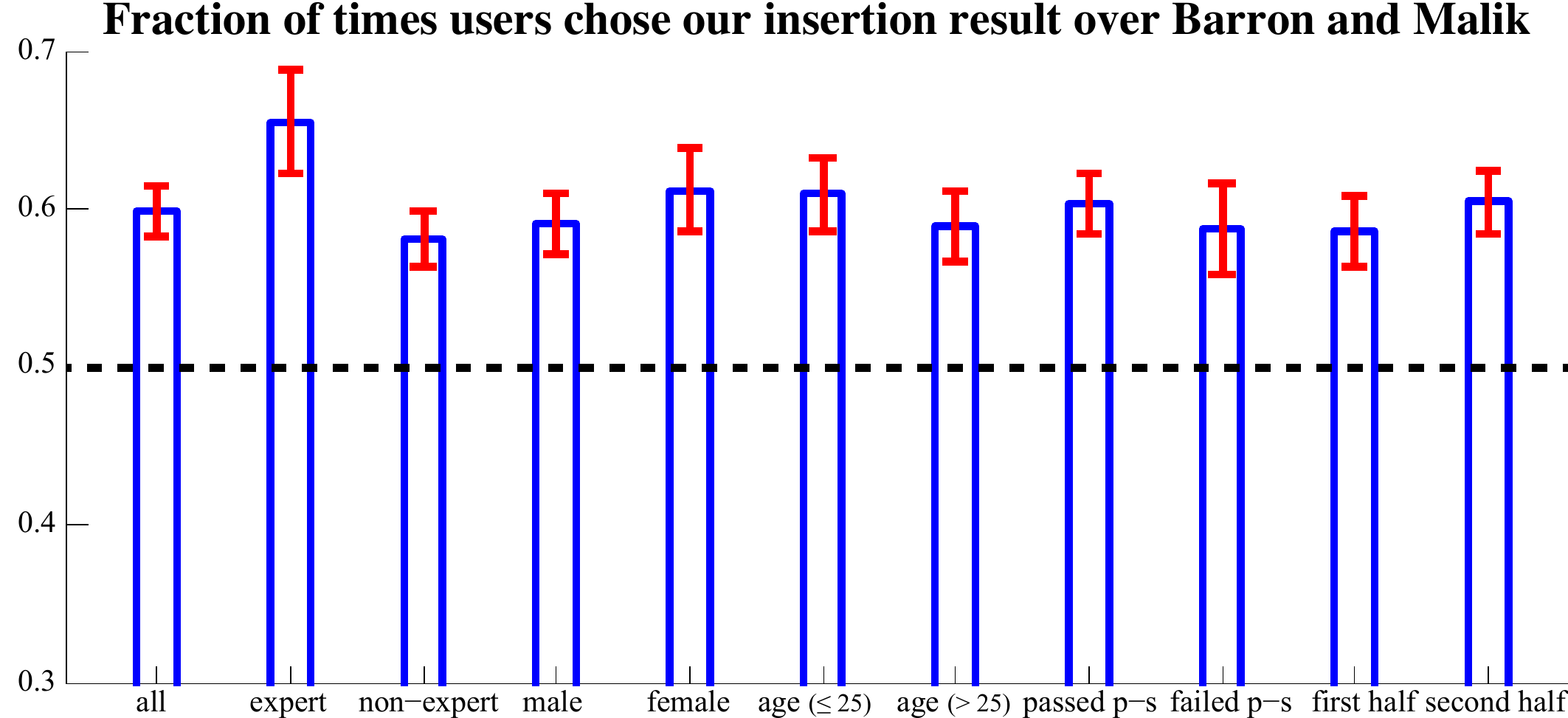}
}
\caption{In a comparison of our results against that by the method of Barron and Malik. our results were chosen as more realistic in 60\% of the trials ($N=1000$). For all subpopulations, our results were preferred well ahead of the other as well. All differences to the dotted line (equal preference) are greater than two standard deviation. The ``expert'' subpopulation chose our insertion results most consistently.
}\label{fig:shadingdecomp:study_vs_barron}
\vspace{-5mm}
\end{figure}

\begin{figure}[h]
\centering{
\includegraphics[width=0.75\linewidth]{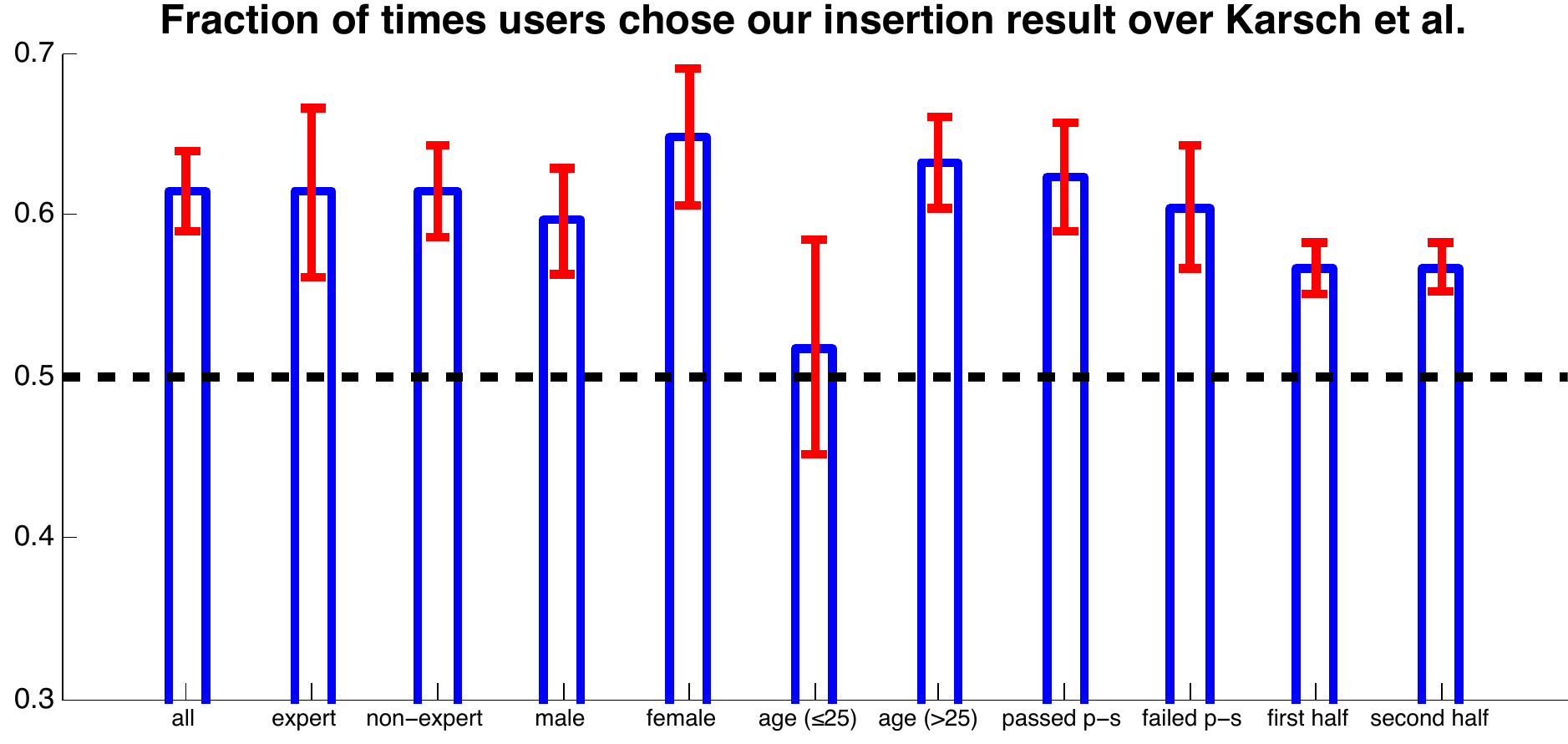}
}
\caption{In a comparison of our results against that by the method of Karsch et al. The advantage our method holds over that of Karsch et al. is similar to the advantage over Barron and Malik.
}\label{fig:shadingdecomp:study_vs_karsch}
\vspace{-5mm}
\end{figure}

\begin{figure}[!th]
\centering{
\subfigure[Input]{
\includegraphics[height=1.3in]{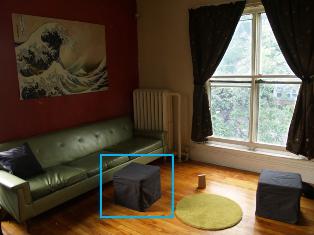}\hspace{-1mm}}
\subfigure[Shape]{
\includegraphics[height=1.3in]{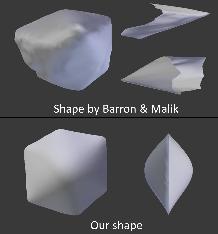}\hspace{-1mm}}
\subfigure[Result of B \& M]{
\includegraphics[height=1.3in]{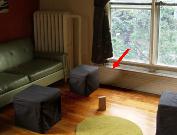}\hspace{-1mm}}
\subfigure[Our result]{
\includegraphics[height=1.3in]{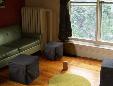}}
}
\caption{In this example, we built models from the cube in the input image (cyan box) and inserted it back into the scene. Sophisticated SfS methods (in this case, Barron and Malik~\protect\cite{Barron2012B}) can have large error and unstable boundaries that violates the generic view assumption. For object insertion, lighting {\it on} the object is important, but it is equally important that cast shadows and interreflected light look correct; shape errors made by complex SfS methods typically exacerbate errors both {\it on} and {\it around} the object (see cast shadows in c). Our shape is simple but behaves well in many situations and is typically robust to such errors (d). Best viewed in color at high-resolution.}
\label{fig:shadingdecomp:cube}
\vspace{-5mm}
\end{figure}

\vspace{2mm}
\noindent{\bf Results and discussion} Overall, our user study showed that subjects confused our insertion result with a real image 44\% of 1040 viewed image pairs (task 1); an optimal result would be 50\%. We also achieve better confusion rates than the insertion results of Barron and Malik (task 2, 42\%), and perform well ahead of the method of Barron and Malik in a head-to-head comparison (task 3, see Fig.~\ref{fig:shadingdecomp:study_vs_barron}), as well as a head-to-head comparison with the method of Karsch et al~\cite{Karsch:SA2011} (task 4, see Fig.~\ref{fig:shadingdecomp:study_vs_karsch}).

Table~\ref{tab:study_vs_real} demonstrates how well images containing inserted objects (using either our method or Barron and Malik) hold up to real images (tasks 1 and 2). See Figure~\ref{fig:shadingdecomp:studypics} (bottom) for an example trial from task 1. We observe better confusion rates (e.g. our method is confused with real images more than the method of Barron and Malik) overall and in each subpopulation except for the population who failed the perspective and shading tests in the questionnaire.

We also compare our method and the method of Barron and Malik head-to-head by asking subjects to choose those most realistic image when shown two similar results side-by-side (see Fig.~\ref{fig:shadingdecomp:studypics} top for an example trial). Figure~\ref{fig:shadingdecomp:study_vs_barron} summarizes our findings. Overall, users chose our method as more realistic in a side-by-side comparison on average 60\% of the time in 1000 trials. In all subject subpopulations, our method was preferred by a large margin to the method of Barron and Malik; each subpopulation was at least two standard deviations away from being ``at chance'' (50\% -- see the red bars and black dotted line in Fig.~\ref{fig:shadingdecomp:study_vs_barron}). Most interestingly, the expert subpopulation preferred our method by an even greater margin (66\%), indicating that our method may appear more realistic to those who are good judges of realism.

Karsch et al.~\cite{Karsch:SA2011} performed a similar study to evaluate their 3D synthetic object insertion technique, in which subjects were shown similar pairs of images, except the inserted objects were synthetic models. In their study, subjects chose the insertion results only 34\% of the time, much lower than the two insertion methods in this study, a full 10 points lower than our method and 8 points lower than the method of Barron and Malik in our study. While the two studies were not identical and performed by different populations, the results are nonetheless intriguing. We postulate that this large difference is due to the nature of the objects being inserted: we use {\it real} image fragments that were formed under real geometry, complex material and lighting, sensor noise, and so on; they use 3D models in which photorealism can be extremely difficult. By inserting image fragments instead of 3D models, we gain photorealism in a data-driven manner. This postulation is validated by our head-to-head comparison in task 4. See results in Fig.~\ref{fig:shadingdecomp:study_vs_karsch}.

\subsection{Conclusion and future work}
We have proposed a new representation suitable for relighting image fragments and an effective workflow for inserting objects photorealistically into new images. Our models are simple yet robust to errors that make existing SfS methods infeasible for relighting tasks. Through both quantitative and most importantly human subject studies, we found that our method is preferable to other methods for object relighting, and images created with our system are confusable (nearly at chance) with real images.

\begin{figure}[h]
\centering
 \begin{minipage}{\linewidth}
  \begin{minipage}{0.495\linewidth}
  \centerline{\bf Ground truth}
  \includegraphics[width=\linewidth]{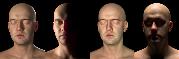}
  \end{minipage}
  \hfill
  \begin{minipage}{0.495\linewidth}
  \centerline{\bf Our results}
  \includegraphics[width=\linewidth]{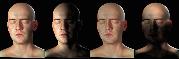}
  \end{minipage}
 \end{minipage}
  \caption{A failure example under extreme lighting conditions. The left group shows a 3D model lit under four lighting directions: 3-quarter (3Q), left, front, top. The right group shows our results. Our model appears realistic when the lighting is not strongly directed (3Q; front), but looks unnatural in harsh conditions (left; top).
  }\label{fig:shadingdecomp:failurecase}
\end{figure}

Due to the simple nature of our shape representation, the model can fail under extreme lighting conditions, i.e., strong point light source from extreme directions, or in the case of complex shapes (e.g. arm chairs or people in certain poses). See Figure~\ref{fig:shadingdecomp:failurecase} for a failure example on human faces. Large areas of strong shadow or highlight in the input image can also cause performance degrade for two reasons: (a) the current albedo-shading procedure works poorly for these cases, (b) the strong shadows and highlights will be (partly) kept in the shading detail layers and appear in subsequent relighting results. Fortunately, the visual system is insensitive to inaccuracies of small shadows and highlights. Nonetheless, the model is best extracted from input images of diffuse objects under multi-source lighting environments.

\chapter{Blind recovery of spatially varying materials from a single image} \label{chap:mats}

\comment{
\begin{abstract}
We propose a new technique for estimating spatially varying parametric materials from a single image of an object with unknown shape in unknown illumination. Our method uses a low-order parametric reflectance model, and incorporates strong assumptions about lighting and shape. We develop new priors about how materials mix over space, and jointly infer all of these properties from a single image. This produces a decomposition of an image which corresponds, in one sense, to microscopic features (material reflectance) and macroscopic features (weights defining the mixing properties of materials over space). 

We have built a large dataset of real objects rendered with different material models under different illumination fields for training and ground truth evaluation.  Extensive experiments on both our synthetic dataset images as well as real images show that (a) our method recovers parameters with reasonable accuracy; (b) material parameters recovered by our method give accurate predictions of new renderings of the object; and (c) our low-order reflectance model still provides a good fit to many real-world reflectances. We also show that our technique has applications in both computer graphics (material transfer and synthesis) and computer vision (material recognition and classification).
\end{abstract}
}


\begin{figure*}
\begin{minipage}{\textwidth}
\includegraphics[width=\textwidth]{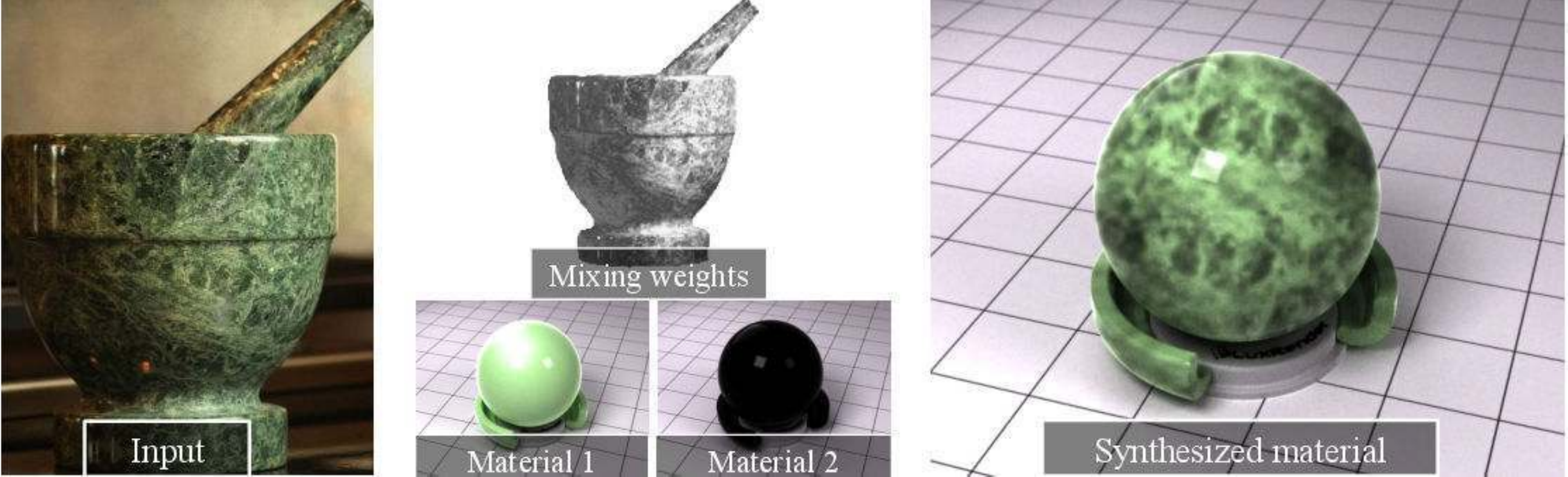}
\caption{From a single photograph, our method estimates spatially varying materials (diffuse reflectance and specular parameters). The input image is decomposed into $k$ low-order, parametric materials (Material 1 and 2) and a set of per-pixel material mixing coefficients (Mixing weights); shape and illumination is jointly inferred. This decomposition can be transferred to new shapes (Synthesized material).
We also show that our decompositions can be used to generate new materials and discriminate against subtle material distinctions.
}
\label{fig:materials:teaser}
\end{minipage}
\end{figure*}

\section{Introduction}
Humans are quite good at guessing an object's material based on appearance alone~\cite{Adelson99lightnessperception}. However, material\footnote{Throughout our work, we abbreviate ``material reflectance'' with ``material.''} estimation from a single photograph remains a challenging and unsolved problem in computer vision. Appearance is often considered a function of object shape, incident illumination, and surface reflectance, and many solutions have been proposed addressing the problem of material estimation from a single image if shape and/or illumination are known precisely. 

Romeiro and Zickler first showed how to estimate reflectance under known shape and illumination~\cite{Romeiro:eccv:08}, and Romeiro et al. later extended this work by marginalizing over illumination~\cite{Romeiro:eccv:10}. Generalizing further, Lombardi and Nishino~\cite{Lombardi:eccv:2012} recover reflectance and illumination from an image assuming only that the object's shape is known, and Oxholm and Nishino~\cite{Oxholm:eccv:2012} estimate reflectance and shape under exact lighting. If multiple images are available, it is also possible to recover shape and spatially varying reflectance~\cite{Alldrin:cvpr:2008,Goldman:tpami:2010}. These techniques provide valuable intuition for moving forward, yet they hinge on knowing {\it exact} shape or {\it exact} illumination, or have strict setup requirements (directional light, multiple photos, etc), and require a fundamentally different approach when additional information is not available. 

Such approaches have been proposed by Baron and Malik~\cite{Barron2012B,Barron:cvpr:12}, who use strict priors to jointly recover shape, diffuse albedo, and illumination. However, as in many shape-from-shading algorithms, all surfaces are assumed to be Lambertian. Glossy surfaces are thus impossible to recover and may cause errors in estimation. Furthermore, Lambertian models of material are not suitable for describing a large percentage of real-world surfaces, limiting the scope and applicability of these techniques.

A major concern of prior work is in recovering real-world BRDFs and high-frequency illumination~\cite{Lombardi:eccv:2012,Oxholm:eccv:2012,Romeiro:eccv:10}, or that recovered shapes are integrable and reconstructions are exact (image and re-rendered image match exactly)~\cite{Barron2012B,Barron:cvpr:12}. However, it is well known that recovering these high-parametric solutions is ill-posed in many cases, and precise conditions must be met to estimate these robustly. For example, real-world BRDFs can be extracted from a curved shape and single directional light (known a priori)~\cite{Chandraker:iccv:11}, and surface normals can be found given enough pictures with particular lighting conditions and isotropic (yet unknown) BRDFs~\cite{Alldrin:iccv:2007,Shi:eccv:2012}.

We opt for lower-order representations of reflectance and illumination. Our idea is to reduce the number of parameters we recover, relax the constraints imposed by prior methods, and attempt to recover materials from a more practical perspective. Our main goal is material inference, but we must jointly optimize over shape and illumination since these are unknown to our algorithm. We consider simple models of reflectance and lighting (models often used by artists, where perception is the only metric that matters), and impose only soft constraints on shape reconstruction. Our material model is low-order (only five parameters), allowing us to tease good estimates of materials from images, even if our recovered shape and illumination estimates are somewhat inaccurate. We also show that our model can be extended to spatially varying materials by inferring mixture coefficients (linearly combining 5-parameter materials) at each pixel. Figure~\ref{fig:materials:teaser} demonstrates the results of our estimation technique on a marble mortar and pestle.

\boldhead{Contributions} Our primary contribution is a technique for extracting spatially varying material reflectance (beyond diffuse parameters) directly from an object's appearance in a single photograph {\it without requiring any knowledge of the object's shape or scene illumination}. We use a low-order parameterization of material, and develop a new model of illumination that can be described also with only a few parameters, allowing for efficient rendering. {\it Because our model has few parameters, we tend to get low variance and thus robustness in our material estimates (e.g. bias-variance tradeoff).} By design, our material model is the same that is used throughout the 3D art/design community, and describes a large class of real-world materials (Sec~\ref{sec:param}). We show how to efficiently estimate materials from plausible initializations of lighting and shape, and propose novel priors that are crucial in estimating material robustly (Sec~\ref{sec:est}), and extend this formulation to spatially varying materials in Section~\ref{sec:sv_est}. Our material estimates perform favorably to baseline methods and measure well with ground truth, and we demonstrate results for both synthetic and real images (Sec~\ref{sec:materials:eval}).

These material estimates have applications in both the domains of vision and graphics. For example, the material of a photographed object can be transferred to new objects of different shape, even in new illumination environments (Figs~\ref{fig:materials:teaser} and~\ref{fig:renderPalettes}). We also demonstrate the generative capabilities of our decompositions: by combining different parts of two decompositions, new, unseen materials can be created (Fig~\ref{fig:renderStealGrid}). Finally, we have collected a new dataset which captures nuances in materials (both at the microscopic and macroscopic levels; see Fig~\ref{fig:rig}), and show that our decompositions can be used as features that aid in the automatic classification of these materials (Fig~\ref{fig:confusion} and Table~\ref{tab:classificationQuant}).

\boldhead{Limitations} Since we are using low-order material models that are isotropic and have monochromatic specular components, we cannot hope to estimate BRDFs of arbitrary shape (e.g. as measured by a gonioreflectometer), and there are some materials not encoded by our representation. Our recovered lighting and shape are not necessarily correct with respect to the true lighting/shape, although they are consistent with one another and sometimes give good estimates; as such, we only make claims about the accuracy of our material estimates. We use infinitely distant spherical lighting without considering interreflections, and we do not attempt to solve color constancy issues; lighting environments in our dataset integrate to the same value (per channel). Since we only have a single view of the object, certain material properties (e.g. specularities) may not be visible (depending especially on the coverage of the normals, i.e. flat surfaces provide much less information than curved surfaces). Due to our low-order model and perhaps mixture priors, shading effects can sometimes manifest in the spatial mixture map (Fig~\ref{fig:failure} top). 

\section{Related work}
Many materials can be intuitively thought of as a small set of reflectance functions modulated by spatial weights\ \cite{Goldman:tpami:2010}. For example, the marble in Fig~\ref{fig:materials:teaser} appears to be made from two fundamental materials (a bright green material and a darker material); at each point on the surface, these two materials are appropriately combined. This representation is richer than standard texture representation as it encodes intrinsic reflectance properties of a material at a fine scale, but it is also like a texture representation as it encodes spatial structure at larger scales. In this chapter, we explore the capabilities of this intuitive model for estimating spatially complex materials in {\it single images}. We show that the models produced by our algorithm can be used for re-rendering (material transfer and generation), as well for material discrimination and description.

{\it Material discrimination} is an established topic, with few strong results. There is a clear distinction between material and texture (eg~\cite{Liumat}, Figs 2 and 3), although texture methods generally apply.  Summaries of spatial variation in appearance appear to be useful to humans~\cite{Adelson2} as well as essential in computer vision.  The CURET database~\cite{CAVE_0060} consists of a set of views of 61 nominally flat material surfaces under differing illumination and view conditions.  Leung and Malik show strong discriminative results on this dataset using inferred views\ \cite{LeungMalik1}. Varma and Zisserman show very strong results on this and other datasets, using nearest neighbor methods applied to a patch based representation~\cite{Varma09a}. Liu et al. offer a demanding dataset of material images, and show that the methods of Varma and Zisserman are not particularly strong at  discriminating between these images~\cite{Liumat}.  However, the dataset of Liu  et al. aggressively mixes spatial scales (50 pure material, 50 object scale images per category), and covers a relatively small range of materials.  Liao et al. give a demanding dataset with a less aggressive mixture of scales~\cite{Liao:cvpr13a}, and demonstrate a feature that encodes shading effects from fine scale surface geometry (pits, grooves, etc.). All these methods, however, name materials (i.e. ``stone''; ``water'').

An alternative would be a description: ``glossy,'' \\ ``rough,'' and so on.  Specularity detection is relatively straightforward (e.g.~\cite{Brelstaff}), and may be enough to tell if a material is glossy.   Abe et al. use an SVM ranker on image features to rank materials on ``glossiness,'' ``transparency,'' ``smoothness'' and ``coldness''~\cite{Abe}. Mall and Lobo distinguish between ``wet'' and ``dry'' surfaces using image features~\cite{Mall}.  An important difficulty here is the richness of the available material vocabulary -- how should one choose attributes that describe many materials and can distinguish all cases?  For example, Adelson gives over forty terms used regularly by mineralogists in describing rocks; specialist vocabularies for other cases might be as rich~\cite{Adelson}.   

The different terms appear to arise from complex interactions between surface properties at several scales.  We use ``microscopic'' for the scale around light wavelengths, captured by BRDF style models; and ``macroscopic'' for longer scales on which the material appears to be homogeneous.  We do not believe this is a complete set of scales.  Appearance phenomena can result from subtle interactions between mechanical properties and optical properties, and humans can exploit these cues for inference (for example, figure 7 of~\cite{Adelson}, where it is quite obvious which is hand-cream and which is cream cheese).

In this work, we attempt to categorize an image into our interpretations of microscopic and macroscopic phenomena. Our idea is to estimate a low number of parametric materials, that, when weighted appropriately (over space), give the same appearance as the input photograph. Our parametric materials encode microscopic phenomena, and macroscopic features are encoded using a spatially varying set of weights. Previous methods have used spatially varying weights to define a linear combination of ``basis'' BRDFs~\cite{Lawrence:2006:IST,Alldrin:cvpr:2008}. These methods rely on hundreds to thousands of input images in specific settings, and use a non-parametric BRDF representation (typically requiring many basis materials).

Most similar to our work is the method of Goldman et al. \cite{Goldman:tpami:2010}. They estimate per-pixel mixture weights and a set of parametric materials, but require multiple HDR input images under {\it known} lighting and impose limiting priors on mixture weights. Our method is applicable to single, LDR images (lighting is jointly estimated), and we also develop new priors for better spatial weight encodings. Furthermore, we show applications of our results in new domains (e.g. material generation and classification).

3D texture refers to the strong visual effects produced by surface detail (e.g.~\cite{gkd,CAVE_0060}).  Our representation cannot  wholly capture 3D texture, because it cannot account for (for example) the way that shadows move around small bumps on a surface as the light moves (e.g.~\cite{dananayar}).   Dana et al. represent these effects with a bidirectional texture distribution function (BTF), a table of texture appearances as a function of view and light direction~\cite{CAVE_0060}.    Histograms of BTF elements provide strong discriminative performance on CURET~\cite{btfrec}.  However, 3D texture representations must either be measured goniometrically, or inferred with the aid of comprehensive dictionaries of examples so measured (as in, say,~\cite{LeungMalik1,Varma09a}).  As we show, our representation captures a rich range of spatial effects relatively effectively, and can be inferred from a single image.

Because our task is defined for single image, we must impose aggressive priors on shape, lighting, and material. Baron and Malik~\cite{Barron2012B} use strict priors to jointly recover shape, and illumination, but their materials are confined to be diffuse. In this chapter, we use existing shape and illumination priors, and introduce new priors for spatially varying materials. We show that these priors enable the estimation of spatially varying specular materials for single images without knowing shape or lighting a priori.

\begin{figure}
\begin{center}
\begin{minipage}{.9\columnwidth}
\includegraphics[width=0.49\columnwidth,page=1]{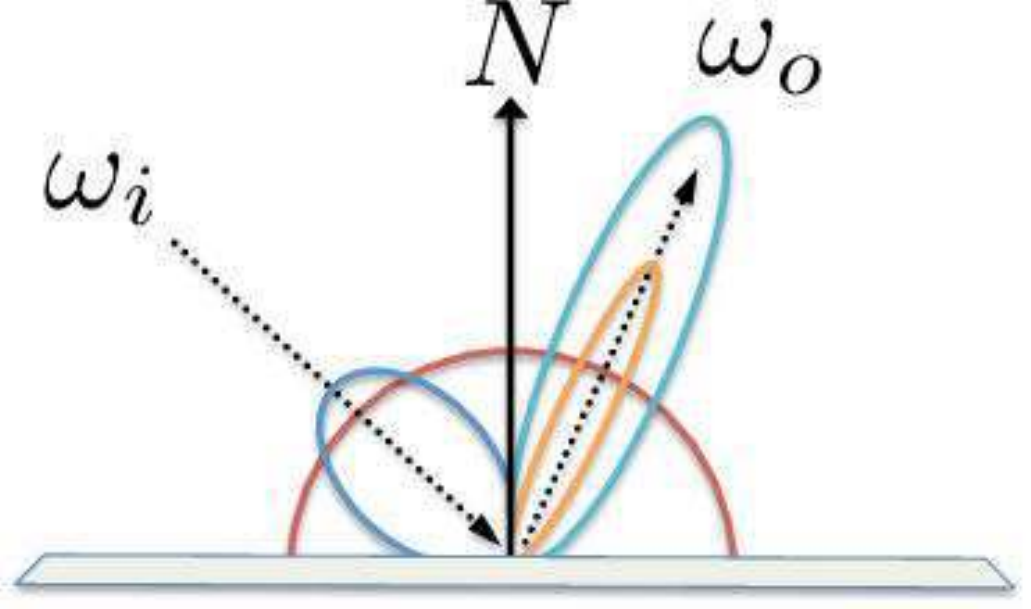} 
\includegraphics[width=0.49\columnwidth,page=2]{./materials/ijcv14-materials/mat_rep.pdf} 
\vspace{-7.5mm} \\
{\color{white}'} \hspace{0.39\columnwidth} \fcolorbox{black}{white}{\Large a}  \hspace{0.42\columnwidth} \fcolorbox{black}{white}{\Large b}\vspace{1mm}\\
\includegraphics[width=0.49\columnwidth]{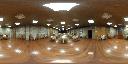}
\includegraphics[width=0.49\columnwidth]{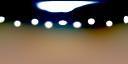} 
\vspace{-7.5mm} \\
{\color{white}'} \hspace{0.39\columnwidth} \fcolorbox{black}{white}{\Large c}  \hspace{0.42\columnwidth} \fcolorbox{black}{white}{\Large d}\vspace{0mm}\\
\centerline{\rule{.99\columnwidth}{0.2mm}}\\
\begin{minipage}{0.325\columnwidth}
\centerline{Measured BRDF (a)}\centerline{Environment map (c)}
\includegraphics[width=1\columnwidth]{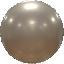}\\
\includegraphics[width=1\columnwidth]{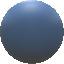}
\end{minipage}
\begin{minipage}{0.325\columnwidth}
\centerline{Our model (b)}\centerline{Environment map (c)}
\includegraphics[width=1\columnwidth]{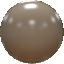}\\
\includegraphics[width=1\columnwidth]{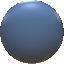}
\end{minipage}
\begin{minipage}{0.325\columnwidth}
\centerline{Our model (b)}\centerline{Our illumination (d)}
\includegraphics[width=1\columnwidth]{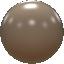}\\
\includegraphics[width=1\columnwidth]{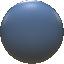}
\end{minipage}
\end{minipage}
\end{center}
\vspace{0mm}
\caption{A general BRDF can be made up of numerous reflection ``lobes'' (a), but in practice (e.g. surface modeling), a simple BRDF model with one diffuse and one specular lobe tends to suffice (b). We use this simple representation, as well as a low-order parameterization of illumination. Our illumination considers a real-world, omnidirectional lighting environment (c), and approximates it with a mixture of Gaussians and spherical harmonics; (d) shows our model fit to (c). We observe only slight perceptual differences when rendering with different combinations of the high- and low-order parameterizations (bottom rows). Our low-order representations of material and illumination not only reduces variance in our estimates, but also allow for efficient estimation.
}
\label{fig:parameterization}
\end{figure}

\section{Low-order reflectance and illumination} 
\label{sec:param}
Many previous methods have attempted to use high-order models of material (e.g. a linear combination of basis functions learned from measured BRDFs~\cite{Romeiro:eccv:10}) and illumination (parameterized with wavelets~\cite{Romeiro:eccv:10} or even on an image grid~\cite{Lombardi:eccv:2012,Oxholm:eccv:2012}, consisting of hundreds to thousands of parameters or more). We propose the use of more rigid models of shape and illumination, which still can describe the appearance of most real world objects, and provide necessary rigidness to estimate materials when neither shape or illumination are exactly known.

\boldhead{Representing material}
We represent materials using an isotropic diffuse and specular BRDF model consisting of only five parameters: diffuse albedo in the red, green, and blue channels ($R_d$), monochromatic specular albedo ($R_s$) and the isotropic ``roughness'' value ($r$), which is the concentration of light scattered in the specular direction, and can considered (roughly) to be the size of the specular ``lobe'' (a smaller roughness value indicates a smaller specular lobe, where $r=0$ encodes a perfect specular reflector).

This type of material model is surprisingly general for its low number of parameters. Ngan et al. have previously shown that such parameterizations provide very good fits to real, measured BRDFs~\cite{ngan:egsr:05}.
Perceptually, this model can also encode a family of isotropic, dielectric surfaces (mattes, plastics, and many other materials in the continuum of perfectly diffuse to near-perfect specular)~\cite{PBRTv2}. There is also compelling evidence that such a material model suffices for photorealistic rendering, as this is the same material parameterization found most commonly in 3D modeling and rendering packages (such as Blender\footnote{\urlwofont{http://wiki.blender.org/index.php/Doc:2.6/Manual/Materials}}, which only considers diffuse and specular reflection for opaque objects), and used extensively throughout the 3D artist/designer community\footnote{\urlwofont{http://www.luxrender.net/forum}}.

We write our BRDF following the isotropic substrate model as described in {\it Physically Based Rendering}~\cite{PBRTv2}, which uses a microfacet reflectance model and assumes the Schlick approximation to the Fresnel effect~\cite{Schlick94aninexpensive}. 

Figure~\ref{fig:parameterization} shows a comparison of what a measured BRDF (a) might look like in comparison to our material parameterization (b). We compare measured BRDFs to our material model (fit using the procedure described in Sec~\ref{sec:materials:eval}) rendered in natural illumination in the bottom row (left two columns). 

\boldhead{Representing illumination}
Consider a single point within a scene and the omnidirectional light incident to that point. This incident illumination can be conceptually decomposed into luminaires (light-emitters) and non-emitting objects.  We consider these two separately, since the two tend to produce visually distinct patterns in object appearance (depending of course on the material). Luminaires will generally cause large, high-frequency changes in appearance (e.g. specular highlights), and non-emitters usually produce small, low-frequency changes. 

Using this intuition, we parameterize each luminaire as a two dimensional Gaussian in the 2-sphere domain (sometimes known as the Kent distribution), and approximate any other incident light (non-emitted) as low-order functions on the sphere using $2^{nd}$ order spherical harmonics\footnote{We assume all lighting comes from an infinite-radius sphere surrounding the object, as done in previous methods~\cite{Barron2012B,Lombardi:eccv:2012,Oxholm:eccv:2012,Romeiro:eccv:10})}. 

Such a parameterization has very few parameters relative to a full illumination environment (or environment map): six per light source (two for direction $L^{(d)}$, one for each intensity $L^{(I)}$, concentration $\kappa$, ellipticalness $\beta$, and rotation about the direction $\gamma$) and 27 spherical harmonic coefficients (nine per color channel), but more importantly, this parameterization still enables realistic rendering at much higher efficiency. Rendering efficiency is crucial as to our procedure, as each function evaluation in our optimization method (Sec~\ref{sec:est}) requires rendering.

Our lighting environments maintain only high frequencies in regions of emitting sources, and is encoded by low-frequency spherical harmonics everywhere else. However, rendering with full versus approximate (our) lighting produces similar results (bottom middle vs bottom right). For additional discussion, see Sec~\ref{sec:est}.

\begin{figure*}
\centerline{\large True \hspace{23mm} Starting point \hspace{12mm} Optimized estimates \hspace{15mm} Final result}
\hspace{0mm} Input image \hspace{5mm} Normals \hspace{6mm} Rendered \hspace{7mm} Normals  \hspace{6mm} Rendered \hspace{7mm} Normals \hspace{4mm} True ${\bf M}$ (input) \hspace{2mm} Recovered ${\bf M}$  \hspace*{2mm} \\
\setlength\fboxrule{1pt}
\setlength\fboxsep{1pt}
\begin{overpic}[width=\linewidth]{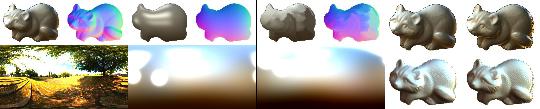}
\put(391,53){\fcolorbox{black}{white}{\footnotesize Original illumination}}
\put(395,0){\fcolorbox{black}{white}{\footnotesize Novel illumination}}
\end{overpic}
\caption{Results from our optimization procedure. On the left is the input image (top left), true surface normals (top right), and true illumination (below), followed to the right by estimates that we begin our optimization with. The estimated rendering (using Eq~\ref{eq:render}), estimated surface normals, and estimated illumination are displayed in the third column. The rightmost column shows the true material (left) and our estimated material (right) rendered onto the true shape in the original lighting environment (top), and rendered in novel lighting (bottom). Our initialization is described in Sec~\ref{sec:init}, and uses no prior information about the input.
}
\label{fig:optimization}
\end{figure*}

\section{Estimating specular reflectance parameters}
\label{sec:est}
Our idea is to jointly recover material, shape, and illumination, such that a rendering of these estimates produces an image that is similar to the input image. Following Barron and Malik~\cite{Barron2012B}, we also enforce a strong set of priors to bias our material estimate towards plausible results. 



Our goal is to recover a five dimensional set of material parameters $\ve{M} = (R_d^{(r)},R_d^{(g)},R_d^{(b)}, R_s,r)$, while jointly optimizing over illumination $\ve{L}$ surface normals $\ve{N}$. Following notation in section~\ref{sec:param}, we denote $R_d$ as RGB diffuse reflectance, $R_s$ as monochromatic specular reflectance, and $r$ as the roughness coefficient (smaller $r \Rightarrow$ narrower lobe $\Rightarrow$ shinier material). Illumination $\ve{L} = \{L,s\}$ is parameterized as a mixture of $m$ Gaussians in the 2-sphere domain $L = \{L_1,\ldots L_k\}$ with direction $L^{(d)}_i$, intensity $L^{(I)}_i$, concentration $L^{(\kappa)}_i$, ellipticalness $L^{(\beta)}_i$, and rotation $L^{(\gamma)}_i$ ($i \in \{1,\ldots,k\}$). Statistics of real illumination environments are nonstationary and can contain concentrated bright points~\cite{dror2004sta}; the Gaussian mixture aim to represent these peaks. Indirect light $s$ is represented as a 9 $\times$ 3 matrix of $2^{nd}$ order spherical harmonic coefficients (9 per color channel). $\ve{N}$ is simply a vector of per-pixel surface normals parameterized by azimuth and elevation directions.

We phrase our problem as a continuous optimization by solving the parameters which minimize the following:
\begin{align}
\label{eq:obj}
\hspace{5mm} \displaystyle \underset{{\ve{M},\ve{N},\ve{L}}}{\operatorname{argmin}} \ \ &E_{\text{rend}}(\ve{M},\ve{N},\ve{L} ) + E_{\text{mat}}(\ve{M}) \ + \nonumber \\ 
  &E_{\text{illum}}(\ve{L} ) + E_{\text{shape}}(\ve{N})  \nonumber  \\ \nonumber \\ 
\text{subject}&\text{ to } 0 \leq \ve{M}^{(i)} \leq 1,\ i\in\{1,\ldots,5\},
\end{align}
where $E_{\text{rend}}$ is the error between a rendering of our estimates and the input image, and $E_{\text{mat}},E_{\text{illum}}$, and $E_{\text{shape}}$ are priors that we place on material, illumination, and shape respectively. In the remainder of this section, we discuss the rendering term and the prior terms. Figure~\ref{fig:optimization} shows the result of our optimization technique at various stages for a given input.

\boldhead{Rendering error} Our optimization is guided primarily by a term that penalizes pixel error between the input image and a rendering of our estimates. The term itself is quite simple, but efficiently optimizing an objective function (which includes the rendering equation) can be challenging. Writing $I$ as the input, we define the term as the average squared error for each pixel: 
\begin{equation}
E_{\text{rend}}(\ve{M},\ve{N},\ve{L} ) = \sum_{i\in \text{pixels}} \sigma_i^{rend} || I_i - f(\ve{M},\ve{N}_i,\ve{L} ) ||^2,
\end{equation}
where $f(\ve{M},\ve{N},\ve{L} )$ is our rendering function, and $\sigma_i^{rend} =  I_i^2$ re-weights the error to place more importance on brighter points (primarily specularities).

Notice that we do not strictly enforce equality (as in~\cite{Barron2012B,Barron:cvpr:12}), as this soft constraint allows more flexibility during the optimization, and because our parameterizations are too stiff for equality to hold. This has the added benefit of reducing variance in our estimates.

As in any iterative optimization scheme, each iteration requires a function evaluation (and most likely a gradient or even hessian depending on the method). If chosen na\"{i}vely, $f$ can take hours or longer to evaluate and differentiate, and here we describe how to construct $f$ so that this optimization becomes computationally tractable.

The key to efficiency is in our low-order parameterization of illumination. By considering emitting and non-emitting sources separately, we treat our rendering as two sub-renders: one is a ``full'' render using the emitting  luminaires (which are purely directional since we assume the light is at infinity), and the other is a diffuse-only render using all other incident light (reflected by non-emitters).

Denoting $\Omega_e$ and $\Omega_n$ as set of ``emitting'' and ``non-emitting'' light directions respectively, $l(\omega)$ as the light traveling along direction $\omega$, and $f_M$ as the BRDF defined by material $M$, we write our rendering function as
\begin{align}
f^{(e)}_i &=  \int_{\Omega_e} f_M(\omega,v) l(\omega) \max(\omega \cdot \ve{N}_i,0) d\omega \nonumber \\
f^{(n)}_i &=   \int_{\Omega_n} l(\omega) \max(\omega \cdot \ve{N}_i,0) d\omega \nonumber \\
f(\ve{M},\ve{N},\ve{L} )_i &= f^{(e)}_i + R_d f^{(n)}_i,
\label{eq:render}
\end{align}
for the $i^{th}$ image pixel and a particular view direction $v$. Notice that $f^{(n)}_i$ is simply irradiance over the non-emitting regions of the sphere, and is modulated by diffuse reflectance ($R_d$) since Lambertian BRDFs are constant.

We can compute both of these efficiently, because $\Omega_e$ is typically small (most lighting directions are occupied by negligible Gaussian components), and it is well known that diffuse objects can be efficiently rendered through spherical harmonic projection~\cite{Ravi:sg:2001}.
In terms of previous notation, directional sources $L$ are used in the full render ($f^{(e)}_i$), and $s$ is used for the diffuse-only render ($f^{(n)}_i$). 

The intuition behind such a model is that indirect light contributes relatively low-frequency effects to an object's appearance, and approximating these effects leads to only slight perceivable differences~\cite{Ramamoorthi04:SignalProcessing}.  A variation of this intuition is used for efficiently choosing samples in Monte Carlo ray tracing (e.g. importance sampling~\cite{PBRTv2}), which causes problems in continuous optimization techniques since rendering is then non-deterministic. 


\boldhead{Material prior} The rigidity of our material model (5 parameters to describe the entire surface), is a strong implicit prior in itself, but we also must deal with the ambiguity that can exist between diffuse and specular terms. For example, if a specular lobe ($r$) is large enough, then the specular albedo and diffuse albedo can be confused (e.g. dark specular albedo/bright diffuse albedo may look the same as bright specular albedo/dark diffuse albedo). Thus, we add a simple term to discourage large specular lobes, persuading the diffuse component to pick up any ambiguity between it and the specular terms:
\begin{equation}
E_{\text{mat}}(\ve{M}) = \lambda_m r^2.
\label{eq:matprior}
\end{equation}
The only material parameter that is constrained is specular lobe size, and $\lambda_m = 1$ in our work. 

\boldhead{Illumination prior}
We develop our illumination prior by collecting statistics from spherical HDR imagery found across the web (more details in Sec~\ref{sec:materials:eval}). Each image gives us a sample of real-world illumination, and to see how each sample relates to our illumination parameters, we fit our lighting model (SO(2) Gaussians + 2${}^{nd}$ order spherical harmonics) to each spherical image. Fitting is done using constrained, non-linear least squares, and the number of Gaussians (corresponding roughly to luminaires) is determined by the number of peaks in the HDR image (smoothed to suppress noise). Priors are developed by clustering the Gaussian parameters, and through principal component analysis on the spherical harmonic coefficients.

Denote $\bar\kappa_j$, $\bar\beta_j$ as the means of the $j^{th}$ clusters (clustered independently using $k$-means) for the concentration, and ellipticalness of Gaussian parameters from our fitting process. Intuitively, these cluster centers give a reasonable sense of the shape of luminaires found in typical lighting environments, and we enforce our estimated sources to have shape parameters similar to these:
\begin{equation}
E_{\text{illum}}^{\text{means}}(L_i) = \mathcal{S}( \{ |L^{(\kappa)}_i - \bar\kappa_j| \}_{i=1}^k ) + \mathcal{S}( \{ |L^{(\beta)}_i - \bar\beta_j| \}_{i=1}^k ),
\end{equation}
where $\mathcal{S}$ is the softmin function (differentiable min approximation) and $|\cdot|$ is a differentiable approximation to the absolute value (e.g. $\sqrt{x^2+\epsilon}$). 

We also find the principal components (per channel channel) of the spherical harmonic coefficients fit to our data. During estimation, we reparameterize the estimated SH coefficients using weight vectors $w_{\{r,g,b\}}$, principal component matrices $S_{\{r,g,b\}}$, and means of all fit SH components $\mu_{\{r,g,b\}}$: $s(w) = [\mu_r + S_r w_r, \mu_g + S_g w_g, \mu_b + S_b w_b]$. We impose a Laplacian prior on the weight vector:
\begin{equation}
E_{\text{illum}}^{\text{pca}}(w) = \sum_{i\in\text{weights}} | w |.
\end{equation}
This coerces the recovered SH components to lie near the dataset mean, and slide along prominent directions in the data. We found that seven principal components (per channel) roughly explained over 95\% of our data (eigenvalue sums contain $>$95\% of the mass), and we discard the two components corresponding to the smallest eigenvalues (then $w_{\{r,g,b\}} \in \mathbb{R}^7$ and $S_{\{r,g,b\}} \in \mathbb{R}^{9 \times 7}$).

We also impose a gray world assumption, namely that each color channel should integrate (over the sphere of directions) to roughly the same value. Because we only have a single view of an object, some portions of the lighting sphere have significantly more influence than others; e.g. the hemisphere behind the object is mostly unseen and has smaller influence than the hemisphere in front. We weight the integration appropriately so that the dominant hemisphere has more influence (using $W_\theta = \cos(\frac{\theta-\pi}{2})$, where $\theta$ is the angle between the view direction and the direction of integration). This integration translates to a simple inner product (due to the nice properties of spherical harmonics), making the prior easy to compute:
\begin{eqnarray}
E_{\text{illum}}^{\text{gray}}(s) = & || G^T s_r - G^T s_g || + || G^T s_g - G^T s_b || \nonumber \\
 &+ || G^T s_r - G^T s_b ||,
\end{eqnarray}
where $G$ is the pre-computed integral of $2^{nd}$ order spherical harmonic basis functions (weighted by $W_\theta$), and $s_{\{r,g,b\}}$ are the current estimates of spherical harmonic coefficients.

Our prior is then a weighted sum of these three terms:
\begin{equation}
E_{\text{illum}}({\bf L}) = \lambda_m \sum_{i=1}^m E_{\text{illum}}^{\text{means}}(L_i) + \lambda_p E_{\text{illum}}^{\text{pca}}(w) + \lambda_g  E_{\text{illum}}^{\text{gray}}(s),
\end{equation}
keeping in mind ${\bf L} = \{L,s(w)\}$, and $w$ as PCA weights described above. We set $\lambda_m = \lambda_p = \lambda_g = 0.1$.

\boldhead{Shape prior}
We also optimize over a grid of surface normals, and impose typical shape-from-contour constraints: smoothness, integrable shape, and boundary normals are assumed perpendicular to the view direction. Let $N_i = (N_i^x, N_i^y, N_i^z)$, and $\hat N$ be the set of normals perpendicular to the occluding contour and view direction. We write the prior as:
\begin{eqnarray}
E_{\text{shape}}({\bf N}) =& \displaystyle\sum_{i \in\text{pixels}} \lambda_s \eta_i^{s} || \nabla N_i || + \lambda_I ||\nabla_y \frac{N^x_i}{N^z_i} -  \nabla_x \frac{N^y_i}{N^z_i} || \nonumber \\ 
&+ \lambda_{c} \sum_{c \in\text{boundary pixels}} ||N_c - \hat N_c||, 
\end{eqnarray}
where the first term encodes smoothness where the input is also smooth (modulated by image dependent weights $\eta$), the second term enforces integrability, and the third ensures that boundary normals are perpendicular to the viewing direction. We set the weights as $\lambda_s=1$, $\lambda_I=\lambda_c=0.1$.

\subsection{Initialization}
\label{sec:init}
Initial estimates of shape come from a na\"{i}ve shape-from-contour algorithm (surface assumed tangent to view direction at the silhouette, and smooth elsewhere), and light is initialized with the mean of our dataset (if applicable; leaving out the illumination that generated the input image). We estimate an initial $R_d$ by rendering irradiance with initial estimates of shape and lighting, dividing by the input image to get per-pixel albedo estimates, and averaging the RGB channels; $R_s, r$ are set as small constants (0.01 for our results). Full details of our initialization procedure can be found in supplemental material.

\subsection{Undoing estimation bias} 
\label{sec:bias}
Our low-parametric models tend to introduce bias into our estimates, but at the same time reduce estimation variance; e.g. bias-variance tradeoff). However, we have found that our priors produce consistent estimation bias: we typically see a smaller specular lobe and specular albedo due most likely to our material prior (Eq~\ref{eq:matprior}). We may also observe omitted-variable bias for images with materials not encoded by our model, but we do not address here.

Past methods point out that there are clear visual distinctions between different types and levels of gloss~\cite{fleming2003rea,Sharan:josaa:2008,Wills:tog:2009}, and we use the input image coupled with our estimates to develop an ``un-biasing'' function. We develop a simple regression method (simple methods should suffice since the bias appears to be consistent) which works well for removing bias and produces improved results.  Our goal is to find a linear prediction function that takes a vector of features to unbiased estimates of $R_d, R_s$ and $r$. Our features consist of our estimates of specular albedo and specular lobe size, as well as histogram features computed on the resulting rendered image, normal map, input image, and the error image (rendered minus input); features are computed for both raw and gradient images. Given a set of results from our optimization technique with ground truth material parameter (obtained, e.g., from our dataset in Sec~\ref{sec:materials:eval}), we compute a bias prediction function by solving an L${}_1$ regression problem (with L${}_2$ regularization). For more details, see the supplemental material.

%
%

\section{Recovering spatially varying reflectance}
\label{sec:sv_est}

We propose an extension of Eq~\ref{eq:obj} for estimating spatial mixtures of materials. First, we define our appearance model simply as a spatially varying linear combination of renderings. Radiance at pixel $i$ is defined as:
\begin{equation}
\sum_{j\in \text{materials}} m_{i,j} f(\ve{M}_j,\ve{N}_i,\ve{L} ),
\label{eq:apperancemodel}
\end{equation}
where $m_{i,j}$ is the $j^{th}$ mixture weight at pixel $i$, and $\ve{M}_j$ is the $j^{th}$ material. The rendering error term for estimating spatial materials then becomes:
\begin{eqnarray}
 E_{\text{rend}}^{\text{mix}}(\ve{M}_1,\ldots,\ve{M}_k,\ve{N},\ve{L}, m) =  & \nonumber \\
 \hspace{-2mm} \sum_{i\in \text{pixels}} \sigma_i^{rend} \Big|\Big| I_i -  \hspace{-4mm} \displaystyle\sum_{j\in \text{materials}} m_{i,j} & f(\ve{M}_j,\ve{N}_i,\ve{L} ) \Big|\Big|^2.
\end{eqnarray}

We define three properties that the spatial maps ($m$) must adhere to: unity, firmness\footnote{We define the firmness prior as the decisiveness of mixture weights to snap to 0 or 1, and in this sense it has no relation to tactile properties.}, and smoothness. First, the unity prior ensures that the mixture weights must be nonnegative and sum to one at every pixel:
\begin{equation}
 \forall i,j \ \ m_{i,j} > 0, \hspace{5mm} \forall i \ \ \sum_{j} m_{i,j} = 1.
\end{equation} 
As noted by Goldman et al.~\cite{Goldman:tpami:2010}, this prevents overfitting and removes certain ambiguities during estimation.

We place another prior on the ``firmness'' of our mixture maps. For certain objects, many patches on the surface are dominated by a single material (e.g. checkerboard); for others, the surface is roughly uniform over space (e.g. soap can be made of a diffuse layer and a glossy film which are both present over the whole surface); there are even materials ranging in between (e.g. marble). We would like a structured way of controlling which type of spatial mixture we produce, and we do so by imposing an exponential prior on each mixture element:
\begin{equation}
E_{\text{firm}}^{\text{mix}}(m) = \sum_{i,j} m_{i,j}^\alpha,
\label{eq:firm}
\end{equation}
where $\alpha > 0$ controls how firm a mixture will be. For example, with the unity constraint, $\alpha>1$ encourages uniform mixture weights (not firm, e.g. soap), and  $\alpha<1$ encourages mixture weights to be near zero or one (firm, e.g. checkerboard). For results in this chapter, we use $\alpha=0.5$. Notice that for $\alpha<1$ this function is no longer convex, although in practice our optimization still seems to fare well. 
Our prior is more general (and controllable) than the method of Goldman et al.~\cite{Goldman:tpami:2010}, which assumes that each pixel is the linear combination of {\it at most} two materials.

Finally, we encourage spatial smoothness of the mixtures, as nearly all mixed-materials contain spatial structure:
\begin{equation}
E_{\text{smooth}}^{\text{mix}}(m) = \sum_{i,j} ||\nabla_x m_{i,j}|| + ||\nabla_y m_{i,j}||,
\label{eq:smoothness}
\end{equation}
where $\nabla_x$ and $\nabla_y$ are spatial gradient operators in the image domain. 

By inserting our new rendering term and mixture priors into the objective function for single materials (Eq~\ref{eq:obj}), we define a new optimization problem for estimating spatially varying materials:
\begin{align}
 \hspace{0mm} \displaystyle \underset{{\ve{M}_1,\ldots,\ve{M}_k,\ve{N},\ve{L}}}{\operatorname{argmin}} \ \ & E_{\text{rend}}^{\text{mix}}(\ve{M}_1,\ldots,\ve{M}_k, \ve{N},\ve{L},m) +  E_{\text{mat}}(\ve{M})  +\nonumber \\
& \hspace{-10mm} E_{\text{illum}}(\ve{L} ) + E_{\text{shape}}(\ve{N}) + E_{\text{firm}}^{\text{mix}}(m)  + E_{\text{smooth}}^{\text{mix}}(m),  \nonumber \\
& \hspace{-10mm}  \text{subject to:    \ \ \ } 0 \leq \ve{M}_j^{(i)} \leq 1,\ i\in\{1,\ldots,5\}, \ \forall j, \nonumber \\
& \hspace{8.5mm}  \forall i,j \ \ m_{i,j} > 0, \nonumber \\ 
& \hspace{8.5mm} \forall i \ \ \sum_{j} m_{i,j} = 1.
\label{eq:obj_mix}
\end{align}
Solving this objective function can be difficult, but we have had success using constrained quasi-Newton methods (L-BFGS Hessian).

Our optimization results in a decomposition of the input image into $k$ materials $\ve{M}$, a set of per-pixel weights for each material $m$, per-pixel surface normals $\ve{N}$, and illumination parameters $\ve{L}$. In this work, we focus on the correctness of our mixture materials and their applications.

\begin{figure*}[t]
\includegraphics[width=.32\linewidth]{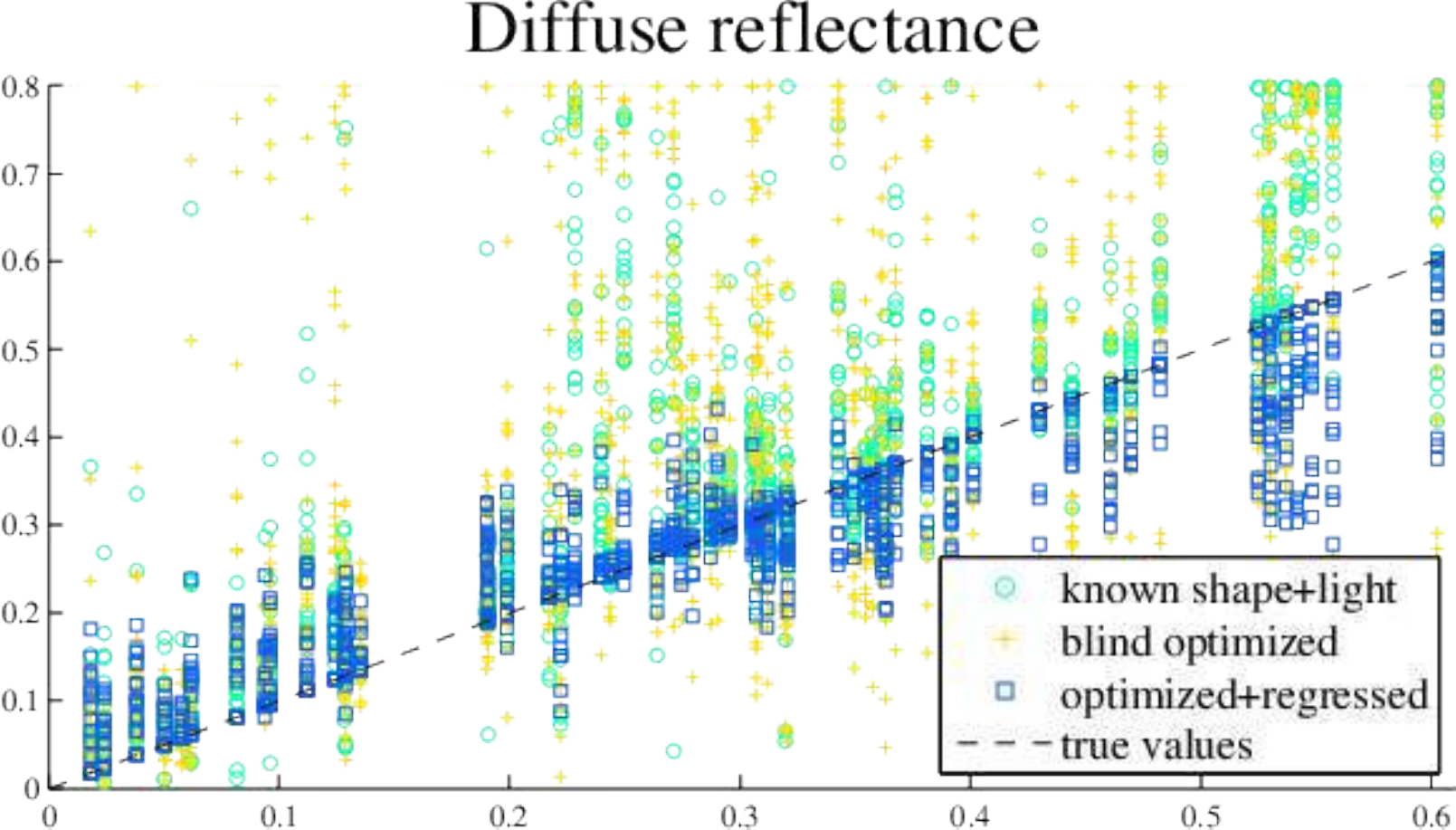}
\includegraphics[width=.32\linewidth]{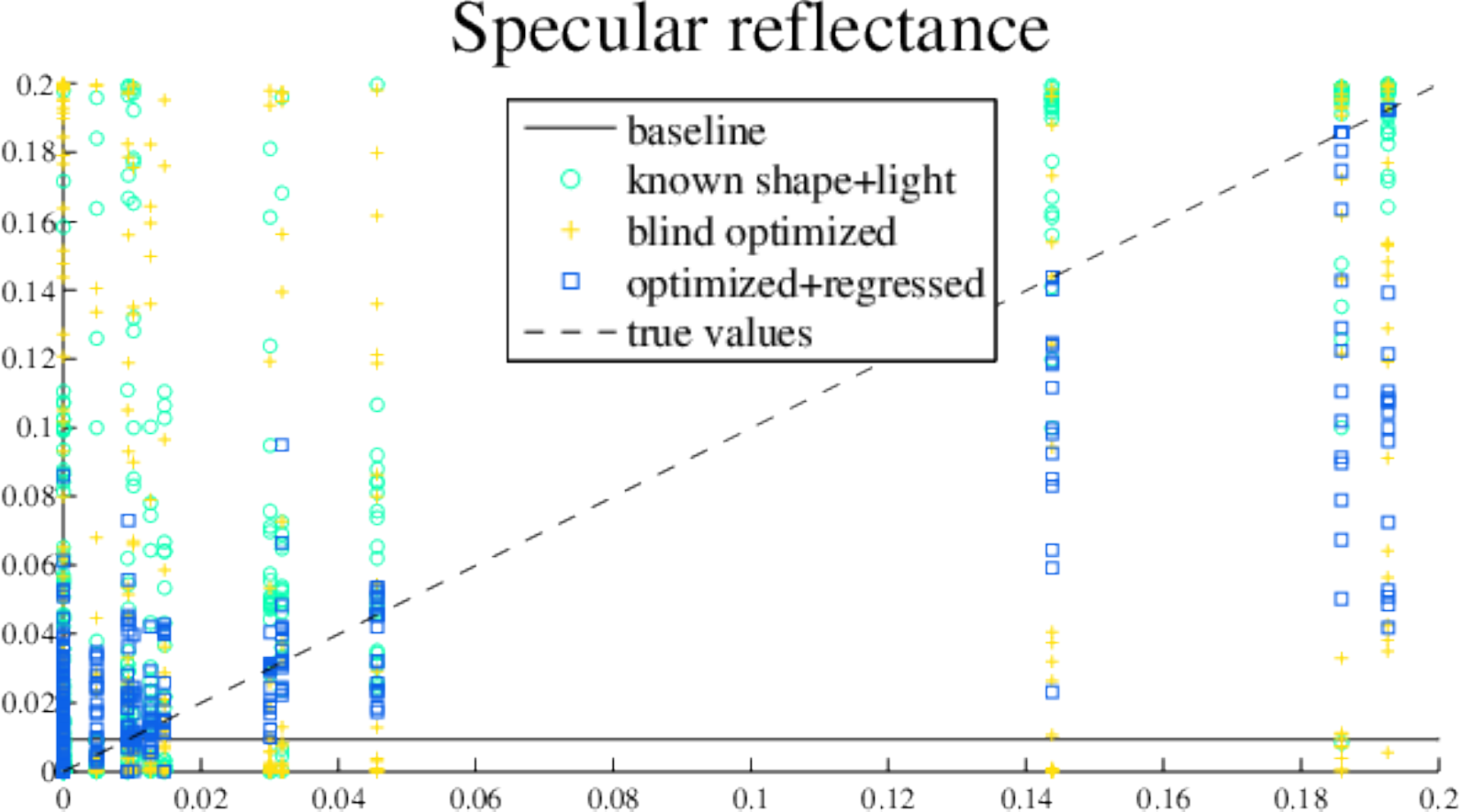}
\includegraphics[width=.32\linewidth]{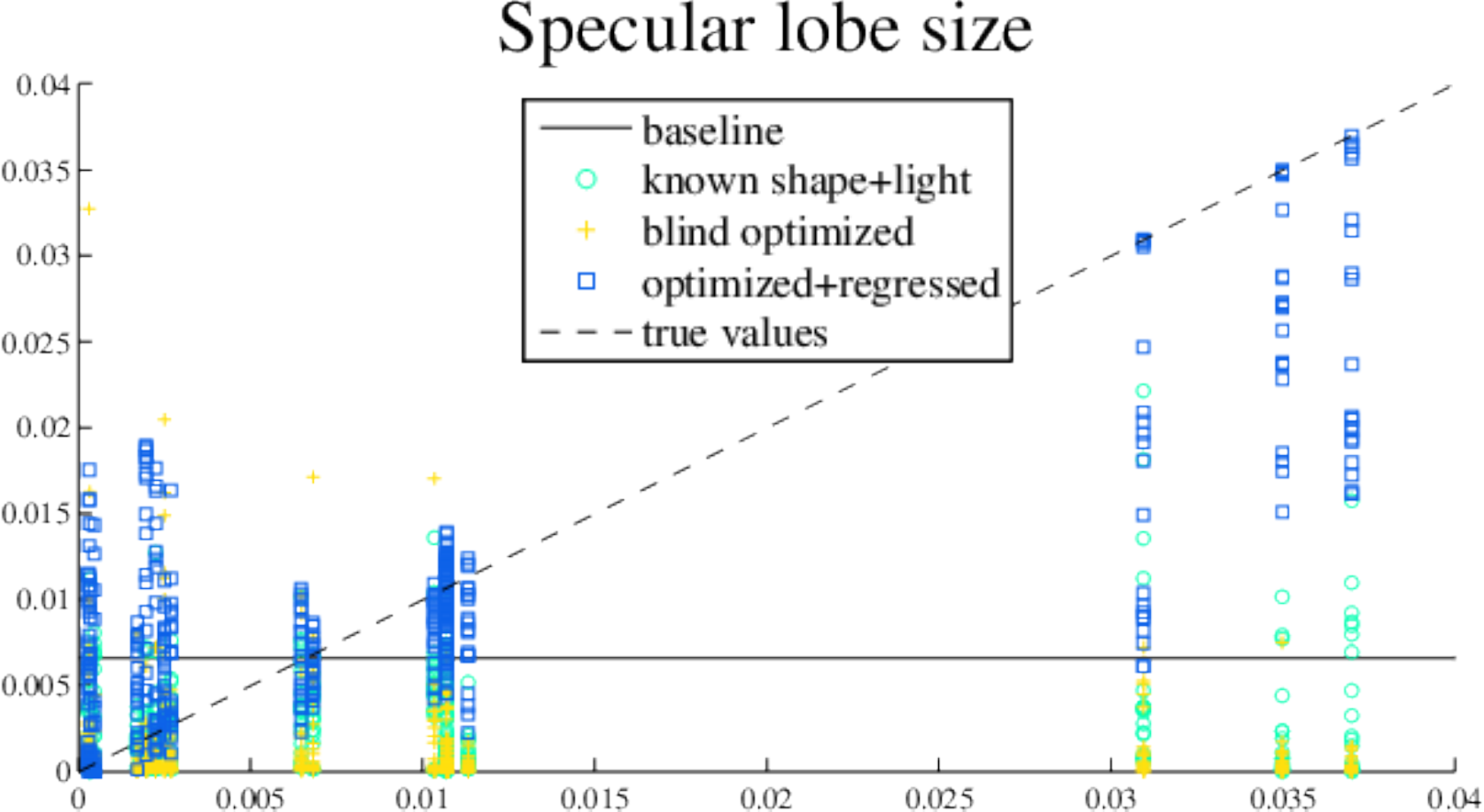}
\caption{Errors in material estimates for each image in our dataset. Each plot shows the true material value on the horizontal axis plotted against our estimate of diffuse reflectance ($R_d$), specular reflectance $R_s$, and specular lobe size $r$ (left to right). We show the results for our baseline, the material produced given accurate initial shape and lighting, our blind optimization technique (blind optimized), and the material regressed by un-biasing our optimization results (blind regressed); details in Sec~\ref{sec:materials:eval}.
}
\label{fig:mat_err}
\end{figure*}

\begin{figure*}[t]
\centerline{
\includegraphics[width=0.45\linewidth]{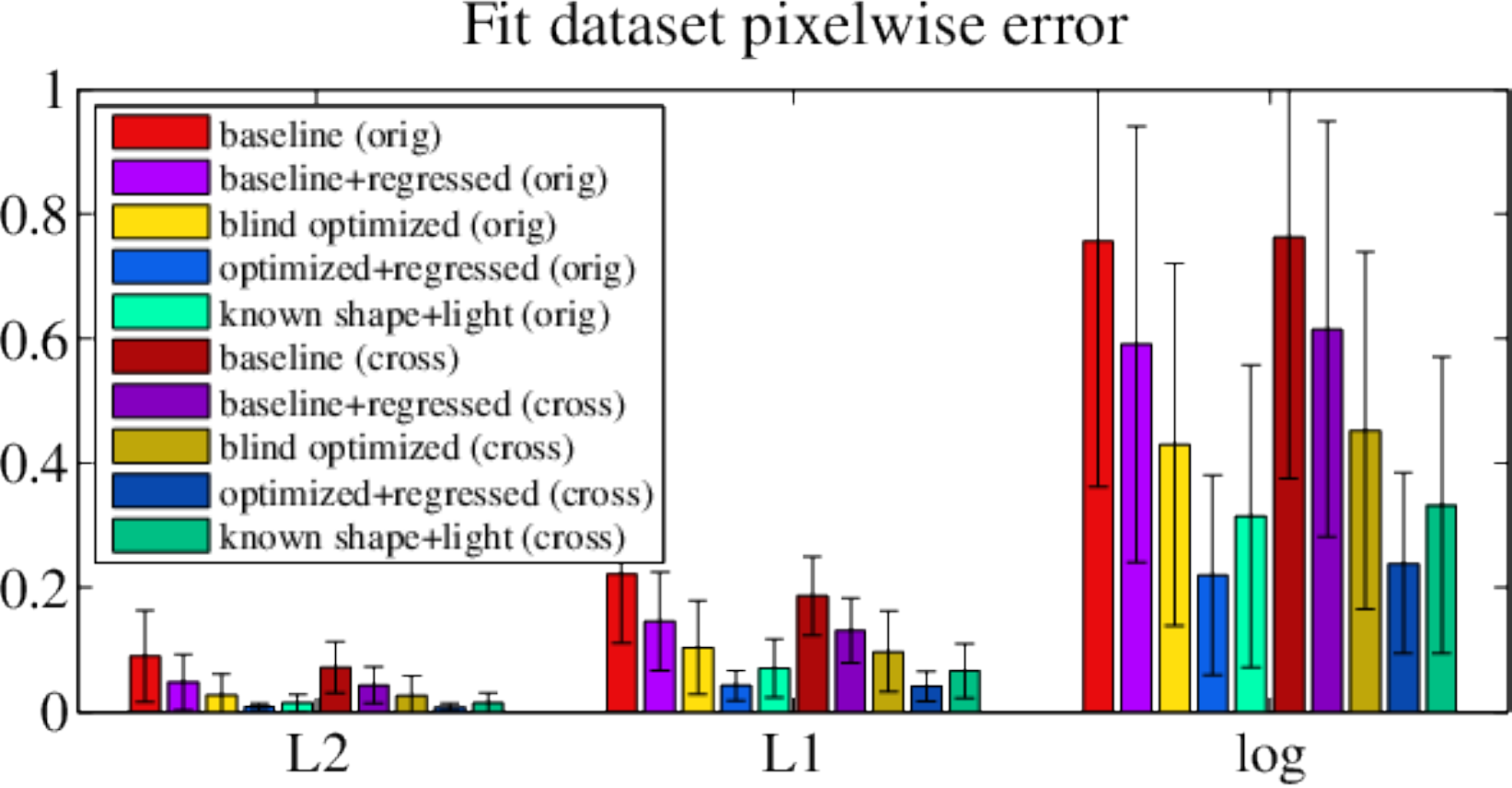}
\hspace{5mm}
\includegraphics[width=0.45\linewidth]{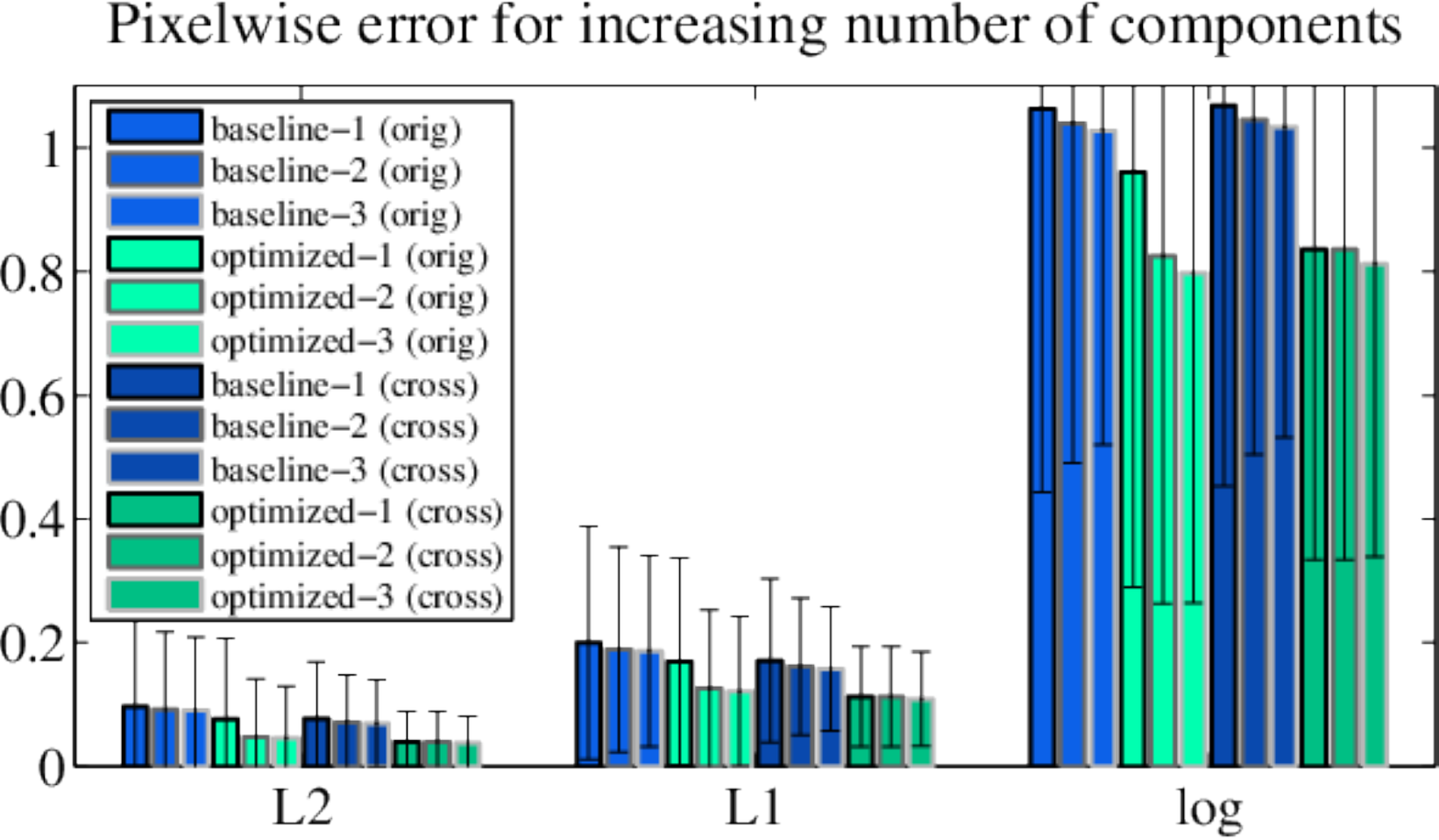}
}
\caption{We compare the average per-pixel error of the input image and a re-rendered image with estimated material ({\it but with the true shape and true lighting that produced the input image}) for various techniques and for both versions of our dataset; see Sec~\ref{sec:materials:eval} for details. We compute errors in the original illumination (orig), and averaged over six novel illumination environments (cross), for three different metrics: L2 and L1 norm, and the absolute log difference, and show the mean over the dataset (error bars indicate one standard deviation). Our full method (optimized+regressed) achieves low error relative to others. We also observe similar (yet slightly worse) error on our measured dataset, indicating that, for a variety of cases, our a) our method can handle real-world materials, and b) that our material model is capable of visually reproducing complex reflectance functions.
}
\label{fig:rendererror}
\end{figure*}

\section{Experiments}
\label{sec:materials:eval}

We evaluate the results of our method for objects with homogeneous (spatially uniform) materials in Section~\ref{sec:materials:eval:uniform}, as well as our inhomogeneous (spatially varying) material results in Section~\ref{sec:materials:eval:sv}. We report results for both a dataset we collected containing ground truth material information, as well as for the Drexel Natural Illumination dataset.

\subsection{Homogeneous materials}
\label{sec:materials:eval:uniform}

For evaluation and training our bias predictors (Sec~\ref{sec:bias}), we have collected a dataset consisting of  400 images rendered with real world shapes, materials, and illumination environments (all chosen from well-established benchmark datasets). We use the 20 ground truth shapes available in the MIT Intrinsic Image dataset~\cite{grosse09intrinsic}, and render each of these objects with 20 of the materials approximated from the MERL BRDF dataset~\cite{Matusik:sg:03}, for a total of 400 images. We use 100 different illumination environments (50 indoor, 50 outdoor) found across the web, primarily from the well known ICT light probe gallery\footnote{\urlwofont{http://gl.ict.usc.edu/Data/HighResProbes}} and the sIBL archive\footnote{\urlwofont{http://www.hdrlabs.com/sibl/archive.html}}. We ensure that each object is rendered in 10 unique indoor and 10 unique outdoor lighting environments, permuted such that each illumination environment is used exactly four times throughout the dataset. Each lighting environment is white balanced and has the same mean (per channel).

Our dataset has two ``versions.'' The first version of our dataset ({\bf fit dataset}) is rendered using our low-order reflectance model (we approximate MERL BRDFs by fitting our own 5-parameter material model to the measured data, and render using our fits). The resulting images are highly realistic, and allow us to both compare our material estimates with ground truth, and regress bias prediction functions (as in Sec~\ref{sec:bias}).

The second version ({\bf measured dataset}) is rendered using only {\it measured} BRDFs (from the MERL dataset); these images are truly realistic as the shape, material, and lighting are all sampled directly from real-world data. Furthermore, these images are synthesized using a physical renderer and thus include shadows and bounced light. This dataset gauges how well our method can generalize to real images and reflectances not encoded by our model.

\begin{figure*}
{\large \hspace{39mm} \underline{Fit dataset} \hspace{53mm} \underline{Measured dataset}}

{ \hspace{24mm} Best results \hspace{12mm} Median results \hspace{25mm} Best results \hspace{10mm} Median results \hspace{10mm} }\\
\begin{minipage}{.09\linewidth}
\hfill
  \begin{sideways}{\large \hspace{2mm} \underline{Novel illumination} \hspace{2mm} \underline{Original illumination}}\end{sideways}
  \begin{sideways}{ \hspace{0mm} true \hspace{2mm} opt+reg \hspace{4mm} opt \hspace{2mm}  true\small{(input)} \hspace{1mm} opt+reg \hspace{4mm} opt}\end{sideways}
  \hspace{2mm}
\end{minipage}
\begin{minipage}{.45\linewidth}
\includegraphics[height=\linewidth]{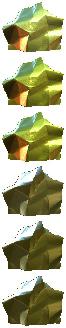}
\includegraphics[height=\linewidth]{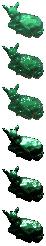}
\includegraphics[height=\linewidth]{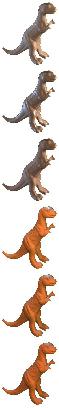}
\includegraphics[height=\linewidth]{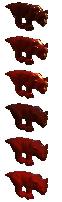}
\end{minipage}
\hfill
\begin{minipage}{.45\linewidth}
\includegraphics[height=\linewidth]{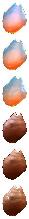}
\includegraphics[height=\linewidth]{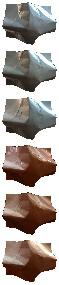}
\includegraphics[height=\linewidth]{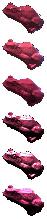}
\includegraphics[height=\linewidth]{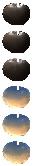}
\end{minipage}
\caption{Qualitative results on both versions of our dataset Materials are estimated using our blind optimized (opt) and optimized+regressed (reg) methods, and compared to ground truth (true). The true original illumination image is also the input for estimating material.  Notice that our technique can recover both glossy and matte materials, performs well even for these complex shapes. Our method attains visually pleasing results even for complex reflectance functions not encoded by our model (e.g. measured dataset) even in {\it new}  lighting conditions.
}
\label{fig:qualitative}
\end{figure*}

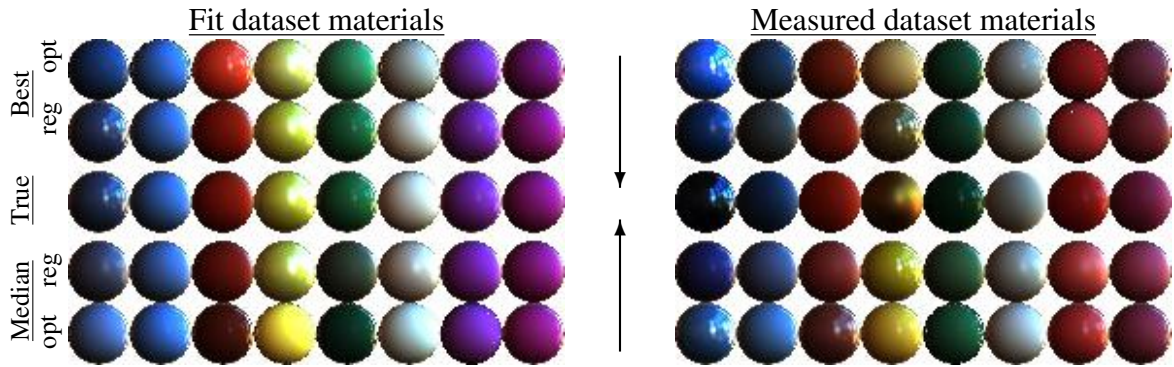
\begin{figure*}
\begin{minipage}{.10\linewidth}
\hfill
\begin{sideways}
{ \hspace{2mm} \underline{Median} \hspace{3mm} \underline{True} \hspace{6mm} \underline{Best} \hspace{8mm} }\end{sideways}
\begin{sideways}
{ \hspace{1mm} opt \hspace{2mm} reg \hspace{14mm} reg \hspace{2mm} opt }
\end{sideways}
\end{minipage}
\begin{minipage}{.40\linewidth}
\centerline{\large\underline{Fit dataset materials}}
\includegraphics[width=\linewidth]{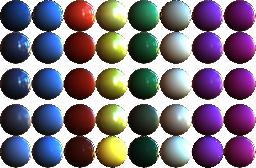}
\end{minipage}
\hfill
\begin{picture}(20,100)
\thicklines
\put(10,-60){\vector(0,1){50}}
\put(10,52){\vector(0,-1){50}}
\end{picture}
\hfill
\begin{minipage}{.40\linewidth}
\centerline{\large\underline{Measured dataset materials}}
\includegraphics[width=\linewidth]{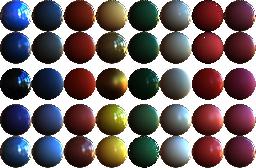}
\end{minipage}
\caption{Comparison of estimated materials rendered in novel lighting. The true materials lie on the middle row alongside our per-material best and median optimized (opt) and regressed (reg); arrows indicate the direction in which materials should improve. We achieve very good results for input images that are well described by our model in the fit dataset (rows 2 and 4 generally look like row 3), and even in many cases for measured BRDFs. However, low-order model bias prevents our method from capturing certain materials well (e.g. column 4; measured dataset).
}
\label{fig:spheres}
\end{figure*}

\boldhead{Results}
We generate results using the optimization procedure described in Sec~\ref{sec:est}, followed by our bias regression method as in Sec~\ref{sec:bias}. Bias prediction functions are learned through leave-one-out cross validation. 

In this section, we report results from our optimization technique ({\bf blind optimized}), and after bias regression ({\bf optimized+regressed}). For comparison, we compute a {\bf  baseline} material estimate which computes the $R_d$ by averaging the image pixels in each channel, and $R_s$ and $r$ from the average found in our material dataset, and then regress and apply bias predictors to the baseline estimates ({\bf baseline+regressed}). We also compare to materials achieved by our optimization assuming the shape and illumination are known\footnote{Known lighting is fit to our parameterization and may still be some distance from ground truth} and fixed ({\bf known shape+light}); hence only the material is optimized. Results using this procedure gauge the difficulty of our optimization problem, and shows how much our optimization can improve with more sophisticated initialization procedures.

On our fit dataset,  our full method (optimized + regressed) is capable of recovering highly accurate material parameters. Figure~\ref{fig:mat_err} plots the true material from our ``fit'' dataset against our estimated parameters for each of the 400 images. A perfect material estimate would lie along the diagonal (dashed line). Overall, we see a linear trend in our diffuse results, and that our bias regression can significantly improve our optimized estimates of specular reflectance and specular lobe size (and even better than shape+light).

We also develop two ways of measuring visual error in our materials. We define {\bf original illumination} as the average pixel error from comparing the input image with the image produced by rendering our {\it estimated} material onto the {\it true} shape in the {\it true} lighting (which are known for our all images in our dataset). This is a harsh test, as any errors in material must manifest themselves once rendered with the true shape and light. The second metric ({\bf cross rendered}) is even more telling: we compare renderings of the input object with the a) true material and b) our estimated material in six novel illumination environments not present in our dataset and compute average pixel error. This measure exposes material errors across unique, unseen illumination.

Using these measures, our full method achieves low error for both versions of our dataset. Figure~\ref{fig:rendererror} shows these error measures for three different metrics (per-pixel L2 and L1 norms, and absolute log difference), and optimized+regressed performs the best overall for both datasets.  This indicates that both our optimization and regression are crucial components, and one is not dominating inference since optimized+regressed consistently outperforms baseline+regressed. Known+shape light also performs well, indicating that our optimization procedure might improve if better initializations are available.

We demonstrate that in many cases our method can do a very well at visually reproducing both measured and fit reflectances, even in novel illuminations. We show qualitative results for both versions of our dataset in Figs~\ref{fig:qualitative} and~\ref{fig:spheres} -- these are some of our best and median results.  Our material estimates are typically visually accurate in original and novel illumination, even for many of the measured BRDFs in our measured dataset. We also observe that our regression generally helps for both datasets, indicating that our learned bias predictors may generalize to complex materials and real-world images. However, it is clear that our results degrade for complex reflectance functions that lie well outside our model (Fig~\ref{fig:spheres}, measured dataset columns 1+4).

\begin{figure*}
\begin{minipage}{.34\linewidth}
  \begin{minipage}{.49\linewidth}
  \centerline{Original light}
  \centerline{\small \hspace{3mm} true \hspace{4.5mm} estimated}
  \includegraphics[width=\linewidth]{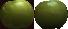}
  \centerline{ 
  \includegraphics[width=0.7\linewidth]{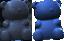}
  }
  \end{minipage}
  \begin{minipage}{.49\linewidth}
  \centerline{Novel light}
  \centerline{\small \hspace{3mm} true \hspace{4.5mm} estimated}
  \includegraphics[width=\linewidth]{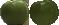}
  \centerline{ 
  \includegraphics[width=0.8\linewidth]{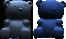} 
  }
  \end{minipage}
\end{minipage}
\hfill
\begin{minipage}{.25\linewidth}
  \begin{minipage}{.49\linewidth}
  \centerline{Original light}
  \centerline{\small \hspace{4mm} true \hspace{1mm} estimated}
  \includegraphics[width=\linewidth]{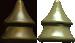}
 \centerline{ 
 \includegraphics[width=0.65\linewidth]{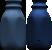}
 }
  \end{minipage}
  \begin{minipage}{.49\linewidth}
  \centerline{Novel light}
  \centerline{\small \hspace{4mm} true \hspace{1mm} estimated}
  \includegraphics[width=\linewidth]{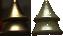}
  \centerline{
  \includegraphics[width=0.65\linewidth]{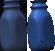}
  }
  \end{minipage}
\end{minipage}
\hfill
\begin{minipage}{.38\linewidth}
  \begin{minipage}{.49\linewidth}
  \centerline{Original light}
  \centerline{\small \hspace{2.5mm} true \hspace{4.5mm} estimated}
  \includegraphics[width=\linewidth]{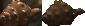}
  \includegraphics[width=\linewidth]{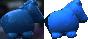}
  \end{minipage}
  \begin{minipage}{.49\linewidth}
  \centerline{Novel light}
  \centerline{\small \hspace{2.5mm} true \hspace{4.5mm} estimated}
  \includegraphics[width=\linewidth]{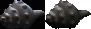}
  \includegraphics[width=\linewidth]{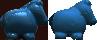}
  \end{minipage}
\end{minipage}
\caption{Results on real data from the Drexel Natural Illumination dataset. This dataset contains real images and corresponding ground truth shape and lighting information. We estimate materials from one picture, and render the material using the true shape and light for the original illumination and another illumination from the dataset (novel light); we compare to the real picture of the object in both scenes (original and novel). Even in the presence of slight spatial variation (e.g.  top left; apple) and complex reflectance (top middle) our method can still recover decent estimates. Still, addressing these issues is key to generalizing our method's applicability.
}
\label{fig:natgeom}
\end{figure*}

Finally, we demonstrate our method's capability on {\it real images} from the Drexel Natural Illumination dataset in Fig~\ref{fig:natgeom}. Our model appears somewhat robust to spatially varying reflectance in these images, but suffers from the complexity of the imaged reflectances and because we assume only a single material is present; this suggests ideas for future work.

\subsection{Inhomogeneous materials}
\label{sec:materials:eval:sv}

For ground truth evaluation, we use again use the {\bf measured dataset}. We use our mixture estimation procedure to estimate $k=\{2, 3\}$ materials\footnote{We use a spatial mixture for homogeneous materials as our mixture maps generalize current literature. They capture spatial variation in material (as in~\cite{Goldman:tpami:2010}), but we use them to also encode any kind of surface variation not well-captured due to long-standing SFS assumptions. E.g., integrability of normals doesn't allow for accurate recovery of certain detail (see Fig~\ref{fig:confusion} ``fiber''); lighting models are imprecise; direct lighting is assumed; material models can't capture subsurface scattering/iridescence/etc. A linear combination of these materials produces similar effects that seem {\it perceivably} good enough for our applications.} for each dataset image, and compare to the results in of our method for $k=1$. For additional comparison, we compute a baseline material estimate by clustering the image into $k$ components (using $k$-means); computing diffuse albedo (per component) by averaging the image pixels in each channel, and the specular components are fixed to a small yet reasonable value.

We measure error by rendering our {\it estimated} material onto the {\it true} shape in the {\it true} lighting (which are known for our all images in the dataset), and compare this to the input image. We do the same test, but for six novel lighting environments not found in the dataset (e.g. estimated material versus true material in novel light). We denote these as ``orig'' and ``cross'' lighting respectively. These are harsh tests of generalization, as any errors in material must manifest themselves once rendered with the true shape and light, and the ``cross'' measure exposes material errors across unique and unseen illumination.

\begin{figure}
\centerline{\includegraphics[width=.6\linewidth]{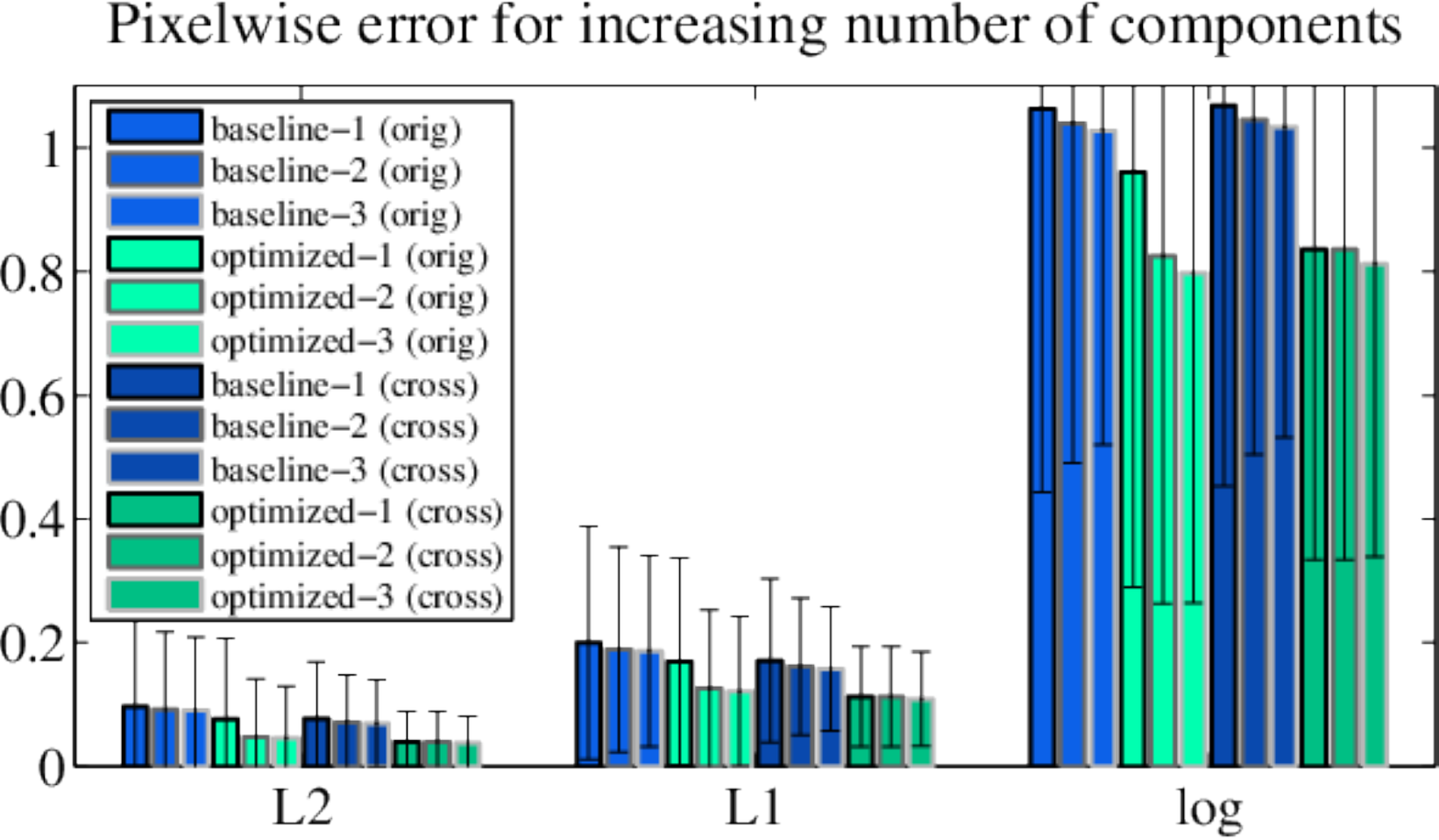}}
\caption{Quantitative results on our ``measured'' dataset. We measure error as the pixelwise difference from comparing a rendering our {\it estimated} material(s) and a rendering of the true, measured BRDF in the {\it true} original lighting (orig) and novel lighting (cross) using {\it true} shape. Errors are reported for L2 and L1 norms, and the absolute log difference. Our mixture materials (optimized-$\{2,3\}$) consistently outperform single material estimation (optimized-$1$), and are always better than the baseline estimates (see text for details).
}
\label{fig:quantPBRTdata}
\end{figure}

\begin{figure*}[t]
\begin{minipage}{\linewidth}
  \hfill
  \begin{minipage}{.45\linewidth}
  \centerline{\Large Original light}
  \centerline{\hspace{0mm} 1 material \hspace{3mm} 2 materials \hspace{3mm} 3 materials \hspace{7.5mm} true \hspace{5mm}}
  \includegraphics[width=\linewidth]{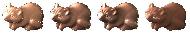}
  \includegraphics[width=\linewidth]{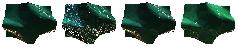}
  \includegraphics[width=\linewidth]{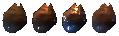}
  \includegraphics[width=\linewidth]{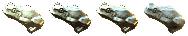}
  \end{minipage}
  \hfill
  \begin{minipage}{.45\linewidth}
  \centerline{\Large Novel light}
  \centerline{\hspace{0mm} 1 material \hspace{3mm} 2 materials \hspace{3mm} 3 materials \hspace{7.5mm} true \hspace{5mm}}
  \includegraphics[width=\linewidth]{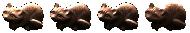}
  \includegraphics[width=\linewidth]{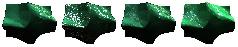}
  \includegraphics[width=\linewidth]{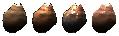}
  \includegraphics[width=\linewidth]{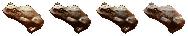}
  \end{minipage}
  \hspace*{\fill}
\end{minipage}
\caption{Best (rows 1,2) and median (rows 3,4) results from the ``measured dataset'' which contains physically rendered rendered objects with measured BRDFs. Typically these materials are not well encoded by our low-order material model with 1 mixture component, but by increasing the number of mixture components, we extend our model's ability to generalize to complex reflectance functions. We show our estimated materials for one, two, and three mixture components, and compare these to the ground truth result (also the input image) in both the original and novel illumination environments. In nearly every case (and not just these examples), adding more mixture components improves the estimate. The improvement between two and three components is not as drastic as the improvement between one and two, coinciding with quantitative results in Fig~\ref{fig:quantPBRTdata}.}
\label{fig:qualPBRTdata}
\end{figure*}

Fig~\ref{fig:quantPBRTdata} shows quantitative results averaged over the entire dataset for L2, L1, and absolute log difference error metrics. Our mixture materials (optimized-$\{2,3\}$) consistently outperform single material estimation (opt- imized-$1$), and are always better than the baseline estimates. 

We observe a similar trend in our qualitative results (Fig~\ref{fig:qualPBRTdata}). Because we are attempting to estimate true, measured BRDFs which may lie outside of our 5-parameter material model, estimation may not work well with a single material. However, by adding multiple materials, we typically get improved results, even in novel illumination. This indicates that our mixture weights are typically robust to shading artifacts such as shadows and specularities. It is clear that adding more components helps, although the distinction between $k=2,3$ is subtle (both qualitatively and quantitatively).

\section{Applications}
We show two applications of our decomposition: material classification and appearance transfer/generation.

\begin{figure}
\includegraphics[width=\linewidth,clip=true, trim=55 75 0 0]{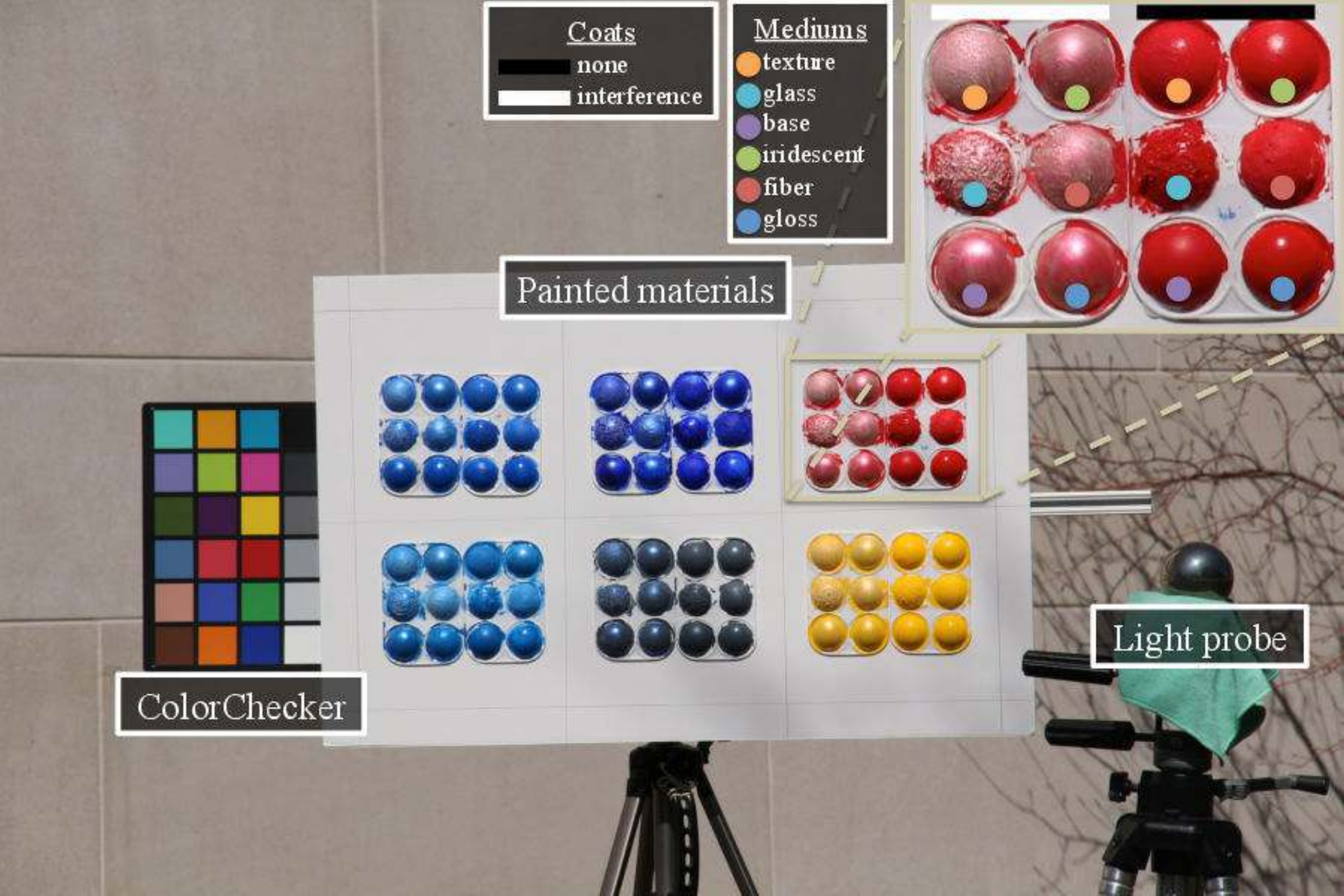}
\caption{Our setup for capturing our material dataset. We create 72 materials by taking all combinations of six colors, two coats, and six mediums. Each 3$\times$4 block of swatches is a unique color, and within each block (enlarged in the top right), there are six swatches with no coat (left half), and six with a translucent ``interference'' coat (right half); for each coat, there are six different mediums. We imaged this grid of materials in three distinct illumination conditions (both indoor and outdoor) and at nine viewing angles per scene. Each image is captured at three different exposures, and contains a color checker and light probe for acquiring true illumination (although we do not use HDR or true lighting for estimating materials). Best viewed in color at high resolution. }
\label{fig:rig}
\end{figure}

\subsection{Automatic material classification}
\label{sec:classify}
We hypothesize that our material decompositions provide some intrinsic information about the picture (including, by our definition, micro- and macroscopic features). To test this, we use our decompositions to derive features for image classification, and evaluate this task on our new material dataset containing fine distinctions at the microscopic, macroscopic, and mesoscopic scales. Fig~\ref{fig:rig} shows the materials and setup for imaging the the dataset. There are six unique paint colors, six mediums (texture, glass, base, iridescent, fiber, gloss), and two coats (none, interference). By taking all combinations of these (colors by mediums by coats), we create 72 unique materials, and paint them onto curved\footnote{In a single photo, flat surfaces typically contain significantly less information about a material due to the limited coverage of surface normals. This is also one reason we do not use existing datasets such as CURET~\cite{CAVE_0060}.} surfaces (the underside of paint palettes, which are flattened hemispheres). We photograph the materials in three scenes (outdoors with strong sunlight; indoors with strong direct light; and indoors with diffuse and bounced light) and at nine different viewing angles per scene; each material appears 27 times in the dataset.

\begin{figure*}[t]
\includegraphics[width=\linewidth]{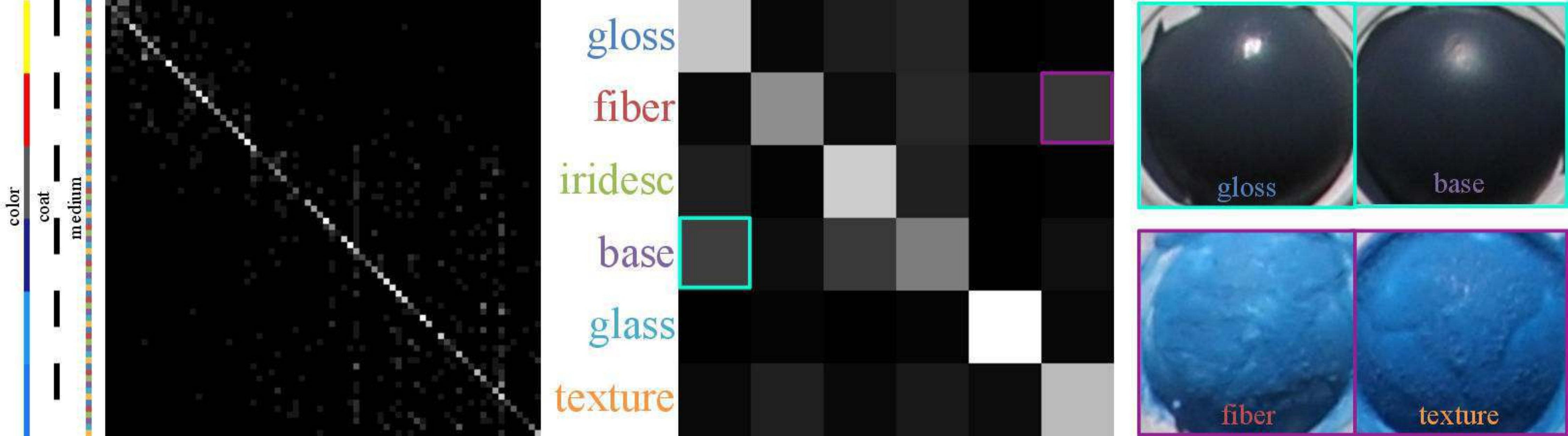}
\centerline{\hspace{22mm} One vs all (+mat: 47.5\%) \hspace{22mm} Medium (+mat+N+mix: 65.0\%)\hspace{12mm} Confusion examples \hspace*{13mm}}
\caption{Confusion matrices (normalized for contrast) for two of our classification tasks and the average classification accuracy (averaged over ten training folds). One versus all classification tries to classify a given material swatch as having a particular color, coat, and medium (72 classes), and medium classification attempts to predict which medium a given swatch contains (6 classes). Most confusion occurs between objects of similar color, or when two mediums look similar in different lighting. Several swatches are shown on the right; these are two examples of classes which were confused by our classifier. Best viewed in color. }
\label{fig:confusion}
\end{figure*}

\begin{table}
{ \centerline{\LARGE Material Classification Accuracy}
\centerline{
\large
\begin{tabular}{| l || c | c | c |} \hline
& \Large coat & \Large medium & \Large one vs all \\ \hline 
 \Large img & .90 $\pm$ .01 & .63 $\pm$ .02 & .44 $\pm$ .045  \\ \hline 
 \Large +mat & .91 $\pm$ .01 & .65 $\pm$ .01 & {\bf .48 $\pm$ .04} \\ \hline 
 \Large +mat+N & {\bf .92 $\pm$ .01} & .65 $\pm$ .01 & .46 $\pm$ .04\\ \hline
 \Large +mat+mix & .90 $\pm$ .01 & .64 $\pm$ .02 & .42 $\pm$ .02 \\ \hline 
 \Large +mat+N+mix & .91 $\pm$ .01 & {\bf .65 $\pm$ .01} & .41 $\pm$ .03 \\ \hline 
\end{tabular}
}
\vspace{1mm}
}
\caption{Classification results for our three tasks (coat, medium, and one versus all classification).  We generate new image descriptors based on the results of our material estimation (see text for details), and use these to help classify image swatches in our dataset. Our descriptors (+mat*) consistently outperform the baseline descriptor (img), and these improvements are statistically significant for coat and medium classification ($p=0.0064$ and $0.0174$ respectively using a two-sample t-test). Mixture weights sometimes degrade classification accuracy, most likely because our method estimates spatial mixture weights consistently for different mediums across different colors/coats (bottom rows, columns 1+3).
}
\label{tab:classificationQuant}
\end{table}

One might (correctly) point out that all of these ``materials'' are actually just the same material: paint! However, upon even a brief inspection, it is clear that each swatch can be distinguished from all other swatches (and some differences are more pronounced than others). Because of these fine distinctions, we expect that current methods will have a hard time correctly classifying these swatches, and propose three challenging tasks: a) {\bf coat}, b) {\bf medium}, and c) {\bf one versus all} classification. More specifically, given an image, we attempt to automatically predict a) whether it has an interference coat or not, b) which medium it contains, and c) precisely which color, medium, and coat was used to create the material.

\begin{figure*}
\begin{center}
\begin{minipage}{1.2\linewidth}
\begin{minipage}{.16\linewidth} \hfill \end{minipage}
\begin{minipage}{.03\linewidth} \hfill \end{minipage}
\begin{minipage}{.64\linewidth}
  \begin{center}  \hspace{0.09\linewidth}
  glass \hfill base \hfill  texture \hfill fiber \hspace*{0.09\linewidth} \\
  \hspace{0.07\linewidth}
  \includegraphics[width=0.1\linewidth]{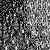} \hfill
  \includegraphics[width=0.1\linewidth]{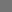} \hfill
  \includegraphics[width=0.1\linewidth]{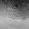} \hfill
  \includegraphics[width=0.1\linewidth]{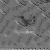} 
  \hspace*{0.07\linewidth}
  \end{center}
\end{minipage} \\
\begin{minipage}{.16\linewidth}
       \begin{overpic}[width=0.79\linewidth]{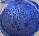}
      \put(1,3){ \color{white}cerulean-glass}
      \end{overpic} \\
       \begin{overpic}[width=0.79\linewidth]{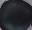}
      \put(8,3){ \color{white}gray-base}
      \end{overpic} \\
       \begin{overpic}[width=0.79\linewidth]{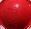} 
      \put(7,3){ \color{white}red-texture}
      \end{overpic} \\
       \begin{overpic}[width=0.79\linewidth]{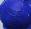}
      \put(11,3){ \color{white}blue-fiber}
      \end{overpic} \\
 \end{minipage}
\begin{minipage}{.03\linewidth}  
    \hfill
  \begin{sideways}
      \hspace{0mm} blue \hspace{18mm} red \hspace{17mm} gray \hspace{12mm} cerulean
  \end{sideways}
 \hspace*{2mm}
\end{minipage}
\begin{minipage}{.64\linewidth}
\vspace{-4mm}
  \includegraphics[width=\linewidth,clip=true, trim=0 240 320 0]{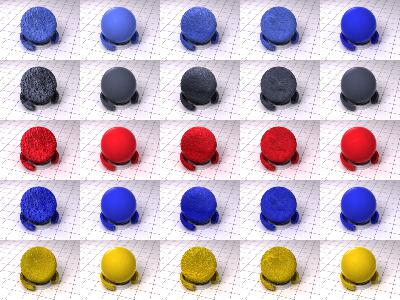}
\end{minipage}
\end{minipage}
\end{center}
\caption{Material transfer and generation for four material swatches in our dataset (gloss and iridescent mediums are excluded as they have little to no spatial structure and appear similar to ``base''). We decompose {\it single images} (on left) into two material components and a spatial mixture map. Then, we synthesize new materials by taking all combinations of the inferred materials and the derived mixture weights, and render these combinations onto spheres in novel illumination (using LuxRender: \urlwofont{http://luxrender.net}). Images along the diagonal show a transfer material result for a given picture on the left. The off-diagonals show the generative capabilities of our algorithm: by combining multiple decompositions (materials + mixing weights), we can generate new, unseen materials. We expect that full 3D textures will give better results, but it is currently impossible to estimate 3D textures from a single picture. Best viewed in color at high resolution.
}
\label{fig:renderStealGrid}
\end{figure*}

As a baseline for classification, we generate a bag of words model using PHOW features extracted from the swatch images and train a kernel SVM (linearized with a homogeneous kernel map~\cite{vedaldi2012efficient}) using VLfeat~\cite{vedaldi08vlfeat} for each class (two classes for coat classification; six for medium classification; and 72 for one vs all). We denote the baseline descriptor as {\bf img}, and create the following new descriptors by supplementing the baseline descriptor with results from our decomposition. We use material parameters ({\bf+mat}), material and surface normals ({\bf+mat+N}), material and spatial mixing weights ({\bf+mat+mix}), and material, normals, and mixing weights ({\bf+mat+N+mix}). Material parameters are concatenated into a feature vector, and a new bag of words model using PHOW features are trained for both the normals and mixing weights (reshaped into images with 3 and $k$ channels respectively). The final descriptor for a given method is the concatenation of all BoW histograms and material parameters (re-weighted appropriately). We train each classifier using half of the dataset and test on the other half and perform 10-fold cross validation. For all of our estimation results, we use two mixture components ($k=2$).

Results are shown in Table~\ref{tab:classificationQuant}. Our features clearly aid in this task, and outperform the baseline (img) in almost every instance.  The differences between our best results and the baseline are statistically significant for coat and medium classification ($p=0.0064$ and $0.0174$ respectively using a two-sample t-test), but not for one versus all ($p=0.0866$; assuming a 5\% significance threshold). 

A notable case where our features degrade classification is when adding mixture maps to the one vs all task. There is perhaps a good reason for this: a red fiber swatch should be classified differently than a gray fiber swatch (for example), but using mixture maps as features might make classification worse because each fiber swatch should have roughly the same mixture map (overemphasizing feature similarity).

 Fig~\ref{fig:confusion} shows the confusion matrix for one versus all and medium classification, as well as two examples of confused swatches in the medium case. Most confusion occurs between objects of similar color, or when two mediums look similar in different lighting.

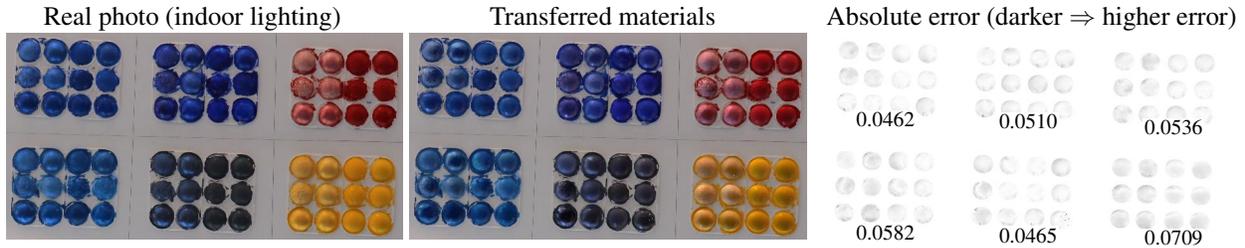
\begin{figure*}
\begin{minipage}{1.0\linewidth}
\hspace{5mm}Real photo (indoor lighting) \hspace{18mm} Transferred materials \hspace{13mm} Absolute error (darker $\Rightarrow$ higher error)  \\
\includegraphics[width=0.32\linewidth]{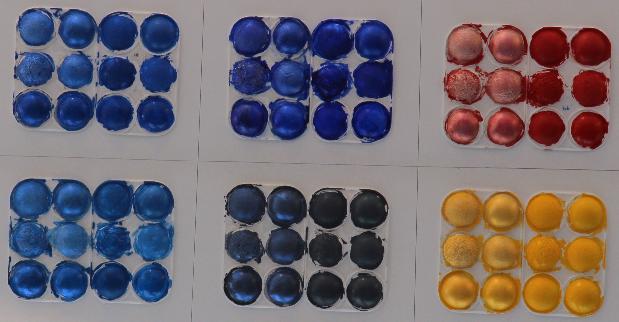}
\includegraphics[width=0.32\linewidth]{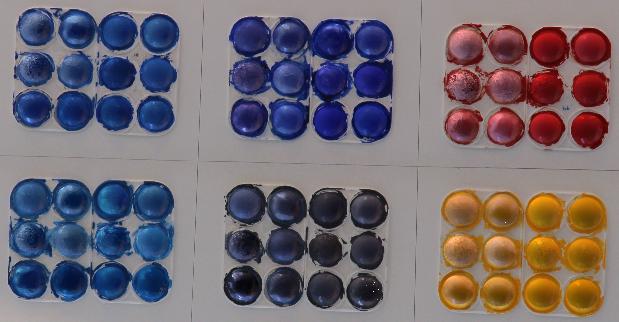}
\setlength\fboxrule{0pt}
\setlength\fboxsep{0.5pt}
\begin{overpic}[width=0.32\linewidth]{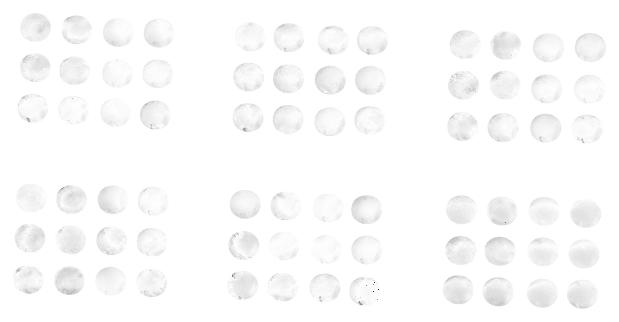}
\put(11,43){\fcolorbox{black}{white}{\footnotesize0.0462}}
\put(65,42){\fcolorbox{black}{white}{\footnotesize0.0510}}
\put(120,40){\fcolorbox{black}{white}{\footnotesize0.0536}}
\put(11,-0){\fcolorbox{black}{white}{\footnotesize0.0582}}
\put(65,-1){\fcolorbox{black}{white}{\footnotesize0.0465}}
\put(120,-2){\fcolorbox{black}{white}{\footnotesize 0.0709}}
\end{overpic}
\vspace{2mm}
\end{minipage}
\caption{Material transfer results on our dataset. We estimate materials for all 72 swatches in two of the dataset photos (indoors--left picture and outdoors--shown in Fig~\ref{fig:rig}) We transfer the materials estimated from the outdoor image onto the indoor image (keeping the material weights from the indoor image the same), and render the result using captured light probe data from out dataset (middle). On the right is a visualization of the absolute error between the left and middle images (average absolute pixel error displayed under each block). These results show our algorithm's ability to estimate correct materials that render well in unseen and substantially different illumination.  Best viewed in color at high resolution.
}
\label{fig:renderPalettes}
\end{figure*}

\subsection{Material transfer and generation}
\label{sec:generate}
Once we have decomposed an image into its materials and spatial mixing weights, we can apply this intrinsic material information to new surfaces as in Fig~\ref{fig:materials:teaser}. Applying the materials (microstructure) to a novel object is straightforward, but transferring the mixture weights (macrostructure) can be challenging in certain cases (e.g. when a mapping from one surface to another is not easily computed). 

We propose a straightforward solution: choose a small patch of the image defined by the mixture weights that is nearly fronto-parallel (determined from our predicted surface normals; to avoid foreshortening), and synthesize a larger texture (seeded with the small patch) using existing methods; e.g.~\cite{Efros99}.  Then, map the surface of the object that the material will be transferred to onto a plane (also using existing methods; e.g.~\cite{Sheffer:2006}); this mapping defines correspondences between the synthesized mixture weights and the new mesh. We generate all of our transfer/generation results using this technique, and more sophisticated methods are clear directions for future work.

We also propose a generative material modeling strategy: besides transferring a complete mixture material, we can combine estimates from multiple images to create new materials (e.g. materials from one and mixture weights from another, and so on).

Generative results (as well as direct transfer results) are shown in Fig~\ref{fig:renderStealGrid}. We have decomposed four swatches from our dataset (all unique colors and mediums and spanning the three illumination environments in our dataset) using $k=2$ mixture components. We apply each set of materials to each synthesized mixture, and render the result onto spheres.  We assert that our estimated materials correspond to microstructure and mixing weights correspond to macrostructure, which appears correct for these results (microstructure varies vertically, macrostructure horizontally). 


We also show transfer results on our material dataset in Fig~\ref{fig:renderPalettes}. For these results, only material is transferred. We estimate all 72 materials from two images of our dataset (one photographed outdoors, one photographed indoors), and then transfer the 72 materials from the outdoor photo to the indoor photo (keeping mixture weights from the indoor photo). We render the results using the illumination captured for the indoor image using the light probes in our dataset, and display the error alongside. This is a very demanding task, and shows how well our estimates generalize when transferred to radically different illumination.

\section{Conclusion}

We have demonstrated a new technique for estimating spatially varying parametric materials from an image of a single object of unknown shape in unknown illumination, going beyond the typical Lambertian assumptions made by existing shape-from-shading techniques. Strong priors and low-order parameterizations of lighting and material are key in providing enough constraints to make this inference tractable. Such rigid parameterizations often lead to estimation bias, and we also present a simple yet powerful technique for removing this bias.

\begin{figure}
\begin{center}
\begin{minipage}{0.8\linewidth}
\begin{minipage}{0.32\linewidth}  \centerline{ Input}
 \includegraphics[width=1\linewidth]{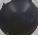} 
  \end{minipage}
\begin{minipage}{0.32\linewidth}  \centerline{Rendered}
 \includegraphics[width=1\linewidth]{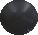}
 \end{minipage}
\begin{minipage}{0.32\linewidth} \centerline{ Mixing weights}
 \includegraphics[width=1\linewidth]{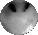} 
  \end{minipage} \\
  \centerline{\hrulefill}
\includegraphics[width=\linewidth]{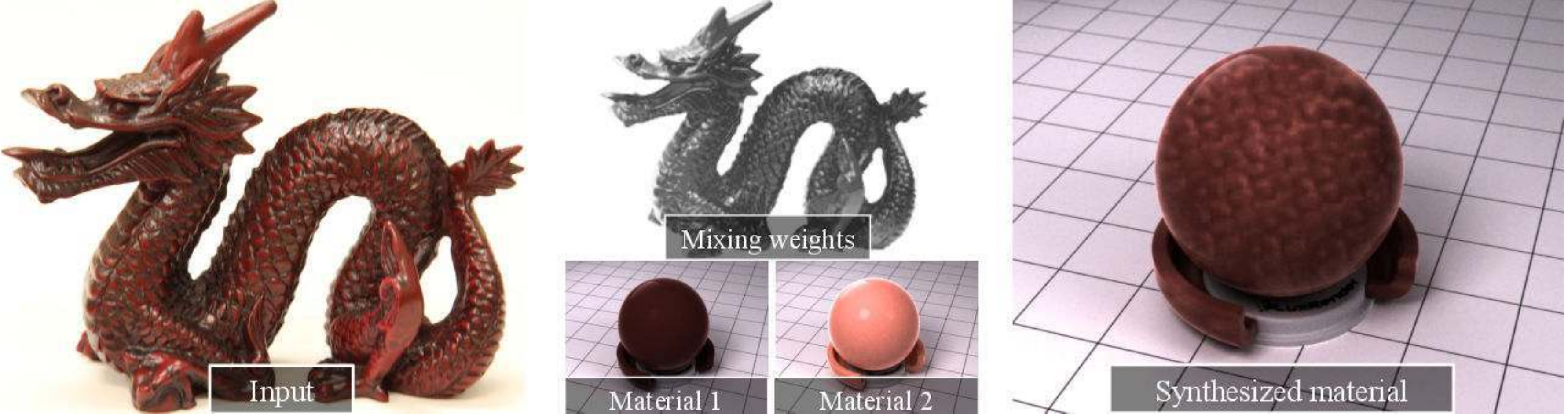}
\end{minipage}
\end{center}
\caption{Failure examples. The top row demonstrates an incorrect mixture map estimate: specularities have been detected as a separate material. Our resulting material looks correct in the original light, but uses two diffuse materials to describe a single shiny material. This typically occurs a) when too many mixture components are used or b) if the firmness exponent ($\alpha$, Eq.~\ref{eq:firm}) is set too low; better priors may discourage this. The bottom row shows material transfer for the dragon image, but the result looks too flat; adjusting the the mesostructure of the rendered surface may help.
}
\label{fig:failure}
\end{figure}

Our results suggest that material recovery is not necessarily dependent upon the joint recovery of accurate shape and illumination; {\it as long as the shape and illumination are consistent with each other, materials can still be robustly estimated}. This is encouraging from a material inference standpoint, as even the best shape-from-shading algorithms still produce flawed estimates in many scenarios.

As far as we know, our method is the first to estimate parametric material models without assuming shape or illumination is known a priori. We believe that our method provides good initial evidence that solving this problem is in fact feasible, and provides a foundation for estimating materials from photographs alone.

Our material decompositions enable new applications in both the realms of computer vision and computer graphics. We have shown that our decompositions are discriminative, and can be used to generate descriptors that improve automatic classification techniques. Classification results are computed on our new, challenging dataset of materials, and we attain reasonably accurate performance.

We also show that our decompositions can be transferred to new shapes, imbuing them with similar appearance as the input image. Furthermore, our decompositions are also generative, and can be used to create new materials by simultaneously transferring decompositions from multiple objects (e.g. mixing weights from one, materials from another). Our re-rendering results do not incorporate any information from our estimated surface normals, and the spatial frequency of our mixture weights are defined by the input image resolution (some artifacts visible in Fig~\ref{fig:materials:teaser}); intelligently incorporating and up sampling these estimates are reasonable directions for future work.


\chapter{Automatic illumination inference for physically grounded image editing} \label{chap:illum}

\section{Introduction} 

\begin{figure}[t]
\begin{center}
\includegraphics[width=.495\linewidth]{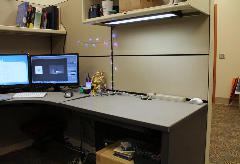}
\includegraphics[width=.495\linewidth]{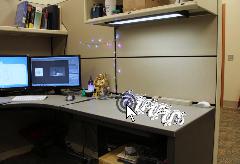}\\
\includegraphics[width=.495\linewidth]{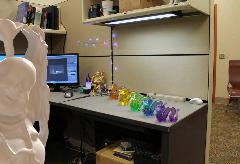}
\includegraphics[width=.495\linewidth]{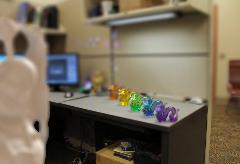}
\end{center}
\vspace{-4mm}
\caption{From a single LDR photograph, our system automatically estimates a 3D scene model without any user interaction or additional information. These scene models facilitate photorealistic, physically grounded image editing operations, which we demonstrate with an intuitive interface. With our system, starting from a legacy photograph (top left), a user can simply drag-and-drop 3D models into a picture (top right), render objects seamlessly into photographs with a single click (bottom left), adjust the illumination, and refocus the image in real time (bottom right). Best viewed in color at high-resolution.}
\label{fig:autoillum:teaser}
\end{figure}

\begin{figure*}
\centerline{\includegraphics[width=0.90\textwidth]{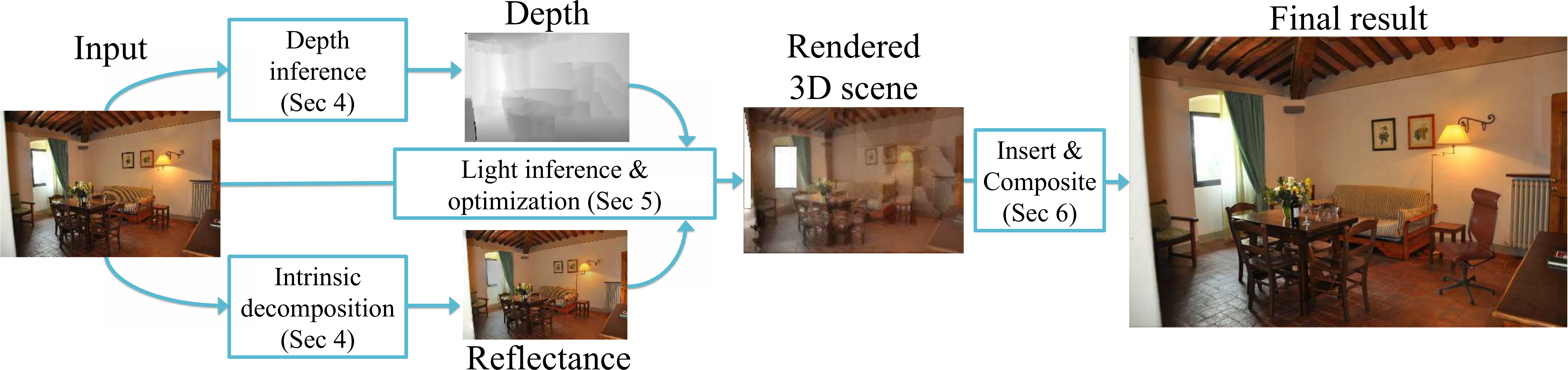}}
\vspace{-2mm}
\caption{Our system allows for physically grounded image editing (e.g., the inserted dragon and chair on the right), facilitated by our automatic scene estimation procedure. To compute a scene from a single image, we automatically estimate dense depth and diffuse reflectance (the geometry and materials of our scene).  Sources of illumination are then inferred without any user input to form a complete 3D scene, conditioned on the estimated scene geometry. Using a simple, drag-and-drop interface, objects are quickly inserted and composited into the input image with realistic lighting, shadowing, and perspective. {\it Photo credits: {\footnotesize \textcopyright{}}Salvadonica Borgo.}
}
\label{fig:overview}
\end{figure*}

Many applications require a user to insert 3D characters, props, or other synthetic objects into images. In many existing photo editors, it is the artist's job to create photorealistic effects by recognizing the physical space present in an image. For example, to add a new object into an image, the artist must determine how the object will be lit, where shadows will be cast, and the perspective at which the object will be viewed. In this chapter, we demonstrate a new kind of image editor -- one that computes the physical space of the photograph automatically, allowing an artist (or, in fact, anyone) to make physically grounded edits with only a few mouse clicks.

Our system works by inferring the physical scene (geometry, illumination, etc.) that corresponds to a single LDR photograph. This process is fully automatic, requires no special hardware, and works for legacy images.  We show that our inferred scene models can be used to facilitate a variety of physically-based image editing operations. For example, objects can be seamlessly inserted into the photograph, light source intensity can be modified, and the picture can be refocused on the fly. Achieving these edits with existing software is a painstaking process that takes a great deal of artistry and expertise. 

In order to facilitate realistic object insertion and rendering we need to hypothesize camera parameters, scene geometry, surface materials, and sources of illumination. To address this, we develop a new method for both single-image depth and illumination inference. We are able to build a full 3D scene model without any user interaction, including camera parameters and reflectance estimates. 


\boldhead{Contributions}
Our primary contribution is a completely automatic algorithm for estimating a full 3D scene model from a single LDR photograph. Our system contains two technical contributions: illumination inference and depth estimation. We have developed a novel, data-driven illumination estimation procedure that automatically estimates a physical lighting model for the entire scene (including out-of-view light sources). This estimation is aided by our single-image light classifier to detect emitting pixels, which we believe is the first of its kind. We also demonstrate state-of-the-art depth estimates by combining data-driven depth inference with geometric reasoning.

We have created an intuitive interface for inserting 3D models seamlessly into photographs, using our scene approximation method to relight the object and facilitate drag-and-drop insertion. Our interface also supports other physically grounded image editing operations, such as post-process depth-of-field and lighting changes. In a user study, we show that our system is capable of making photorealistic edits: in side-by-side comparisons of ground truth photos with photos edited by our software, subjects had a difficult time choosing the ground truth.

\boldhead{Limitations} 
this method works best when scene lighting is diffuse, and therefore generally works better indoors than out (see our user studies and results in Sec~\ref{sec:study}). Our scene models are clearly not canonical representations of the imaged scene and often differ significantly from the true scene components. These coarse scene reconstructions suffice in many cases to produce realistically edited images. However, in some case, errors in either geometry, illumination, or materials may be stark enough to manifest themselves in unappealing ways while editing. For example, inaccurate geometry could cause odd looking shadows for inserted objects, and inserting light sources can exacerbate geometric errors. 
Also, our editing software does not handle object insertion {\it behind} existing scene elements automatically, and cannot be used to deblur an image taken with wide aperture. A Manhattan World is assumed in our camera pose and depth estimation stages, but this method is still applicable in scenes where this assumption does not hold (see Fig~\ref{fig:autoillum:results}).

\section{Related work}
In order to build a physically based image editor (one that supports operations such as lighting-consistent object insertion, relighting, and new view synthesis), it is requisite to model the three major factors in image formation: geometry, illumination, and surface reflectance. Existing approaches are prohibitive to most users as they require either manually recreating or measuring an imaged scene with hardware aids~\cite{Debevec:98,Yu:1999,Boivin:2001,Debevec05:Parthenon}.
In contrast, Lalonde et al.~\cite{lalonde2007} and Karsch et al.~\cite{Karsch:SA2011} have shown that even coarse estimates of scene geometry, reflectance properties, illumination, and camera parameters are sufficient for many image editing tasks. Their techniques require a user to model the scene geometry and illumination -- a task that requires time and an understanding of 3D authoring tools. While our work is similar in spirit to theirs, our technique is fully automatic and is still able to produce results with the same perceptual quality.

Similar to this method, Barron and Malik~\cite{Barron2013A} recover shape, surface albedo and illumination for entire scenes, but their method requires a coarse input depth map (e.g. from a Kinect) and is not directly suitable for object insertion as illumination is only estimated near surfaces (rather than the entire volume).

A technique for manipulating and rearranging 3D objects within photographs was presented by Kholgade et al.~\cite{OM3D2014}. Similar to the method presented in this chapter, their work also estimates geometry, lighting, and materials, but at an object level rather than for the entire scene. Such methods could be viewed as complementary to our work; these two methods combined could be used for both inserting, rearranging, and deleting objects from photographs.

\boldhead{Geometry} User-guided approaches to single image modeling~\cite{Horry97:TiP,Criminisi00:SVMetrology,Oh01:IBM} have been successfully used to create 3D reconstructions that allow for viewpoint variation. Single image depth estimation techniques have used learned relationships between image features and geometry to estimate depth~\cite{Torralba02depthestimation,Hoiem05:Popup,Saxena:pami09,Liu:cvpr10,Karsch:ECCV12}. Our depth estimation technique improves upon these methods by incorporating geometric constraints, using intuition from past approaches which estimate depth by assuming a Manhattan World (\cite{Delage:ISRR:05} for single images, and ~\cite{Furukawa:09,Gallup:CVPR:10} for multiple images).

Other approaches to image modeling have explicitly parametrized indoor scenes as 3D boxes (or as collections of orthogonal planes)~\cite{Lee:CVPR09,hedau2009iccv,Karsch:SA2011,Schwing:ECCV12,NedovicTPAMI2010}. In our work, we use appearance features to infer image depth, but augment this inference with priors based on geometric reasoning about the scene. 

The contemporaneous works of Satkin et al.~\cite{satkin_bmvc2012} and Del Pero et al.~\cite{pero:cvpr:13} predict 3D scene reconstructions for the rooms and their furniture. Their predicted models can be very good semantically, but are not suited well for our editing applications as the models are typically not well-aligned with the edges and boundaries in the image.



	
\boldhead{Illumination} Lighting estimation algorithms vary by the representation they use for illumination. Point light sources in the scene can be detected by analyzing silhouettes and shading along object contours~\cite{Johnson05:Lighting,LopezMoreno2010698}. Lalonde et al.~\cite{Lalonde:2009vp} use a physically-based model for sky illumination and a data-driven sunlight model to recover an environment map from time-lapse sequences. In subsequent work, they use the appearance of the sky in conjunction with cues such as shadows and shading to recover an environment map from a single image~\cite{Lalonde:2009vp}. Nishino et al.~\cite{Nishino04:EyesRelighting} recreate environment maps of the scene from reflections in eyes. Johnson and Farid~\cite{Johnson07:Complex} estimate lower-dimensional spherical harmonics-based lighting models from images. Panagopoulos et al.~\cite{Panagopoulos:cvpr:11} show that automatically detected shadows can be used to recover an illumination environment from a single image, but require coarse geometry as input.

While all these techniques estimate physically-based lighting from the scene, Khan et al.~\cite{Khan:2006:IME:1179352.1141937} show that wrapping an image to create the environment map can suffice for certain applications. 

Our illumination estimation technique attempts to predict illumination both within and outside the photograph's frustum with a data-driven matching approach; such approaches have seen previous success in recognizing scene viewpoint~\cite{Xiao:CVPR:12} and view extrapolation~\cite{Zhang:CVPR:13}.

We also attempt to predict a one-parameter camera response function jointly during our inverse rendering optimization. Other processes exist for recovering camera response, but require multiple of images~\cite{Diaz:2013wk}.

\boldhead{Materials} In order to infer illumination, we separate the input image into diffuse albedo and shading using the Color Retinex algorithm~\cite{grosse09intrinsic}. We assume that the scene is Lambertian, and find that this suffices for our applications. Other researchers have looked at the problem of recovering the Bi-directional Reflectance Density Function (BRDF) from a single image, but as far as we know, there are no such methods that work automatically and at the scene (as opposed to object) level. These techniques typically make the problem tractable by using low-dimensional representations for the BRDF such as spherical harmonics~\cite{Ramamoorthi04:SignalProcessing}, bi-variate models for isotropic reflectances~\cite{Romeiro:eccv:08,Romeiro:eccv:10}, and data-driven statistical models~\cite{Lombardi:eccv:2012,Lombardi12:MultiMaterial}. All these techniques require the shape to be known and in addition, either require the illumination to be given, or use priors on lighting to constrain the problem space. 


\boldhead{Perception} Even though our estimates of scene geometry, materials, and illumination are coarse, they enable us to create realistic composites. This is possible because even large changes in lighting are often not perceivable to the human visual system. This has been shown to be true for both point light sources~\cite{LopezMoreno2010698} and complex illumination~\cite{Ramanarayanan07:VisualEq}.

\section{Method overview}
\label{sec:overview}
this method consists of three primary steps, outlined in Fig~\ref{fig:overview}. First, we estimate the physical space of the scene (camera parameters and geometry), as well as the per-pixel diffuse reflectance (Sec~\ref{sec:recon}). Next, we estimate scene illumination (Sec~\ref{sec:illum}) which is guided by our previous estimates (camera, geometry, reflectance). Finally, our interface is used to composite objects, improve illumination estimates, or change the depth-of-field (Sec~\ref{sec:interface}). We have evaluated this method with a large-scale user study (Sec~\ref{sec:study}), and additional details and results can be found in the corresponding supplemental document. Figure~\ref{fig:overview} illustrates the pipeline of our system. 

\boldhead{Scene parameterization} Our geometry is in the form of a depth map, which is triangulated to form a polygonal mesh (depth is unprojected according to our estimated pinhole camera). Our illumination model contains polygonal area sources, as well as one or more spherical image-based lights.

While unconventional, our models are suitable for most off-the-shelf rendering software, and we have found our models to produce better looking estimates than simpler models (e.g. planar geometry with infinitely distant lighting).

\boldhead{Automatic indoor/outdoor scene classification} As a pre-processing step, we automatically detect whether the input image is indoors or outdoors. We use a simple method: $k$-nearest-neighbor matching of GIST features~\cite{Oliva:2001wt} between the input image and all images from the indoor NYUv2 dataset and the outdoor Make3D Dataset. We choose $k=7$, and decide use majority-voting to determine if the image is indoors or outdoors (e.g. if 4 of the nearest neighbors are from the Make3D dataset, we consider it to be outdoors). More sophisticated methods could also work.

this method uses different training images and classifiers depending on whether the input image is classified as an indoor or outdoor scene. 

\section{Single image reconstruction}
\label{sec:recon}

\begin{figure}
\centerline{\includegraphics[width=0.50\linewidth]{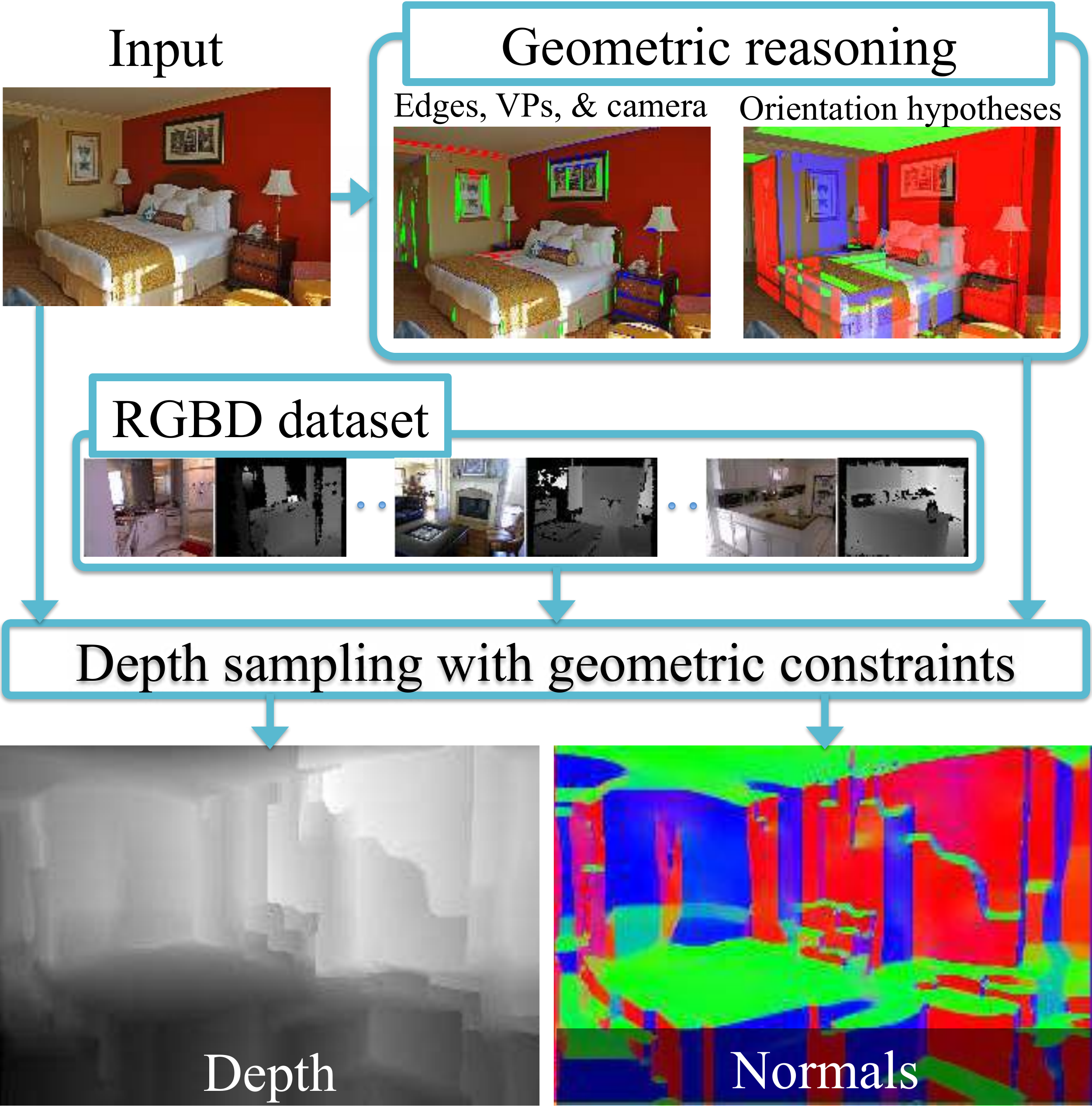}}
\caption{Automatic depth estimation algorithm. Using the geometric reasoning method of Lee et al.~\protect\cite{Lee:CVPR09}, we estimate focal length and a sparse surface orientation map. Facilitated by a dataset of RGBD images, we then apply a non-parametric depth sampling approach to compute the per-pixel depth of the scene. The geometric cues are used during inference to enforce orientation constraints, piecewise-planarity, and surface smoothness. The result is a dense reconstruction of the scene that is suitable for realistic, physically grounded editing. {\it Photo credits: Flickr user {\footnotesize \textcopyright{}}``Mr.TinDC''.}
}
\label{fig:depth_overview}
\end{figure}

The first step in our algorithm is to estimate the physical space of the scene, which we encode with a depth map, camera parameters, and spatially-varying diffuse materials.  Here, we describe how to estimate these components, including a new technique for estimating depth from a single image that adheres to geometric intuition about indoor scenes.

\boldhead{Single image depth estimation} 
Karsch et al.~\cite{Karsch:ECCV12} describe a non-parametric, ``depth transfer'' approach for estimating dense, per-pixel depth from a single image. While shown to be state-of-the-art, this method is purely data-driven, and incorporates no explicit geometric information present in many photographs. It requires a database of RGBD (RGB+depth) images, and transfers depth from the dataset to a novel input image in a non-parametric fashion using correspondences in appearance. This method has been shown to work well both indoors and outdoors; however, only appearance cues are used (multi-scale SIFT features), and we have good reason to believe that adding geometric information will aid in this task.

A continuous optimization problem is solved to find the most likely estimate of depth given an input image. In summary, images in the RGBD database are matched to the input and warped so that SIFT features are aligned, and an objective function is minimized to arrive at a solution. We denote $\ve{D}$ as the depth map we wish to infer, and following the notation of Karsch et al., we write the full objective here for completeness:
\begin{equation}
\argmin_{\ve{D}} E(\ve{D}) = \sum_{i\in \text{pixels}} E_t(\ve{D}_i) + \alpha E_s(\ve{D}_i) + \beta E_p(\ve{D}_i),
\end{equation}
where $E_t$ is the data term (depth transfer), $E_s$ enforces smoothness, and $E_p$ is a prior encouraging depth to look like the average depth in the dataset. We refer the reader to~\cite{Karsch:ECCV12} for details.

Our idea is to reformulate the depth transfer objective function and infuse it with geometric information extracted using the geometric reasoning algorithm of Lee et al.~\cite{Lee:CVPR09}. Lee et al. detect vanishing points and lines from a single image, and use these to hypothesize a set of sparse surface orientations for the image. The predicted surface orientations are aligned with one of the three dominant directions in the scene (assuming a Manhattan World).


\begin{figure}
\centerline{\includegraphics[width=.40\linewidth]{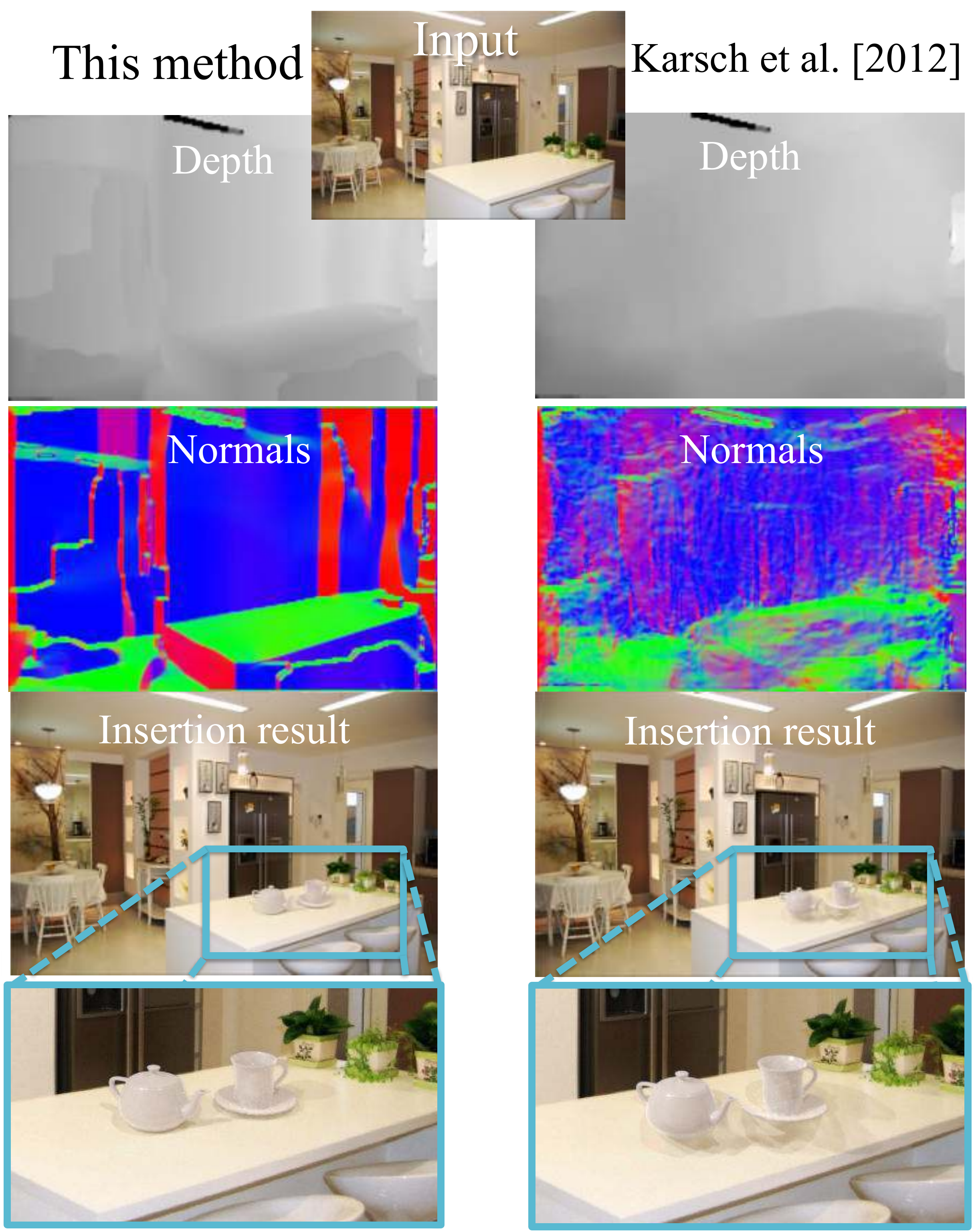}}
\caption{Comparison of different depth estimation techniques. Although the depth maps of our technique and of Karsch et al.~\protect\cite{Karsch:ECCV12} appear roughly similar, the surface orientations provide a sense of how distinct the methods are. Our depth estimation procedure (aided by geometric reasoning) is crucial in achieving realistic insertion results, as the noisy surface orientations from Karsch et al. can cause implausible cast shadows and lighting effects.
}
\label{fig:depth_comparison}
\end{figure}

We remove the image-based smoothness ($E_s$) and prior terms ($E_p$), and replace them with geometric-based priors. We add terms to enforce a Manhattan World ($E_m$), constrain the orientation of planar surfaces ($E_o$), and impose 3D smoothness ($E_{3s}$, spatial smoothness in 3D rather than 2D):
\begin{align}
\label{eq:depth}
\argmin_{\ve{D}} E_{geom}(\ve{D}) = \sum_{i\in \text{pixels}} &E_t(\ve{D}_i) + \lambda_m E_m(N(\ve{D})) \ + \\
&\lambda_o E_o(N(\ve{D})) + \lambda_{3s} E_{3s}(N(\ve{D})), \nonumber
\end{align}
where the weights are trained using a coarse-to-fine grid search on held-out ground truth data (indoors: $\lambda_m = 100$, $\lambda_o = 0.5$, $\lambda_{3s} = 0.1$, outdoors: $\lambda_m = 200$, $\lambda_o = 10$, $\lambda_{3s} = 1$); these weights dictate the amount of influence each corresponding prior has during optimization. Descriptions of these priors and implementation details can be found in the supplemental file.

Figure~\ref{fig:depth_overview} shows the pipeline of our depth estimation algorithm, and Fig~\ref{fig:depth_comparison} illustrates the differences between this method and the depth transfer approach; in particular, noisy surface orientations from the depth of Karsch et al.~\cite{Karsch:ECCV12} lead to unrealistic insertion and relighting relights.

In supplemental material, we show additional results, including state-of-the-art results on two benchmark datasets using our new depth estimator. 

\boldhead{Camera parameters}
It is well known how to compute a simple pinhole camera (focal length, $f$ and camera center, $(c_0^x, c_0^y)$) and extrinsic parameters from three orthogonal vanishing points~\cite{hartley2004} (computed during depth estimation), and we use this camera model at render-time. 

\boldhead{Surface materials} We use Color Retinex (as described in~\cite{grosse09intrinsic}), to estimate a spatially-varying diffuse material albedo for each pixel in the visible scene. 

\section{Estimating illumination}
\label{sec:illum}

\begin{figure*}
\centerline{\includegraphics[width=.95\linewidth]{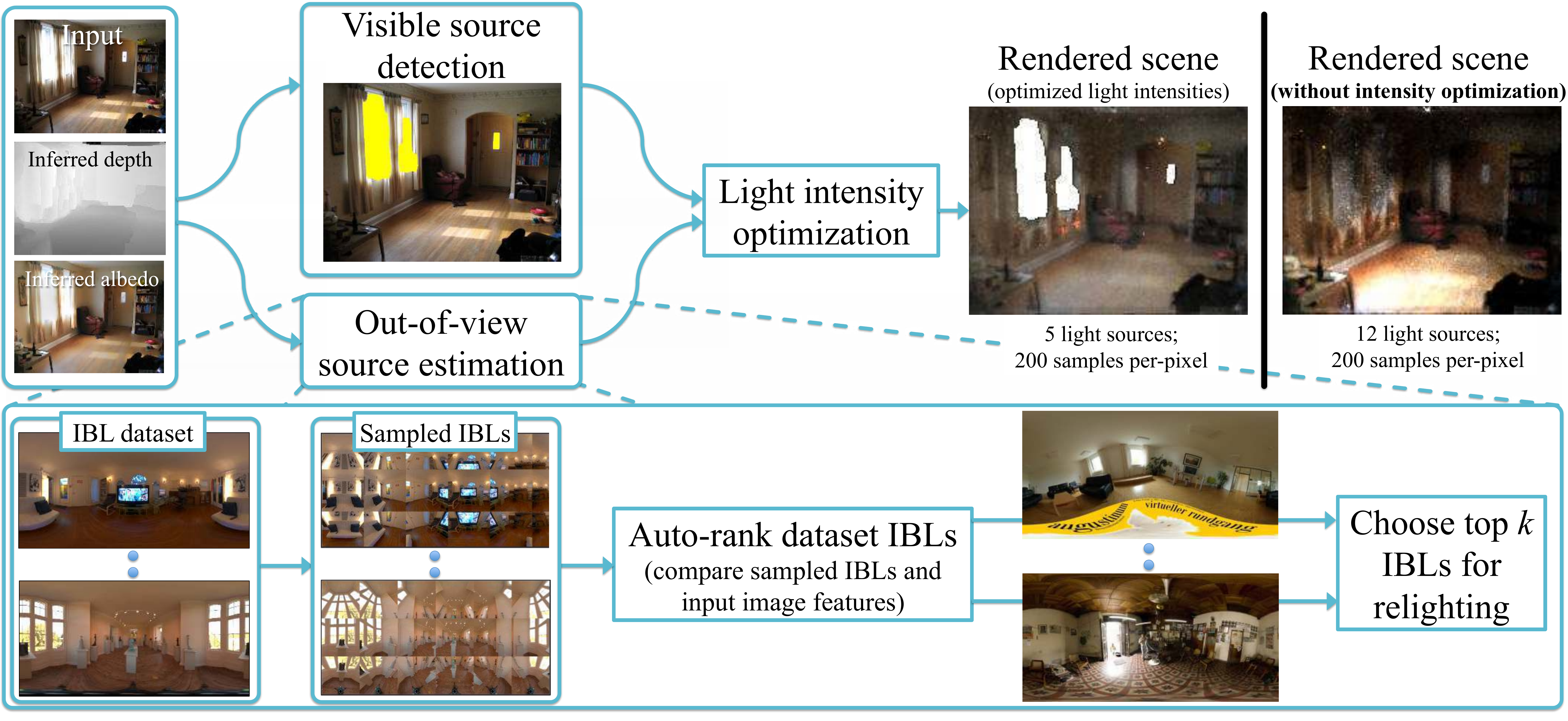}}
\vspace{-3mm}
\caption{Overview of our lighting estimation procedure. Light sources are first detected in the input image using our light classifier. To estimate light outside of the view frustum, we use a data-driven approach (utilizing the SUN360 panorama dataset). We train a ranking function to rank IBLs according to how well they ``match'' the input image's illumination (see text for details), and use the top $k$ IBLs for relighting. Finally, source intensities are optimized to produce a rendering of the scene that closely matches the input image. Our optimization not only encourages more plausible lighting conditions, but also improves rendering speed by pruning inefficient light sources.
}
\label{fig:light_overview}
\end{figure*}

We categorize luminaires into {\it visible} sources (sources visible in the photograph), and {\it out-of-view} sources (all other luminaires). Visible sources are detected in the image using a trained ``light classifier'' (Sec~\ref{sec:illum:inside}). Out-of-view sources are estimated through a data-driven procedure (Sec~\ref{sec:illum:outside}) using a large dataset of annotated spherical panoramas (SUN360~\cite{Xiao:CVPR:12}).  The resulting lighting model is a hybrid of area sources and spherical emitting sources. Finally, light source intensities are estimated using an optimization procedure which adjusts light intensities so that the rendered scene appears similar to the input image (Sec~\ref{sec:illum:render}). Figure~\ref{fig:light_overview} illustrates this procedure.

\boldhead{Dataset} We have annotated light sources in 100 indoor and 100 outdoor scenes from the SUN360 dataset. The annotations also include a discrete estimate of distance from the camera (``close'': 1-5m, ``medium'': 5-50m, ``far'': $>$50m, ``infinite'': reserved for sun).  Since SUN360 images are LDR and tonemapped, we make no attempt to annotate absolute intensity, only the position/direction of sources. Furthermore, the goal of our classifier is only to predict location (not intensity). This data is used in both our in-view and out-of-view techniques below. 


\subsection{Illumination visible in the view frustum}
\label{sec:illum:inside}
To detect sources of light in the image, we have developed a new light classifier. For a given image, we segment the image into superpixels using SLIC~\cite{Achanta:ijcv:12}, and compute features for each superpixel. We use the following features: the height of the superpixel in the image (obtained by averaging the 2D location of all pixels in the superpixel), as well as the features used by Make3D\footnote{17 edge/smoothing filter responses in YCbCr space are averaged over the superpixel. Both the energy and kurtosis of the filter responses are computed (second and fourth powers)  for a total of 34 features, and then concatenated with four neighboring (top,left,bottom,right) superpixel features. This is done at two scales (50\% and 100\%), resulting in $340 (=34\times 5 \times 2$) features per superpixel.} ~\cite{Saxena:pami09}. 

Using our annotated training data, we train a binary classifier to predict whether or not a superpixel is emitting/reflecting a significant amount of light using these features. For this task, we do not use the discrete distance annotations (these are however used in Sec~\ref{sec:illum:outside}). A classification result is shown in Figure~\ref{fig:light_overview}. In supplemental material, we show many more qualitative and quantitative results of our classifier, and demonstrate that our classifier significantly outperforms baseline detectors (namely thresholding).

For each detected source superpixel (and corresponding pixels), we find their 3D position using the pixel's estimated depth and the projection operator $K$. Writing $D$ as the estimated depth map and $(x,y)$ as a pixel's position in the image plane, the 3D position of the pixel is given by: 
\begin{equation}
\ve{X} = D(x,y) K^{-1} [x,y,1]^T,
\end{equation}
where $K$ is the intrinsic camera parameters (projection matrix) obtained in Section~\ref{sec:recon}. We obtain a polygonal representation of each light source by fitting a 3D quadrilateral to each cluster (oriented in the direction of least variance). Notice that this only provides the position of the visible light sources; we describe how we estimate intensity in Section~\ref{sec:illum:render}. Figure~\ref{fig:light_overview} (top) shows a result of our light detector.

One might wonder why we train using equirectangular images and test using rectilinear images. Dror et al.~\cite{dror2004sta} showed that many image statistics computed on equirectangular images follow the same distributions as those computed on rectilinear images; thus features computed in either domain should be roughly the same. We have also tested this method on both kinds of images (more results in the supplemental file), and see no noticeable differences.

\subsection{Illumination outside of the view frustum}
\label{sec:illum:outside}

Estimating lighting from behind the camera is arguably the most difficult task in single-image illumination estimation. We use a data-driven approach, utilizing the extensive SUN360 panorama dataset of Xiao et al.~\cite{Xiao:CVPR:12}. However, since this dataset is not available in HDR, we have annotated the dataset to include light source positions and distances.

Our primary assumption is that if two photographs have similar appearance, then the illumination environment beyond the photographed region will be similar as well. There is some empirical evidence that this can be true (e.g. recent image-extrapolation methods~\cite{Zhang:CVPR:13}), and studies suggest that people hallucinate out-of-frame image data by combining photographic evidence with recent memories~\cite{Intraub:JEP:1989}. 

Using this intuition, we develop a novel procedure for matching images to luminaire-annotated panoramas in order to predict out-of-view illumination. We sample each IBL into $N$ rectilinear projections (2D) at different points on the sphere and at varying fields-of-view, and match these projections to the input image using a variety of features described below (in our work, $N=10$ stratified random samples on the sphere with azimuth $\in [0, 2\pi)$, elevation $\in [-\frac{\pi}{6},\frac{\pi}{6}]$, FOV $\in [\frac{\pi}{3},\frac{\pi}{2}]$). See the bottom of Fig~\ref{fig:light_overview} for an illustration. These projections represent the images a camera with that certain orientation and field of view would have captured for that particular scene. By matching the input image to these projections, we can effectively "extrapolate" the scene outside the field of view and estimate the out-of-view illumination.

Given an input image, our goal is to find IBLs in our dataset that emulate the input's illumination. Our rectilinear sampled IBLs provide us with ground truth data for training: for each sampled image, we know the corresponding illumination. Based on the past success of rank prediction for data-driven geometry estimation~\cite{satkin_bmvc2012}, we use this data and train an IBL rank predictor (for an input image, rank the dataset IBLs from best to worst).

\boldhead{Features} After sampling the panoramas into rectilinear images, we compute seven features for each image: geometric context~\cite{Hoiem:iccv:2005}, orientation maps~\cite{Lee:CVPR09}, spatial pyramids~\cite{Lazebnik:cvpr:06}, HSV histograms (three features total), and the output of our light classifier (Sec~\ref{sec:illum:inside}). 

We are interested in ranking pairs of images, so our final feature describes how well two particular images match in feature space. The result is a 7-dimensional vector where each component describes the similarity of a particular feature (normalized to [0,1], where higher values indicate higher similarity). Similarity is measured using the histogram intersection score for histogram features (spatial pyramid -- sp -- and HSV -- h,s,v -- histograms), and following Satkin et al.~\cite{satkin_bmvc2012}, a normalized, per-pixel dot product (averaged over the image) is used for for other features (geometric contact -- gc, orientation maps -- om, light classifier -- lc).  

More formally, let $F_i, F_j$ be the features computed on images $i$ and $j$, and $x^i_j$ be the vector that measures similarity between the features/images:
\begin{align}
x^i_j = [&nd(F_i^{\text{gc}},F_j^{\text{gc}}), nd(F_i^{\text{om}}, F_j^{\text{om}}), nd(F_i^{\text{lc}}, F_j^{\text{lc}}), \nonumber \\ 
&hi(F_i^{\text{sp}}, F_j^{\text{sp}}), hi(F_i^{\text{h}}, F_j^{\text{h}}), hi(F_i^{\text{s}}, F_j^{\text{s}}), hi(F_i^{\text{v}}, F_j^{\text{v}})]^T,
\label{eq:features}
\end{align}
where $nd(\cdot), hi(\cdot)$ are normalized dot product and histogram intersection operators respectively. In order to compute per-pixel dot products, images must be the same size. To compute features on two images with unequal dimension, we downsample the larger image to have the same dimension as the smaller image.

\boldhead{Training loss metric} 
In order to discriminate between different IBLs, we need to define a distance metric that measure how similar one IBL is to another. We use this distance metric as the loss for our rank training optimization where it encodes the additional margin when learning the ranking function (Eq~\ref{eq:ranking}).

A na\"{\i}ve way to measure the similarity between two IBLs is to use pixel-wise or template matching~\cite{Xiao:CVPR:12}. However, this is not ideal for our purposes since it requires accurate correspondences between elements of the scene that may not be important from a relighting aspect. Since our primary goal is re-rendering, we define our metric on that basis of how different a set of canonical objects appear when they are illuminated by these IBLs. In particular, we render nine objects with varying materials (diffuse, glossy, metal, etc.) into each IBL environment (see supplemental material). Then, we define the distance as the mean L2 error between the renderings.

One caveat is that our IBL dataset isn't HDR, and we don't know the intensities of the annotated sources. So, we compute error as the minimum over all possible light intensities. Define $\mathbf{I}_i$ and $\mathbf{I}_j$ as two IBLs in our dataset, and $I_i=[I_i^{(1)},\ldots,I_i^{(n)}], I_j=[I_j^{(1)},\ldots,I_j^{(m)}]$ as column-vectorized images rendered by the IBLs for each of the IBLs' sources (here $\mathbf{I}_i$ has $n$ sources, and $\mathbf{I}_j$ has $m$). Since a change in the intensity of the IBL corresponds to a change of a scale factor for the rendered image, we define the distance as the minimum rendered error over all possible intensities ($y_i$ and $y_j$):
\begin{equation}
d(\mathbf{I}_i,\mathbf{I}_j) = \min_{y_i, y_j} || I_i y_i -I_j y_j ||, \text{ s.t. } ||[y_i^T,y_j^T] || = 1.
\label{eq:metric}
\end{equation}
The constraint is employed to avoid the trivial solution, and we solve this using SVD.

\boldhead{Training the ranking function}
Our goal is to train a ranking function that can properly rank images (and their corresponding IBLs) by how well their features match the input's. Let $w$ be a linear ranking function, and $x^i_j$ be features computed between images $i$ and $j$. Given an input image $i$ and any two images from our dataset (sampled from the panoramas) $j$ and $k$, we wish to find $w$ such that $w^T x^i_j > w^T x^i_k$ when the illumination of $j$ matches the illumination of $i$ better than the illumination of $k$.

To solve this problem, we perform a standard 1-slack, linear SVM-ranking optimization~\cite{Joachims:kdd:06}:
\begin{equation}
\argmin_{w,\xi} ||w||^2 + C \xi, \ \text{ s.t. } w^T x^i_j \geq w^T x^i_k + \delta^i_{j,k}  - \xi, \ \xi\geq 0,
\label{eq:ranking}
\end{equation}
where $x$ are pairwise image similarity features (Eq~\ref{eq:features}), and $\delta^i_{j,k} = \max(d(\mathbf{I}_i,\mathbf{I}_k)-d(\mathbf{I}_i,\mathbf{I}_j),0)$ is a hinge loss to encourage additional margin for examples with unequal distances (according to Eq~\ref{eq:metric}, where $\mathbf{I}_i$ is the IBL corresponding to image $i$).

\boldhead{Inference} To predict the illumination of a novel input image ($i$), we compute the similarity feature vector (Eq~\ref{eq:features}) for all input-training image pairs ($x^i_j, \ \forall j$), and sort the prediction function responses ($w^Tx^i_j$) in decreasing order. Then, we use choose the top $k$ IBLs (in our work, we use $k=1$ for indoor images, and $k=3$ outdoors to improve the odds of predicting the correct sun location). Figure~\ref{fig:light_overview} (bottom) shows one indoor result using this method (where $k=2$ for demonstration).

\subsection{Intensity estimation through rendering}
\label{sec:illum:render}
Having estimated the location of light sources within and outside the image, we must now recover the relative intensities of the sources. Given the exact geometry and material of the scene (including light source positions), we can estimate the intensities of the sources by adjusting them them until a rendered version of the scene matches the original image. While we do not know exact geometry/materials, we assume that our automatic estimates are good enough, and could apply the above rendering-based optimization to recover the light source intensities; Fig~\ref{fig:light_overview} shows this process. Our optimization has two goals: match the rendered image to the input, and differing from past techniques~\cite{Boivin:2001,Karsch:SA2011}, ensure the scene renders efficiently.

Define $L_i$ as the intensity of the $i^{th}$ light source, $I$ as the input image, and $R(\cdot)$ as a scene ``rendering'' function that takes a set of light intensities and produces an image of the scene illuminated by those lights (i.e., $R(L)$). We want to find the intensity of each light source by matching the input and rendered images, so we could solve $\argmin_{L} \sum_{i \in \text{pixels}} ||I_i - R_i(L)||$. However, this optimization can be grossly inefficient and unstable, as it requires a new image to be rendered for each function evaluation, and rendering in general is non-differentiable. However, we can use the fact that light is additive, and write $R(\cdot)$ as a linear combination of ``basis'' renders~\cite{Schoeneman:1993,Nimeroff:94}. We render the scene (using our estimated geometry and diffuse materials) using only one light source at a time (i.e., $L_k = 1, L_j = 0 \ \forall j \neq k$, implying $L = e_k$). This results in one rendered image per light source, and we can write a new render function as $R'(w) = C \left(\sum_k w_k R(e_k)\right)$, where $C$ is the camera response function, and $R(e_i)$ is the scene rendered with only the $i^{th}$ source. In our work, we assume the camera response can be modeled as an exponent, i.e., $C(x) = x^\gamma$.  This allows us to rewrite the matching term above as
\begin{equation}
Q(w,\gamma) = \sum_{i \in \text{pixels}} \left| \left| I_i - \left[\sum_{k \in \text{sources}} w_k R_i(e_k)\right]^\gamma \right| \right|.
\label{eq:light_opt2}
\end{equation}
Since each basis render, $R(e_k)$, can be precomputed prior to the optimization, $Q$ can be minimized more efficiently than the originally described optimization.

We have hypothesized a number of light source locations in Secs~\ref{sec:illum:inside} and~\ref{sec:illum:outside}, and because our scene materials are purely diffuse, and our geometry consists of a small set of surface normals, there may exist an infinite number of lighting configurations that produce the same rendered image. Interestingly, user studies have shown that humans cannot distinguish between a range of illumination configurations~\cite{Ramanarayanan07:VisualEq}, suggesting that there is a family of lighting conditions that produce the same perceptual response. This is actually advantageous, because it allows our optimization to choose only a small number of ``good'' light sources and prune the rest. In particular, since our final goal is to relight the scene with the estimated illumination, we are interested in lighting configurations that can be rendered faster. We can easily detect if a particular light source will cause long render times by rendering the image with the given source for a fixed amount of time, and checking the variance in the rendered image (i.e., noise); we incorporate this into our optimization. By rendering with fewer sources and sources that contribute less variance, the scenes produced by this method render significantly faster than without this optimization (see Fig~\ref{fig:light_overview}, right).

Specifically, we ensure that only a small number of sources are used in the final lighting solution, and also prune problematic light sources that may cause inefficient rendering. We encode this with a sparsity prior on the source intensities and a smoothness prior on the rendered images: 
\begin{equation}
P(w) = \sum_{k \in \text{sources}} \left[ ||w_k||_1 + w_k \sum_{i \in \text{pixels}} || \nabla R_i(e_k) || \right].
\label{eq:light_priors}
\end{equation}
Intuitively, the first term coerces only a small number of nonzero elements in $w$, and the second term discourages noisy basis renders from having high weights (noise in a basis render typically indicates an implausible lighting configuration, making the given image render much more slowly).

\begin{figure*}
\setlength\fboxrule{1pt}
\setlength\fboxsep{0pt}
\begin{overpic}[width=.24\linewidth,clip,trim=0 0 40 0]{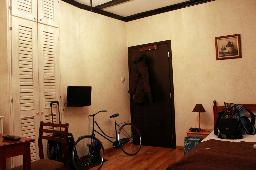}
\put(0,52){\fcolorbox{white}{white}{\includegraphics[width=.10\linewidth,clip,trim=0 0 40 0]{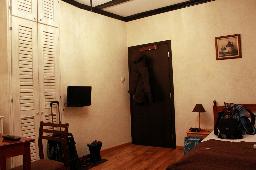}}} 
\setlength\fboxsep{2pt}
\put(0,2){\colorbox{white}{\Large (a)}}
\end{overpic}
\setlength\fboxsep{0pt}
\begin{overpic}[width=.24\linewidth,clip,trim=0 0 40 0]{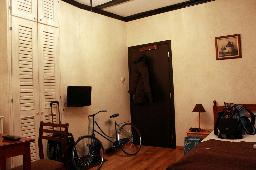} 
\setlength\fboxsep{2pt}
\put(0,2){\colorbox{white}{\Large (b)}}
\end{overpic}
\hfill %
\setlength\fboxsep{0pt}
\begin{overpic}[width=.24\linewidth]{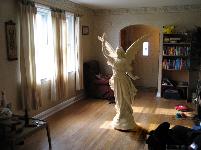}
\put(0,49){\fcolorbox{white}{white}{\includegraphics[width=.10\linewidth]{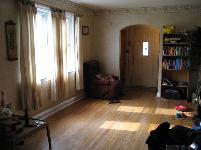}}} 
\setlength\fboxsep{2pt}
\put(0,2){\colorbox{white}{\Large (c)}}
\end{overpic}
\setlength\fboxsep{0pt}
\begin{overpic}[width=.24\linewidth]{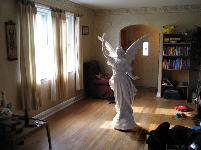} 
\setlength\fboxsep{2pt}
\put(0,2){\colorbox{white}{\Large (d)}}
\end{overpic}
\vspace{1mm} \\
\setlength\fboxsep{0pt}
\begin{overpic}[width=.33\linewidth]{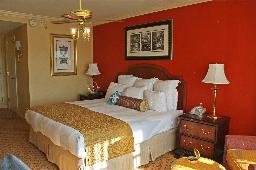}
\put(94,63){\fcolorbox{white}{white}{\includegraphics[width=.13\linewidth]{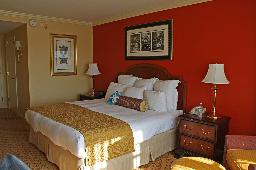}}}
\end{overpic}
\setlength\fboxsep{2pt}
\begin{overpic}[width=.33\linewidth]{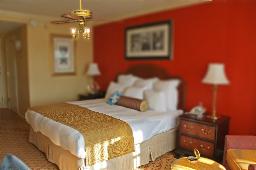} \put(0,2){\colorbox{white}{\Large (e)}}\end{overpic}
\setlength\fboxsep{2pt}
\begin{overpic}[width=.33\linewidth]{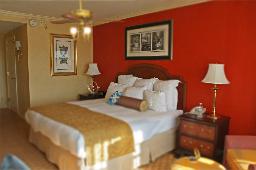} \put(0,2){\colorbox{white}{\Large (f)}}\end{overpic}
\caption{Our user interface provides a drag-and-drop mechanism for inserting 3D models into an image (input image overlaid on our initial result). Our system also allows for real-time illumination changes without re-rendering. Using simple slider controls, shadows and interreflections can be softened or hardened (a,b), light intensity can be adjusted to achieve a different mood (c,d), and synthetic depth-of-field can be applied to re-focus the image (e,f). See the accompanying video for a full demonstration. Best viewed in color at high-resolution. {\it Photo credits: {\footnotesize \textcopyright{}}Rachel Titiriga (top) and Flickr user {\footnotesize \textcopyright{}}``Mr.TinDC'' (bottom).}
}
\label{fig:interface}
\end{figure*}

Combining the previous equations, we develop the following optimization problem:
\begin{align}
\argmin_{w,\gamma} \ \ Q(w,\gamma) + \lambda_{P} P(w) + \lambda_{\gamma} ||\gamma-\gamma_0||, \nonumber \\
\text{s.t. } w_k \geq 0 \ \forall k,\ \gamma>0,
\label{eq:light_opt3}
\end{align}
where $\lambda_{P} = \lambda_{\gamma} = 0.1$ are weights, and $\gamma_0=\frac{1}{2.2}$. We use a continuous approximation to the absolute value ($|x| \approx \sqrt{x^2+\epsilon}$), and solve using the active set algorithm~\cite{Nocedal2006NO}. The computed weights ($w$) can be directly translated into light intensities ($L$), and we now have an entire model of the scene (geometry, camera, and materials from Sec~\ref{sec:recon}, and light source positions/intensities as described above). 

this method has several advantages to past ``optimization-through-rendering'' techniques~\cite{Karsch:SA2011}. First, our technique has the ability to discard unnecessary and inefficient light sources by adding illumination priors (Eq~\ref{eq:light_priors}). Second, we estimate the camera response function jointly during our optimization, which we do not believe has been done previously. 
Finally, by solving for a simple linear combination of pre-rendered images, our optimization procedure is much faster than previous methods that render the image for each function evaluation (e.g., as in~\cite{Karsch:SA2011}). Furthermore, our basis lights could be refined or optimized with user-driven techniques guided by aesthetic principles (e.g.~\cite{Boyadzhiev:SG:2013}).

\section{Physically grounded image editing}
\label{sec:interface}

One of the potential applications of our automatically generated 3D scene models is physically grounded photo editing. In other words, we can use the approximate scene models to facilitate physically-based image edits. For example, one can use the 3D scene to insert and composite objects with realistic lighting into a photograph, or even adjust the depth-of-field and aperture as a post-process.  

There are many possible interactions that become available with our scene model, and we have developed a prototype interface to facilitate a few of these. Realistically inserting synthetic objects into legacy photographs is our primary focus, but our application also allows for post-process lighting and depth-of-field changes. To fully leverage our scene models, we require physically based rendering software. We use LuxRender\footnote{\url{http://www.luxrender.net}}, and have built our application on top of LuxRender's existing interface. Figure~\ref{fig:interface} illustrates the possible uses of the interface, and we refer the reader to the accompanying video for a full demonstration. 

\boldhead{Drag-and-drop insertion} Once a user specifies an input image for editing, our application automatically computes the 3D scene as described in sections~\ref{sec:recon} and~\ref{sec:illum}. Based on the inserted location, we also add additional geometric constraints so that the depth is flat in a small region region around the base of the inserted object. For a given image of size $1024\times768$, this method takes approximately five minutes on a 2.8Ghz dual core laptop to estimate the 3D scene model, including depth and illumination (this can also be precomputed or computed remotely for efficiency, and only occurs once per image). Next, a user specifies a 3D model and places it in the scene by simply clicking and dragging in the picture (as in Fig~\ref{fig:autoillum:teaser}). Our scene reconstruction facilitates this insertion: our estimated perspective camera ensures that the object is scaled properly as it moves closer/farther from the camera, and based on the surface orientation of the clicked location, the application automatically re-orients the inserted model so that its up vector is aligned with the surface normal. Rigid transformations are also possible through mouse and keyboard input (scroll to scale, right-click to rotate, etc).

Once the user is satisfied with the object placement, the object is rendered into the image\footnote{Rendering time is clearly dependent on a number of factors (image size, spatial hierarchy, inserted materials, etc), and is slowed by the fact that we use an unbiased ray-tracer. These results took between 5 minutes and several hours to render, but this could be sped up depending on the application and resources available.}. We use the additive differential rendering method~\cite{Debevec:98} to composite the rendered object into the photograph, but other methods for one-shot rendering could be used (e.g. Zang et al.~\cite{Zang:2012}). This method renders two images: one containing synthetic objects $I_{obj}$, and one without synthetic objects $I_{noobj}$, as well as an object mask M (scalar image that is 0 everywhere where no object is present, and (0, 1] otherwise). The final composite image $I_{final}$ is obtained by
\begin{equation}
I_{final} = M \odot I_{obj} + (1-M) \odot (I_{b}+I_{obj}-I_{noobj}),
\label{eq:composite}
\end{equation}
where $I_{b}$ is the input image, and $\odot$ is the entry-wise product. For efficiency (less variance and overhead), we have implemented this equation as a surface integrator plugin for LuxRender. Specifically, we modify LuxRender's bidirectional path tracing implementation~\cite{PBRTv2} so that pixels in $I_{final}$ are intelligently sampled from either the input image or the rendered scene such that inserted objects and their lighting contributions are rendered seamlessly into the photograph. We set the maximum number of eye and light bounces to 16, and use LuxRender's default Russian Roulette strategy (dynamic thresholds based on past samples).

The user is completely abstracted from the compositing process, and only sees $I_{final}$ as the object is being rendered. The above insertion method also works for adding light sources (for example, inserting and emitting object). Figure~\ref{fig:autoillum:teaser} demonstrates a drag-and-drop result.

\boldhead{Lighting adjustments} Because we have estimated sources for the scene, we can modify the intensity of these sources to either add or subtract light from the image. Consider the compositing process (Eq~\ref{eq:composite} with no inserted objects; that is, $I_{obj} = I_{noobj}$, implying $I_{final} = I_{b}$. Now, imagine that the intensity of a light source in $I_{obj}$ is increased (but the corresponding light source in $I_{noobj}$ remains the same).  $I_{obj}$ will then be brighter than $I_{noobj}$, and the compositing equation will reflect a physical change in brightness (effectively rendering a more intense light into the scene).  Similarly, decreasing the intensity of source(s) in $I_{obj}$ will remove light from the image. Notice that this is a physical change to the lighting in the picture rather than tonal adjustments (e.g., applying a function to an entire image). 

This technique works similarly when there are inserted objects present in the scene, and a user can also manually adjust the light intensities in a scene to achieve a desired effect (Fig~\ref{fig:interface}, top). Rather than just adjusting source intensities in $I_{obj}$, if sources in both $I_{obj}$ and $I_{noobj}$ are modified equally, then only the intensity of the inserted object and its interreflections (shadows, caustics, etc) will be changed (i.e. without adding or removing light from the rest of the scene).

By keeping track of the contribution of each light source to the scene, we only need to render the scene once, and lighting adjustments can be made in real time without re-rendering (similar to~\cite{Loscos:1999,Gibson:2000}). Figure~\ref{fig:interface} shows post-process lighting changes; more can be found in supplemental material.

\boldhead{Synthetic depth-of-field} Our prototype also supports post-focusing the input image: the user specifies a depth-of-field and an aperture size, and the image is adaptively blurred using our predicted depth map ($\ve{D}$). Write $I_{final}$ as the composite image with inserted objects (or the input image if nothing is inserted), and $G(\sigma)$ as a Gaussian kernel with standard deviation $\sigma$. We compute the blur at the $i^{th}$ pixel as
\begin{equation} 
I_{dof,i} = I_{final,i} \star G( a|\ve{D}_i -  d| ),  
\end{equation}
where $d$ is the depth of field, and $a$ corresponds to the aperture size. Figures~\ref{fig:autoillum:teaser} and~\ref{fig:interface} (bottom) shows a post-focus result. 

\section{Evaluation}
\label{sec:study}

We conducted two user studies to evaluate the ``realism'' achieved by our insertion technique. Each user study is a series of two-alternative forced choice tests, where the subject chooses between a pair of images which he/she feels looks the most realistic.

In the first study (Sec~\ref{sec:study:real}), we use real pictures. Each pair of images contains one actual photograph with a small object placed in the scene, and the other is photo showing a similar object inserted synthetically into the scene (without the actual object present).

The second study (Sec~\ref{sec:study:synthetic}) is very similar, but we use highly realistic, synthetic images rather than real pictures. For one image, a synthetic object is placed into the full 3D environment and rendered (using unbiased, physically-based ray tracing); in the other, the same 3D scene is rendered without the synthetic object (using the same rendering method), and then synthetically inserted {\it into the photograph} (rather than the 3D scene) using this method.

Finally, we visualize the quantitative accuracy achieved by our estimates in Section~\ref{sec:study:quant}.

\subsection{Real image user study}
\label{sec:study:real}
\boldhead{Experimental setup}  We recruited 30 subjects for this study, and we used a total of 10 synthetic objects across three unique, real scenes. Each subject saw each inserted object only once, and viewed seven to nine side-by-side trial images; one was a real photograph containing a real object, and the other is a synthetic image produced by this method (containing a synthetic version of the real object in the real photo). Subjects were asked to choose the image they felt looked most realistic. Trials and conditions were permuted to ensure even coverage. An example trial pair is shown in Fig~\ref{fig:userstudy}, and more can be found in supplemental material.

\boldhead{Conditions} We tested six binary conditions that we hypothesized may contribute to a subject's performance: 
{\bf expert} subject; {\bf realistic shape}; {\bf complex material}; {\bf automatic} result; {\bf multiple objects} inserted; and whether the trial occured in the {\bf first half} of a subject's study (to test whether there was any learning effect present). Subjects were deemed experts if they had graphics/vision experience (in the study, this meant graduate students and researchers at a company lab), and non-experts otherwise. The authors classified objects into realistic/synthetic shape and complex/simple material beforehand. We created results with our automatic drag-and-drop procedure as one condition (``automatic''), and using our interface, created an improved (as judged by the authors) version of each result by manually adjusting the light intensities (``refined''). 

\boldhead{Results} On average, the result generated by our {\bf automatic}  method was selected as the real image in 34.1\% of the 232 pairs that subjects viewed. The refined condition achieved 35.8\% confusion, however these distributions do not significantly differ using a two-tailed t-test. An optimal result would be 50\%. For comparison, Karsch et al.~\cite{Karsch:SA2011} evaluated their semiautomatic method with a similar user study, and achieved 34\% confusion, although their study (scenes, subjects, etc) were not identical to ours.

We also tested to see if any of the other conditions (or combination of conditions) were good predictors of a person's ability to perform this task. We first used logistic regression to predict which judgements would be correct using the features, followed by a Lasso ($L_1$ regularized logistic regression~\cite{Tibshirani}, in the Matlab R2012b implementation) to identify features that predicted whether the subject judged a particular pair correctly.   We evaluated deviance (mean negative log-likelihood on held out data) with 10-fold cross validation.  The lowest  deviance regression ignores all features; incorporating the ``expert'' and ``realistic shape'' features causes about  a 1\% increase in the deviance; incorporating other features causes the deviance to increase somewhat.  This strongly suggests that none of the features actually affect performance.   

\begin{figure}[t]
\begin{overpic}[width=.49\linewidth,clip,trim=120 60 60 60]{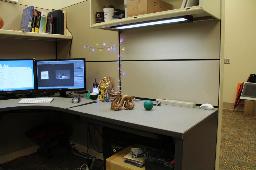}\put(0,0){\colorbox{white}{\large A1}}
\end{overpic}
\begin{overpic}[width=.49\linewidth,clip,trim=120 60 60 60]{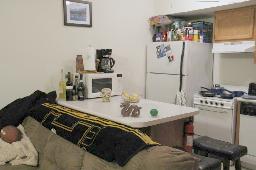}\put(0,0){\colorbox{white}{\large A2}}
\end{overpic}
\vspace{1mm} \\
\begin{overpic}[width=.49\linewidth,clip,trim=0 0 0 0]{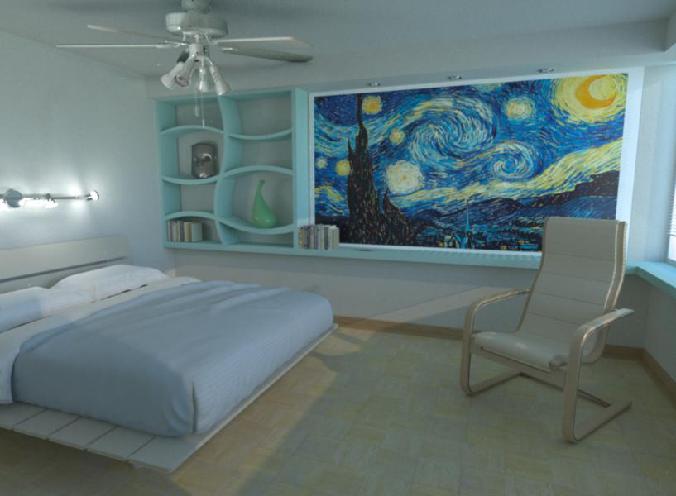}\put(0,2){\colorbox{white}{\large B1}}
\end{overpic}
\begin{overpic}[width=.49\linewidth,clip,trim=0 0 0 0]{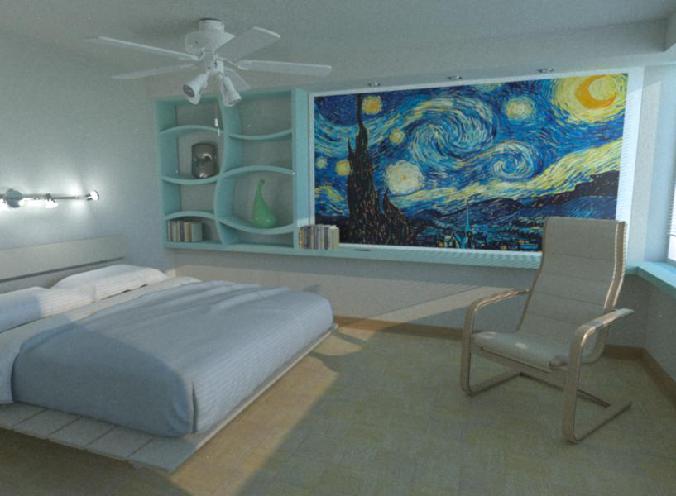}\put(0,2){\colorbox{white}{\large B2}}
\end{overpic}
\caption{Example trials from our ``real image'' user study (top) and our ``synthetic image'' user study (bottom). In the studies, users were shown two side-by-side pictures; one photograph is real (or rendered photorealistically with an exact 3D scene), and the other has synthetic objects inserted into it with this method. Users were instructed to choose the picture from the pair that looked the most realistic. For each row, which of the pair would you choose? Best viewed in color at high-resolution.  {\it Scene modeling credits (bottom): Matthew Harwood.}
}
\begin{turn}{180}\centerline{\it A1, B2: real/fully rendered; A2, B1: created with this method}\end{turn}
\label{fig:userstudy}
\end{figure}

\subsection{Synthetic image user study}
\label{sec:study:synthetic}
Our ``real image'' study provides encouraging initial results, but it is limited due to the nature of data collection. 
Collecting corresponding real and synthetic objects with high-quality geometry and reflectance properties can be extremely difficult, so we confined our real image study to small, tabletop accessories (household items and 3D-printable items).

In this follow-up study, we utilize highly realistic, synthetic 3D scenes in order to more extensively evaluate this method (objects with larger-pixel coverage and varying materials/geometry, diverse lighting conditions, indoors/outdoors).

\boldhead{Experimental setup} We collected four synthetic (yet highly realistic\footnote{We pre qualified these images as highly-confusable with real pictures in a preliminary study; see supplemental material.}) scenes -- three indoor, one outdoor. For each scene, we inserted three realistic 3D models using conventional modeling software (Blender, \url{http://www.blender.org/}), and rendered each scene using LuxRender under three lighting conditions (varying from strongly directed to diffuse light), for a total of 36 (= 4 scenes $\times$ 3 objects $\times$ 3 lighting conditions) unique and varied scenes (viewable in supplemental material). Next, we used this method to insert the same 3D models in roughly the same location into the empty, rendered images, resulting in 36 ``synthetic'' insertion results corresponding to the 36 ground truth rendered images. 

For the study, each participant viewed 12 pairs\footnote{Image pairs are randomly permuted/selected so that each of the 12 objects appears exactly once to each subject.} of corresponding images, and was asked to select which image he/she felt looked the most realistic (two-alternative forced choice). For example, a subject might see two identical bedroom scenes with a ceiling fan, except in one picture, the fan had actually been inserted using this method (see Fig~\ref{fig:userstudy}).

We polled 450 subjects using Mechanical Turk. In an attempt to avoid inattentive subjects, each study also included four ``qualification'' image pairs (a cartoon picture next to a real image) placed throughout the study in a stratified fashion. Subjects who incorrectly chose any of the four cartoon picture as realistic were removed from our findings (16 in total, leaving 434 studies with usable data).

At the end of the study, we showed subjects two additional image pairs: a pair containing rendered spheres (one a physically plausible, the other not), and a pair containing line drawings of a scene (one with proper vanishing point perspective, the other not). For each pair, subjects chose the image they felt looked most realistic. Then, each subject completed a brief questionnaire, listing demographics, expertise, and voluntary comments.

\begin{table}
\centerline{\bf \Large Fraction of times subjects chose a synthetic insertion result}
\centerline{\bf \Large over the ground truth insertion in our user study}
\begin{center}
\Large
\begin{tabular}{| l c || c | c | c |} \hline
Condition & N & ours & Khan & Lalonde/matching \\ \hline

indoor & 1332 & \cellcolor{darkyellow} {\bf .377} & .311 & .279 \\
outdoor & 444 &  \cellcolor{darkyellow} .288 & {\bf .297} & .240 \\  \hline

diffuse light & 1184 &  \cellcolor{darkyellow} {\bf .377} & \cellcolor{darkyellow}  .332 &  \cellcolor{darkyellow} .238 \\ 
directional light & 592 &  \cellcolor{darkyellow} {\bf .311} &  \cellcolor{darkyellow} .257 & \cellcolor{darkyellow}  .330 \\ \hline

simple material & 1036 & {\bf .353} & .301 & \cellcolor{darkyellow}.248 \\
complex material & 740 & {\bf .357  }  &    .315     &  \cellcolor{darkyellow} .299\\  \hline

large coverage  & 1332 & {\bf .355} & \cellcolor{darkyellow} .324 & .277 \\
small coverage  & 444 & {\bf .356} & \cellcolor{darkyellow}  .253 & .245 \\  \hline


good composition  & 1036 & {\bf .356} & .312 & .265 \\
poor composition  & 740 & {\bf .353} & .300 & .274 \\  \hline

good perspective  & 1036 & \cellcolor{darkyellow}  {\bf .378} & .316 & .287 \\
poor perspective  & 740 & \cellcolor{darkyellow}  {\bf .322} & .295 & .244 \\  \hline

male  & 1032 & {\bf .346} & .295 & .267 \\
female  & 744 & {\bf .367} & .319 & .272 \\  \hline

age ($\leq$25)  & 468 & {\bf .331} &  .294  & \cellcolor{darkyellow}   .234 \\
age ($>$25) & 1164 &  {\bf .363} & .312 & \cellcolor{darkyellow} .284 \\  \hline

color normal  & 1764 & {\bf .353} & .308 & .267 \\
not color normal & 12 & {\bf .583} & .167 & .361 \\  \hline

passed p-s tests & 1608 & \cellcolor{darkyellow}  {\bf .341} & \cellcolor{darkyellow}  .297 & \cellcolor{darkyellow}  .260 \\
failed p-s tests & 168 &  \cellcolor{darkyellow}  .482 & \cellcolor{darkyellow}   {\bf .500} & \cellcolor{darkyellow}  .351 \\  \hline 

non-expert  & 1392 & {\bf .361} & .312 & .275 \\
expert  & 384 & {\bf .333} & .288 & .252 \\  \hline 
\hline
overall & 1776 &  {\bf .355} & .307 & .269  \\ \hline

\end{tabular}
\end{center}
\vspace{0mm}
\label{tab:study_results_conditions}
\caption{
Highlighted blocks indicate that there are significant differences in confusion when a particular condition is on/off ($p$ value $< 0.05$ using a 2-tailed t-test). For example, in the top left cell, the ``ours indoor'' distribution was found to be significantly different from the ``ours outdoor'' distribution. For the Lalonde/matching column, the method Lalonde et al. is used for outdoor images, and the template matching technique is used indoors (see text for details). The best method for each condition is shown in bold. N is the total number of samples. Overall, our confusion rate of 35.5\% better than the baselines (30.7\% and 26.9\%) by statistically significant margins.
}
\end{table}

\boldhead{Methods tested} We are primarily interested in determining how well this method compares to the ground truth images, but also test other illumination estimation methods as baselines.  We generate results using the method of Khan et al.~\cite{Khan:2006:IME:1179352.1141937} (projecting the input image onto a hemisphere, duplicating it, and using this as the illumination environment), the method of Lalonde et al.~\cite{Lalonde:2009vp} (for outdoor images only), and a simplified IBL-matching technique that finds a similar IBL to the input image by template matching (similar to~\cite{Xiao:CVPR:12}; we use this method indoors only).

For each method, only the illumination estimation stage of our pipeline changes (depth, reflectance, and camera estimation remain the same), and all methods utilize our lighting optimization technique to estimate source intensity (Sec~\ref{sec:illum:render}).

\boldhead{Conditions} Because we found no condition in our initial study to be a useful predictor of people's ability to choose the real image, we introduce many new (and perhaps more telling) conditions in this study: {\bf indoor/outdoor} scene, {\bf diffuse/direct lighting} in the scene, {\bf simple/complex material} of inserted object, {\bf good/poor composition} of inserted object, and if the inserted object has {\bf good/poor perspective}. We also assigned each subject a set of binary conditions: {\bf male/female}, {\bf age $<25$ / $\geq 25$}, {\bf color-normal / not color-normal}, whether or not the subject correctly identified both the physically accurate sphere {\it and} the proper-perspective line drawing at the end of the study ({\bf passed/failed perspective-shading (p-s) tests}), and also {\bf expert/non-expert} (subjects were classified as experts only if they passed the perspective-shading tests {\it and} indicated that they had expertise in art/graphics).

\boldhead{Results and discussion}
Overall, our synthetic image study showed that people confused our insertion result with the true rendered image in over 35\% of 1776 viewed image pairs (an optimal result would be 50\%). We also achieve better confusion rates than the methods we compared to, and Table~\ref{tab:study_results_conditions} shows these results. 

\begin{figure}
\begin{center}
\begin{minipage}{1.0\linewidth}
\begin{center}
\begin{minipage}{0.325\linewidth} 
\centerline{\Large Ground truth} \vspace{0.5mm}
\includegraphics[width=\linewidth,clip=true,trim=0 40 0 130]{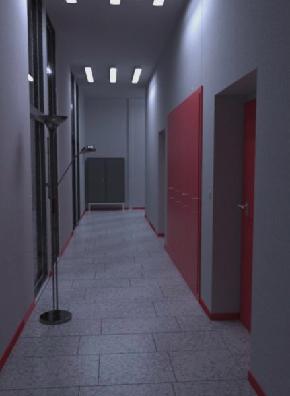}\\
\includegraphics[width=\linewidth,clip=true,trim=0 200 250 0]{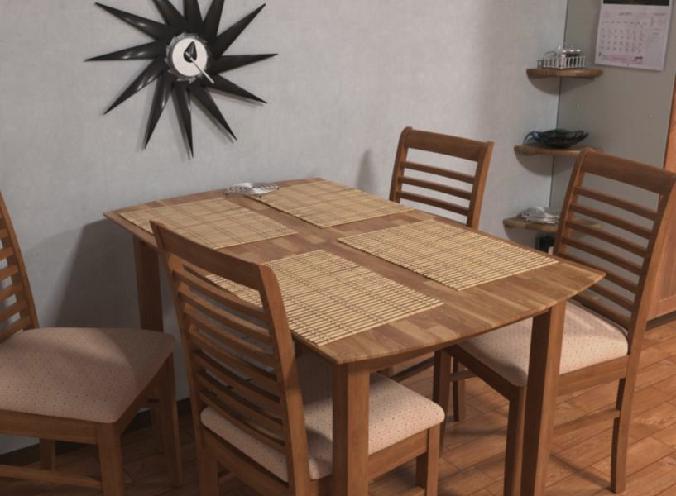}\\
\includegraphics[width=\linewidth]{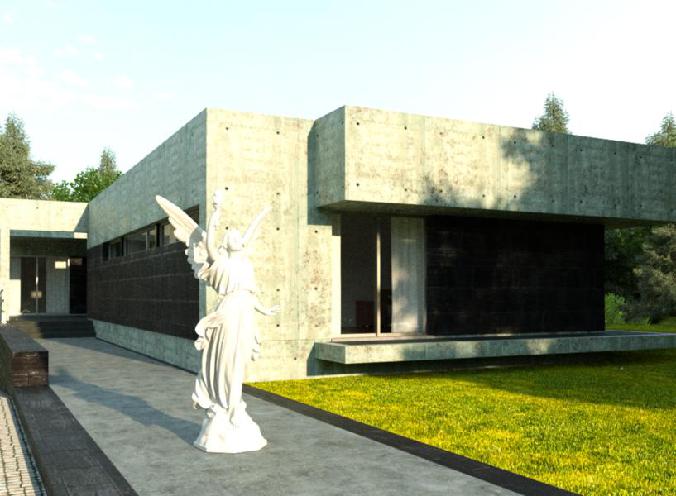}
\end{minipage}
\begin{minipage}{0.325\linewidth} 
\centerline{\Large Our result} \vspace{0.5mm}
\includegraphics[width=\linewidth,clip=true,trim=0 40 0 130]{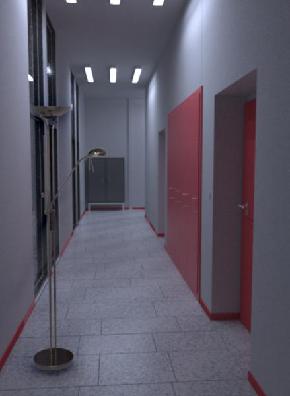}\\
\includegraphics[width=\linewidth,clip=true,trim=0 200 250 0]{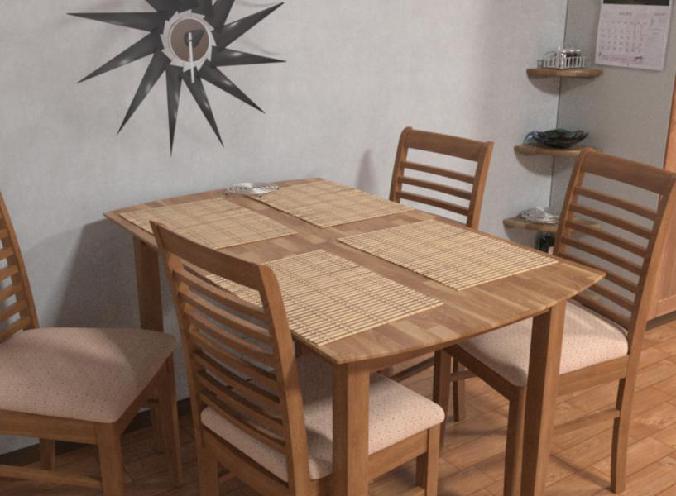}\\
\includegraphics[width=\linewidth]{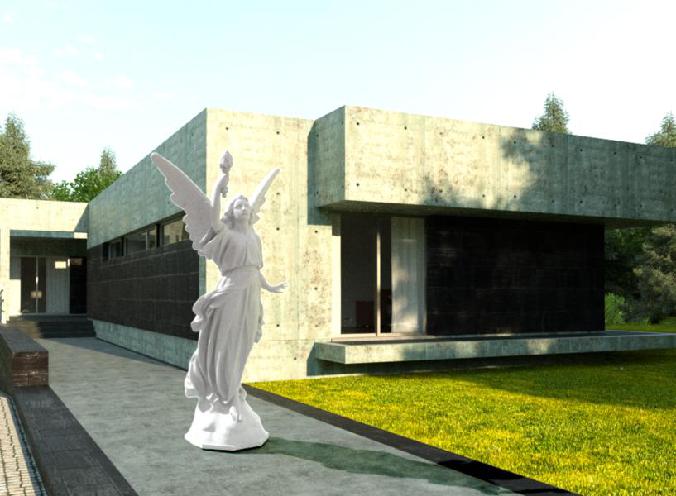}
\end{minipage}
\begin{minipage}{0.325\linewidth} 
\centerline{\Large Khan et al.~\cite{Khan:2006:IME:1179352.1141937} lighting}
\includegraphics[width=\linewidth,clip=true,trim=0 40 0 130]{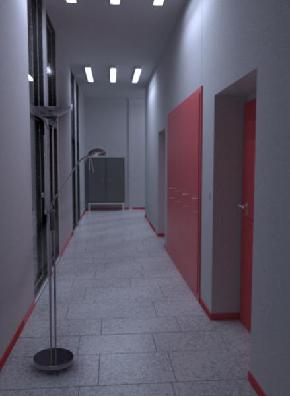}\\
\includegraphics[width=\linewidth,clip=true,trim=0 200 250 0]{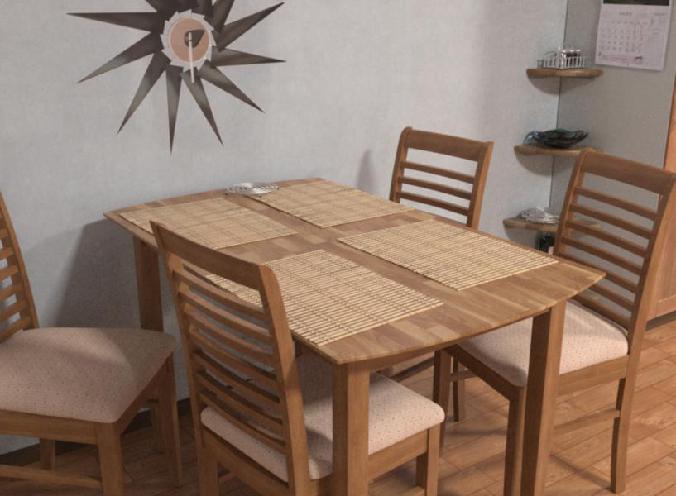}\\
\includegraphics[width=\linewidth]{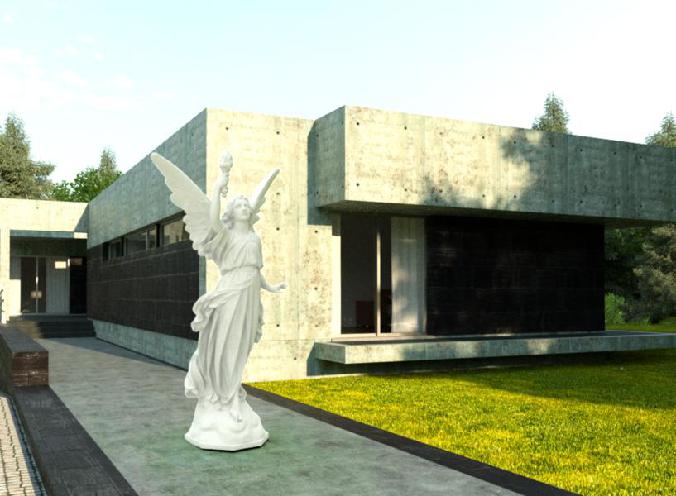}
\end{minipage}
\end{center}
\end{minipage}
\end{center}
\caption{Example results used in our synthetic user study (corresponding to the scenes shown in Figure~\protect\ref{fig:groundtruth}). We compared the ground truth renderings (left) to our insertion method (middle) and others (e.g. using the method of Khan et al. for relighting, right). {\it Scene modeling credits (top to bottom): {\footnotesize \textcopyright{}}Simon Wendsche, {\footnotesize \textcopyright{}}Peter Sandbacka, {\footnotesize \textcopyright{}}Eduardo Camara.}
}
\label{fig:studysynth_results}
\end{figure}

Although the differences between this method and the other methods might appear smaller ($\scriptstyle\mathtt{\sim}$5-10 percentage points), these differences are statistically significant using a two-tailed test. Furthermore, we are only assessing differences in light estimation among these method (since it is our primary technical contribution); every other part of our pipeline remains constant (e.g. camera and depth estimation, as well as the light intensity optimization). In other words, we do not compare directly to other techniques, rather these other lighting techniques are aided by bootstrapping them with the rest of our pipeline.

We also analyze which conditions lead to significant differences in confusion for each method, indicated by highlighted cells in the table. As expected, this method works best indoors and when the lighting is not strongly directed, but still performs reasonably well otherwise (over 31\% confusion). Also, this method looks much more realistic when inserted objects have reasonable perspective.

Perhaps most interestingly, this method is almost confusable at chance with ground truth for people who have a hard time detecting shading/perspective problems in photographs (failed p-s tests), and the population that passed is significantly better at this task than those that failed. This could mean that such a test is actually a very good predictor for whether a person will do well at this task, or it could be viewed as a secondary ``qualification'' test (i.e. similar yet possibly more difficult than the original cartoon-real image qualification tests). Either way, for the population that passed these tests, the confusion rates are very similar to the overall mean (i.e. results are very similar even if this data is disregarded).

All other conditions did not have a significant impact on the confusion rate for this method. Again, we observe that the ``expert'' subjects did not perform significantly better ($p$ value = 0.323) than ``non-expert'' subjects (by our classification), although they did choose the ground truth image more times on average.

\subsection{Ground truth comparison}
\label{sec:study:quant}

We also measure the accuracy of our scene estimates by comparing to ground truth. In Figure~\ref{fig:studysynth_results}, we show three images from our ``synthetic'' user study: rendered images (ground truth) are compared with insertion results using our technique as well as results relit using the method of Khan et al.~\cite{Khan:2006:IME:1179352.1141937} (but this method is used for all other components, e.g. depth, light optimization, and reflectance). 

Figure~\ref{fig:groundtruth} shows our inverse rendering estimates (depth, reflectance, and illumination) for the same three scenes as in Figure~\ref{fig:studysynth_results}. These images illustrate typical errors produced by our algorithm: depth maps can be confused by textured surfaces and non-planar regions, reflectance estimates may fail in the presence of hard shadows and non-Lambertian materials, and illumination maps can contain inaccurate lighting directions or may not appear similar to the actual light reflected onto the scene. These downfalls suggest future research directions for single image inference.

Notice that while in some cases our estimates differ greatly from ground truth, our insertion results can be quite convincing, as demonstrated by our user studies. This indicates that the absolute physical accuracy of inverse rendering is not a strong indicator of how realistic an inserted object will look; rather, we believe that relative cues are more important. For example, inserted objects typically cast better shadows in regions of planar geometry and where reflectance is constant and relatively correct (i.e. erroneous up to a scale factor); strong directional illuminants need not be perfectly located, but should be consistent with ambient light. Nonetheless, 3D object insertion will likely benefit from improved estimates.

For quantitative errors and additional results, we refer the reader to supplemental material.

\newcommand{\includestudydecomposition}[8]{
\begin{center}
\begin{minipage}{#6\linewidth}
\begin{center}
\begin{minipage}{0.2\linewidth} 
\vspace{13mm}\centerline{\Large #1}\vspace{13mm} 
\centerline{Rendered image} 
\includegraphics[width=\linewidth,clip=true,trim=#2 #3 #4 #5]{./auto_illum/fig/userstudy-synthetic/img/#1-#7.jpg}
\end{minipage}
\begin{minipage}{0.2\linewidth} 
\centerline{True depth}
\includegraphics[width=\linewidth,clip=true,trim=#2 #3 #4 #5]{./auto_illum/fig/userstudy-synthetic/decompositions/depth/#1-true.jpg}\\
\centerline{Estimated depth} 
\includegraphics[width=\linewidth,clip=true,trim=#2 #3 #4 #5]{./auto_illum/fig/userstudy-synthetic/decompositions/depth/#1-#7-ours.jpg}
\end{minipage}
\begin{minipage}{0.2\linewidth} 
\centerline{True reflectance} \vspace{0.5mm}
\includegraphics[width=\linewidth,clip=true,trim=#2 #3 #4 #5]{./auto_illum/fig/userstudy-synthetic/decompositions/refl/#1-true.jpg}\\
\centerline{Estimated reflectance}\vspace{0.5mm}
\includegraphics[width=\linewidth,clip=true,trim=#2 #3 #4 #5]{./auto_illum/fig/userstudy-synthetic/decompositions/refl/#1-#7-ours.jpg}
\end{minipage}
\begin{minipage}{#8\linewidth} 
\centerline{True illumination} \vspace{0.5mm}
\includegraphics[width=\linewidth]{./auto_illum/fig/userstudy-synthetic/decompositions/illum/#1-#7-true.jpg}\\
\centerline{Estimated illumination}\vspace{0.5mm}
\includegraphics[width=\linewidth]{./auto_illum/fig/userstudy-synthetic/decompositions/illum/#1-#7-ours.jpg}
\end{minipage}
\end{center}
\end{minipage}
\end{center}
}

\begin{figure*}
\includestudydecomposition{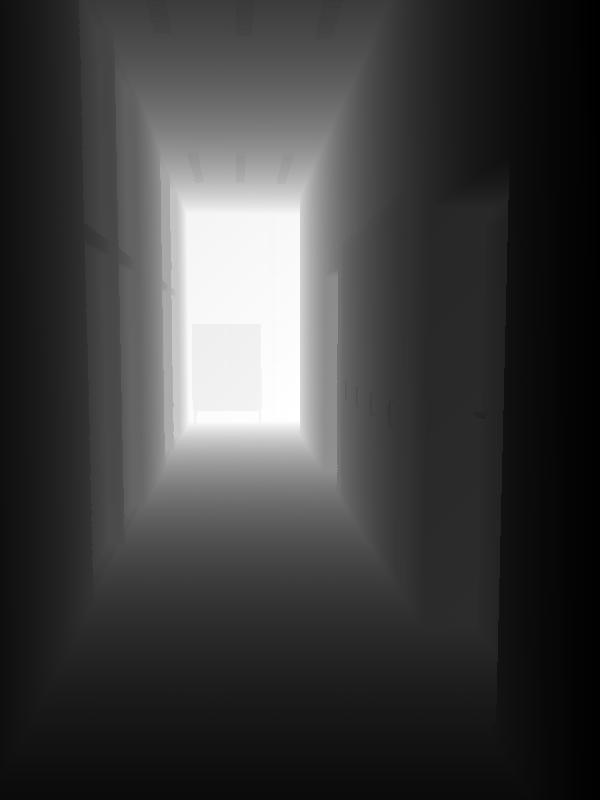}{0}{80}{0}{200}{0.85}{C}{0.345}
\vspace{3mm}
\centerline{\rule{0.95\linewidth}{0.1mm}}
\vspace{3mm}
\includestudydecomposition{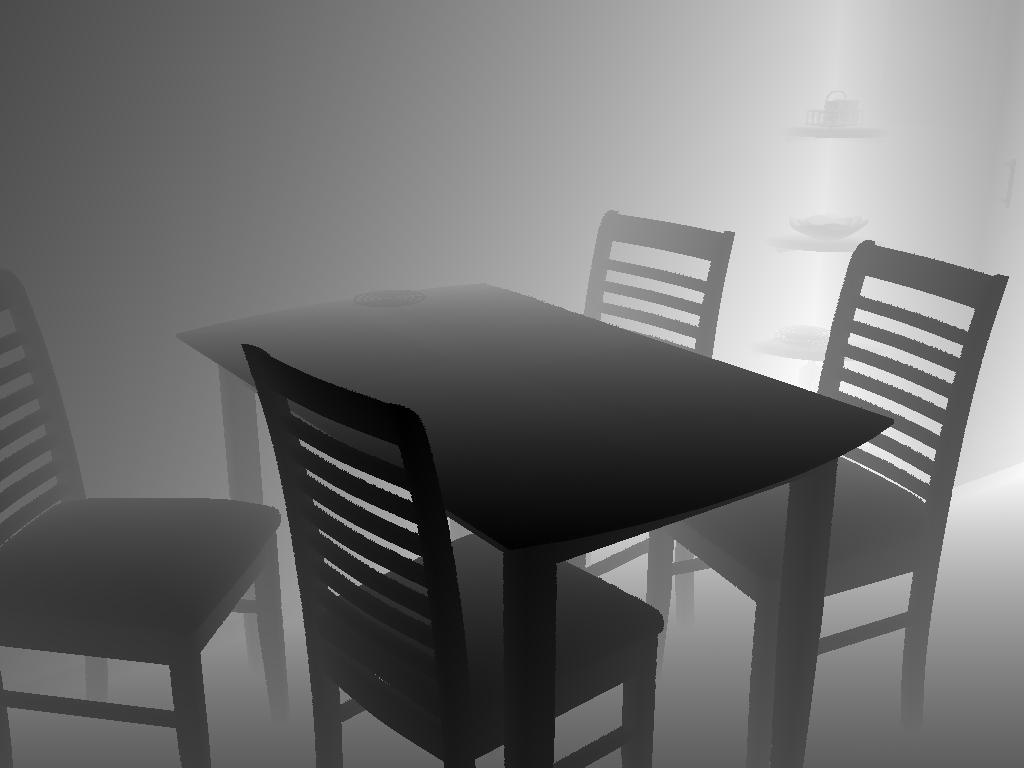}{0}{0}{0}{0}{1}{B}{0.295}
\vspace{3mm}
\centerline{\rule{0.95\linewidth}{0.1mm}}
\vspace{3mm}
\includestudydecomposition{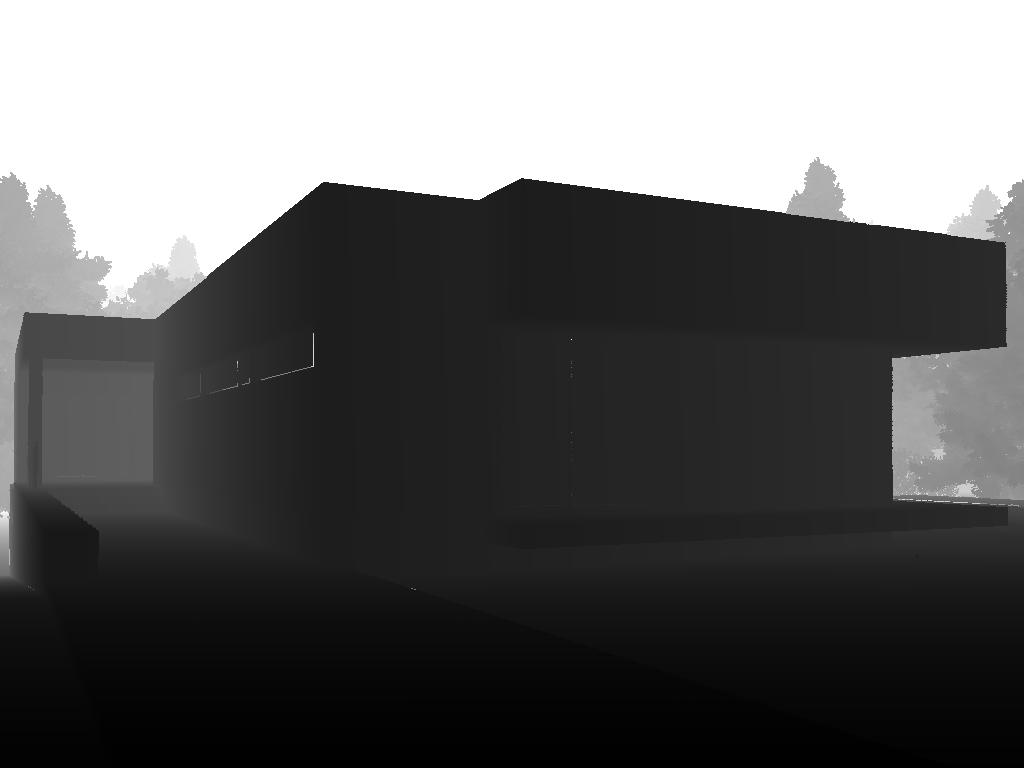}{0}{0}{0}{0}{1}{A}{0.295}
\caption{Comparison of ground truth scene components (depth, diffuse reflectance, and illumination) with our estimates of these components. Illumination maps are tonemapped for display. See Figure~\protect\ref{fig:studysynth_results} and the supplemental document for insertion results corresponding to these scenes. {\it Scene modeling credits (top to bottom): {\footnotesize \textcopyright{}}Simon Wendsche, {\footnotesize \textcopyright{}}Peter Sandbacka, {\footnotesize \textcopyright{}}Eduardo Camara.}}
\label{fig:groundtruth}
\end{figure*}

\section{Results and conclusion}
\label{sec:results}

\begin{figure*}
\setlength\fboxsep{0pt}
\setlength\fboxrule{1pt}
\begin{centering}
\begin{minipage}{1\linewidth}
\begin{overpic}[width=.32\linewidth,clip=true,trim=0 50 0 0]{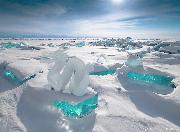}
\put(88,57){\fcolorbox{white}{white}{\includegraphics[width=0.13\linewidth,clip,trim=0 10 0 0]{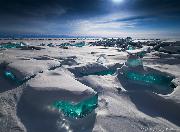}}}
\end{overpic}
\hfill
\begin{overpic}[width=.32\linewidth,clip=true,trim=0 20 0 72]{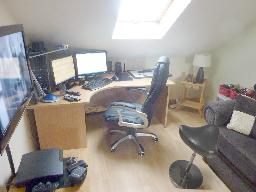}
\put(0,1){\fcolorbox{white}{white}{\includegraphics[width=0.13\linewidth,clip,trim=0 20 0 72]{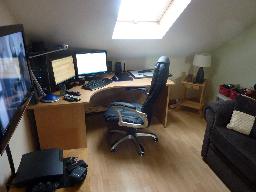}}}
\end{overpic}
\hfill
\begin{overpic}[width=.32\linewidth,clip=true,trim=0 0 0 35]{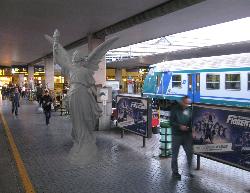}
\put(87,52){\fcolorbox{white}{white}{\includegraphics[width=0.13\linewidth,clip,trim=0 0 0 0]{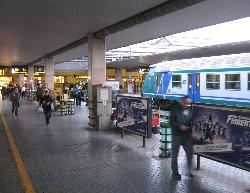}}}
\end{overpic}
\\
\begin{overpic}[width=.32\linewidth,clip=true,trim=0 0 0 0]{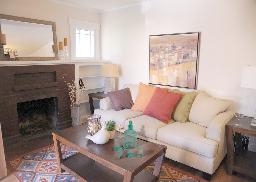}
\put(0,1){\fcolorbox{white}{white}{\includegraphics[width=0.13\linewidth,clip,trim=0 0 0 25]{./auto_illum/fig/results_supp/martini.jpg}}}
\end{overpic}
\hfill
\begin{overpic}[width=.32\linewidth,clip,trim=18 18 18 160]{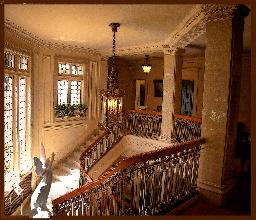}
\put(89,65){\fcolorbox{white}{white}{\includegraphics[width=.13\linewidth,clip,trim=18 18 18 188]{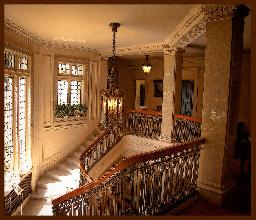}}}
\end{overpic}
\hfill
\begin{overpic}[width=.32\linewidth,clip,trim=67 18 0 22]{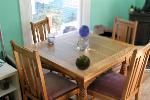}
\put(0,1){\fcolorbox{white}{white}{\includegraphics[width=.13\linewidth,clip,trim=67 18 0 22]{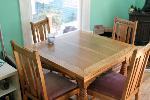}}}
\end{overpic}
\end{minipage}
\end{centering}
\vspace{-3mm}
\caption{Additional results produced by this method (original picture is overlaid on top of the result). this method is capable of producing convincing results both indoors and outdoors, and can be robust to non-manhattan scenes. this method can be applied to arbitrary images, and makes no explicit assumptions about the scene geometry; virtual staging is one natural application of our system. Best viewed in color at high-resolution. {\it Photo credits (left to right, top to bottom): {\footnotesize \textcopyright{}}Alexey Trofimov, {\footnotesize \textcopyright{}}Sean MacEntee, Flickr users {\footnotesize \textcopyright{}}``AroundTuscany,'' {\footnotesize \textcopyright{}}``Wonderlane,'' {\footnotesize \textcopyright{}}``PhotoAtelier,'' and {\footnotesize \textcopyright{}}Brian Teutsch}.
}
\label{fig:autoillum:results}
\end{figure*}

We show typical results produced by our system in Fig~\ref{fig:autoillum:results}. this method is applicable both indoors and outdoors and for a variety of scenes. Although a Manhattan World is assumed at various stages of our algorithm, we show several reasonable results even when this assumption does not hold. Failure cases are demonstrated in Fig~\ref{fig:failure_examples}. Many additional results (varying in quality and scene type) can be found in the supplemental document. The reader is also referred to the accompanying video for a demonstration of our system.

As we found in our user study, this method is better suited for indoor scenes or when light is not strongly directional. In many cases, people confused our insertion results as real pictures over one third of the time. For outdoor scenes, we found that simpler illumination methods might suffice (e.g.~\cite{Khan:2006:IME:1179352.1141937}), although our geometry estimates are still useful for these scenes (to estimate light intensity and to act as shadow catchers).

We would like to explore additional uses for our scene models, such as for computer gaming and videos. It would be interesting to try other physically grounded editing operations as well; for example, deleting or moving objects from a scene, or adding physically-based animations when inserting objects (e.g., dragging a table cloth over a table). Extending this method to jointly infer a scene all at once (rather than serially) could lead to better estimates. Our depth estimates might also be useful for inferring depth order and occlusion boundaries.

Judging by our synthetic image user study, this method might also be useful as a fast, incremental renderer. For example, if a 3D modeler has created and rendered a scene, and wants to insert a new object quickly, this method could be used rather than re-rendering the full scene (which, for most scenes, should incur less render time since our estimated scene may contain significantly fewer light sources and/or polygons).

We have presented a novel image editor that, unlike most image editors that operate in 2D, allows users to make physically meaningful edits to an image with ease. Our software supports realistic object insertion,  on-the-fly lighting changes, post-process depth of field modifications, and is applicable to legacy, LDR images. These interactions are facilitated by our automatic scene inference algorithm, which encompasses our primary technical contributions: single image depth estimation, and data-driven illumination inference. Results produced by our system appear realistic in many cases, and a user study provides good evidence that objects inserted with our fully automatic technique look more realistic than corresponding real photographs over one third of the time.

\begin{figure*}
\setlength\fboxrule{1pt}
\setlength\fboxsep{0pt}
\begin{minipage}{.32\linewidth}
\begin{overpic}[width=\linewidth,clip=true,trim=0 0 0 0]{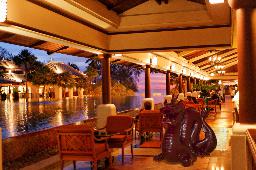}
\put(0,1){\fcolorbox{white}{white}{\includegraphics[width=0.33\linewidth,clip,trim=0 0 0 0]{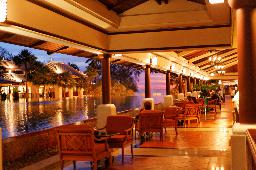}}}
\end{overpic}\\
\begin{overpic}[width=\linewidth,clip=true,trim=0 100 0 0]{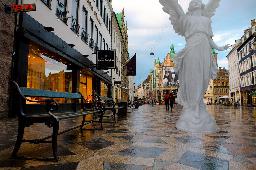}
\put(0,1){\fcolorbox{white}{white}{\includegraphics[width=0.33\linewidth,clip,trim=0 0 0 0]{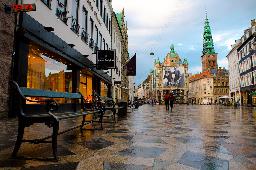}}}
\end{overpic}
\end{minipage}
\hfill
\begin{minipage}{.32\linewidth}
\begin{overpic}[width=\linewidth,clip=true,trim=0 65 0 0]{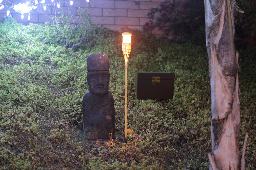}
\put(0,1){\fcolorbox{white}{white}{\includegraphics[width=0.33\linewidth,clip,trim=0 0 0 0]{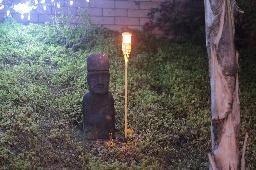}}}
\end{overpic}\\
\begin{overpic}[width=\linewidth,clip=true,trim=0 0 0 130]{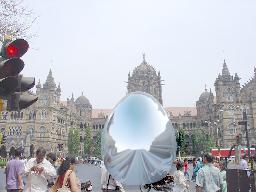}
\put(100,55){\fcolorbox{white}{white}{\includegraphics[width=0.33\linewidth,clip,trim=0 0 0 0]{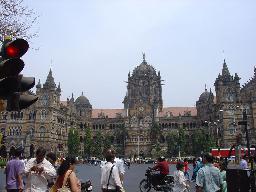}}}
\end{overpic}
\end{minipage}
\hfill
\begin{minipage}{.32\linewidth}
\begin{overpic}[width=\linewidth,clip=true,trim=0 70 0 0]{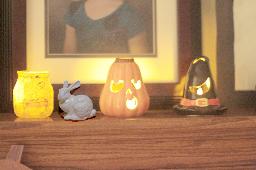}
\put(100,1){\fcolorbox{white}{white}{\includegraphics[width=0.33\linewidth,clip,trim=0 0 0 0]{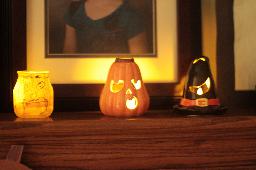}}}
\end{overpic}\\
\begin{overpic}[width=\linewidth,clip=true,trim=0 70 0 350]{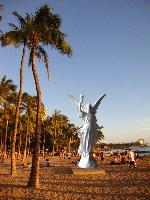}
\put(0,0){\fcolorbox{white}{white}{\includegraphics[width=0.33\linewidth,clip,trim=0 0 0 0]{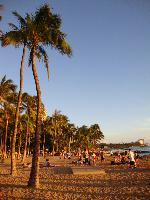}}}
\end{overpic}
\end{minipage}
\vspace{-1mm}
\caption{Failure cases. In some images, placement can be difficult due to inaccurate geometry estimates (top row), e.g. the dragon's shadow (top left) is not visible due to poor depth estimates, and the chest (top middle) rests on a lush, slanted surface, and our depth is flat in the region of the chest. Insertions that affect large portions of the image (e.g. statue; bottom left) or reflect much of the estimated scene (e.g. curved mirrors; bottom middle) can sometimes exacerbate errors in our scene estimates. Light estimates and color constancy issues lead to less realistic results (top right), especially outdoors (bottom right). Note that the people in the bottom right image have been manually composited over the insertion result. {\it Photo credits (left to right, top to bottom): {\footnotesize \textcopyright{}}Dennis Wong, {\footnotesize \textcopyright{}}Parker Knight, {\footnotesize \textcopyright{}}Parker Knight, {\footnotesize \textcopyright{}}Kenny Louie, Flickr user {\footnotesize \textcopyright{}}``amanderson2,'' and {\footnotesize \textcopyright{}}Jordan Emery.}
}
\label{fig:failure_examples}
\end{figure*}

\chapter{Advanced physically grounded imaging editing}
\label{chap:applications}

In this chapter, we demonstrate methods for going beyond physically grounded object insertion. In Section~\ref{sec:apps}, we describe several methods for using our scene estimates to achieve both object removal and modification, including physical interactions with scene elements in photographs (e.g. a ball bouncing on a bed). Our scene estimates also make it possible to automatically predict where newly inserted object should be placed. In Section~\ref{sec:constructaide}, we demonstrate the use of inverse rendering in other contexts, including automatic visualization and construction site monitoring.

\section{Beyond object insertion}
\label{sec:apps}

\begin{figure*}
\includegraphics[width=.245\linewidth]{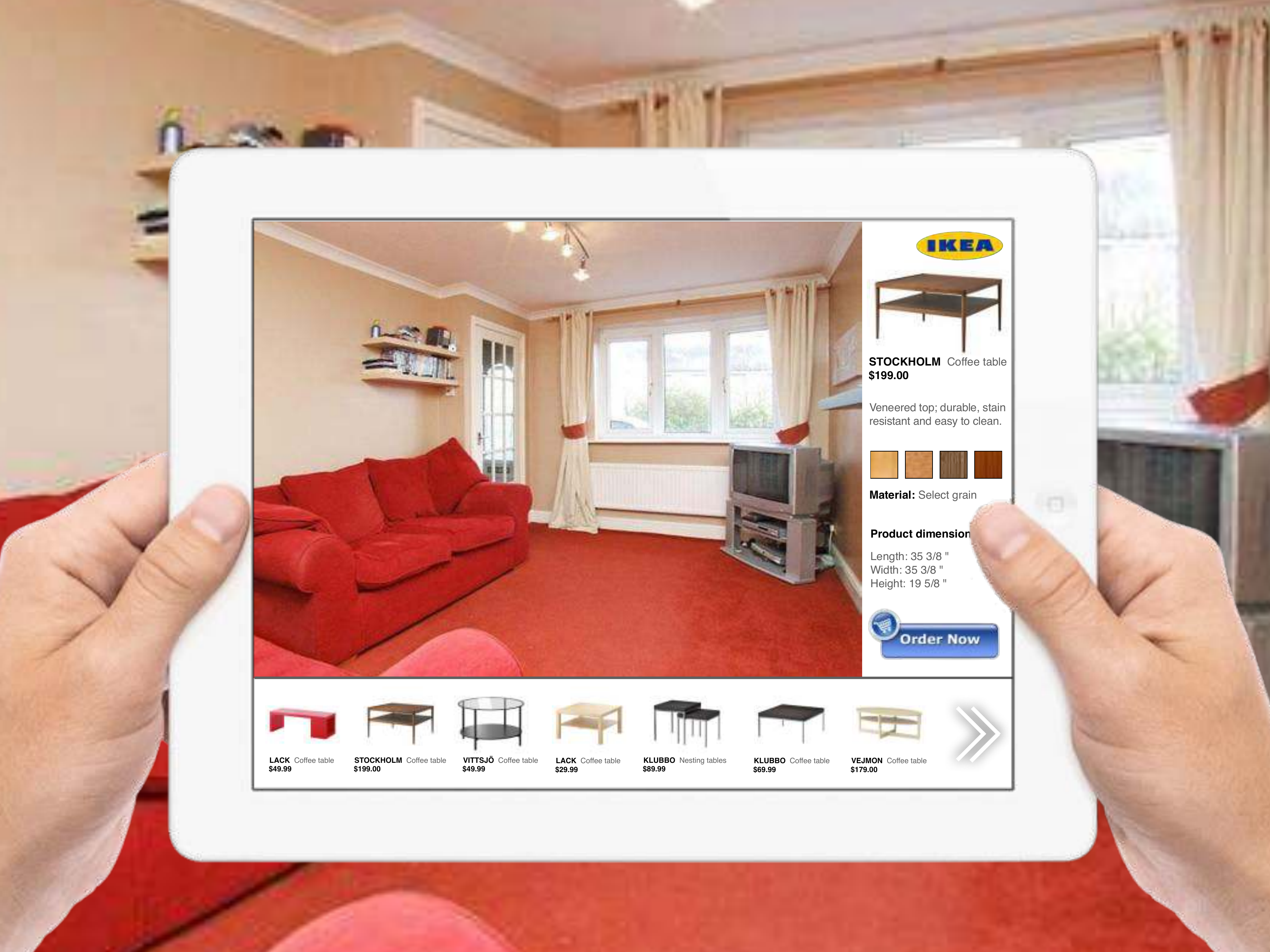}
\includegraphics[width=.245\linewidth]{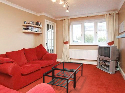}
\includegraphics[width=.245\linewidth]{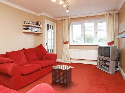}
\includegraphics[width=.245\linewidth]{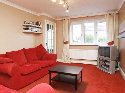}
\caption{By combining physically grounded image editing with geometric and semantic data, rooms can be designed and redecorated automatically. The existing layout of the room is assessed, and new objects and their placements are suggested to the user by realistically inserting objects into the image. Mockup software shown on left, followed by automatically generated furniture layouts/renderings on right. This type of automatic content creation could also be used to generate extensive datasets to facilitate object detection and recognition algorithms. ({\it App visualization: Scott Satkin}.)}
\label{fig:scott}
\end{figure*}

\subsection{Automatic object insertion and placement}

Given a 3D scene, there are established methods for automatically placing objects in scenes~\cite{Yu:sg11,2012-scenesynth}. We can apply these methods directly to our scenes (e.g. estimated as in Chapter~\ref{chap:illum} or using other geometry estimation techniques~\cite{satkin_bmvc2012}) to allow for automatic object placement, redecoration, and so on. We can then use our preliminary works (Chapters~\ref{chap:rsolp} and~\ref{chap:illum}) to render and composite the automatically placed 3D models realistically into photographs.

Figure~\ref{fig:scott} shows an example of automatically inserting a coffee table into a scene. We envision an application that allows a user to take a picture of a room in their home, and then immediately see many different aesthetic configurations of the room with various pieces and styles of furniture. This all becomes possible once we have reasonable inverse rendering estimates.

\subsection{Removing large objects from complex scenes}

\begin{figure*}
\begin{minipage}{.31\linewidth}
\centerline{Input Image}
\includegraphics[width=\linewidth]{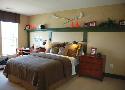}
\centerline{Labelled geometry}
\includegraphics[width=\linewidth]{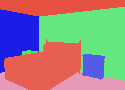}
\end{minipage}
\begin{minipage}{.31\linewidth}
\centerline{Nightstand removed (our method)}
\includegraphics[width=\linewidth]{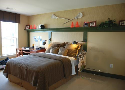}
\centerline{Labelled geometry (nightstand removed)}
\includegraphics[width=\linewidth]{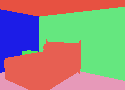}
\end{minipage}
\hfill
\begin{minipage}{.31\linewidth}
\centerline{State-of-the-art (content aware fill~\cite{Barnes:2010:TGP})}
\includegraphics[width=\linewidth]{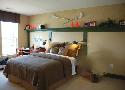}\\
\end{minipage}
\caption{Example of physically grounded object removal. We use the depth and geometry to select an object for removal and intelligently fill in the hole left by removing that object. Even with a basic implementation, we achieve higher quality results than existing techniques.}
\label{fig:michael}
\end{figure*}

We also envision a new technique for seamlessly deleting objects from images. Many algorithms exist for inpainting or synthesizing holes in image (i.e. as created by a ``deleted'' object), yet none work terribly well for complex scenes with many occlusions and structured, non-stochastic patterns/elements. The best technique is currently implemented in Adobe Photoshop (``Content-Aware Fill''). This method is a combination of two state-of-the-art methods on filling image holes~\cite{Barnes:2009:PAR,Wexler:2007}, yet it still fails in many cases, especially when the object is not small or not on a uniform/stochastic background (see Figure~\ref{fig:michael} right).

The use of semantic image labels (obtained automatically~\cite{Silberman:ECCV12,satkin_bmvc2012}) and depth can make the hole-filling process better. To remove an object from the RGB image, we first remove its geometry from the scene (Fig~\ref{fig:michael} bottom); then, we fill in the hole region using the the labelled geometry image as a guide with standard techniques~\cite{Barnes:2009:PAR,Wexler:2007}. As part of this process, image patches can be rectified to ensure filling is done from a frontal view to avoid distortion affects. This approach can also be extended by incorporating existing, anotated RGB-D datasets; for example, to give further intuition on how to inpaint regions that might not be filled well given a lack of data in the input.

\subsection{Physically-based interaction and scene modification}

Our 3D scene estimates can also be useful for interacting with photos in physically meaningful ways beyond inserting and removing objects. Given the ability to insert and remove objects, it is natural to use any combination of these operations to rearrange a scene. However, what if our goal is to interact with the scene and the objects present in the image? This requires even more information than just geometry, lighting, and materials. We would need to know the tactile properties of these objects, and perhaps many other physical properties (mass, volume, etc).

\begin{figure*}
\includegraphics[width=.245\linewidth]{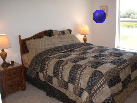}
\includegraphics[width=.245\linewidth]{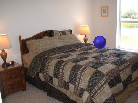}
\includegraphics[width=.245\linewidth]{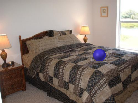}
\includegraphics[width=.245\linewidth]{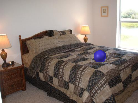}\\
\includegraphics[width=.245\linewidth,clip=true,trim=240 140 80 50]{rgbd/raf/anim1.png}
\includegraphics[width=.245\linewidth,clip=true,trim=240 140 80 50]{rgbd/raf/anim2.png}
\includegraphics[width=.245\linewidth,clip=true,trim=240 140 80 50]{rgbd/raf/anim3.png}
\includegraphics[width=.245\linewidth,clip=true,trim=240 140 80 50]{rgbd/raf/anim4.png}
\caption{Depth data coupled with semantic information (e.g. knowledge that the object in the middle of the scene is a bed, and that beds are soft) allows us to realistically insert physically based animations into scenes. This sequence shows a ball falling onto a bed (changes are subtle but realistic; bottom row shows closeup frames).}
\label{fig:raf}
\end{figure*}

As an example, say we wanted to animate a virtual actor walking to a couch and sitting down on it. How should the couch deform to the actor's body? How soft should it be? How do we modify the affected couch pixels to make this look realistic?

Using semantic image labels and depth can enable such a process. The semantic information provides clues about the physical properties of objects in the scene (hard, soft, hollow, etc) which can be used to drive physically based animations. One example of a ball falling onto a bed is shown in Figure~\ref{fig:raf}. After the animation is inserted, we use our existing techniques (Chapters~\ref{chap:rsolp} and~\ref{chap:illum}) to render and composite the animated object realistically into the photographs.


\section{ConstructAide: Analyzing and visualizing construction sites with photographs and building models}
\label{sec:constructaide}
\begin{figure*}
\vspace{-5mm}
\includegraphics[page=1,width=0.105\linewidth,clip=true,trim=350 -50 0 0]{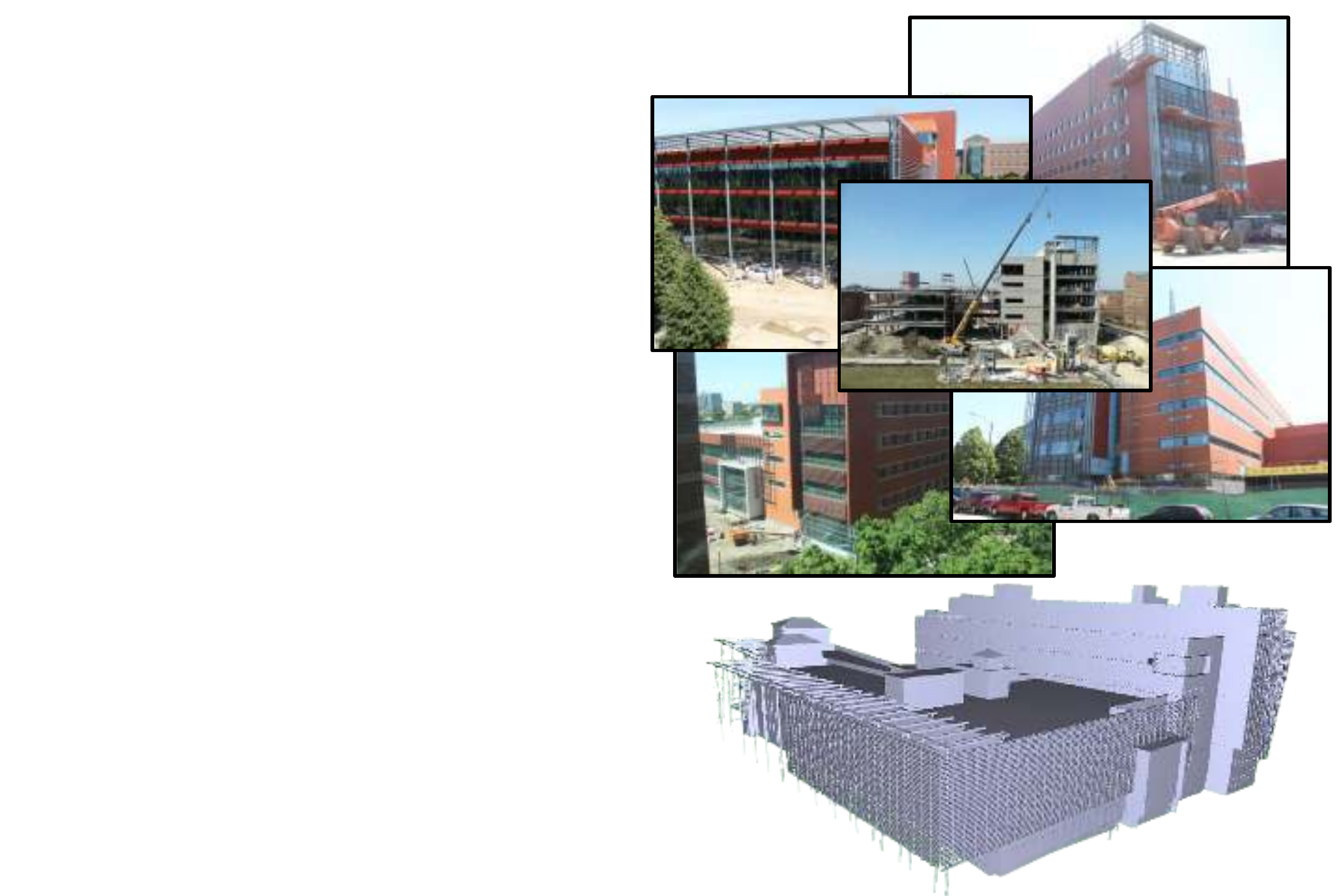}
\includegraphics[page=6,width=0.2175\linewidth,clip=true,trim=0 -50 0 0]{constructaide/fig/teaser-new.pdf}
\includegraphics[page=5,width=0.2175\linewidth,clip=true,trim=40 0 80 0]{constructaide/fig/teaser-new.pdf}
\includegraphics[page=3,width=0.2175\linewidth,clip=true,trim=40 0 80 0]{constructaide/fig/teaser-new.pdf}
\includegraphics[page=4,width=0.2175\linewidth,clip=true,trim=40 0 80 0]{constructaide/fig/teaser-new.pdf}
\\
\begin{minipage}{0.105\linewidth}\centerline{\large (a)}\end{minipage}
\begin{minipage}{0.2175\linewidth}\centerline{\large (b)}\end{minipage}
\begin{minipage}{0.2175\linewidth}\centerline{\large (c)}\end{minipage}
\begin{minipage}{0.2175\linewidth}\centerline{\large (d)}\end{minipage}
\begin{minipage}{0.2175\linewidth}\centerline{\large (e)}\end{minipage}
\caption{Our system aligns sets of photographs with 4D building models (a) to allow for new modes of construction-site interaction and visualization (using the ConstructAide GUI, b), such as architectural renderings (c), 4D navigation (d), and performance monitoring (e).}
\label{fig:constructaide:teaser}
\end{figure*}

\subsection{Introduction}
\begin{figure*}[t]
\begin{minipage}{0.48\linewidth}
\includegraphics[width=.495\linewidth]{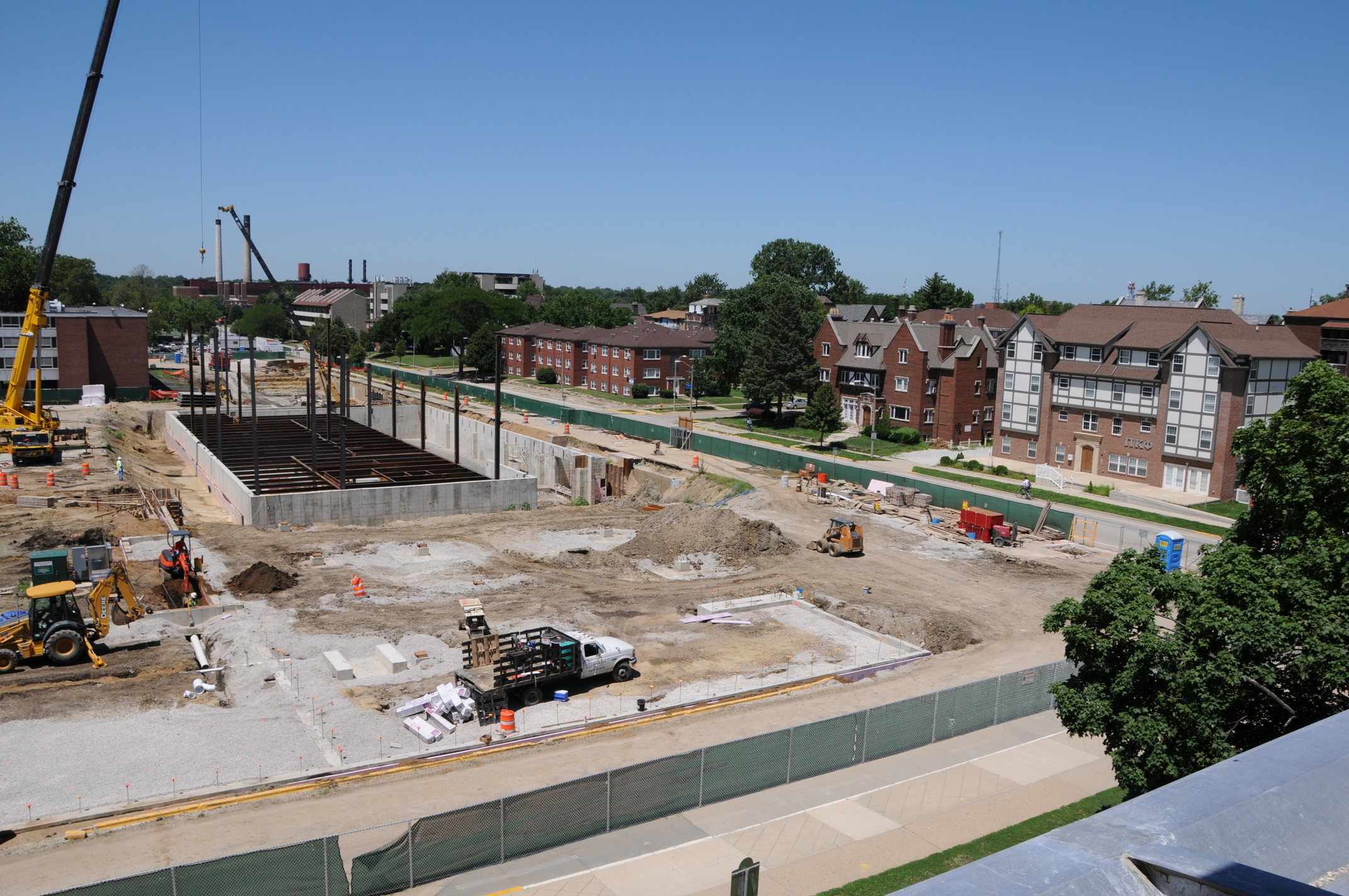}
\includegraphics[width=.495\linewidth]{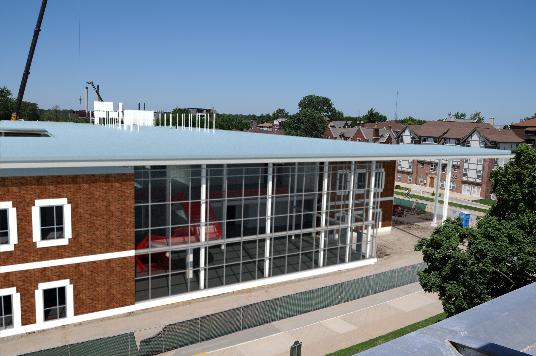}\\
\includegraphics[width=.495\linewidth]{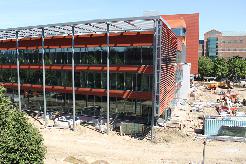}
\includegraphics[width=.495\linewidth]{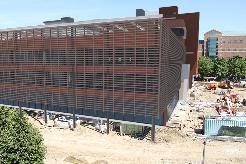}
\caption{Our software automatically renders and composites photorealistic visualizations of the building model into in-progress construction photos. Original photos on left, visualizations on right.}
\label{fig:constructaide:rendering}
\end{minipage}
\hfill
\begin{minipage}{0.48\linewidth}
\includegraphics[width=0.495\linewidth]{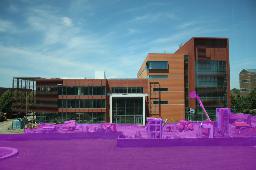}
\includegraphics[width=0.495\linewidth]{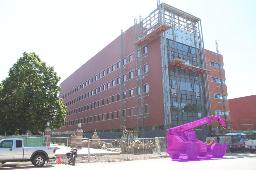}\\
\includegraphics[width=0.495\linewidth]{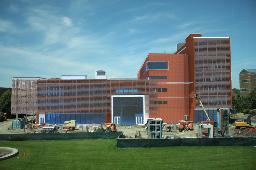}
\includegraphics[width=0.495\linewidth]{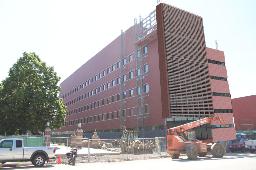}
\caption{Occlusions can be identified by analyzing the structure-from-motion point cloud and/or past imagery, allowing for both occlusion removal or advanced compositing and rendering.
}
\label{fig:constructaide:occlusions}
\end{minipage}
\end{figure*}

On construction sites, visualization tools for comparing 3D architectural/ construction models with actual performance are an important but often unfeasible commodity for project managers~\cite{turkan2012automated}. In this chapter, we develop a new, interactive method for 4D (3D+time) visualization of these models using photographs from standard mobile devices. {Our system works with unordered photo collections of any size (one picture to hundreds or more). Aligning photographs to the models to enable a suite of architectural and construction task related interactions: \vspace{-1mm}

\begin{itemize}
\item {\bf Photorealistic visualization}. Automatically create architectural renderings overlaid realistically onto photographs (Fig~\ref{fig:constructaide:rendering}) and identify and segment occluding elements on the job site (e.g. construction equipment and vehicles, Fig~\ref{fig:constructaide:occlusions}). \vspace{-1mm}
\item {\bf Performance monitoring}. Track the current state of construction to determine components which have been constructed late, on time, or constructed according to the building plan (or not) (Fig~\ref{fig:constructaide:progress}). Annotations made on one site photo are automatically transferred to other site photos (both new and existing) for fast annotation and collaborative editing/analysis. \vspace{-1mm}
\item {\bf 4D navigation}. Selectively view portions of a photographed scene at different times (past, present and future, Fig~\ref{fig:constructaide:4d}).
\end{itemize}}

{
This system is our primary contribution. We also demonstrate a new, user-assisted Structure-from-Motion method, which leverages 2D-3D point correspondences between a mesh model and one image in the collection.} We propose new objective functions for the classical point-$n$-perspective and bundle adjustment problems, and demonstrate that our SfM method outperforms existing approaches.

\subsubsection{Design Considerations}
\boldhead{Architectural Visualizations} A common and costly problem for designing new buildings or renovating existing facilities is misinterpretation of the building design intents. Our system allows for interactive visualizations of architectural models using photographs taken from desired viewpoints, and conveys a greater spatial awareness of a finished project. It also encouraging homeowners and facility managers to \textit{interact} with the design model, creating the ability to \textit{touch}. Building fa\c cades, architectural patterns, and materials can all be experimented with and quickly altered, and customized to individual preference from any desired viewpoint. It also promotes efficiency in professional practices by shortening the duration of design development and coordination processes.

\boldhead{Construction Visualizations} On-demand access to project information during the construction phase has a significant potential for improving decision-making during on-site activities. Visualizing 4D models with photographs provides an unprecedented opportunity for site personnel to visually interact with project documents, geo-localize potential errors or issues, and quickly disseminate this information to other users across the project. It can also facilitate field reporting and quality inspections as it allows iterations of work-in-progress and inspections to be properly logged. A time-lapse sequence of rendered images can also act as rich workflow guidelines (especially when contractors require detailed and step-by-step instructions), facilitate onsite coordination tasks, and minimize changes of requests for information from the architects. Facility owners and managers can also easily review their project at any time during the construction phase. These minimize inefficiencies that cause downtime, leading to schedule delays or cost overruns. 

\boldhead{Facility Management Visualizations} The ability to illustrate what elements lay within and behind finished surfaces (e.g., a wall) and interact with them -- either through photos previously captured during the construction/renovation phase or 3D architectural model -- during the operation phase of existing facilities is of tremendous value to facility managers. Joint rendering of envisioned construction versus actual construction can facilitate inventory control tasks and simplify recordings related to repair histories.

Existing tools for addressing these needs fall into two categories: one group of tools (e.g., Studio Max, MicroStation) allow users to interactively insert 3D models into single or time-lapse photos. The second group are mobile augmented reality systems that rely on radio-frequency based location tracking, fiducial markers, or on-board sensors to track location and orientation of user and superimpose 3D models into live video streams. 
There are also challenges in storing and frequently updating large 3D models, together with relevant project information, on mobile devices. 
All these challenge frequent application of visualization tools for site monitoring purposes, and thus may minimize opportunities for detecting and communicating performance deviations before they result in schedule delays or cost overruns.

\subsection{Related work}

The problem of registering large numbers of unordered ground photos, time-lapse videos, and aerial imagery with 3D architectural models has received tremendous interest in the civil engineering, computer graphics, and computer vision communities. Significant success has been reported with semi-automated systems for registering 3D architectural/construction models with time-lapsed videos~\cite{golparvar2009visualization,Kahkonen}, and using radio-frequency based location tracking or fiducial markers for augmented reality visualization of 3D CAD models for head-mounted displays~\cite{Behzadan:2005,dunston2003mixed,wang2009augmented,hammad2009distributed} and more recently commodity smartphones~\cite{cote2013live,hakkarainen2009software,irizarry2012infospot,woodward2010mixed,lee2011augmented,shin2010technology,yabuki2010collaborative}. 

Among related work, D4AR modeling~\cite{JCEM2011} is the most closely related to ours. Using an unordered collection of site photos, the underlying geometrical model of a construction site is captured using a pipeline of Structure-from-Motion (SfM)~\cite{SNAVELY-IJCV08} and Multi-View Stereo~\cite{Furukawa:2010}. By solving the similarity transformation between the 3D CAD model and the point cloud using a few user inputs, the point cloud is transformed into the CAD coordinate system, allowing the CAD models to be seen through SfM images. Using the ``traffic light'' metaphor as in~\cite{aravind2007} the photos can be augmented with color coded CAD elements. 

\begin{figure}[t]
\begin{minipage}{0.48\linewidth}
\includegraphics[width=\linewidth]{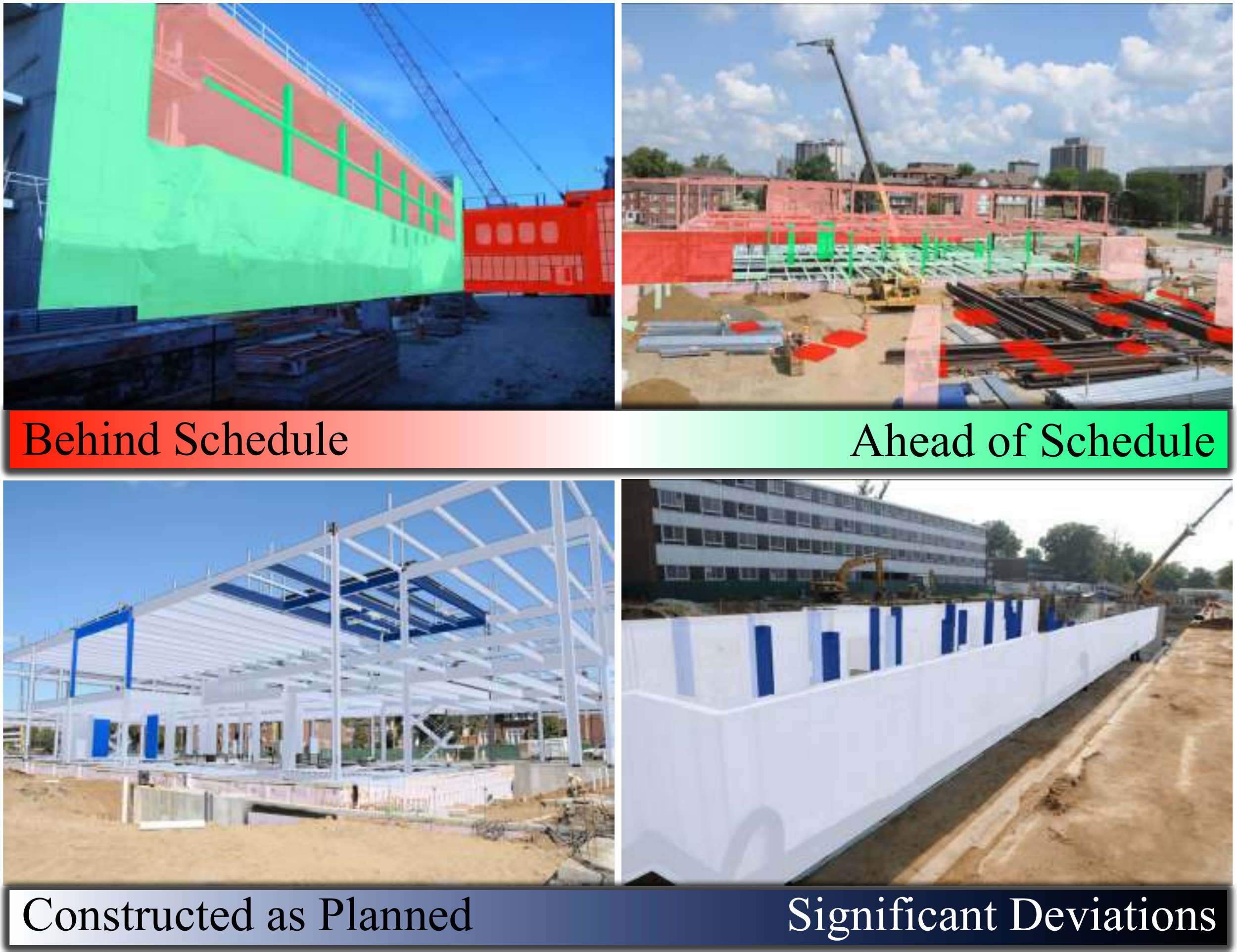}
\caption{Construction photos can be annotated using a palette of smart-selection tools, indicating which regions/building components need review or are behind or ahead of schedule. Annotations are automatically transferred among all views (including those at previous/future dates), allowing field engineers and project managers to collaborate and save time in creasing visualizations.
}
\label{fig:constructaide:progress}
\end{minipage}
\hfill
\begin{minipage}{0.48\linewidth}
\includegraphics[width=\linewidth]{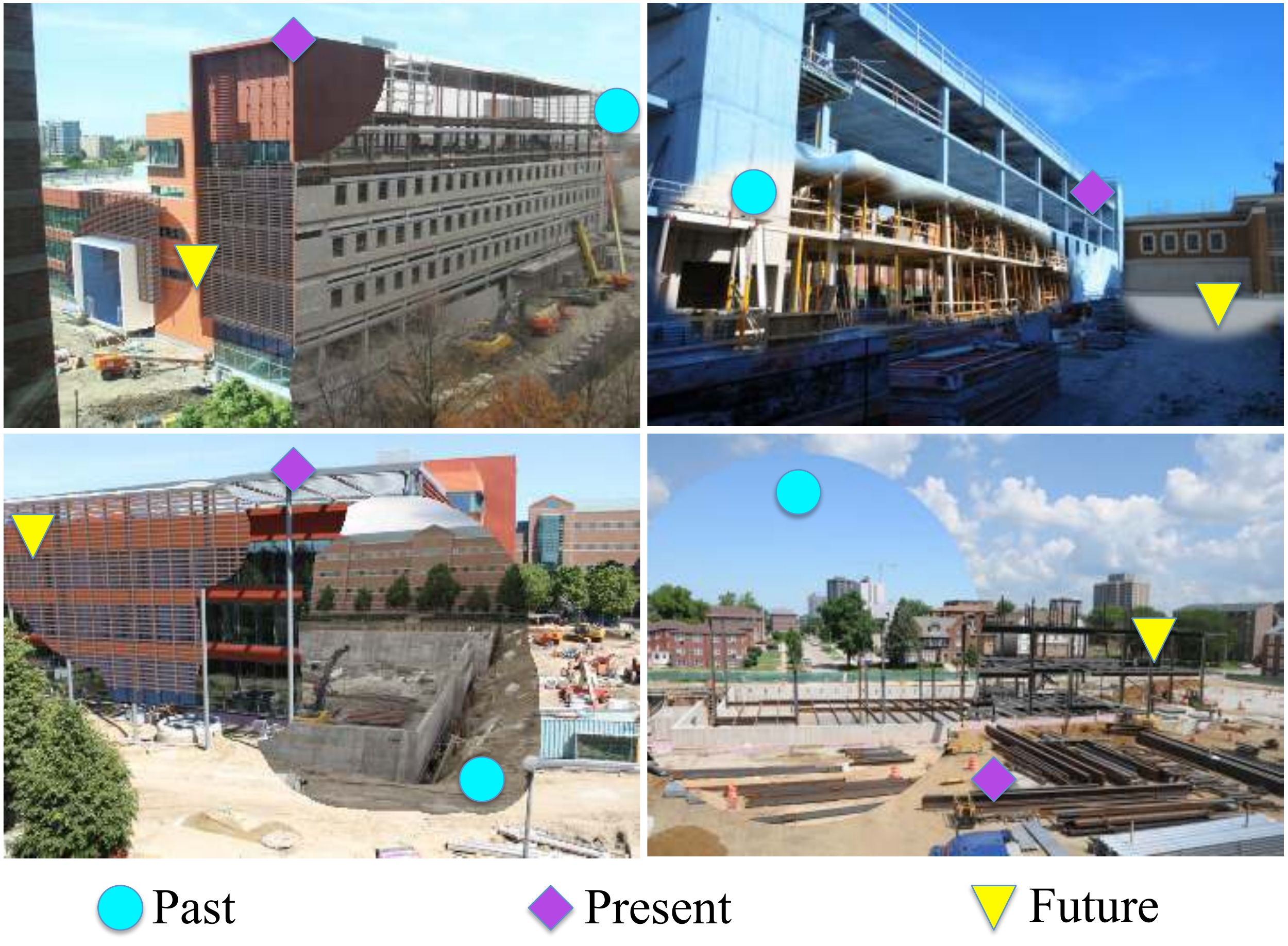}
\caption{Our system allows users to navigate image and construction models in 4D. Here, a user has selected to visualize both past and present information on each photograph.}
\label{fig:constructaide:4d}
\end{minipage}
\end{figure}
Other works on construction site visualization techniques include manual occlusion management~\cite{zollman_occlusion} and ``x-ray'' metaphors~\cite{zollmann_xray}. While these methods are intended for augmented reality / head-mounted displays, our work improves on these techniques by automatically estimating occlusions and allowing for more control in the visualization process.

Our semi-automated method for registering 3D models to photographs is inspired by past work on registration~\cite{franken2005minimizing,PintusFast}, architectural reconstruction~\cite{Wan201214,Werner:2002}, and automatic camera pose estimation/calibration from unordered photo collections~\cite{SNAVELY-IJCV08,SGSS-siggraph08}. Such methods, including ours, are known as ``incremental SfM'' (adding one or a few photo(s) at time), and recent methods demonstrate improvements by solving the SfM problem at once~\cite{disco12pami}. Our method is also related to the user-guided SfM method of Dellepiane et al.~\cite{semiauto_sfm}, although the inputs and goals our systems are different (input comes in the form of taking additional photos to guide reconstruction). 

Several methods exist for aligning 3D models automatically in photographs~\cite{Huttenlocher_reg,Lowe_reg}, and more recently Russell et al.~\cite{Russell-3DRR11} demonstrate a technique for automatically aligning 3D models in paintings. However, such methods are not suitable for construction sites since the 3D model and photographs rarely correspond in appearance (e.g., construction photos contain many occlusions, missing or added elements not present in the 3D model; the level of detail in 3D model may not match the level of detail in actual elements on site). Although unrelated to construction, Bae et al.~\cite{Bae:2010} demonstrated a method for merging modern and historical photos from similar viewpoints to allow for temporal navigation.

Many techniques exist for 3D architectural and mesh modeling from photographs~\cite{videotrace,sinhaArch,ImageRemodel,xu_sig11,Debevec:facade}. Our method instead relies on an existing, semantic 3D CAD model (known as a building information model, BIM). BIM are widely available as majority of building jobs require such models prior to construction.

Several methods leverage 3D models for photo editing, rendering, and visualization~\cite{DeepPhoto,city4d,view4d}. Most similar to our technique, Schindler and Dellaert in particular describe a method for navigating historical photo collections, including 4D interactions. Matzen and Snavely also describe a method for creating 4D visualizations of urban sites from unorganized user photographs~\cite{Matzen:eccv:14}. Distinct from other techniques, our method utilizes a semantic 3D model to enable 4D visualizations from arbitrary viewpoints (provided enough images exist), as well as tools for collaborative analysis and visualization, occlusion identification, and photorealistic rendering.

\begin{figure*}[t]
\includegraphics[width=\linewidth]{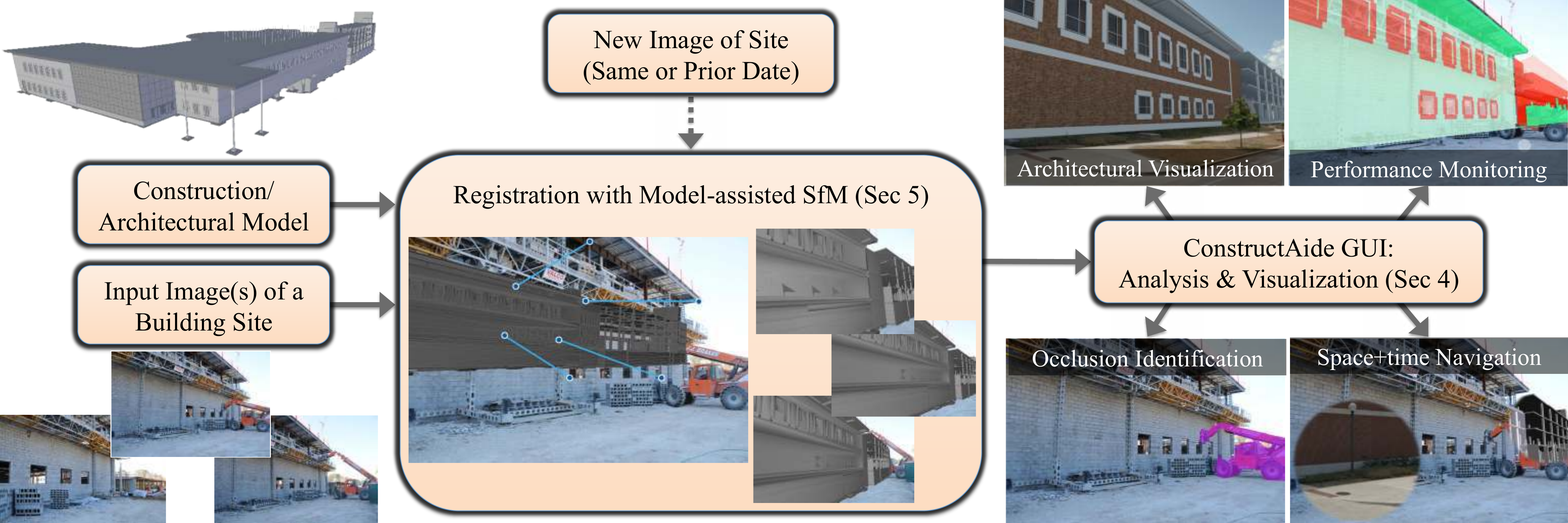}
\caption{Method overview. Our system takes as input a 3D model and one or more photos of a construction site. The model and photos are  aligned using our Model-assisted SfM approach: one image is registered by specifying 2D-3D correspondences, and other images are then registered automatically. Leveraging aligned photo-mesh data, we extract automatic estimates of occlusion, rendering information, and selection aids. Our interface then allows users to explore the model and photo data in 4D and create informative/ photorealistic visualizations.}
\label{fig:constructaide:overview}
\end{figure*}

\subsection{System overview}

Our approach, outlined in Figure~\ref{fig:constructaide:overview}, takes advantage of a small amount of user input to register all photos with the underlying 3D architectural/construction model. We only ask the user to specify a few correspondences between an image and the underlying 3D model, providing a registration between the model and the photo. Our system then registers other images automatically using our proposed Structure-from-Motion (SfM) formulation (Sec~\ref{sec:MeshSFM}). New photos of the same site -- taken at either an earlier or later date -- can also be registered with no additional interaction.


Once the 3D model is registered with the photograph(s), we preprocess the images to estimate timelapses from unordered photo sets (Sec~\ref{sec:4D:timelapse}), static and dynamic occlusions (Sec~\ref{sec:4D:occ}), and light/material models for rendering (Sec~\ref{sec:4D:buildinginfo}). Our user interface (Sec~\ref{sec:4D:GUI}) provides simple visualization metaphors that enable a user to interact with and explore the rich temporal data from the photo sets and architectural/construction models. For example, a user can quickly select elements from the photographs at any point in time, hide/show the elements, visualize the construction progress or analyze errors. Annotations and visualizations are automatically transferred across views, allowing for real-time, collaborative analysis and viewing. Finally, photorealistic architectural renderings can be produced without the user ever using CAD, 3D modeling or rendering software: the model is rendered using the extracted material/lighting model and composited back into the photograph automatically.

{
\subsubsection{Assumptions and Input Requirements}
ConstructAide relies on accurate BIM (semantically rich CAD models), including complete, up-to-date mesh models and scheduling information.} {We require BIM as these allow us to easily hide/show parts of the 3D model based on construction schedules, allowing for easy component selection and annotation at any phase of the build process. Furthermore, the information contained in BIM allows us to automatically produce photorealistic renderings composited onto in-progress construction photographs.}
{In order to best traverse and visualize the construction process using our system's tools, a spatially and temporally dense set of photographs is required. However, results can still be achieved with fewer images; registration, photorealistic rendering, and performance monitoring can be done with a single photograph, and occlusion identification and 4D navigation are possible with two or more photos (given the view location/orientations are similar). As more photographs are collected and added, our system enables more automation: Mesh-SfM automatically registers new images, annotations can be transferred across views, and occlusion estimates improve.
}
{The photo collections in this chapter typically contain 10-25 photos, which we found is sufficient to obtain reasonable results.}

\subsection{ConstructAide system}
\label{sec:4D}
Our interface requires as input one or more photos of the job site, a 3D building model, and an accurate registration of the model to each of the photos. In this section, we assume that the registration process is complete (Sec~\ref{sec:MeshSFM} describes our registration approach, but other approaches or ground truth data could be used if available). Our system allows users to virtually explore the job site in both space and time, analyze and assess job progress, and create informative visualizations for the construction team. We first preprocess the input (images and 3D model) as in Sec~\ref{sec:4D:preprocess}, and then the user can begin virtually interacting with the construction site (Sec~\ref{sec:4D:GUI}).

\subsubsection{Preprocessing}
\label{sec:4D:preprocess}
To enable more efficient interactions, we first process the registered data to extract information useful for selection, visualization, and rendering. For example, converting unordered collections into time-lapse data, identifying and removing occlusions, and extracting rendering information from building models enable users to navigate and visualize data with ease, allowing for valuable job-site visualizations to be created in minutes.

\paragraph{Converting Unordered Image Sets into Time-lapses} 
\label{sec:4D:timelapse}
The first step in this process is, for each image, to identify other images that were taken from roughly the same viewpoint, determined by how well a single homography can model matched features in every pair of images. We have already computed this data for registering the construction models to the photos (described in Sec~\ref{sec:MeshSFM}), and there is no need to recompute homographic transformations. Once similar-viewpoint pairs are identified, the homography is used to transform one image into the other's view; we do this at each camera location and for all nearby viewpoints, resulting in pixel-aligned temporal information. If no nearby viewpoints are found, this image cannot be traversed temporally in 2D (however, the registered 4D mesh can still be traversed).

\begin{figure}[t]
\includegraphics[width=0.49\linewidth,clip=true,trim=0 0 0 300]{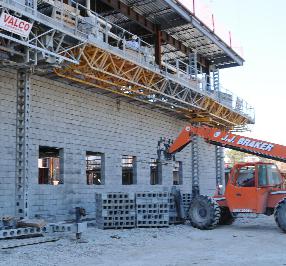}
\begin{overpic}[width=0.49\linewidth,clip=true,trim=0 0 0 300]{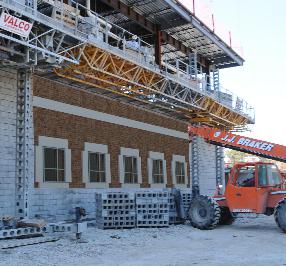}
\setlength\fboxsep{0pt}
\setlength\fboxrule{0pt}
\put(92,2){\fbox{\includegraphics[width=0.1\linewidth]{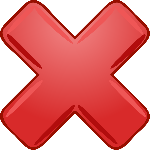}}}
\end{overpic}
\\
\includegraphics[width=0.49\linewidth,clip=true,trim=0 0 0 300]{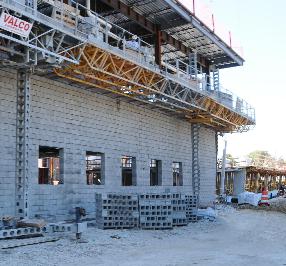}
\begin{overpic}[width=0.49\linewidth,clip=true,trim=0 0 0 300]{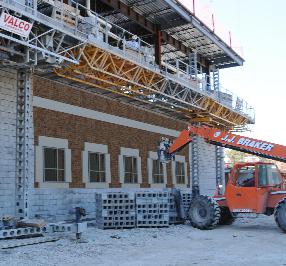}
\setlength\fboxsep{0pt}
\setlength\fboxrule{0pt}
\put(90,0){\fbox{\includegraphics[width=0.12\linewidth]{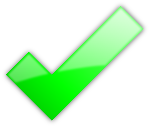}}}
\end{overpic}
\caption{Our method reasons about dynamic occlusions (such as the red telehandler pictured in the top left) by sampling similar viewpoints at different times so that depth layers are not confused in visualizations (top right). A clean background image is computed automatically (bottom left), and an occlusion mask is created by comparing the original image with the background image, allowing for building elements to be visualized with proper occlusion (bottom right).}
\label{fig:constructaide:dynamicOcc}
\end{figure}

\begin{figure*}[!ht]
\setlength\fboxsep{0pt}
\setlength\fboxrule{0pt}
\includegraphics[width=0.16\linewidth,clip=true,trim=0 0 0 30]{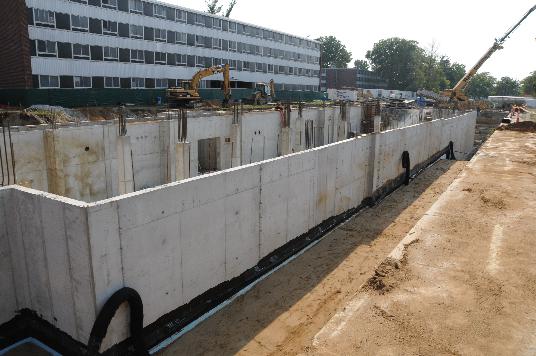}
\begin{overpic}[width=0.16\linewidth,clip=true,trim=0 0 0 30]{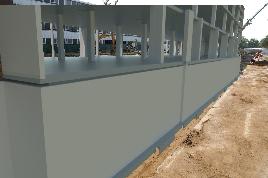}
\put(62,2){\fbox{\includegraphics[width=0.03\linewidth]{constructaide/fig/dynamicOcc/x.png}}}
\end{overpic}
\begin{overpic}[width=0.16\linewidth,clip=true,trim=0 0 0 30]{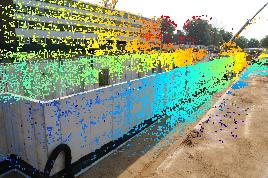}
\put(20,0){\fbox{\includegraphics[width=0.12\linewidth]{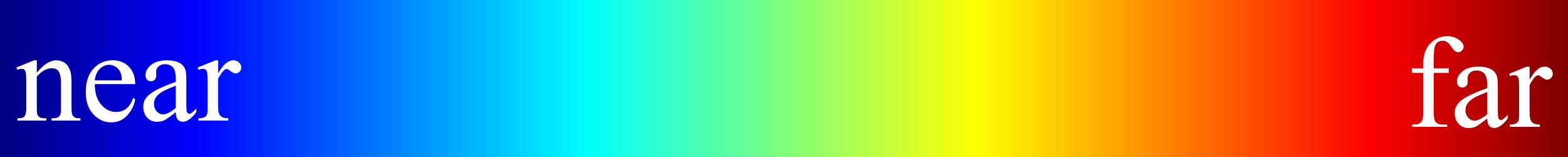}}}
\end{overpic}
\includegraphics[width=0.16\linewidth,clip=true,trim=0 0 0 30]{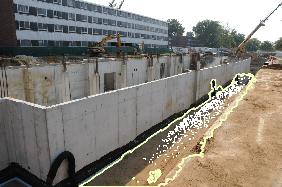}
\hfill
\begin{overpic}[width=0.16\linewidth,clip=true,trim=0 0 0 30]{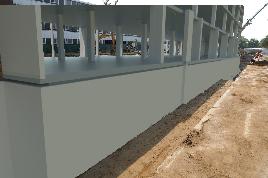}
\put(60,1){\fbox{\includegraphics[width=0.04\linewidth]{constructaide/fig/dynamicOcc/check.png}}}
\end{overpic}
\begin{overpic}[width=0.16\linewidth,clip=true,trim=0 0 0 30]{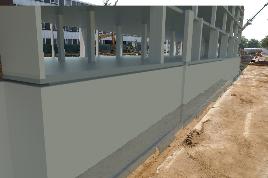}
\put(60,1){\fbox{\includegraphics[width=0.04\linewidth]{constructaide/fig/dynamicOcc/check.png}}}
\end{overpic}
\\
\begin{minipage}{0.16\linewidth}\centerline{(a)}\end{minipage}
\begin{minipage}{0.16\linewidth}\centerline{(b)}\end{minipage}
\begin{minipage}{0.16\linewidth}\centerline{(c)}\end{minipage}
\begin{minipage}{0.16\linewidth}\centerline{(d)}\end{minipage}
\hfill
\begin{minipage}{0.16\linewidth}\centerline{(e)}\end{minipage}
\begin{minipage}{0.16\linewidth}\centerline{(f)}\end{minipage}
\caption{Our system attempts to detect and static occlusions (i.e. the basement of this building model is hidden by the ground in (a) by comparing the mesh model (b) the point cloud estimated with SfM (c). 3D points that are measured to be in front of the model (see text for details) are then propagated and smoothed based on image appearance, resulting in an occlusion mask (d). The occlusion mask can be used to properly hide basement elements (e), or create an x-ray type visualization (f).}
\label{fig:constructaide:staticOcc}
\end{figure*}

\begin{figure*}[t!]
\includegraphics[width=.16\linewidth]{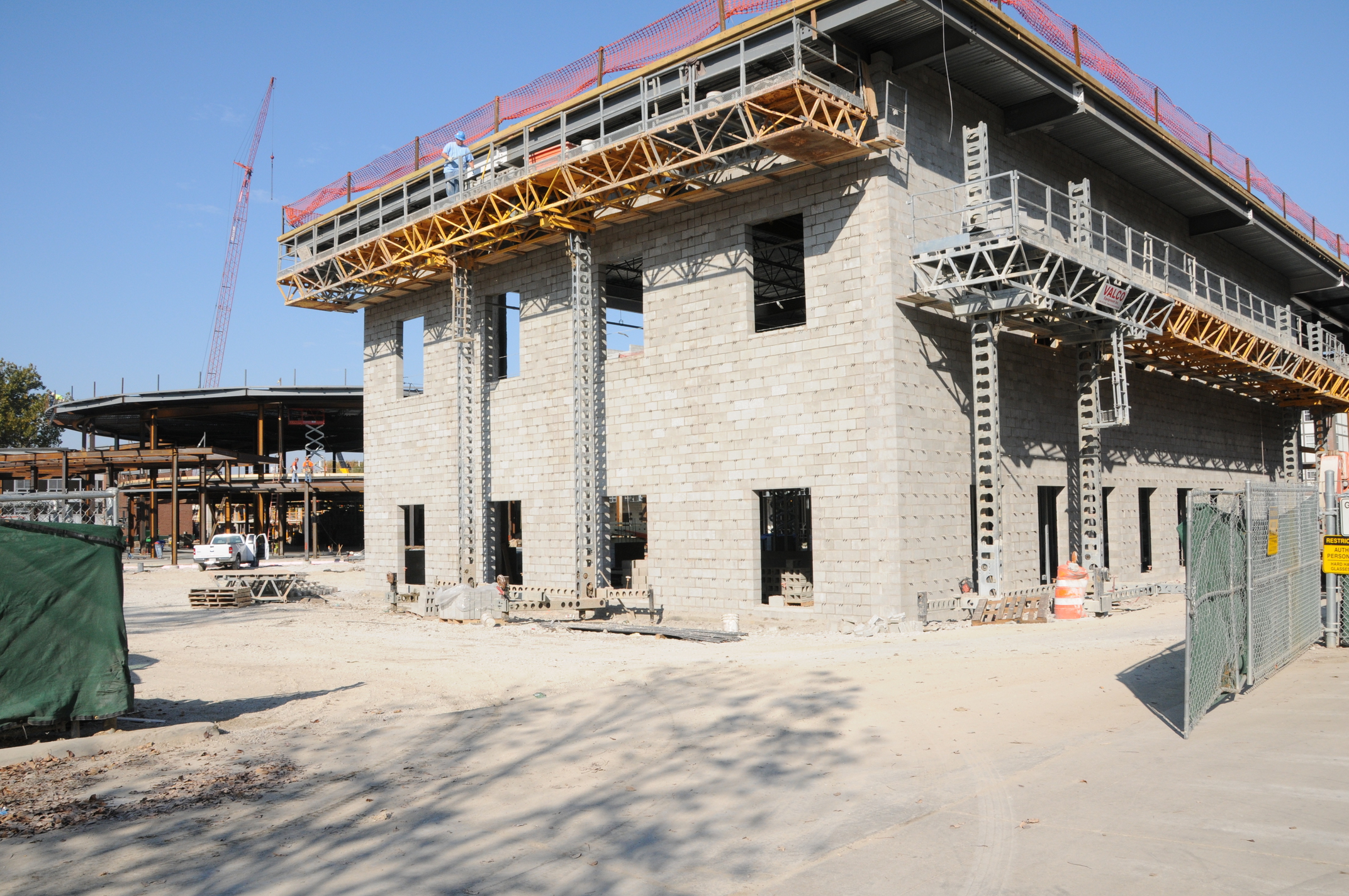}
\includegraphics[width=.16\linewidth]{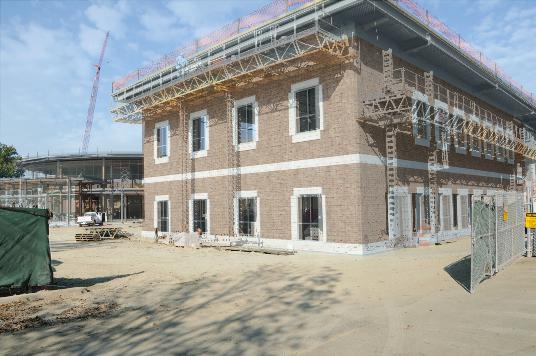}
\includegraphics[width=.16\linewidth]{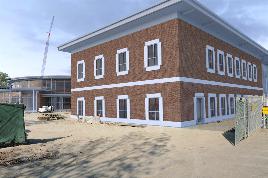}
\includegraphics[width=.16\linewidth]{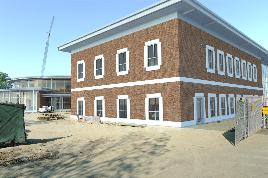}
\includegraphics[width=.16\linewidth]{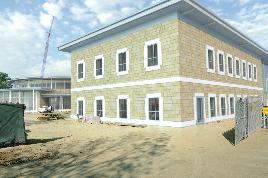}
\includegraphics[width=.16\linewidth]{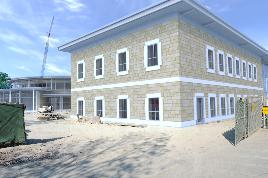}
\caption{Our system can be used to create photorealistic architectural visualizations automatically rendered {\it into} the photograph without the use or knowledge of any CAD, modeling, or rendering software. Here, we show a construction image, followed by a blended architectural render, and four different material / lighting choices for the scene. Occlusion information is computed automatically; errors and other objects can be corrected/added using our efficient selection tools (e.g., bottom row; the truck and crane were added manually, and we make no attempt to modify the shadow cast by the fence). Time lapses and changes to materials can be rendered with ease by swapping out preset HDRI light sources and materials.}
\label{fig:constructaide:timelapse}
\end{figure*}

\paragraph{Occlusion Identification} 
\label{sec:4D:occ}
We attempt to automatically identify troublesome occlusions that can lead to unappealing visualizations. For example, a truck may be idle temporarily in front of a fa\c cade (dynamic occlusion), or certain building components may be built beneath the ground or behind other non-building structures (static occlusions) -- see Figs~\ref{fig:constructaide:dynamicOcc} and ~\ref{fig:constructaide:staticOcc}. Such occlusions can be a nuisance when creating visualizations, and manually removing them may take time and expertise.

We handle the two types of occlusion (dynamic: moving equipment and workers; static: immobile elements blocking camera's field of view) separately. For dynamic occlusions, we assume that the occluding object is only in place temporarily, and thus that it does not occupy the same pixels in a majority of the aligned time lapse data (computed in Sec~\ref{sec:4D:timelapse}). We then find the ``background'' image by computing the per-pixel median of the time lapse (disregarding the image containing the occlusion); if our assumption holds, the moving object will be removed. To identify the pixels of the dynamic object, we compute the squared pixel-wise difference (in HSV) between the original image and the background, smooth the result with the cross-bilateral filter~\cite{cross_bf}, and threshold the smoothed result, keeping pixels greater than 0.05 in any channel. Fig~\ref{fig:constructaide:dynamicOcc} demonstrates this process. As few as one additional image can suffice for detecting dynamic occlusions, but more can be used if available. 

For static occlusions, we attempt to identify pixels in an image which are spatially in front of the 3D model, e.g., a fence might block a fa\c cade, or the ground may occlude the model's basement. Our idea is to make use of the 3D model and the sparse set of 3D points computed during our SfM procedure (Sec~\ref{sec:MeshSFM}). For each of these 3D points $p$ project onto the 3D model, we predict whether or not this point is in front of the model by evaluating the following heuristic:

\begin{equation}
[p - p_{\text{model}} > 0.3] \vee [\cos^{-1}(n(p)^T n(p_{\text{model}})) > \pi/6],
\end{equation}

where $p_{model}$ is the 3D location corresponding to the point on the mesh $p$ projects to, and $n(p)$ calculates the surface normal at $p$ (estimated using nearby points in the point cloud). In other words, if $p$ is closer to the camera by more than 0.3m, or normals differ by more than $30^\circ$, we assume the mesh must be occluded at this pixel. The normal criterion ensures that occluding points within the 0.3m threshold are oriented properly (otherwise they belong to the ``occlusion set'').

Since our point cloud is sparse, the binary occlusion predictions will be too. To obtain a dense occlusion mask, we flood superpixels (computed using SLIC~\cite{Achanta:ijcv:12,vedaldi08vlfeat}) with the sparse occlusion estimates (if a superpixel contains an occluded pixel, it becomes part of the occlusion mask); finally we smooth this mask using a cross-bilateral filter. Our approach is shown in Fig~\ref{fig:constructaide:staticOcc}.

In the event of failure either due to not enough images / triangulated points or misestimation, the user can correct errors using selection and editing tools in our interface.

\paragraph{Utilizing Other Building Information} 
\label{sec:4D:buildinginfo}
Architectural and construction models -- commonly known as Building Information Models (BIM) -- contain rich semantic information about element interconnectivity and materials. We leverage these in our interface to improve the user's experience. Building elements are clustered by \textit{primitive}, \textit{group}, and \textit{type} to accelerate selection in the photograph, scheduling info is used to create ``snapshots'' of the model's geometry at various points in time, building element material names are used to generate renderable, computer graphics materials, and GPS coordinates are used to acquire sun position (e.g. using publicly available lookup tables \url{http://aa.usno.navy.mil/data/docs/AltAz.php}).

\subsubsection{User Interface}
\label{sec:4D:GUI}
Now that the meshes (from building information model) and photos are aligned and visualization tools have been prepared, a user can interact with our system using a simple user interface. Selections in an image can be made by one of many unique ``marquee'' tools: 3D building elements can be selected individually as well as grouped by type or material, and individual faces/primitives can also be selected. These semantic tools accompany standard selection tools (lasso, brush, etc; see supplemental video). Occlusion masks obtained during our occlusion identification step are also used to create grouped pixel regions for efficiently selecting such regions. Once a selection is made in the image, the user can perform the following functions: 

\begin{itemize}

\item \textbf{Photorealistic Visualization:} A user can specify a subset of the underlying mesh model (using our selection tools), and seamlessly render visible/selected mesh components into the image. Geometry, lighting, and materials are known in advance (as in Sec~\ref{sec:4D:buildinginfo}), so the model can be rendered with no user interaction, and composited back into the photo using the technique of Karsch et al.~\cite{Karsch:SA2011}. We demonstrate a rendered result In Fig~\ref{fig:constructaide:timelapse}, we demonstrate a time-lapsed architectural visualization created with our software. Architectural rendering is performed using LuxRender (\url{http://www.luxrender.net/}), and preview rendering is done in our OpenGL interface.

\item \textbf{Performance Monitoring:} Based on scheduling data and the progress of construction visible in the image(s), a user can assess the progress of a region in the image by adding annotations. A color label can be given to indicate whether the component was built ahead of schedule (green), on time (semi-transparent white), or behind schedule (red), as shown in Fig~\ref{fig:constructaide:progress}. Annotating deviations in the builing process is also possible; darker blue overlays indicate components have not been built according to plan or need review. Any visualizations and notes added to an image are propagated to all other registered views/images, allowing for efficient annotation and real-time, collaborative analysis.

\item \textbf{4D Navigation:} Users can scroll through both the spatial and temporal extent of the photo collection. For navigating in time, a user can selectively peer forward or backward in time, revealing past image data or future renderings of the 3D model (in the region selected by the user).
\end{itemize}

For a demonstration of our GUI tools, see our supplemental video.


Construction site imagery typically comes from a few static/mounted cameras that record time-lapses or from project participants photographing the site at regular intervals (e.g. tens to hundreds of photos every few weeks). Thus, data is typically dense temporally, but the photos usually are not spatially dense and as such have wide baselines.  Our system and interface can work with any sampling of photographs, but temporal navigation and occlusion detection is more compelling with a dense time-sampling of images.

\subsubsection{Domain Expert Evaluation}
We interviewed five domain experts\footnote{All participants have advanced degrees in construction monitoring and have worked for 2-8 years as field engineers for numerous construction sites.} with experience in construction progress monitoring and assessment. All subjects were males between the ages of 24 and 38, and no compensation was given.

In preparation to the interview, we asked participants about existing solutions for progress monitoring. All subjects responded that it is current practice to simply take notes and photographs (e.g. on mobile devices) and hand-annotate these with progress and financial information. Three participants used the words ``subjective'' or ``unreliable'' to describe this process; one participant noted that these methods are ``highly subjective since the results are based on [one field engineer's] judgement and experience.'' All but one participant mentioned that current practices are tedious and are not scalable. 

Following the pre-interview, we demonstrated ConstructAide to each subject. Four of the five subjects mentioned that smart-selection tools and sharing annotations would reduce the bottlenecks of current techniques, allowing for faster and more accurate on-site inspections. Subjects generally felt that the 4D navigation feature was most useful, e.g. ``space-time navigation is very useful for change management and  assessing/verifying contractor claims. You need as-built evidence for claims.'' However, three of the participants also noted that assessing building deviations in a photographical interface may not be as useful for certain tasks since ``millimeter accuracy may be required, and not all approved building changes are reflected in the BIM.'' 

Participants described further automation as desired features, e.g. ``automatic progress and error assessment using the BIM and pictures.'' One user described his ideal progress monitoring tool during the pre-interview: ``In a perfect world, field reporting would automatically synch in the cloud and be accessible anywhere using smartphones. Tools would be easy to use, objective, and based on quantitative measures.''

\begin{figure}
\centerline{User study questionnaire average responses}
\includegraphics[width=\linewidth]{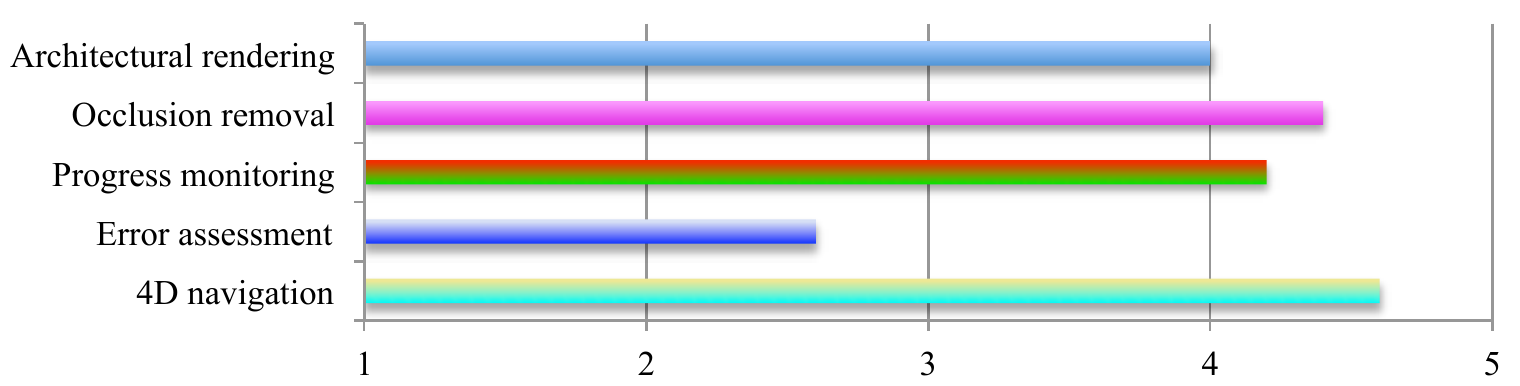}
\caption{Responses from the exit questionnaire of our user study averaged over all subjects. Each question posited the usefulness of a given feature using a Likert scale, e.g. ``Rate the usefulness of the 4D navigation tool.  (1 = poor, 5 = excellent)''}
\label{fig:constructaide:studyresponses}
\end{figure}

\begin{figure}[t]
\includegraphics[width=.325\linewidth]{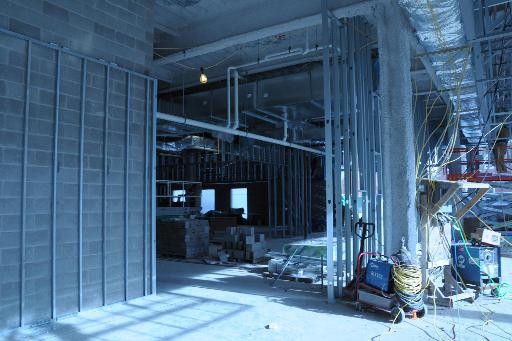}
\includegraphics[width=.325\linewidth]{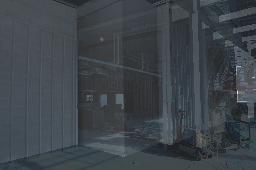}
\includegraphics[width=.325\linewidth]{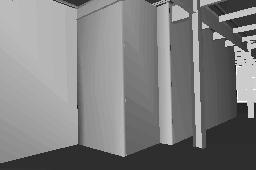}\\
\includegraphics[width=.325\linewidth,clip=true, trim=0 0 0 0]{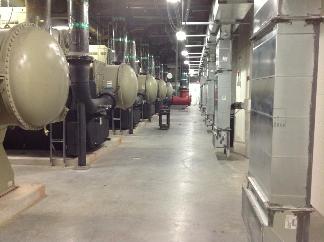}
\includegraphics[width=.325\linewidth,clip=true, trim=0 0 0 0]{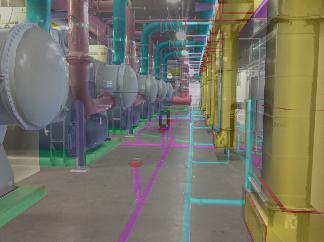}
\includegraphics[width=.325\linewidth,clip=true, trim=0 0 0 0]{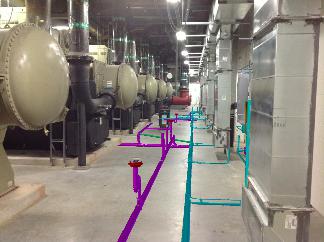}\\
\includegraphics[width=.325\linewidth,clip=true, trim=0 0 0 0]{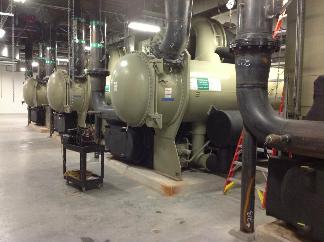}
\includegraphics[width=.325\linewidth,clip=true, trim=0 0 0 0]{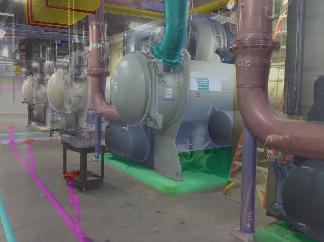}
\includegraphics[width=.325\linewidth,clip=true, trim=0 0 0 0]{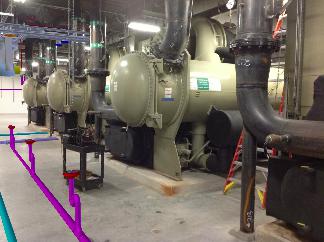}
\caption{Our system is applicable indoors and can register images with textureless surfaces (top row), repetitive structures, and significant occlusion (middle/bottom rows). Hidden or highly cluttered building systems, such as mechanical, electrical, and plumbing components can be visualized and annotated. Components can be filtered based on visibility and type (middle and bottom right).}
\label{fig:constructaide:indoor}
\end{figure}

Subjects also completed an exit Likert questionnaire about the features of ConstructAide (e.g. ``Rate the usefulness of the [\ \ \ ] tool (1=poor, 5=excellent)''). Responses are summarized in Fig~\ref{fig:constructaide:studyresponses}. Consistent with interview responses, subject found the 4D navigation feature to be the most useful and the construction error assessment tools to be the least useful. Occlusion removal and progress monitoring features were generally well-liked. Finally, all users responded ``yes'' to the exit question ``would you use this software for your field of work?''

{
\subsubsection{Applications}

Beyond the primary use of construction monitoring, we also demonstrate other practical use-cases for ConstructAide: 

\boldhead{Architectural visualizations for homeowners and clients} Pre-visualization of interior and exterior finishes, colors, or styles. Our system allows buyers take images from their desired viewpoints and obtain photorealistic renderings of different alternatives on their desired building components (e.g. choosing different tiles/ceramics and wallpapers for a bathroom). Our system eliminates the need for the physical mockups or reviewing samples, and instead allows the end users to review alternatives from one or many viewpoints of their choice (see Figure \ref{fig:constructaide:timelapse}).

\boldhead{Performance, liability and dispute resolution} Visual records of a project's lifecycle can be extremely useful for handling legal and contractual disputes, but analyzing and jointly visualizing the photographs with project 3D models has previously been difficult. ConstructAide provides new solutions in this space, as echoed by subjects during our pilot interview.

\boldhead{Facility management} Despite efforts to link product data and maintenance schedules to BIM, these models are rarely used for operation and maintenance purposes. One issue is that today's smartphones do not have the capacity to directly provide access to BIM when/where such information is needed. The application of BIM with mobile augmented reality has been limited to proof-of-concepts due to challenges of using location tracking systems indoors.  Our solution can be used alongside current practices by allowing users take pictures and immediately analyze and visualize the site on their smart devices. To minimize scene clutter in presence of a large number of building elements in a model, our tools can filter visualizations based on element type or the task in hand. We show this method works both for indoors (Fig.~\ref{fig:constructaide:indoor}) as well as outdoor scenes.

}

\subsection{Model-assisted structure-from-motion}
\label{sec:MeshSFM}

The availability of inexpensive and high-resolution mobile devices equipped with cameras, in addition to the Internet has enabled contractors, architects, and owners the ability to capture and share hundreds of photos on their construction sites on a daily basis. These site images are plagued with problems that can be difficult for existing SfM techniques, such as large baselines, moving objects (workers, equipment, etc.), and the constantly changing geometry/appearance of a construction site. {Furthermore, current SfM approaches are designed to work with hundreds of photographs of a static scene in which there is very high spatial density among the camera positions and view directions. Taking this many photos regularly of a construction site is not practical; a majority of site photos come from a few fixed-position, time-lapse cameras.} To overcome these issues, we propose a user-assisted SfM pipeline in which the user provides an accurate initial camera pose estimate (through mesh-image correspondences) which drives the remainder of the registration process. 

\begin{figure}[t]
\includegraphics[width=\linewidth]{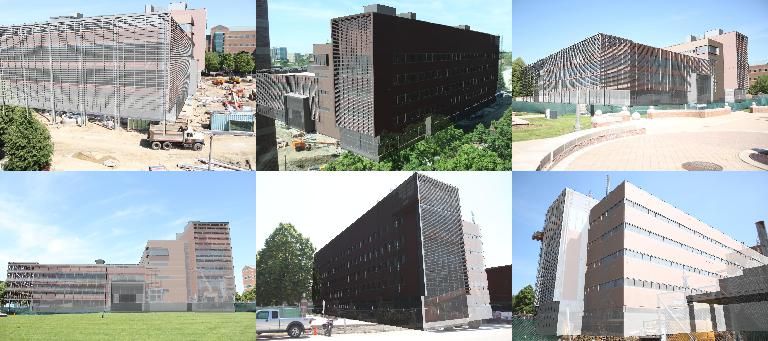}
\caption{Registrations estimated with Model-assisted SfM.}
\label{fig:constructaide:registration}
\end{figure}

As in typical SfM algorithms, the result is a set of camera intrinsic and extrinsic parameter estimates as well as a sparse point cloud (triangulated feature correspondences). We are primarily interested in the camera parameters as this provides a registration of the 3D model to each photo. To a lesser extent, we use the point cloud when determining static occlusions (Sec~\ref{sec:4D:occ}). Fig~\ref{fig:constructaide:registration} shows example registrations obtained using our method on a construction dataset.

Our goal is to find the proper camera parameters (intrinsic and extrinsic) as to register the virtual cameras with the {\it Euclidean} 3D model for each image. Here, we use the term {\it Euclidean} to represent the similarity transformation that maps an image-based 3D model into a measurable coordinate system for Engineering applications. We model intrinsic parameters using a three parameter pinhole model with variable focal length and two radial distortion coefficients, and assume that the principal point is in the center of the image and that pixels are square.

To begin the registration process, a user chooses one image from the collection (denoted throughout as an {\it anchor camera}) and selects 2D locations in the image and corresponding 3D points on the mesh model\footnote{Using the semantic/timing information present in the building information model, our interface allows for future building components to be hidden so that model can be easily made to match a photo taken at any point during the construction progress}. Our interface facilitates this selection by allowing the users to quickly navigate around the mesh. Given at least four corresponding points, we solve for the six-parameter extrinsic parameters of the camera -- three rotation and three translation parameters -- that minimize reprojection error using Levenberg-Marquardt (also called the Perspective-$n$-Point, or P$n$P problem). During this optimization, we fix the intrinsic parameters to have no radial distortion, and the focal length is obtained either from EXIF data or initialized such that the field of view is 50${}^\circ$. Prior to optimization, the camera parameters are initialized using the pose of of the model in the GUI (see supplemental video).

Choosing the anchor camera is important: this image should contain a sufficient view of the mesh such that many corresponding points in the image and on the model are clearly visible. This is typically straightforward as construction imagery focuses on a particular object (e.g. a building) and empirically we find many user-selected points to be visible in nearly all frames.

Knowledge of mesh-to-photo registration for {\it one} image doesn't help the SfM process as much as one might expect, but it does eliminate the coordinate system ambiguity (gauge transformation), and we later show how the 2D-to-3D correspondences can constrain and improve SfM estimates (see Tables~\ref{tab:quant_construction} and ~\ref{tab:middlebury_mvs}). Photos of construction sites are typically object/building-centric, so many cameras will be viewing the same object (many SfM methods cannot make this assumption, i.e. when reconstructing Rome~\cite{buildingRome}). Thus, the anchor camera can constrain many other images in the collection.


\begin{algorithm}[t]
 \KwIn{3D model, set of unregistered images $\mathcal{U}$}
 \KwOut{set of registered images $\mathcal{R}$}
 $\mathcal{R} \gets \varnothing$\\
 Choose initial image $I$ from $\mathcal{U}$\\
 Select 2D-3D point correspondences between I and 3D model\\
 Solve for camera parameters\\
 $\mathcal{R}\gets \mathcal{R} \cup I$, \ $\mathcal{U}\gets \mathcal{U} \setminus I$\\
 Compute feature matches for each image pair in collection\\
 \While{$\mathcal{U} \neq \varnothing$ }{
  \ForEach{$R \in \mathcal{R}, $U$ \in \mathcal{U}$}{
    \If{Homography between $R$ and $U$ fits well}{
      Transfer selected 2D correspondences from $R$ to $U$
      Solve for $U$'s camera, $\mathcal{R}\gets \mathcal{R} \cup U$, \ $\mathcal{U}\gets \mathcal{U} \setminus U$
    }
  }
  Perform constrained bundle adjustment on all images in $\mathcal{R}$
  Identify $U'\in\mathcal{U}$ with most triangulated features (tracks)\\
  \eIf{$U'$ has enough matched tracks}{
   Triangulate tracks in $\mathcal{R}$ corresponding to features in $U'$
   }{
   Identify $U'\in\mathcal{U}$ with fewest matched tracks\\
   Select correspondences between $U'$ and 3D model\\
  }
  Solve for $U'$ camera, $\mathcal{R}\gets \mathcal{R} \cup U'$, \ $\mathcal{U}\gets \mathcal{U} \setminus U'$\\
  Perform constrained bundle adjustment on all images in $\mathcal{R}$
 }
\caption{Model-assisted SfM}
\label{alg:overview}
\end{algorithm}

\begin{table*}[t]
{\footnotesize
\begin{tabular}{|c c|| c|c|c|c || c|c|c|c || c|c|c|c |} \hline
& & \multicolumn{4}{c||}{Rotational error (degrees)} & \multicolumn{4}{c||}{Translational error (meters)} & \multicolumn{4}{c|}{Reprojection error (\% width)} \\ 
Dataset & N & Ours & VSfM  & PS & PS-ICP & Ours & VSfM & PS & PS-ICP & Ours & VSfM & PS & PS-ICP \\ \hline \hline
Northwest A & 15 & \hili{0.67} & 2.28 & 8.79 & 79.40 & \hili{1.91} & 2.51 & 6.99 & 10.19 & \hili{9.34} & 22.70 & 23.26 & 52.96\\ \hline
Northwest B & 160 & 0.36 & \hili{0.30} & 0.31 & 5.13 & \hili{0.22} & 0.24 & 0.24 & 2.43 & 1.41 & \hili{0.87} & 0.94 & 17.93\\ \hline
West & 26 & \hili{1.20} & 1.81 & 1.67 & 20.02 & \hili{0.21} & 0.53 & 1.16 & 1.97 & \hili{1.37} & 1.96 & 3.32 & 20.25\\ \hline
Northeast & 22 & \hili{0.17} & 1.23 & 1.21 & 6.22 & \hili{0.15} & 1.34 & 1.14 & 9.08 & \hili{0.50} & 3.54 & 3.12 & 17.65\\ \hline
Basement & 10 & \hili{1.70} & 137.90 & 12.45 & 3.29 & \hili{0.44} & 8.15 & 1.56 & 1.55 & \hili{1.53} & 45.61 & 8.22 & 9.67\\ \hline
Southeast & 25 & \hili{0.31} & 0.73 & 0.94 & 5.00 & \hili{0.07} & 0.72 & 1.97 & 2.16 & \hili{0.52} & 1.86 & 3.63 & 9.49\\ \hline
\end{tabular}
}
\caption{Comparison of our method against existing approaches using real-world construction data on an instructional facility. N is the number of images.}
\label{tab:quant_construction}
\end{table*}

After the first image is registered with the mesh model, we proceed by iteratively registering other images in the collection. Our approach is similar to many existing structure-from-motion algorithms, but with several important differences that leverage the anchor camera. Following existing structure-from-motion methods~\cite{SNAVELY-IJCV08}, we detect and match interest points across all images in the collection, and prune the matches by estimating the Fundamental matrix between image pairs using RANSAC. Different from existing methods, we then search for images which match the anchor image well up to a single homography (80\% of matched features are required as inliers), warp the selected 2D points from the anchor image to these images, and solve the P$n$P problem for each of these images using the known 3D correspondences to register nearby images (excluding points that fall outside the image; if fewer than four remain, we do not register the image). This is particularly useful for construction images, as many can be taken from roughly the same viewpoint with only focal length and rotational differences, such as those from a mounted camera. 

Among all of the registered images, we perform one round of {\it constrained bundle adjustment}. As in most reconstruction approaches, our bundle adjustment optimizes over extrinsic/intrinsic camera parameters and triangulated 3D points; however, points triangulated using the anchor camera are constrained to lie along the anchor camera's ray, and we do not adjust the pose of the anchor camera (but intrinsics may change); refer to the supplemental document for details. We do not triangulate matched features corresponding to rays less than two degrees apart to avoid issues of noise and numerical stability. If no matches are triangulated, bundle adjustment is skipped. Constrained bundle adjustment is detailed in \hyperlink{appendix:constrainedBA}{Appendix B}.

One point of consideration is that the content of construction images will change drastically over time, even from similar viewpoints. However, many elements in these photos remain constant (ground/road, background trees/buildings) and we have found -- based on our datasets -- that features on these objects seem generally sufficient drive our SfM over long timescales. However, if automatic registration
fails, new anchor image(s) can easily be added.

\begin{table*}
{\small
\begin{center}
\begin{tabular}{|c c|| c|c|c || c|c|c || c|c|c |} \hline
& & \multicolumn{3}{c||}{Rotational error (degrees)} & \multicolumn{3}{c||}{Translational error (unitless)} & \multicolumn{3}{c|}{Reprojection error (\% width)} \\ 
Dataset & \# images & Ours &  VSfM  & PS & Ours & VSfM & PS & Ours & VSfM & PS\\ \hline \hline
temple (medium) & 47 & \hili{2.03} & 2.47 & 2.84 & 0.05 & \hili{0.02} & 0.04 & \hili{1.94} & 2.32 & 2.02\\ \hline
temple (small) & 16 & \hili{2.34} & 2.72 & 3.35 & \hili{0.05} & 0.11 & 0.07 & 2.01 & \hili{1.91} & 2.13\\ \hline
dino (medium) & 48 & 2.43 & \hili{0.79} & 6.27 & \hili{0.06} & 0.09 & 0.33 & 0.75 & \hili{0.72} & 1.78\\ \hline
dino (small) & 16 & 3.06 & 6.46 & \hili{0.76} & \hili{0.08} & 0.50 & 0.30 & \hili{1.04} & 4.01 & 1.13\\ \hline
\end{tabular}
\end{center}
}
\vspace{-1mm}
\caption{Evaluation using ground truth data from the Middlebury Multi-View Stereo dataset.}
\label{tab:middlebury_mvs}
\end{table*}

Next, we search for other images with a sufficient number of features corresponding to existing tracks, i.e. matched features common to two or more registered images; such features can be triangulated. We choose the image that has the fewest matches over a threshold (60 in our implementation) to ensure a good match and potentially wide baseline. This camera is registered by solving a constrained P$n$P problem using its 2D matches corresponding to the triangulated 3D tracks, made robust with RANSAC (inliers considered within 1\% of the image width). We also use the anchor camera to improve the P$n$P solution: using the Fundamental matrix between the anchor camera image and the image that is currently being registered, epipolar lines are computed corresponding to the user-selected 2D locations in the anchor image; the corresponding 3D mesh locations are then constrained to lie nearby these lines (based on reprojection error). Given a set of $k$ 3D points $X=\{X_1, \ldots, X_k\}$ and their corresponding projected pixel locations $u=\{u_1, \ldots, u_k\}$ and epipolar lines $e=\{e_1, \ldots, e_k\}$, we search for a 3D rotation ($R$) and translation ($t$) that jointly minimizes reprojection error as well as the point-to-line distance from projected points to their corresponding epipolar lines:
\begin{align}
\argmin_{R,t} \sum_i || x_i - u_i || + pld(x_i, e_i), \nonumber \\
\text{where: } x_i = \text{project}(R X_i + t, f)
\label{eq:contrainedPnP}
\end{align}
where $\text{project}(X,f)$ projects 3D locations into the plane according to focal length $f$, and $pld(x,l)$ computes the shortest distance from pixel location $x$ to the line specified by $l$. In our experience, this strategy helps avoid errors due to noisy camera estimates and triangulations.

In the case that not enough features in unregistered images match existing tracks in registered images, we choose the image with the {\it least} matched track of feature points. The user then specifies 2D locations in this image corresponding to 3D mesh locations selected in the anchor image\footnote{This is purely an image-based task, as the 3D positions do not need to be specified again.}, and this image is registered again by solving P$n$P. This happens typically if the image graph, or sets of tracks through the image collection, is disjoint. The image with the least matched tracks is chosen with the goal of connecting the graph, or at the very least, adding an image with large baseline. Since the user assisted in registering the chosen image, this camera is also designated as an anchor camera. After this camera is registered, another round of constrained bundle adjustment is performed. Until all images have been registered, this process is repeated. See Algorithm~\ref{alg:overview} for an overview.

\subsubsection{Experiments}

We hypothesize that knowing at least one camera's pose (as in our method) should aid camera pose and reconstruction estimates, as compared to blind, automatic SfM techniques. To test our hypothesis (and accuracy of registration), we compared our estimates to ground truth camera poses as well as camera pose estimates from established SfM methods. In total, we tested 10 different photo collections falling into two categories: real-world construction site images and object-centric images from the Middlebury Multiview Stereo dataset~\cite{middlebury_mvs}. We chose this data for several reasons: construction site data allows us to quantify error on real-world sites, the data vary widely in appearance, baseline, and number of photos, testing the limits of our method, and we require a corresponding mesh-model (available for our construction data, and obtainable for the Middlebury data). We compare our method to Wu's VisualSfM~\cite{vsfm1,vsfm2} and Photosynth\footnote{\url{http://photosynth.net}. Camera information extracted using the Photosynth Toolkit: \url{https://code.google.com/p/visual-experiments/}}. While both methods are based on the method of Snavely et al.~\cite{phototourism}, we found the estimates to be quite different in some cases most likely due to differences in implementation (e.g. Photosynth uses a different feature matching scheme than VisualSFM\footnote{\url{http://en.wikipedia.org/wiki/Photosynth}}). 

\boldhead{Construction Site Evaluation} 
We first test our method on real-world construction data. Ground truth camera pose estimates do not exist for this data, so we create ground truth data by manually calibrating five of the images in each dataset (images are chosen for dataset coverage). Corresponding 2D and 3D locations are chosen by hand, allowing us to solve for the true camera pose. At least four pairs of corresponding points must be chosen, and the set of chosen points should not be co-planar to avoid projective ambiguity. As our method requires the same ground truth calibration for at least one of the images (during initialization), we ensure that the images calibrated in our method are not used when creating ground truth (and thus not compared to). 

{In order to test the limits of each of the evaluated algorithms, we ensure that our datasets vary significantly in spatial and temporal sampling of photos. Each dataset required between two and four anchor images depending on the spatial density.} For each photo collection, we process the images with our model-assisted SfM technique (Sec~\ref{sec:MeshSFM}) as well as VisualSfM and Photosynth (denoted as {\bf VSfM} and {\bf PS} onward). Since the models produced by VSfM and PS are not in the same coordinate system as the ground truth data, we align them with a simple procedure: (a) triangulate a set of points (hand-selected for accuracy) using both the ground truth cameras and VSfM's cameras, (b) find the similarity transformation (scale, rotation, translation) that minimizes the squared distance between the point sets, and (c) apply this transformation to VSfM's cameras. The same procedure is applied to the result from PS. For nearly all datasets, the mean squared error is $<0.01$m, ensuring a good fit. There is no need to adjust the pose estimates from our method as our estimates are already in the 3D model's coordinate system.

For additional comparison, we also match the coordinate system of PS results to the ground truth by matching all triangulated features with points sampled from the 3D model using the iterative closest point algorithm; we call this method {\bf PS-ICP}. 
 
Between each of the methods and the ground truth cameras, we compute three error measures: rotational difference (angle between viewing directions), translational difference (distance between camera centers, in meters), and reprojection error of seven hand-selected ground truth 3D locations (selected for wide coverage of the model). Table~\ref{tab:quant_construction} shows the results of this experiment on the six construction site photo collections. The errors shown are averaged over all five of the ground truth calibrations. 

\boldhead{Middlebury Evaluation}
We also test our method and others against ground truth camera pose from the Middlebury Multiview Stereo datasets. We investigate four of the datasets (dino and temple datasets, the medium and small collections), and compare our method with VisualSfM (VSfM) and Photosynth (PS). As in the construction data experiment, we compute rotational, translational, and reprojection error. Since we now have ground truth data for each of the images in the dataset, we compute the average error over all images in the dataset (excluding any that were not successfully registered by a particular algorithm). Table~\ref{tab:middlebury_mvs} shows the results.

\boldhead{Discussion}
In both experiments, we observe that our model-assisted SfM technique typically outperforms existing methods in the three error measures (although the compared methods do not have access to a 3D model during registration). Incorporating 3D model data into the SfM process can be beneficial at a low cost to the user, even if the model is incomplete/inexact. We see that the results are fairly consistent across the two experiments, indicating that our method might be suitable for ``object-sized'' data as well.

These experiments suggest that our method may perform better than other techniques for smaller image collections with wider baselines. For example, the Basement sequence contains only ten photos from significantly different viewpoints. Our method is able to achieve low error while other techniques cannot handle this sparse sampling of photographs. Thus, our system is uniquely suited for a small number of viewpoints with a dense time sampling (e.g. a few time-lapse cameras around a construction site) -- annotation, rendering, occlusion detection, and temporal navigation are all still possible given these kinds of datasets. These scenarios are what our system is designed for, whereas existing SfM techniques require spatially dense images (also suggested quantitatively by our evaluation). The proposed interface and visualizations are more compelling with dense time data but possible otherwise.

\begin{figure}[h]
\includegraphics[width=\linewidth]{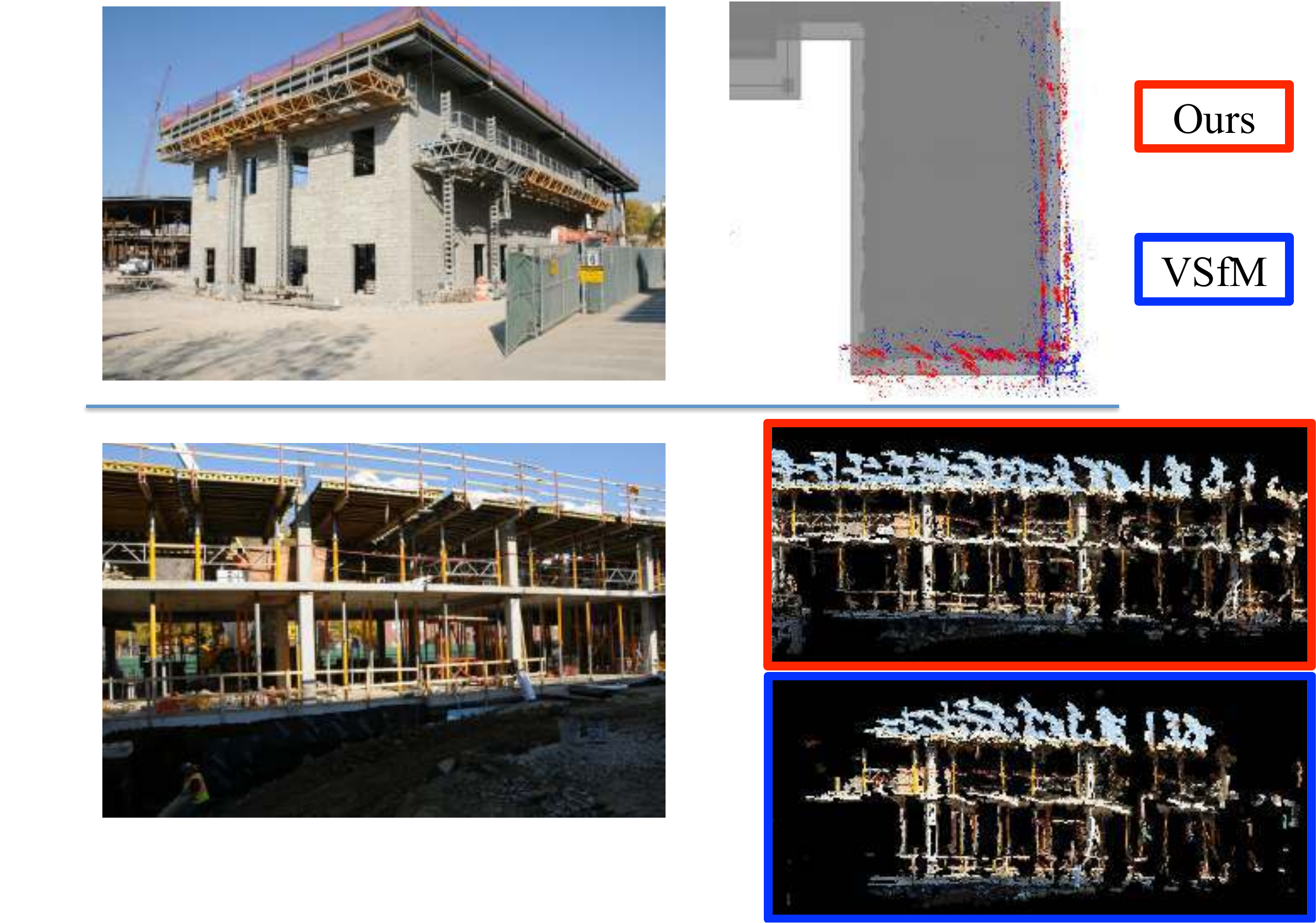}
\caption{Our SfM method typically produces more robust estimates than existing automatic approaches, resulting lower reconstruction error and less drift (top row; VSfM's point cloud propagates error away from the building corner) as well as more dense multi-view stereo reconstructions (bottom row; using the PMVS2 software of Furukawa and Ponce~\protect\cite{Furukawa:2010}).}
\label{fig:constructaide:drift_completeness}
\end{figure}

For larger, more complete collections, existing automatic techniques methods may suffice, although a manual coordinate-system registration process must still be used to bring the cameras into the 3D model's coordinate system. Our system can also handle spatially dense sets (e.g. Table~\ref{tab:quant_construction} Northwest B) but makes no explicit requirements and typically yields improvements and reduced accumulated error compared to existing methods (Fig~\ref{fig:constructaide:drift_completeness}).


\subsection{Limitations}

Currently, our system relies on a small amount of user interaction to provide pixel-precise registration. Our method automates several tasks (such as the handling of occlusions and computing sun direction), but if errors occur, the user must correct these using the smart-selection tools provided by our interface. 

{Our technique requires accurate and complete BIM data which are typically only available for commercial construction sites. To get the most out of our system, the BIMs must also contain semantic information such as materials, scheduling information, and sufficient levels of detail; the input photographs should be both spatially and temporally dense as well. As such, our system may not be suitable for certain photo collections and construction sites.} While our user study provided preliminary information validating our system's usefulness and applicability, a comprehensive study with more users is necessary to evaluate the user interface and practicality in the field. 

\boldhead{Failure cases} As with many structure from motion techniques, our method may fail to register an image in the presence of either inadequate/inaccurate feature matches or a poor initialization. In these cases, we allow the user to fix the registration by selecting 2D point correspondences, as described in Sec~\ref{sec:MeshSFM}. Furthermore, occlusion estimation may fail due to too few nearby photographs. For example, our static occlusion detection can fail if a SfM point cloud is too sparse or unavailable, and our dynamic occlusion detection can fail if no nearby viewpoints/images exist in the photo collection.

\subsection{Conclusion}

We have demonstrated a system that aligns 4D architectural/construction models and photographs with high accuracy and minimal user interaction. Our system is quick and easy to use, and enables many valuable job-site visualization techniques. Our interface can be used to navigate the construction site both forwards and backwards in time, assess construction progress, analyze deviations from the building plans, and create photorealistic architectural visualizations  without the time and training required to learn complicated CAD, modeling and rendering software. Our proposed model-assisted SfM algorithm and outperforms existing techniques achieves precise registration, enabling semantic selection tools and accurate visualization. Using this data, we show that occlusions can be reasoned accurately, and materials and lighting can be extracted from the plan data for rendering. 

{Many technical challenges remain for further automation. For example, assessing construction progress and errors automatically would be very useful for site inspectors. Automatically registering a model to a photo (similar to~\cite{Aubry13}) is another interesting avenue for future work, especially in the presence of incomplete or inaccurate building models. Optimizing these processes (such as SfM) is also necessary for enabling our system on mobile devices so that it can be used in real-time on job sites.
}



\begin{figure*}    
\includegraphics[page=1,width=0.105\linewidth,clip=true,trim=350 -50 0 0]{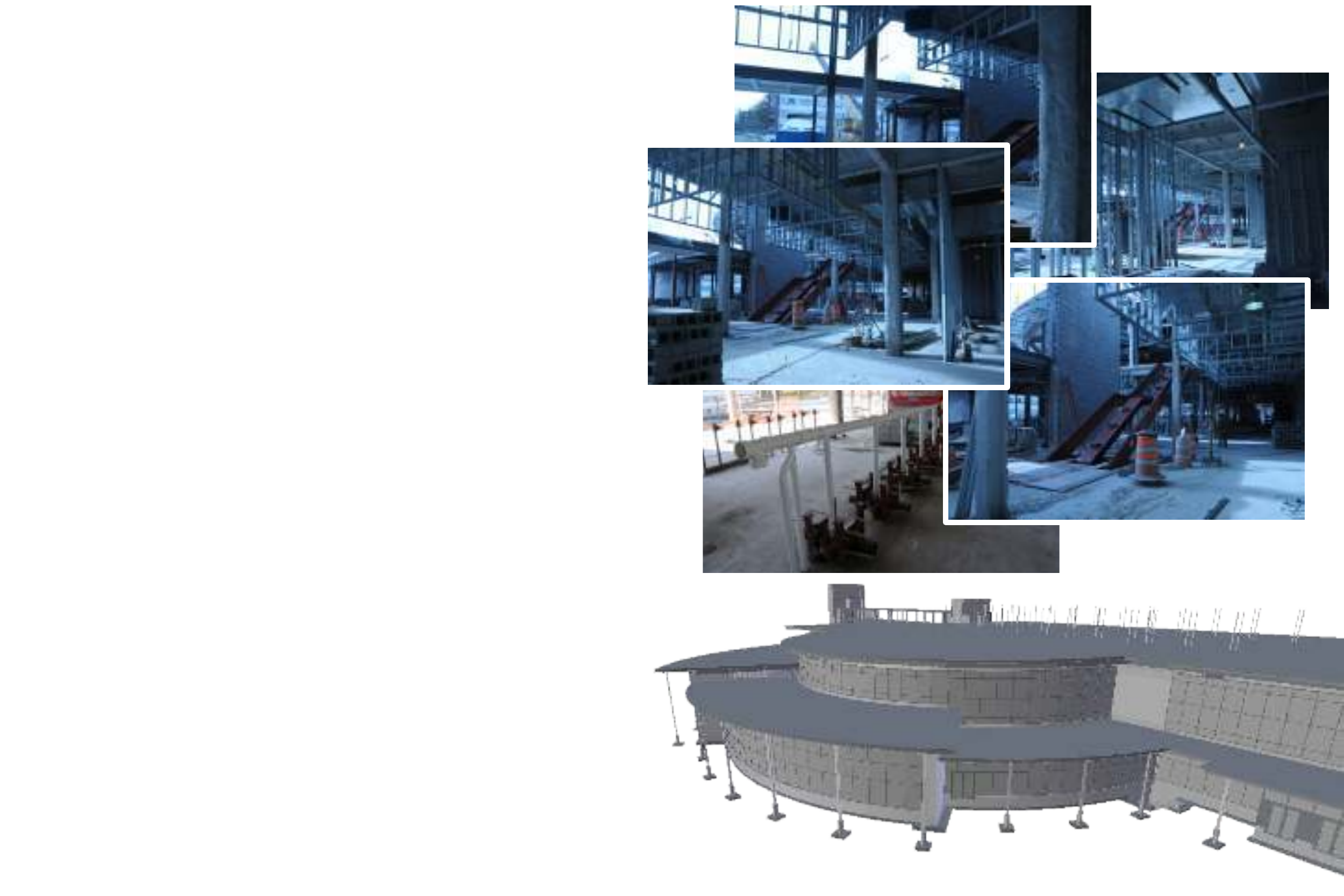}
\includegraphics[page=6,width=0.2175\linewidth,clip=true,trim=0 -50 0 0]{constructaide/fig/teaser.pdf}
\includegraphics[page=5,width=0.2175\linewidth,clip=true,trim=40 0 80 0]{constructaide/fig/teaser.pdf}
\includegraphics[page=3,width=0.2175\linewidth,clip=true,trim=40 0 80 0]{constructaide/fig/teaser.pdf}
\includegraphics[page=4,width=0.2175\linewidth,clip=true,trim=40 0 80 0]{constructaide/fig/teaser.pdf}
\caption{ConstructAide results on the Basement dataset, demonstrating architectural rendering, performance monitoring, and temporal navigation.}
\label{fig:constructaide:teaser2}
\end{figure*}

\begin{figure*}
\centerline{\includegraphics[width=.9\linewidth]{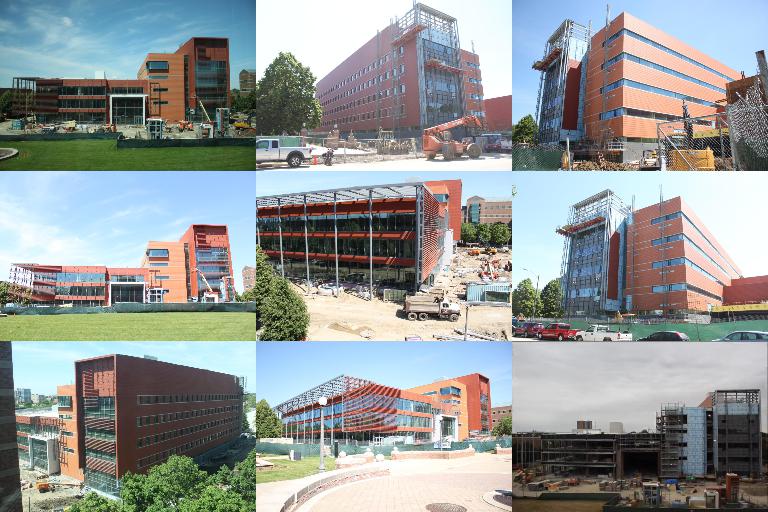}}
\vspace{10mm}
\centerline{\includegraphics[width=.9\linewidth]{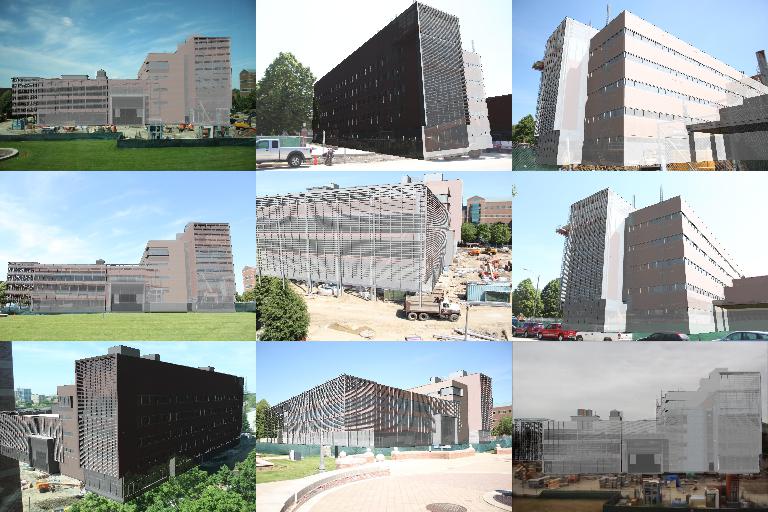}}
\caption{Example registrations estimated with our Model-assisted SfM procedure. Original photos on top, rendered overlay demonstrating registrations on bottom.}
\label{fig:constructaide:registration2}
\end{figure*}

\begin{figure*}
\includegraphics[width=0.33\linewidth]{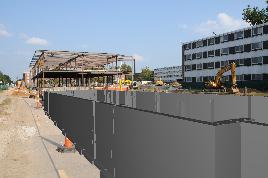}
\includegraphics[width=0.33\linewidth]{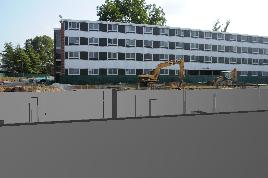}
\includegraphics[width=0.33\linewidth]{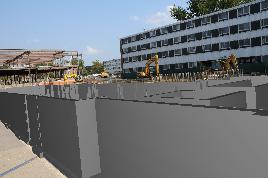}
\\
\includegraphics[width=0.33\linewidth]{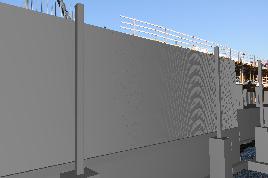}
\includegraphics[width=0.33\linewidth]{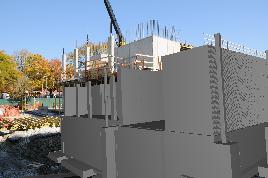}
\includegraphics[width=0.33\linewidth]{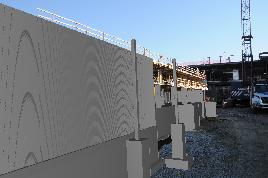}
\caption{Registration results from the RH4 A and RH10 datasets (used in our quantitative evaluation.}
\label{fig:constructaide:qual_construction}
\end{figure*}

\begin{figure*}
\includegraphics[width=\linewidth]{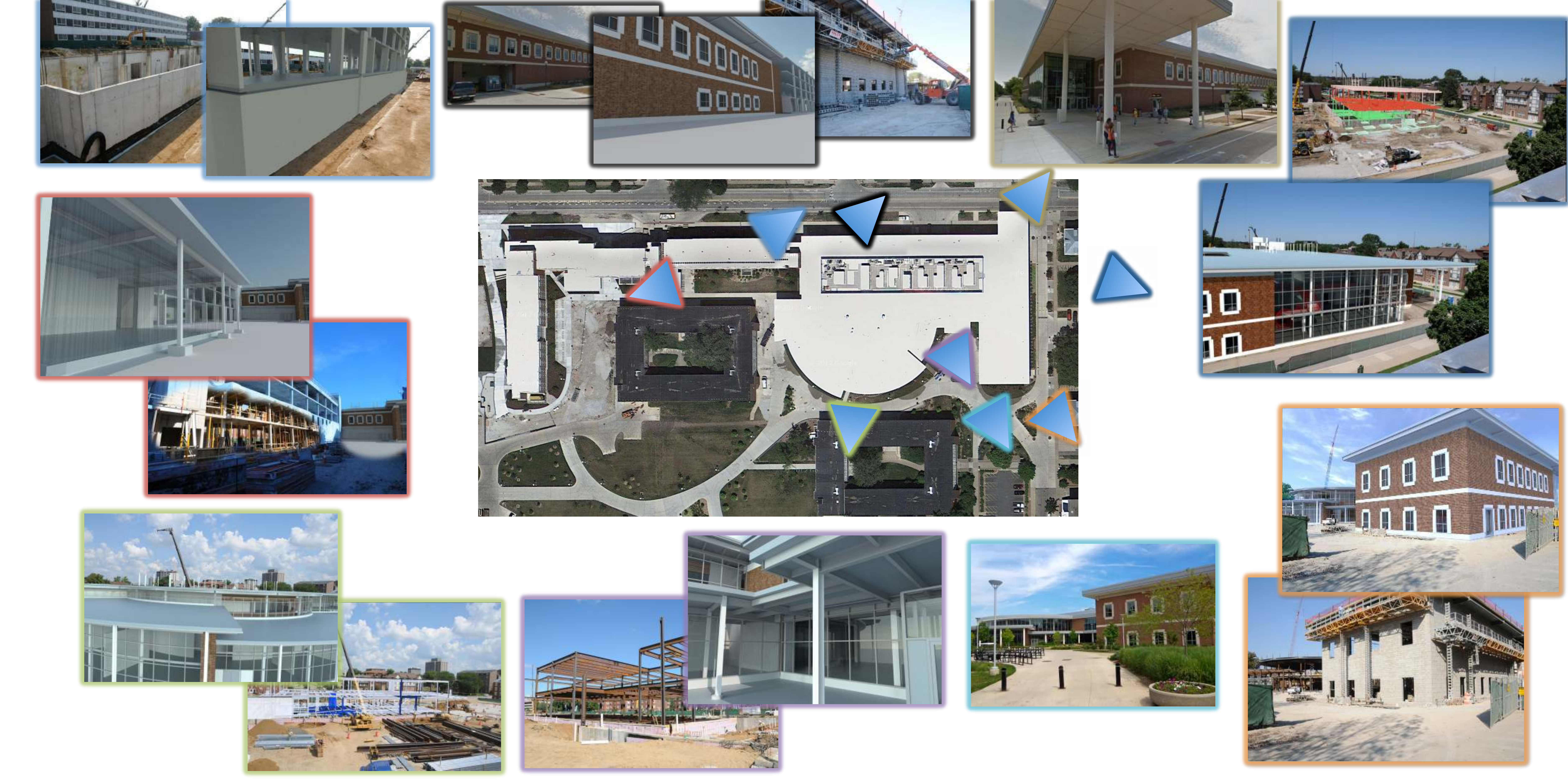}
\caption{Additional results for other construction datasets -- triangles indicate the rough camera pose of the data with respect to an overhead view of the construction sites, and border colors relate the cameras to surrounding photos. Our interface can help in generating job site summaries such as these. In the future, ConstructAide aims to automatically produce such summaries.}
\label{fig:constructaide:map}
\end{figure*}

\chapter{Conclusion}
\label{chap:conclusion}

The methods proposed in this thesis cast new light onto the fundamental computer vision problem of understanding intrinsic scene properties from single images. Our work shows overall improvements for inverse rendering tasks: shape estimation and representation, material inference and classification, and lighting estimation. These methods are useful in many computer graphics, computer vision and machine learning tasks, and we demonstrate that our scene estimates can enable powerful image editing operations with ease. This work has culminated in the first entirely automatic, single image inverse rendering engine, including a photorealistic, drag-and-drop image editor.

There are still many problems left to solve. This thesis focuses exclusively on inverse rendering from single images. Extending inverse rendering to video sequences is a very interesting line of future research. A video sequence contains much more useful information than a single photograph (such as parallax and multiple views of the scene), but at the same time there are many new challenges that arise for interpreting video sequences. This thesis provides some insight on how to begin attacking such problems.

While the estimates our methods produce are ideal for certain applications, they can be inexact and unsuitable for certain tasks such as path planning and canonical reconstruction. Improving inverse rendering for these and other tasks beyond our work will likely require improved scene representations (e.g. how the geometry, BRDFs, light sources, etc, are parameterized) and new joint estimation procedures. For example, if lighting and geometry were estimated simultaneously (rather than sequentially as in Chapter~\ref{chap:illum}), it's likely that errors in geometry estimates could be corrected as to produce a more realistic rendered image, and in turn, the lighting would be less affected by areas of inaccurate geometry. This line of reasoning extends beyond just lighting and geometry -- if all of the inverse rendering parameters were estimated jointly (including materials, camera information, and perhaps semantic labels about what objects are), the estimates are bound to improve under the right priors and parameterizations. These representations and estimators will likely require an immense amount of ground truth data in the form of images with known configurations of geometry/material/light which, at present, do not exist. With enough data, it may even be possible to automatically learn such representations and estimators. There is also a great deal of computational efficiency that must be achieved in order to make inverse rendering practical for many applications (e.g. robotics or others requiring real-time processing). The strides made in inverse rendering over the past few years have paved the way for solving this longstanding problem in computer vision, and future research on this topic will soon allow computers to understand and interpret photographs the same way that humans do.

\backmatter

\bibliographystyle{ieee}
\bibliography{combined.bib}


\chapter{Appendix A: Contour fold constraint for non-linear optimization}
\setcounter{equation}{0} \renewcommand{\theequation}{A.\arabic{equation}}
\hypertarget{appendix:foldConstraint}{}

Consider the $(i)$th point on the contour $C$, parametrized by position $\mathbf{p} = [ \mathbf{p}_x, \, \mathbf{p}_y]$ and tangent vector $\mathbf{u}= [ \mathbf{u}_x,\, \mathbf{u}_y]$, both on the image plane. The sign of the tangent vector is arbitrary. Let us define a vector perpendicular to each tangent vector: $\mathbf{v} = [-\mathbf{u}_y, \, \mathbf{u}_x]$. By default, this fold is convex --- folded in the direction of negative $Z$. To construct a concave fold, we flip the sign of $\mathbf{v}$. With this parametrization, we can find the positions of the points to the left and right of the point in question relative to the contour:
\begin{eqnarray}
\mathbf{p}^{\ell} = [ \round{ \mathbf{p}_x + \mathbf{v}_x}, \, \round{ \mathbf{p}_y + \mathbf{v}_y} ] \\
\mathbf{p}^r = [ \round{ \mathbf{p}_x - \mathbf{v}_x}, \, \round{ \mathbf{p}_y - \mathbf{v}_y} ]
\end{eqnarray}
Given a normal field $N$ we compute the normal of the surface at these ``left'' and ``right'' points:
\begin{eqnarray}
N^{\ell} = [ N_x( \mathbf{p}^{\ell}_x, \mathbf{p}^{\ell}_y), \, N_y( \mathbf{p}^{\ell}_x, \mathbf{p}^{\ell}_y), \, N_z( \mathbf{p}^{\ell}_x, \mathbf{p}^{\ell}_y) ] \\
N^{r} = [N_x( \mathbf{p}^{r}_x, \mathbf{p}^{r}_y), \, N_y(\mathbf{p}^{r}_x, \mathbf{p}^{r}_y), \, N_z( \mathbf{p}^{r}_x, \mathbf{p}^{r}_y) ]
\end{eqnarray}
Consider $c$, the dot product of $[\mathbf{u}_x, \mathbf{u}_y, 0]$ with the cross-product of $\mathbf{n}^{\ell}$ and $\mathbf{n}^r$:
\begin{eqnarray}
c =  \mathbf{u}_x ( N^{\ell}_y N^r_z - N^{\ell}_z N^r_y) + \mathbf{u}_y ( N^{\ell}_z N^{r}_x - N^{\ell}_x N^r_z)
\end{eqnarray}
If $c=1$, then the cross product of the surface normals on both sides of the contour is exactly equal to the tangent vector, and the surface is therefore convexly folded in the direction of the contour. If $c=-1$, then the surface is folded and concave. Of course, If the sign of the contour, and therefore of the $\mathbf{v}$ vector, is flipped, then $c=1$ when the surface is concavely folded, etc. Intuitively, to force the surface to satisfy the fold constraint imposed by the contour, we should force $c$ to be as close to $1$ as possible. This is the insight used in edge constraint of the shape-from-contour algorithm in \cite{MalikMaydan}. But constraining $c=1$ is not appropriate for our purposes, as it ignores the fact that $\mathbf{u}$ and therefore $\mathbf{v}$ lie in an image plane, while the true tangent vector of the contour may not be parallel to the image plane. To account for such contours, we will therefore penalized $c$ for being significantly smaller than $1$. More concretely, we will minimize the following cost with respect to each contour pixel:
\begin{eqnarray}
f_{\mathit{fold}}(N(Z)) = \sum_{i \in C} \max(0, \epsilon - \supi{c}),
\end{eqnarray}
where $\epsilon = {1 \over \sqrt{2}}$. This is a sort of $\epsilon$-insensitive hinge loss which allows for fold contours to be oriented as much as $45^{\circ}$ out of the image plane. In practice, the value of $\epsilon$ effects how sharp the contours produced by the fold-constraint are --- $\epsilon=0$ is satisfied by a flat fronto-parallel plane, and $\epsilon=1$ is only satisfied by a perfect fold whose crease is parallel with the image plane. In our experience, $\epsilon = {1 \over \sqrt{2}}$ produces folds that are roughly $90^{\circ}$, and which look reasonable upon inspection.

\chapter{Appendix B: Constrained bundle adjustment}
\setcounter{equation}{0} \renewcommand{\theequation}{B.\arabic{equation}}
\hypertarget{appendix:constrainedBA}{}

Bundle adjustment solutions for camera and geometry that differ only by a change of coordinate system (a gauge transformation) must have the same reprojection error.  This effect is an important difficulty for systems that must produce general reconstructions. The effect is particularly pronounced as the percentage of camera pairs that view the same geometry goes down. In some cases even the structure of the gauge group is not clear~\cite{SNAVELY-IJCV08}, and complex strategies apply~\cite{Bartoli:CVPR03,TZRbook}. Our case is simpler: we expect a high percentage of camera pairs to share features, and so we can resolve this issue by fixing the coordinates of one camera, the anchor camera~\cite{Hartley:CVPR92}.

In typical SfM bundle adjustment formulations, reprojection error is minimized by simultaneously adjusting intrinsic and extrinsic camera parameters, and triangulated points $X$. Let $\mathbb{P} = \{\mathbb{P}_1, \ldots, \mathbb{P}_N\}$ be the set of all camera parameters corresponding to the $N$ images, and  tracks${}_i$ be the pixel locations of keypoint tracks in image $i$. The classic bundle adjustment problem is formulated as a nonlinear least squares problem,
\begin{equation}
\argmin_{\mathbb{P},\ X} \sum_{i=1}^N \sum_{u \in \text{tracks}_i} ||\text{project}(\mathbb{P}_i,X_u) - u||,
\label{eq:BA}
\end{equation}
where $X_u$ is a triangulated point corresponding to pixel $u$, and $\text{project}(\cdot)$ is the function that projects 3D locations into 2D according to a set camera parameters.

We formulate a new version of this problem, {\it constrained} bundle adjustment, which leverages one or more calibrated cameras. In our system, the user provides guidance for registering a mesh to the initial image input to our system, giving a very good camera pose estimate for this image. We call this camera an {\it anchor}, and denote its parameters $\mathbb{P}_\dagger$. During bundle adjustment, this anchor camera is used to constrain the 3D points such that any point triangulated using a feature point from the anchor camera must lie along the ray generated by the anchor camera. Therefore, we re-parameterize points as $X_u(t_u) = \mathbb{P}_\dagger^{\text{center}} + t_u \ \mathbb{P}_\dagger^{\text{ray}}(u)$, where $t_u$ is a scalar and $\mathbb{P}_\dagger^{\text{center}}$ is the anchor camera center and $\mathbb{P}_\dagger^{\text{ray}}(u)$ is the ray generated from pixel $u$ in the anchor camera. Our formulation then becomes

\begin{align}
\argmin_{\mathbb{P} \setminus \mathbb{P}_\dagger,\ t} \sum_{i=1}^N & \left[ \sum_{u \in \text{tracks}_\dagger} ||\text{project}(\mathbb{P}_i,X_u(t_u)) - u|| \ + \right.  \nonumber \\
& \left. \sum_{u \in \text{tracks}_i \setminus \text{tracks}_\dagger}  ||\text{project}(\mathbb{P}_i,X_u) - u|| \right].
\label{eq:constrainedBA}
\end{align}

Notice that the anchor's camera parameters are left out of the bundle adjustment, and any tracks that are not seen by the anchor camera revert to the classic bundle adjustment formulation.

From our experience, this formulation typically provides better estimates since the model is constrained by accurate camera parameters. Also, it has an added benefit of having fewer parameters to optimize over, increasing optimization efficiency and reducing variance in the estimates. One downside is that this model can be inflexible if the other initial camera estimates are too poor, and we also propose a ``soft-constrained'' bundle adjustment in these cases:

\begin{equation}
\argmin_{\mathbb{P}\setminus \mathbb{P}_\dagger,\ X} \sum_{i=1}^N  \sum_{u \in \text{tracks}_i} w_i\ ||\text{project}(\mathbb{P}_i,X_u) - u||,
\label{eq:softconstrainedBA}
\end{equation}

where $w_i$ is a scalar weight dependent on each image. We set the anchor image's weight to a large value (e.g. 100), and all other image weights to 1, enforcing the reprojection error for the anchor camera to be much smaller than other cameras. This has a similar effect as Eq.~\ref{eq:constrainedBA}, but allows for flexibility in the 3D point locations.

The user may also guide the registration of other images if certain conditions are met, adding new anchor cameras.

\end{document}